\documentclass{bhamthesis} 
\usepackage{amsthm} 
\usepackage{amssymb} 
\usepackage{amsmath}
\usepackage{pdfpages}
\usepackage{latexsym}
\usepackage{adjustbox,lipsum}
\usepackage{color,soul}
\usepackage{xcolor}
\usepackage{tikz,tikz-3dplot}
\usepackage{pgfplots}
\pgfplotsset{width=7cm,compat=1.3,scale only axis}
\usepackage{tikz-dependency}
\usepackage{cancel}
\usetikzlibrary{plotmarks}
\usepackage{graphicx}
\graphicspath{ {./} }
\usepackage[linesnumbered,lined,boxed,commentsnumbered]{algorithm2e}

\usepackage{titlesec}
\titleformat{\chapter}[display] {\large\bfseries\centering}{\MakeUppercase{\chaptertitlename}\ \thechapter\vspace{3ex}}{0px}{\Large\uppercase}

\newcommand{\blankpage}{
	\newpage
	\thispagestyle{empty}
	\mbox{}
	\newpage
}

\renewcommand{\contentsname}{CONTENTS}
\renewcommand{\listtablename}{LIST OF TABELS}
\renewcommand{\listfigurename}{LIST OF FIGURES}

\title{Semi-Supervised Methods for Out-of-Domain Dependency Parsing}

\author{Juntao Yu}

\school{School of Computer Science}

\begin{document}

\frontmatter
\newcommand{\build}{} 
\newcommand{\rQone}{Could the off-the-shelf dependency parsers be successfully used in co-training for domain adaptation?}
\newcommand{\rQtwo}{Would tri-training be more effective for out-of-domain parsing when off-the-shelf dependency parsers are used?}
\newcommand{\rQthree}{How could self-training be effectively used in out-of-domain dependency parsing?}
\newcommand{\rQfour}{If self-training works for English dependency parsing, can it be adapted to other languages?}
\newcommand{\rQfive}{Can dependency language models be adapted to strong transition-based parsers?}
\newcommand{\rQsix}{Can dependency language models be used for out-of-domain parsing?}
\newcommand{\rQseven}{Quality or quantity of the auto-parsed data, which one is more important to the successful use of dependency language models?}

%\date{\today} 
\date{November 2017}

\maketitle 
\blankpage

\begin{abstract}
Dependency parsing is one of the important natural language processing tasks that assigns syntactic trees to texts. Due to the wider availability of dependency corpora and improved parsing and machine learning techniques, parsing accuracies of supervised learning-based systems have been significantly improved. However, due to the nature of supervised learning, those parsing systems highly rely on the manually annotated training corpora. They work reasonably good on the in-domain data but the performance drops significantly when tested on out-of-domain texts.  To bridge the performance gap between in-domain and out-of-domain, this thesis investigates three semi-supervised techniques for out-of-domain dependency parsing, namely co-training, self-training and dependency language models. Our approaches use easily obtainable unlabelled data to improve out-of-domain parsing accuracies without the need of expensive corpora annotation. The evaluations on several English domains and multi-lingual data show quite good improvements on parsing accuracy. Overall this work conducted a survey of semi-supervised methods for out-of-domain dependency parsing, where I extended and compared a number of important semi-supervised methods in a unified framework. The comparison between those techniques shows that self-training works equally well as co-training on out-of-domain parsing, while dependency language models can improve both in- and out-of-domain accuracies.

\end{abstract}

\begin{dedication}
To my wonderful wife Mingyu Du.
\end{dedication}

\begin{acknowledgements}
Now nearly four years, since I first come to Birmingham, I and my wife had a great time here.  I would take this opportunity to thank all the friends who supported and took care of us during our time in Birmingham.  

First of all, I would like to thank my primary supervisor Bernd Bohnet, who is not only a great supervisor but also a good friend. Four years ago, when Bernd first got me into his group, I have very limited knowledge about the natural language processing and research in general. During those years, through our meetings (majorly in the Costa and recently on the train :) ), he equipped me with all I need for my PhD study.  From how to use Mate, to writing my first paper, preparing my first conference talk, applying for travel funding, applying for jobs and writing this thesis, whenever I needed help, Bernd is always there to support me. Because of Bernd, I had a great time in Birmingham!

I would like also to thank my co-supervisor Mark and John for their supervision and took care of me within the department. For their feedbacks on my research, papers and this thesis.

It would be less joy if we don't have all the friends here in the UK, I would like to thank all the lovely friends for the wonderful time we spend together!

Finally, I would like to thank my family for their support and encouragement. Without their help, I would not even start my degree. I would especially thank my wonderful wife for make every important decision with me and took care of me all the time. For cooking me the delicious food, growing me the beautiful garden, they are huge contributors to the happiness of life and of course the vanilla lattes. For the time you spent to listen the talks start with "my name is" which made you an expert of "self-training"!  For introducing me the work-life balance, for introducing the bolt journal, for an endless list of things you did for me, it is hard to imagine a life without you.
\end{acknowledgements}

\tableofcontents

\listoffigures
\listoftables

\mainmatter

\chapter{Introduction}\label{chapter:intro}
Syntactic parsing is an important natural language processing (NLP) task that focuses on analysing the syntactic structures of sentences. The syntax of a sentence has been found to be important to many other NLP tasks that require deeper analysis of the sentences, such as semantic parsing \cite{surdeanu08conll,hajic09conll}, anaphora resolution \cite{pradhan11conll,pradhan12conll} and machine translation \cite{tiedemann12}. There are two major families of syntactic parsing, the first one is constituency parsing that generates parse trees of sentences according to phrase structure grammars, the other is dependency parsing that assigns head-child relations to the words of a sentence. Initially, the parsing community mainly focused on constituency parsing systems, as a result,Πa number of high accuracy constituency parsers have been introduced, such as the Collins Parser \cite{collins1999phd}, Stanford PCFG Parser \cite{klein03}, BLLIP reranking parser \cite{charniak2005} and Berkeley Parser \cite{petrov07}. In the past decade, dependency-based systems have gained more and more attention \cite{mcdonald2006online,nivre2009non,martins2010turbo,bohnet2013joint,martins2013turning}, as they have a better multi-lingual capacity and are more efficient. For a long period, dependency parsing systems were mainly based on carefully selected feature sets, we denote those systems as conventional dependency parsers. In the recent years, a number of dependency parsing systems based on neural networks have also been investigated, some of which have achieved better accuracies when compared to conventional dependency parsers. We evaluated our approaches only on conventional dependency parsers, as these neural network-based systems were introduced after we finished most of the work. However, the techniques evaluated in this thesis have the potential to be adapted to neural network-based parsers as well.

Many dependency parsers are based on supervised learning techniques, which could produce high accuracy when trained on a large amount of training data from the same domain \cite{mcdonald2006online,nivre2009non,martins2010turbo,bohnet2013joint,martins2013turning}. However, those models trained on the specific training data are vulnerable when dealing with data from domains different from the training data \cite{nivre07conll,petrov2012overview}. One effective way to make models less domain specific is to annotate more balanced corpora. However, the annotation work is very time-consuming and expensive. As a result of these difficulties, only very limited annotations are available to the community. As an alternative to annotating new corpora, domain adaptation techniques have been introduced to train more robust models for out-of-domain parsing. Semi-supervised methods are one family of those techniques that aim to improve the out-of-domain parsing performance by enhancing the in-domain models with a large amount of unlabelled data. Some semi-supervised methods use the unlabelled data as the additional training data,  such as co-training \cite{sarkar01,sagae07,zhang12hit} and self-training \cite{mcclosky2006reranking,reichart2007self,sagae2010self}. Alternatively, other research uses the unlabelled data indirectly. Word clusters \cite{zhou2011exploiting,pekar2014exploring}  and word embeddings \cite{chen2014neural,weiss2015neural} are examples of this direction.

\section{Research Questions}
The focus of this thesis is on using semi-supervised techniques to bridge the accuracies between the in-domain and the out-of-domain dependency parsing. More precisely, this thesis evaluates three important semi-supervised methods, namely co-training, self-training and dependency language models. Two of the methods use unlabelled data directly as additional training data (i.e. co-/self-training). Co-training is a method that has been used in many domain adaptation tasks, it uses multiple learners to derive additional training data from unlabelled target domain data. The successful use of co-training is conditioned on learners being as different as possible. Previous work on parsing with co-training is mainly focused on using learners that are carefully designed to be very different. In this thesis, we use only off-the-shelf dependency parsers as our learners to form our co-training approaches. In total, we evaluate two co-training approaches, the normal co-training (uses two parsers) and the tri-training (uses three parsers). For both approaches, the evaluation learner is retrained on the additional training data annotated identically by two source learners. The normal co-training uses two learners, the evaluation learner is used as one of the source learners, while the tri-training uses three learners, two of which are used as source learners, the third one is used as the evaluation learner. Compare to the normal co-training, tri-training approach allows the evaluation learner to learn from the novel annotations that is \textit{not} predicted by its own. For our evaluation on co-training, we trying to answer the following research questions:

\textbf{Q1.} \rQone
%Could the off-the-shelf dependency parsers be successfully used in co-training for domain adaptation?

\textbf{Q2.} \rQtwo
%Would tri-training be more effective for out-of-domain parsing when off-the-shelf dependency parsers are used?

In contrast to co-training, which retrains the parser on additional training data annotated by multiple learners, self-training retrains the parser on training data enlarged by its own automatically labelled data. Previous research mainly focused on applying self-training to constituency parsers \cite{mcclosky2006reranking,reichart2007self,sagae2010self}. Attempts to use self-training for dependency parsing either need additional classifiers \cite{kawahara2008learning} or only use partial parse trees \cite{chen2008learning}. In this thesis, we aim to find a more effective way to use self-training for dependency parsing. We intend to answer the following research questions for our self-training evaluation:

\textbf{Q3. } \rQthree
%How could self-training be effectively used in out-of-domain dependency parsing?

\textbf{Q4. } \rQfour
%If self-training works for English dependency parsing, can it be adapted to other languages?

To use auto-labelled data as additional training data is effective but comes with consequences. First of all, the re-trained models usually have a lower performance on the source domain data. Secondly, those approaches can only use a relatively small unlabelled data, as training parsers on a large corpus might be time-consuming or even intractable on a corpus of millions of sentences. To overcome those limitations we investigate dependency language models which use the unlabelled data indirectly. Dependency language models (DLM) were previously used by  \newcite{chen2012utilizing} to leverage the performance and the efficiency of a weak second-order graph-based parser \cite{mcdonald2006online}. In this thesis, we adapt this method to a strong transition-based parser \cite{bohnet2013joint} that on its own can produce very promising accuracies. The research questions for this part are as follows:

\textbf{Q5. } \rQfive
%Can dependency language models be adapted to strong transition-based parsers?

\textbf{Q6. } \rQsix
%Can dependency language models be used for out-of-domain parsing?

\textbf{Q7. } \rQseven
%Quality or quantity of the auto-parsed data, which one is more important to the successful use of dependency language models?

\section{Thesis Structure}

After the introduction, in Chapter \ref{chapter:background} we begin by discussing the background knowledge and previous work related to this thesis. This mainly covers two topics, dependency parsing and domain adaptation. We then introduce the Mate parser in detail. Mate is a strong transition-based parser which is used in all of our evaluations. After that, we introduce the corpora and the evaluation/analysis methods.

In Chapter \ref{chapter:cotrain} we introduce our experiments on agreement-based co-training. It first discusses the effect of using different off-the-shelf parsers on a normal agreement-based co-training setting (i.e. only involves two parsers). And then we introduce our experiments on its variant that uses three parsers (tri-training). 

Chapter \ref{chapter:selftrain} and Chapter \ref{chapter:multiselftrain} introduce our confidence-based self-training approaches. In Chapter \ref{chapter:selftrain}, we introduce our evaluations on confidence-based self-training for English out-of-domain dependency parsing. In total, two confidence-based methods are compared in our experiments. Chapter \ref{chapter:multiselftrain} introduces our experiments on multi-lingual datasets. The confidence-based self-training approach is evaluated on nine languages.

Chapter \ref{chapter:dlm} discusses our dependency language models method that is able to improve both in-domain and out-of-domain parsing. The evaluations on English include both in-domain and out-of-domain datasets, in addition to that, we also evaluated on the Chinese in-domain data.

Chapter \ref{chapter:conclusion} provides a summary of the thesis and gives conclusions. 

\section{Published Work}

In total, there are four publications based on this thesis. Each of the publications is related to one chapter of this thesis, \newcite{pekar2014exploring} is related to our evaluation on co-training (Chapter \ref{chapter:cotrain}). \newcite{yu2015iwpt} is made from our English self-training evaluation (Chapter \ref{chapter:selftrain}). \newcite{yu2015depling} is associated with our multi-lingual self-training experiments (Chapter \ref{chapter:multiselftrain}). \newcite{yu2017iwpt} presents our work on dependency language models (Chapter \ref{chapter:dlm}).

\begin{description}
	\item {\bf Juntao Yu} and Bernd Bohnet. 2017. Dependency language models for transition-based dependency parsing. In \emph{Proceeding of the 15th International Conference on Parsing Technologies,} pages 11-17, Pisa, Italy. Association for Computational Linguistics.
	\item {\bf Juntao Yu} and Bernd Bohnet. 2015. Exploring confidence-based self-training for multilingual dependency parsing in an under-resourced language scenario. In \emph{Proceeding of the Third International Conference on Dependency Linguistics,} pages 350-358, Uppsala, Sweden. Uppsala University.
	\item {\bf Juntao Yu}, Mohab Elkaref, and Bernd Bohnet. 2015. Domain adaptation for dependency parsing via self-training. In \emph{Proceeding of the 14th International Conference on Parsing Technologies,} pages 1-10, Bilbao, Spain. Association for Computational Linguistics.
	\item Viktor Pekar, {\bf Juntao Yu}, Mohab Elkaref, and Bernd Bohnet. 2014. Exploring options for fast domain adaptation of dependency parsers. In \emph{Proceedings of the First Joint Workshop on Statistical Parsing of Morphologically Rich Languages and Syntactic Analysis of Non-Canonical Languages,} pages 54-65, Dublin, Ireland. Dublin City University.
	
\end{description}

\section{Chapter Summary}
In this chapter, we first briefly introduced dependency parsing and the problems of out-of-domain parsing that we are trying to address in this thesis. We then discussed the research questions that we intend to answer. The chapter also gave a brief introduction of the thesis structure. Finally, the chapter illustrated the published works based on this thesis.

\chapter{Background and Experiment Set-up}\label{chapter:background}
In this chapter, we first introduce the background and related work of this thesis, which includes a brief introduction of dependency parsing systems, a detailed introduction of the baseline parser \cite{bohnet2013joint} and previous work on out-of-domain parsing (especially those on semi-supervised approaches). We then introduce the corpora that have been used in this thesis. Finally, we introduce the evaluation metric and the analysis methods.
\section{Dependency parsing}\label{section:depparsing}
\begin{figure}[t]
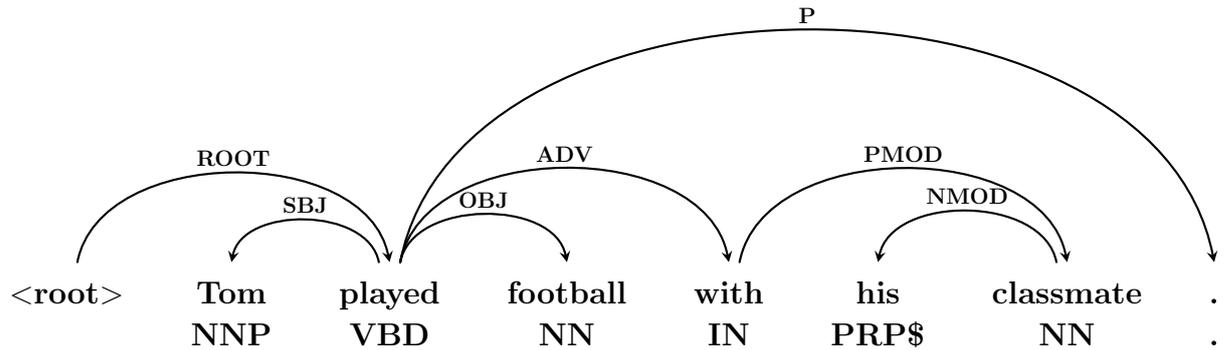

 \begin{center}
 \begin{dependency}[theme=simple,edge vertical padding=1ex,edge style={thick}, label style={font=\bfseries}] 	  
 
 \tikzstyle{word}=[font=\bfseries]
 
 \begin{deptext}[column sep=0.7cm, row sep=0.3ex, nodes={word}]
 $<$root$>$ \& Tom\& played\& football\& with\&his\&classmate\&.\\
 \& NNP\& VBD\&NN\&IN\&PRP\$\&NN\&.\\
 \end{deptext}
 \depedge{1}{3}{ROOT}
 \depedge{3}{2}{SBJ}
 \depedge{3}{4}{OBJ}
 \depedge{3}{5}{ADV}
 \depedge{3}{8}{P}
 \depedge{5}{7}{PMOD}
 \depedge{7}{6}{NMOD}
 \end{dependency}
 \end{center}
\caption{\label{figure:dep-tree} The dependency relations of the sentence (\textit{Tom played football with his classmate .}) parsed by Mate parser.}
\end{figure}

Dependency parsing is one important way to analyse the syntactic structures of natural language. It has been widely studied in the past decade. A dependency parsing task takes natural language (usually tokenised sentence) as input and outputs a sequence of head-dependent relations. Figure \ref{figure:dep-tree} shows the dependency relations of a sentence (\textit{Tom played football with his classmate .}) parsed by an off-the-shelf dependency parser. During the past decade, many dependency parsing systems have been introduced, most of them are graph-based or transition-based systems. The graph-based system solves the parsing problem by searching for maximum spanning trees (MST). A first-order MST parser first assigns scores to directed edges between tokens of a sentence. It then uses an algorithm to search a valid dependency tree with the highest score. By contrast, the transition-based system solves the parsing task as a sequence of transition decisions, in each step the parser deciding the next transition. In Section \ref{graphparser} and \ref{transitionparser} we briefly describe the two major system types. In recent years, deep learning has been playing an important role in the machine learning community. As a result, several neural network-based systems have been introduced, some of them surpassing the state-of-the-art accuracy achieved by the conventional dependency parsers based on perceptions or SVMs. We briefly touch on neural network-based systems in Section \ref{nnparser}, although most of them are still transition/graph-based systems. The evaluation of the neural network-based parsers is beyond the scope of this thesis, as they become popular after most of the work of this thesis has been done. We mainly use the Mate parser \cite{bohnet2013joint}, a transition-based approach that was state-of-the-art at the beginning of this work and whose performance remained competitive even after the introduction of the parsers based on neural network. Section \ref{mateparser} introduces the technical details of the Mate parser.
 
\subsection{Graph-based Systems} \label{graphparser}
\begin{figure}[t]
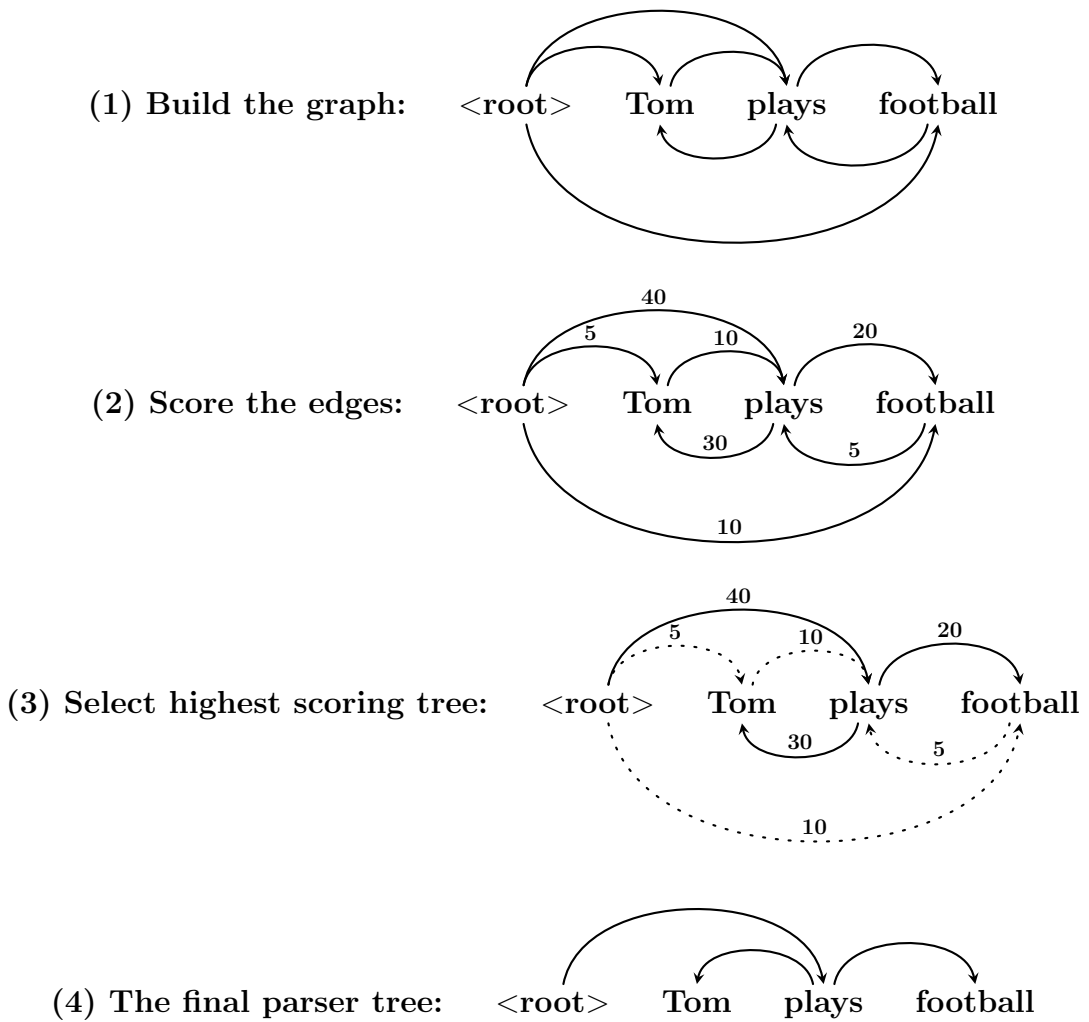

\tikzstyle{word}=[font=\normalsize\bfseries]
\depstyle{dash}{edge style = {thick, loosely dotted}}
 \begin{center}
  \begin{dependency}[theme=simple,hide label,edge style={thick}, label style={font=\bfseries}]    
  \begin{deptext}[column sep=0.5cm, row sep=0.3ex,nodes={word}]
(1) Build the graph:\&$<$root$>$ \& Tom\&plays\& football\\
 \end{deptext}
 \depedge{2}{3}{}
 \depedge{2}{4}{}
 \depedge[edge below]{2}{5}{}
 \depedge[edge below]{4}{3}{}
 \depedge{4}{5}{}
 \depedge[edge below]{5}{4}{}
 \depedge{3}{4}{}
 \end{dependency}
  
\begin{dependency}[theme=simple,edge style={thick}, label style={font=\bfseries}]    
  \begin{deptext}[column sep=0.5cm, row sep=0.3ex,nodes={word}]
(2) Score the edges:\&$<$root$>$ \& Tom\&plays\& football\\
 \end{deptext}
 \depedge{2}{3}{5}
 \depedge{2}{4}{40}
 \depedge[edge below]{2}{5}{10}
 \depedge[edge below]{4}{3}{30}
 \depedge{4}{5}{20}
 \depedge[edge below]{5}{4}{5}
 \depedge{3}{4}{10}
 \end{dependency}

\begin{dependency}[theme=simple,edge style={thick}, label style={font=\bfseries}]    
\begin{deptext}[column sep=0.5cm, row sep=0.3ex,nodes={word}]
(3) Select highest scoring tree:\&$<$root$>$ \& Tom\&plays\& football\\
 \end{deptext}
 \depedge[dash]{2}{3}{5}
 \depedge{2}{4}{40}
 \depedge[dash,edge below]{2}{5}{10}
 \depedge[edge below]{4}{3}{30}
 \depedge{4}{5}{20}
 \depedge[dash,edge below]{5}{4}{5}
 \depedge[dash]{3}{4}{10}
 \end{dependency}

 \begin{dependency}[theme=simple,hide label,edge style={thick}, label style={font=\bfseries}]    
\begin{deptext}[column sep=.5cm, row sep=0.3ex,nodes={word}]
(4) The final parser tree:\&$<$root$>$ \& Tom\&plays\& football\\
 \end{deptext}
 \depedge{2}{4}{40}
 \depedge{4}{3}{30}
 \depedge{4}{5}{20}
 \end{dependency}
 
  \end{center}
\caption{\label{figure:graph_example} Parsing the sentence (\textit{Tom plays football}) with a graph-based dependency parser.}
\end{figure}

The graph-based dependency parser solves the parsing problem by searching for maximum spanning trees (MST). In the following, we consider the first-order MST parser of \newcite{mcdonald05acl}. Let $x$ be the input sentence, $y$ be the dependency tree of $x$, $x_i$ is the $i$th word of $x$, $(i,j) \in y$ is the directed edge between $x_i$ (head) and $x_j$ (dependent). $dt(x)$ is used to represent the set of possible dependency trees of the input sentence where $y \in dt(x)$.  The parser considers all valid directed edges between tokens in $x$ and builds the parse trees in a bottom-up fashion by applying a CKY parsing algorithm. It scores a parse tree $y$ by summing up the scores $s(i,j)$ of all the edges $(i,j) \in y$. The $s(i,j)$ is calculated according to a high-dimensional binary feature representation $f$ and a weight vector $w$ learned from training data $\tau$ ($ \tau = \{(x_t,y_t)\}_{t=1}^{T} $). To be more specific, the score of a parse tree $y$ of an input sentence $x$ is calculated as follows:

$$s(x,y)=\sum_{(i,j) \in y}s(i,j)=\sum_{(i,j) \in y} w * f(i,j) $$

Where $f$ consists of a set of binary feature representations associated with a number of feature templates. For example, an edge $(plays, football)$ with a bi-gram feature template $(head_{word}, dep_{word})$ will give a value of 1 for the following feature representation:

$$f(i,j)=\begin{cases}1\ if\ head_{word}=``plays"\ and\ dep_{word}=``football"\\
	0\ otherwise\end{cases}$$

After scoring the possible parse trees $dt(x)$, the parser outputs the highest-scored dependency tree $y_{best}$. Figure \ref{figure:graph_example} shows an example of a sentence being parsed with a first-order graph-based parser. 

In terms of training, the parser uses an online learning algorithm to learn the weight vector $w$ from the training set $\tau$. In each training step, only one training instance $(x_t, y_t)$ ($(x_t, y_t) \in \tau$) is considered, the $w$ is updated after each step.  More precisely, the Margin Infused Relaxed Algorithm (MIRA) \cite{crammer2006online} is used to create a margin between the score of a correct parse tree $s(x_t,y_t)$ and the incorrect ones $s(x_t,y')$ ($y'\in dt(x_t)$). The loss $L(y_t,y')$ of a dependency tree is defined as the number of incorrect edges. Let $w^{(i)}$, $w^{(i+1)}$ be the weight vector before and after the update of the $i$th training step, $w^{(i+1)}$ is updated subject to keeping the margin at least as large as the $L(y_t,y')$, while at the same time, keeping the norm of the changes to the $w$ as small as possible. A more detailed training algorithm is showed in algorithm \ref{algorithm_mst}.

\begin{algorithm}[h]
\SetAlgoLined
\KwData{$\tau = \{(x_t,y_t)\}_{t=1}^{T}$}
\KwResult{$w$}
$w^0=0; i=0$\;
\For(\tcp*[h]{N training iterations}){$n : 1..N $ }{
\For{$t :1..T$}{
$w^{(i+1)} = update\ w^{(i)}\ to\ min\ ||w^{(i+1)}-w^{(i)}||$\;
$s.t.\ \ s(x_t,y_t)-s(x_t,y') \geq L(y_t,y')$\;
$\forall y' \in dt(x_t)$\;
\BlankLine
$i = i + 1$\;
}
}
\caption{MIRA algorithm for MST parser}\label{algorithm_mst}

\end{algorithm}

The MST parser is later improved by \newcite{mcdonald2006online} to include second-order features, however, the system is still weaker than its successors which also include third-order features \cite{koo10acl}. Other mostly used strong graph-based parsers include Mate graph-based parser \cite{bohnet10} and Turbo Parser \cite{martins2013turning}.

\begin{figure}[t]
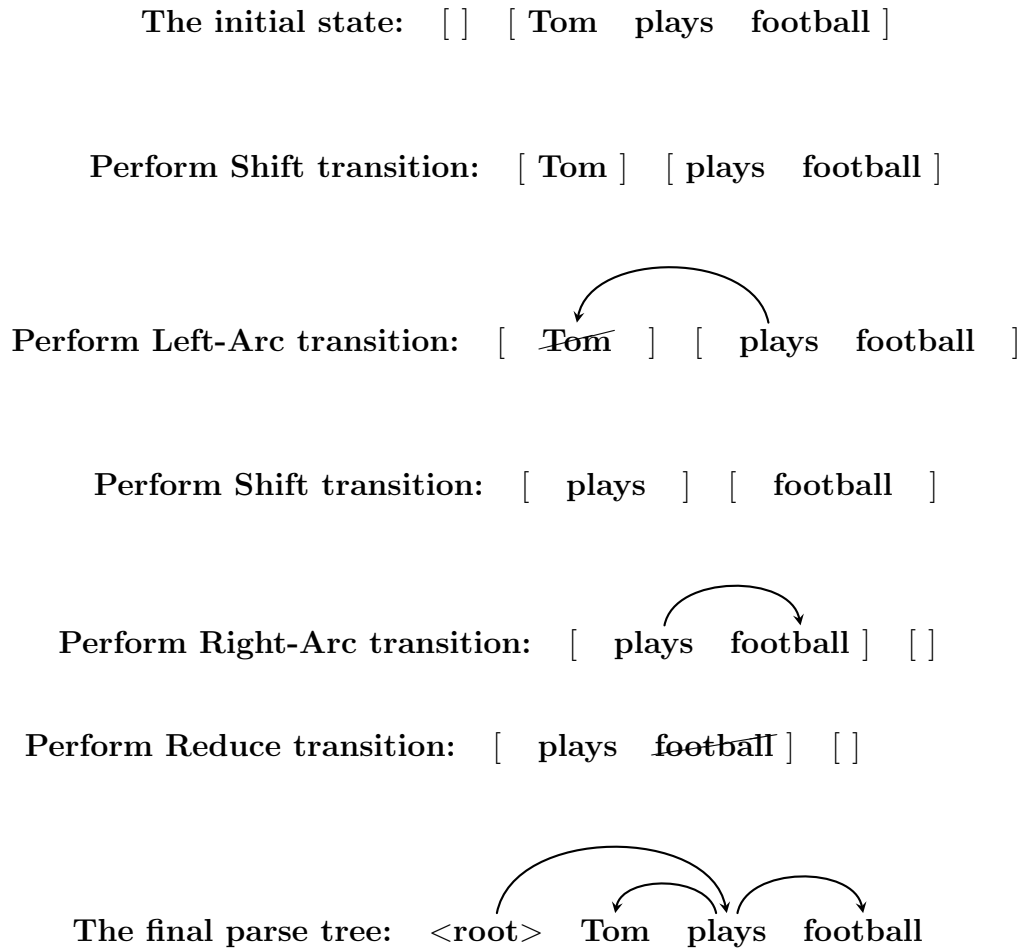

\tikzstyle{word}=[font=\normalsize\bfseries]
 \begin{center}
  \begin{dependency}[theme=simple,edge style={thick}, label style={font=\bfseries}]    
  \begin{deptext}[column sep=0.3cm, row sep=0.3ex,nodes={word}]
The initial state:\& $[$ $]$ \&  $[$ Tom\&plays\& football $]$\\ \\\\
 \end{deptext}
 \end{dependency}

  \begin{dependency}[theme=simple,edge style={thick}, label style={font=\bfseries}]   
  \begin{deptext}[column sep=0.3cm, row sep=0.3ex,nodes={word}]
 Perform Shift transition: \& $[$ Tom  $]$ \&  $[$ plays \& football $]$\\  \\
 \end{deptext}
 \end{dependency}

  \begin{dependency}[theme=simple,hide label,edge style={thick}, label style={font=\bfseries}]   
  \begin{deptext}[column sep=0.3cm, row sep=0.3ex,nodes={word}]
 Perform Left-Arc transition: \& $[$  \& \cancel{Tom}   \& $]$ \&  $[$\& plays \& football \& $]$\\\\\\
 \end{deptext}
 \depedge{6}{3}{SBJ}
 \end{dependency}

  \begin{dependency}[theme=simple,edge style={thick}, label style={font=\bfseries}]   
  \begin{deptext}[column sep=0.3cm, row sep=0.3ex,nodes={word}]
 Perform Shift transition: \& $[$  \& plays \& $]$ \&  $[$ \& football \& $]$\\\\
 \end{deptext}
 \end{dependency}

  \begin{dependency}[theme=simple,hide label,edge style={thick}, label style={font=\bfseries}]   
  \begin{deptext}[column sep=0.3cm, row sep=0.3ex,nodes={word}]
Perform Right-Arc transition: \& $[$  \& plays \& football $]$ \&  $[$ $]$\&\\\\
 \end{deptext}
 \depedge{3}{4}{OBJ}
 \end{dependency}
 
  \begin{dependency}[theme=simple,hide label,edge style={thick}, label style={font=\bfseries}]   
  \begin{deptext}[column sep=0.3cm, row sep=0.3ex,nodes={word}]
Perform Reduce transition: \& $[$  \& plays\& \cancel{football} $]$ \&  $[$ $]$\& \& \& \&\\\\
 \end{deptext}
 \end{dependency}
 
 \begin{dependency}[theme=simple,hide label,edge style={thick}, label style={font=\bfseries}]    
 \begin{deptext}[column sep=0.3cm, row sep=0.3ex,nodes={word}]
The final parse tree:\&  $<$root$>$ \& Tom\&plays\& football\&\\ 
 \end{deptext}
 \depedge{2}{4}{ROOT}
 \depedge{4}{5}{OBJ}
 \depedge{4}{3}{SBJ}
 \end{dependency}
 \end{center}
\caption[Parsing the sentence (\textit{Tom plays football}) with an arc-eager transition-based dependency parser.]{\label{figure:transition_example} Parsing the sentence (\textit{Tom plays football}) with an arc-eager transition-based dependency parser. The square brackets denote the stack (left) and the buffer (right) used by transition-based parser.}
\end{figure}

\subsection{Transition-based Systems} \label{transitionparser}
The transition-based parsers build the dependency trees in a very different fashion compared to graph-based systems. Instead of searching for the maximum spanning trees, transition-based systems parse a sentence with a few pre-defined transitions. The Malt parser \cite{nivre07nle} is one of the earliest transition-based parsers which has been later widely used by researchers. The parser is well engineered and can be configured to use different transition systems. We take the parser's default transition system (arc-eager) as an example to show how the transition-based parser works. The Malt parser starts with an initial configuration and performs one transition at a time in a deterministic fashion until it reaches the final configuration. The parser's configurations are represented by triples $c=(\Sigma, B, A)$, where $\Sigma$ is the stack that stores partially visited tokens, $B$ is a list of remaining tokens that are unvisited, and $A$ stores the directed arcs between token pairs that have already been parsed. The parser's initial configuration consists of an empty $\Sigma$ and an empty $A$, while all the input tokens are stored in $B$. The final configuration is required to have an empty $B$. A set of four transitions (Shift, Left-Arc, Right-Arc and Reduce) are defined to build the parse trees. The Shift transition moves the token on the top of $B$ into $\Sigma$, the Left-Arc transition adds an arc from the top of $B$ to the top of $\Sigma$ and removes the token on the top of $\Sigma$, the Right-Arc transition adds an arc from the top of $\Sigma$ to the top of $B$ and moves the token on the top of $B$ into $\Sigma$, and the Reduce transition simply removes the token on the top of $\Sigma$. More precisely,  table \ref{table-arc-eager-transitions} shows the details of the transitions of an arc-eager system. 

\begin{table}[t]
\begin{center}
\begin{tabular}{|ll|}
\hline \bf Transition&\\
\textsc{Left-Arc}&\( ([\sigma|i], [j|\beta], A) \Rightarrow (\sigma, [j|\beta], A \cup \{(j \to i)\} \)\\
\textsc{Right-Arc}&\( ([\sigma|i], [j|\beta], A) \Rightarrow ([\sigma|i|j], \beta, A \cup \{(i \to j)\}) \)\\
\textsc{Shift}&\( (\sigma, [i|\beta], A) \Rightarrow ([\sigma|i], \beta, A) \)\\
\textsc{Reduce}&\( ([\sigma|i], B, A) \Rightarrow (\sigma, B, A) \)\\
 \hline
\end{tabular}
\end{center}
\caption{\label{table-arc-eager-transitions} Transitions for arc-eager parsing. }
\end{table} 

To train the parser, support vector machine classifier (SVM) with the one-versus-all strategy is used to solve the transition-based parser as a multi-classification problem. In a transition-based parsing scenario, the classes are different transitions. Each of the SVMs is trained to maximise the margin between the target transition and the other transitions, as in the one-versus-all strategy the classes other than the target class are treated the same as the negative examples. Since the data may not be linearly separable, they use in additional a quadratic kernel ($K(x_i,x_j) = (\gamma x_i^T x_j + r)^2$) to map the data into a higher dimensional space. The SVMs are trained to predict the next transition based on a given parser configuration. They used similar binary feature representations as those of the MST parser, in which the features are mapped into a high dimensional vector. The feature templates for the transition-based system are mainly associated with the configurations, for example, a feature between the $\Sigma_{top}$ (the top of the stack) and the $B_{top}$ (the top of the Buffer) is as follows:

$$f_{c_i}=\begin{cases}1\ if\ \Sigma_{top}=``plays"\ and\ B_{top}=``football"\\
	0\ otherwise\end{cases}$$

Figure \ref{figure:transition_example} shows an example of parsing the sentence (\textit{Tom plays football}) with the Malt transition-based parser. 

Benefiting from the deterministic algorithm, the Malt parser is able to parse the non-projective sentences in linear time \cite{nivre2009non}, which is much faster compared to the second-order MST parser's cubic-time parsing \cite{mcdonald2006online}. Although the deterministic parsing is fast, the error made in the previous transitions will largely affect the decisions taken afterwards, which results in a lower accuracy. To overcome this problem beam search has been introduced to the transition-based systems, which leads to significant accuracy improvements \cite{bohnet2013joint}.

\subsection{Neural Network-based Systems}\label{nnparser}
\begin{figure}
\includegraphics[width=\textwidth]{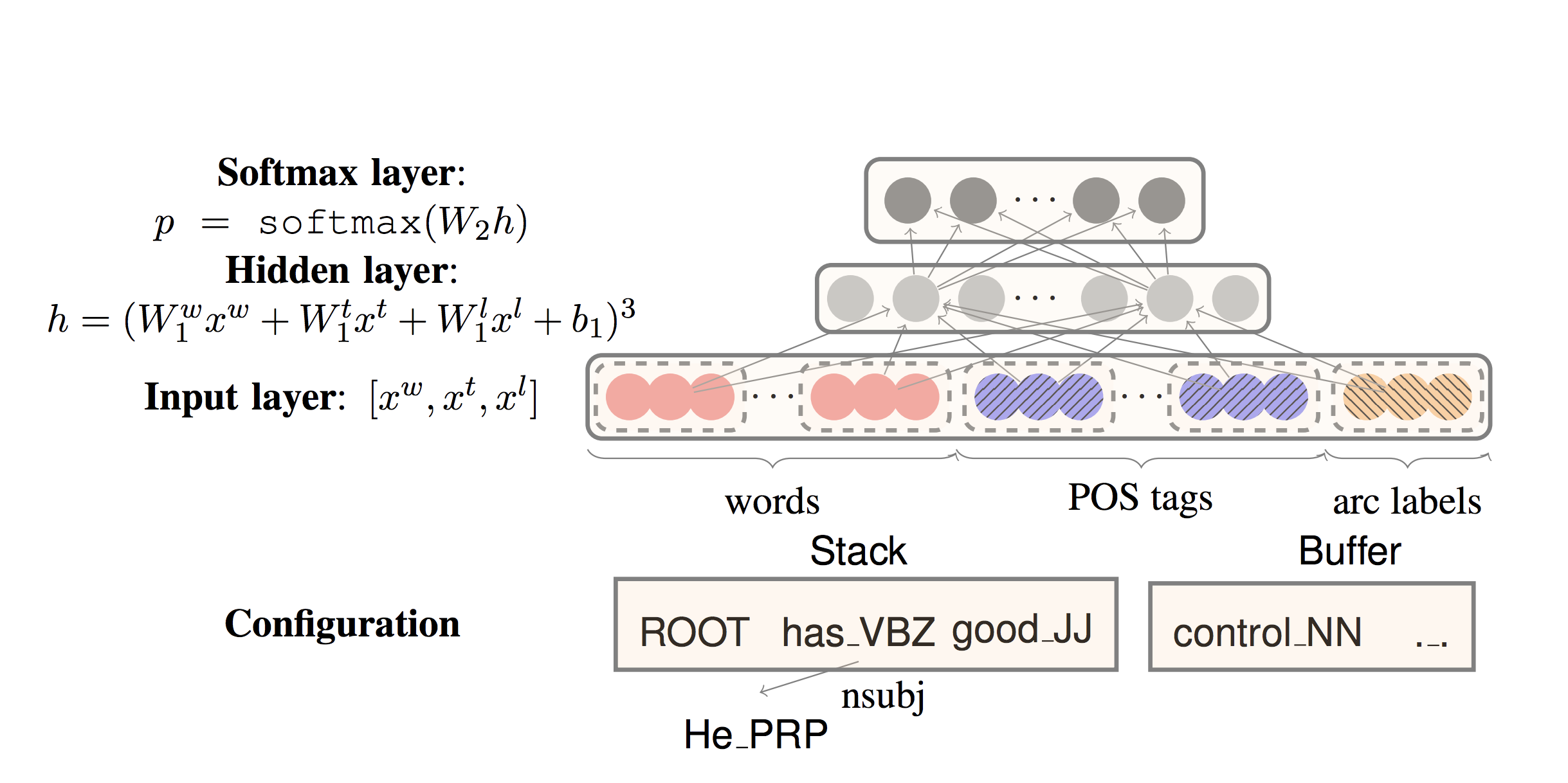}
\caption[Neural Network architecture of Chen and Manning (2014) system]{\label{figure:neural_network_architecture} Neural Network architecture taking from \newcite{chen2014neural} }
\end{figure}

Neural network-based systems have only been recently introduced to the literature. \newcite{chen2014neural} were the first to introduce a simple neural network to a deterministic transition-based parser, yielding good results. The parser used an arc-standard transition system. Similar to arc-eager, the arc-standard is another highly used transition-based system.  Many dependency parsers are based on or have options to use an arc-standard approach, which include the Malt parser we introduced in the previous section (section \ref{transitionparser}) and our main evaluation parser (Mate parser). We will introduce the arc-standard transition system in more detail in section \ref{mateparser}.  

One of the major differences between the neural network based systems and the conventional systems is the use of feature representations. Instead of using the binary feature representations (commonly used by the conventional systems), the neural network based approaches represent the features by embeddings. During training, feature embeddings (e.g. word, part-of-speech embeddings) are capable of capturing the semantic information of the features. Take the part-of-speech tags as an example, adjective tags $JJ, JJR, JJS$ will have similar embeddings. This allows the neural network-based systems to reduce the feature sparsity problem of the conventional parser systems. Conventional parsers usually represent different tokens or token combinations by independent feature spaces, thus are highly sparse. 

Another advantage of using the neural network based approach is that the system allows using the pre-trained word embeddings. Word embeddings extracted from large unlabelled data carry the statistical strength of the words, this could be a better bases for the system when compared to the randomly initialised embeddings. The empirical results confirmed that large improvements can be achieved by using the pre-trained word embeddings. The idea of using the pre-trained word embeddings goes into the same direction of the semi-supervised approaches that use unlabelled data indirectly, such as dependency language models evaluated in this thesis, or word clusters. 

In terms of the network architecture, \newcite{chen2014neural} used a single hidden layer and a softmax layer to predict the next transition based on the current configuration. To map the input layer to the hidden layer they used a cube activation function ($h = (W^wx^w+W^tx^t+W^lx^l+b)^3$), in which $x^w,x^t,x^l$ are feature embeddings of the words, part-of-speech tags and arc labels and $W^w,W^t,W^l$ are the relative weights. Figure \ref{figure:neural_network_architecture}  shows the details of their neural network architecture.

This first attempt of using the neural network for dependency parsing leads to many subsequent research. \newcite{chen2014neural}'s system has been later extended by \newcite{weiss2015neural} who introduced beam search to the system and achieved state-of-the-art accuracy. Since then a number of more complex and powerful neural networks have been evaluated, such as the stack-LSTM \cite{dyer2015slstm} and the bi-directional LSTM \cite{dozat2017deep}. The current state-of-the-art is achieved by the parser of \newcite{dozat2017deep} who used the bi-directional LSTM in their system.

\subsection{The Mate Parser}\label{mateparser}
In this thesis, we mainly used the Mate transition-based parser \cite{bohnet2012eacl,bohnet2012emnlp,bohnet2013joint}. The parser is one of the best performing parsers on the data set of the major shared task (CoNLL 2009) on dependency parsing \cite{hajic09conll} and it is freely available \footnote{https://code.google.com/p/mate-tools/}.
The parser uses the arc-standard transition system, it is also integrated with a number of techniques to maximise the parser's performance. Firstly, the parser employs a beam search to go beyond the greedy approach. Secondly, it uses  an additional optional graph-based model to rescore the beam entries. In their paper \cite{bohnet2012eacl}, they name it completion model as it scores factors of the graph as soon as they are finished by the parser. Furthermore, the parser has an option for joint tagging and parsing \cite{bohnet2012emnlp}. Same as the pipeline system, the tagger model is trained separately from the parser model. However, during the parsing, instead of using only the best-predicted part-of-speech (PoS) tag, they made the n-best ($n>1$) PoS tags of a token available to the parser. The joint system is able to gain a higher accuracy for
both PoS tagging and parsing compared to a pipeline system. In this thesis, we use the Mate parser as our baseline and make the necessary modifications, where appropriate to comply with the requirements of our approaches.

\begin{table}[t]
\begin{center}
\begin{tabular}{|lll|}
\hline \bf Transition&&\bf Condition\\
\textsc{Left-Arc\(_d\)}&\( ([\sigma|i, j], B, A, \pi, \delta) \Rightarrow ([\sigma|j], B, A \cup \{(j, i)\}, \pi, \delta[(j, i) \to d]) \)&\( i \neq 0 \)\\
\textsc{Right-Arc\(_d\)}&\( ([\sigma|i, j], B, A, \pi, \delta) \Rightarrow ([\sigma|i], B, A \cup \{(i, j)\}, \pi, \delta[(i, j) \to d]) \)&\\
\textsc{Shift\(_p\)}&\( (\sigma, [i|\beta], A, \pi, \delta) \Rightarrow ([\sigma|i], \beta, A, \pi[i\top], \delta) \)&\\
\textsc{Swap}&\( ([\sigma|i, j], B, A, \pi, \delta) \Rightarrow ([\sigma|j], [i|\beta], A, \pi, \delta) \)&\( 0<i<j \) \\
 \hline

\end{tabular}
\end{center}
\caption[Transitions for joint tagging and parsing.]{\label{table-transitions} Transitions for joint tagging and parsing taking from \newcite{bohnet2013joint}. $\Sigma$ (the stack) is represented as a list with its head to the right and a tail $\sigma$; The buffer $B$ as a list with its head to the left and tail $\beta$. }
\end{table}

The transition-based part of the parser uses a modified arc-standard transition system. Comparing to the original arc-standard transition system (has only three transitions: Left-Arc, Right-Arc and Shift) of \newcite{nivre2004incrementality}, the Mate parser modified the \textsc{Shift} transition for joint tagging and parsing and included the \textsc{Swap} transition to handling non-projective parsing. More precisely, the parser tags and parses a sentence $x = w_1, ..., w_n$ using a sequence of transitions listed in Table \ref{table-transitions}. An additional artificial token $<$root$>$ $(w_0)$ is added to the beginning of the sentence to allow the parser assigning a \textsc{Root} to the sentence at the last step of the transitions.   
The transitions change the initial configuration ($c_s$) in steps until reaching a terminal configuration ($c_t$). \newcite{bohnet2013joint} used the 5-tuples $C = (\Sigma, B, A, \pi, \delta)$ to represent all configurations, where $\Sigma$ (the stack) and $B$ (the buffer) refers to disjoint sublists of the sentence $x$, $A$ is a set of arcs, $\pi$ and $\delta$ are functions to assign a part-of-speech tag to each word and a dependency label to each arc. The initial configuration ($c_s$) has an empty stack, the buffer consists of the full input sentence $x$, and the arc set $A$ is empty. The terminal configuration ($c_t$) is characterised by an empty stack and buffer, hence no further transitions can be taken. The arc set $A$ consists of a sequence of arc pairs ($i,j$), where $i$ is the head and $j$ is the dependent. They use \textsc{Tree$(x,c)$} to represent the tagged dependency tree defined for $x$ by $c=(\Sigma, B, A, \pi, \delta)$.

\begin{figure}[h!]
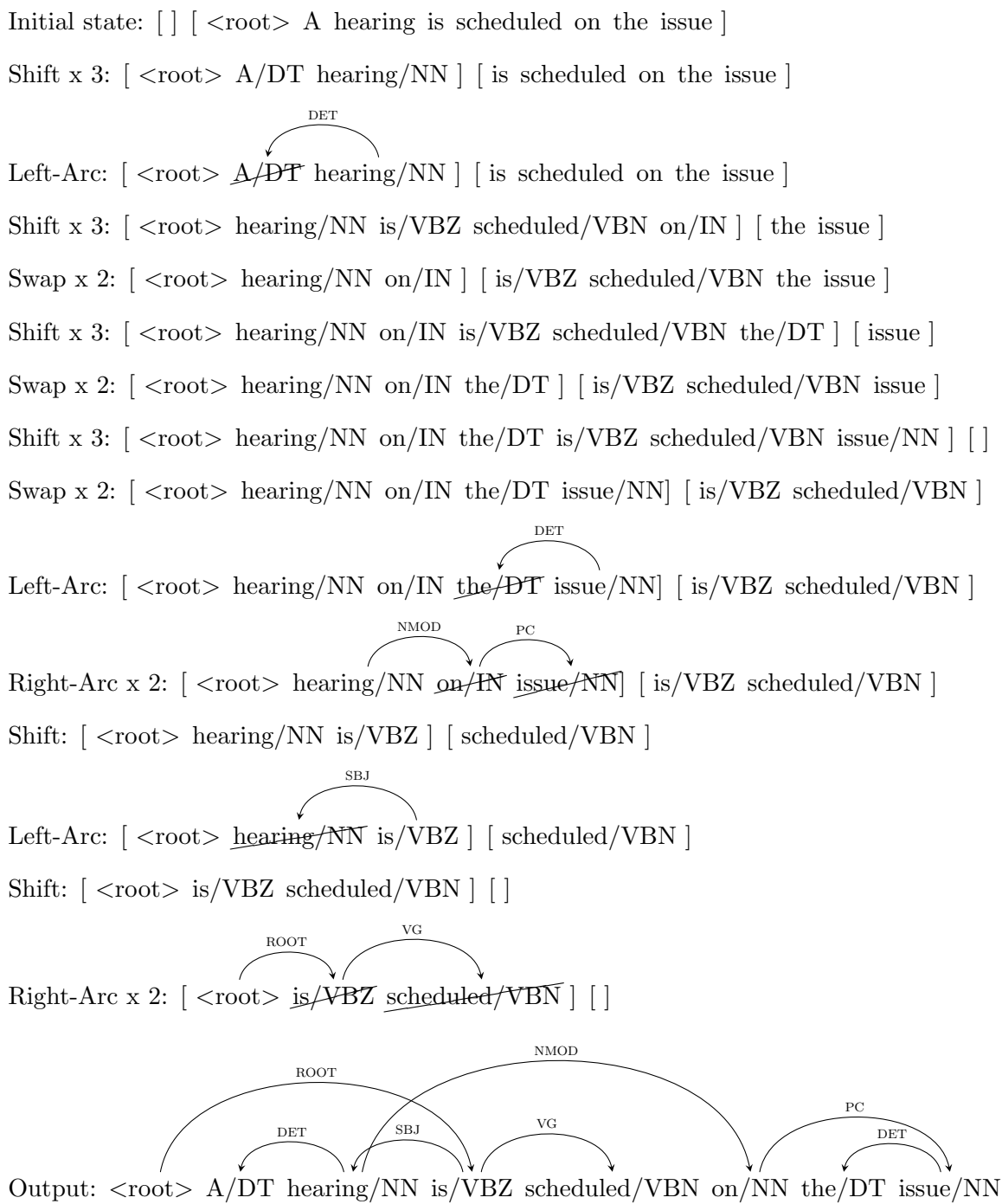

\tikzstyle{word}=[font=\small]
  \begin{dependency}[theme=simple,edge style={thick}, label style={font=\bfseries}]    
  \begin{deptext}[nodes={word}]
Initial state:\& $[$ $]$ \&  $[$ $<$root$>$ \& A\&hearing\& is\&scheduled\&on \&the\&issue $]$\\
 \end{deptext}
 \end{dependency}
  
  \begin{dependency}[theme=simple]   
  \begin{deptext}[nodes={word}]
Shift x 3: \& $[$ $<$root$>$  \&A/DT \& hearing/NN $]$ \&   $[$ is \& scheduled\&on\&the \& issue $]$\\
 \end{deptext}
 \end{dependency}

  \begin{dependency}[theme=simple]   
  \begin{deptext}[nodes={word}]
 Left-Arc: \& $[$ $<$root$>$ \& \cancel{A/DT} \& hearing/NN $]$ \&  $[$ is \& scheduled\&on\&the \& issue $]$\\
 \end{deptext}
 \depedge{4}{3}{DET}
 \end{dependency}

  \begin{dependency}[theme=simple]   
  \begin{deptext}[nodes={word}]
Shift x 3: \& $[$ $<$root$>$  \& hearing/NN  \&   is/VBZ \& scheduled/VBN\&on/IN $]$  \& $[$ the \& issue $]$\\
 \end{deptext}
 \end{dependency}
 
  \begin{dependency}[theme=simple]   
  \begin{deptext}[nodes={word}]
Swap x 2: \& $[$ $<$root$>$  \& hearing/NN  \&on/IN $]$  \& $[$ is/VBZ \& scheduled/VBN \& the \& issue $]$\\
 \end{deptext}
 \end{dependency}
 
  \begin{dependency}[theme=simple]   
  \begin{deptext}[nodes={word}]
Shift x 3: \& $[$ $<$root$>$  \& hearing/NN  \&on/IN \& is/VBZ \& scheduled/VBN \& the/DT $]$  \& $[$  issue $]$\\
 \end{deptext}
 \end{dependency}
 
  \begin{dependency}[theme=simple]   
  \begin{deptext}[nodes={word}]
Swap x 2: \& $[$ $<$root$>$  \& hearing/NN  \&on/IN \& the/DT $]$  \& $[$ is/VBZ \& scheduled/VBN  \& issue $]$\\
 \end{deptext}
 \end{dependency}
 
  \begin{dependency}[theme=simple]   
  \begin{deptext}[nodes={word}]
 Shift x 3: \& $[$ $<$root$>$  \& hearing/NN  \&on/IN \& the/DT \& is/VBZ \& scheduled/VBN  \& issue/NN $]$  \& $[$  $]$\\
 \end{deptext}
 \end{dependency}
 
  \begin{dependency}[theme=simple]   
  \begin{deptext}[nodes={word}]
Swap x 2: \& $[$ $<$root$>$  \& hearing/NN  \&on/IN \& the/DT \& issue/NN$]$  \& $[$ is/VBZ \& scheduled/VBN   $]$\\
 \end{deptext}
 \end{dependency}
 
  \begin{dependency}[theme=simple]   
  \begin{deptext}[nodes={word}]
 Left-Arc: \& $[$ $<$root$>$  \& hearing/NN  \&on/IN \& \cancel{the/DT} \& issue/NN$]$  \& $[$ is/VBZ \& scheduled/VBN   $]$\\
 \end{deptext}
 \depedge{6}{5}{DET}
 \end{dependency}
 
  \begin{dependency}[theme=simple]   
  \begin{deptext}[nodes={word}]
Right-Arc x 2: \& $[$ $<$root$>$  \& hearing/NN  \&\cancel{on/IN} \& \cancel{issue/NN}$]$  \& $[$ is/VBZ \& scheduled/VBN   $]$\\
 \end{deptext}
 \depedge{4}{5}{PC}
 \depedge{3}{4}{NMOD}
 \end{dependency}
 
  \begin{dependency}[theme=simple]   
  \begin{deptext}[nodes={word}]
 Shift: \& $[$ $<$root$>$  \& hearing/NN  \& is/VBZ  $]$  \& $[$ scheduled/VBN   $]$\\
 \end{deptext}
 \end{dependency}

  \begin{dependency}[theme=simple]   
  \begin{deptext}[nodes={word}]
 Left-Arc: \& $[$ $<$root$>$  \& \cancel{hearing/NN}  \&is/VBZ $]$  \& $[$ scheduled/VBN   $]$\\
 \end{deptext}
 \depedge{4}{3}{SBJ}
 \end{dependency}
 
  \begin{dependency}[theme=simple]   
  \begin{deptext}[nodes={word}]
 Shift: \& $[$ $<$root$>$  \& is/VBZ \& scheduled/VBN  $]$  \& $[$  $]$\\
 \end{deptext}
 \end{dependency}

  \begin{dependency}[theme=simple]   
  \begin{deptext}[nodes={word}]
Right-Arc x 2: \& $[$ $<$root$>$  \& \cancel{is/VBZ} \& \cancel{scheduled/VBN}  $]$  \& $[$  $]$\\
 \end{deptext}
 \depedge{2}{3}{ROOT}
  \depedge{3}{4}{VG}
 \end{dependency}
   
  \begin{dependency}[theme=simple]   
 \begin{deptext}[nodes={word}]
Output:\&  $<$root$>$\& A/DT\&hearing/NN\& is/VBZ\&scheduled/VBN\&on/NN \&the/DT\&issue/NN\\ 
 \end{deptext}
 \depedge{2}{5}{ROOT}
 \depedge{4}{3}{DET}
 \depedge{4}{7}{NMOD}
 \depedge{5}{4}{SBJ}
 \depedge{5}{6}{VG}
 \depedge{7}{9}{PC}
 \depedge{9}{8}{DET}
 \end{dependency}
\caption[Parsing the sentence (\textit{A hearing is scheduled on the issue}) with the Mate transition-based dependency parser.]{\label{figure:mate_example} Parsing the sentence (\textit{A hearing is scheduled on the issue}) with the Mate transition-based dependency parser. The square brackets denote the stack (left) and the buffer (right) used by transition-based parser.}
\end{figure}

As shown in Table \ref{table-transitions}, the \textsc{Left-Arc$_d$} adds an arc from the token ($j$) at the top of the stack ($\Sigma$) to the token ($i$) at the second top of the stack and removes the dependent ($i$) from the stack. At the same time, the $\delta$ function assigns a dependency label ($d$) to the newly created arc $(j,i)$. The \textsc{Left-Arc$_d$} transition is permissible as long as the token at the second top of the stack is \textit{not} the $<$root$>$ (i.e. $i \neq 0$).  The \textsc{Right-Arc$_d$} adds a labelled arc from the token ($i$) at the second top of the stack to the token ($j$) at the top of the stack and removes the later. The \textsc{Shift$_p$} transition assigns a PoS tag $p$ to the first node of the buffer and moves it to the top of the stack. The \textsc{Swap} transition that is used to handling non-projective tree extracts the token ($i$) at the second top of the stack and moves it back to the buffer. The \textsc{Swap} transition is only permissible when the top two tokens of the stack are in the original word order (i.e. $i < j$), this prevents the same two tokens from being swapped more than once. In additional, the artificial $<$root$>$ token is \textit{not} allowed to be swapped back to the buffer (i.e. $i > 0$). Figure \ref{figure:mate_example} shows an example of joint tagging and parsing a sentence by the Mate parser.

The graph-based completion model consists of a number of different second- and third-order feature models to rescore the partial parse tree \textsc{Tree$_n(x,c_n)$}. Some feature models are similar to \newcite{carreras07} and \newcite{koo10acl}. Take one of the models $2a$ as an example, which consists of the second-order factors of \newcite{carreras07}:
\begin{enumerate}
	\item The head and the dependent.
	\item The head, the dependent and the right/left-most grandchild in between.
	\item The head, the dependent and the right/left-most grandchild away from the head.
	\item The head, the dependent and between those words the right/left-most sibling.
\end{enumerate}

Feature models are independent to each other and can be easily turned on/off by configuration. The score of a parse tree \textsc{Tree$(x,c)$} or a partial parse tree \textsc{Tree$_n(x,c_n)$} is then defined as the sum of the scores from the both parts: 

$$Score(x,c) = Score_T(x,c) + Score_G(x,c)$$ 

Where $Score_T(x,c)$ is the score of the transition-based part of the parser and $Score_G(x,c)$ is the score from the graph-based completion model.

\begin{algorithm}[t]
\SetAlgoLined
\KwData{$(x, w, b)$}
\KwResult{$\textsc{Tree}(x, h.c)$}
$h_0.c \gets c_s(x)$\; 
$h_0.s \gets 0.0$\; 
$h_0.f \gets \{0.0\}^{dim(w)}$\; 
$\textsc{Beam} \gets [h0]$\;
\While{$\exists c \in \textsc{Beam} : c \notin C_t $}{
$\textsc{Tmp} \gets [\ ]$\;
\For{$h: \textsc{Beam} $ }{
\For{$t \in T: \textsc{Permissible}(c,t)$}{
$h.f \gets h.f + f(h.c,t)$\;
$h.s \gets h.s + f(h.c,t) * w$\;
$h.c \gets t(h.c)$\;
$\textsc{Tmp} \gets \textsc{Insert}(h,\textsc{Tmp})$\;
}
}
$\textsc{Beam} \gets \textsc{Prune}(\textsc{Tmp},b)$\;
}
$h \gets \textsc{Top}(\textsc{Beam})$\;
\textbf{return} $\textsc{Tree}(x, h.c)$\;
\caption{Beam search algorithm for the Mate parser}\label{algorithm_mate_beam_search}
\end{algorithm}

Mate parser uses similar binary feature representations as those of the MST/Malt parser (the features are represented by a high dimensional feature vector ($f$)).  A learned weight vector ($w$) is used with the feature vector ($f$) to score the configurations in conjunction with the next transition. In addition, the parser uses the beam search to mitigate error propagation. Comparing with the deterministic parsing algorithm that only keeps the best partial parse tree, the beam search approach keeps the n-best partial parse trees during the inference. By using the beam search, errors made in the early stage can potentially be recovered in the late stage, as long as the correct configuration has \textit{not} fallen out of the beam.  The beam search algorithm takes a sentence ($x$), the weight vector ($w$) and the beam size parameter ($b$) and returns the best scoring parse tree (\textsc{Tree}$(x, h.c)$). A parse hypothesis ($h$) of a sentence consists of a configuration ($h.c$), a score ($h.s$) and a feature vector ($h.f$). Initially the \textsc{Beam} only consists of the initial hypothesis ($h_0$), in which $h_0$ contains a initial configuration of the sentence ($c_s(x)$), a score of $0.0$ and a initial feature vector ($\{0.0\}^{dim(w)}$). The transitions ($T$) change the hypotheses in steps and create new hypotheses by applying different permissible transitions to them. For each step, the top $b$ scoring hypotheses are kept in the \textsc{Beam}. The beam search terminates when every hypothesis in the \textsc{Beam} contains a terminal configuration ($h.c \in C_t$). It then returns the top scoring parse tree (\textsc{Tree}$(x, h.c)$). Algorithm \ref{algorithm_mate_beam_search} outlines the details of the beam search algorithm used by the Mate parser.

In order to learn the weight vector, the parser goes through the training set ($\tau = \{(x_t,y_t)\}_{t=1}^{T}$) for $N$ iterations. The weight vector is updated for every sentence $x_t$  when an incorrect parse is returned (i.e. the highest scoring parse $y^{*}_{t}$ is different from the gold parse $y_t$). More precisely, the passive-aggressive update of \newcite{crammer2006online} is used:

$$w^{(i+1)}\ =\ w^{(i)}\ +\ \frac{f(x_t,y_t)\ -\ f(x_t,y_{t}^{*})}{||f(x_t,y_t)\ -\ f(x_t,y_{t}^{*})||^2}$$

In this thesis, unless specified, we used the default settings of the parser:
\begin{enumerate}
	\item We use all the graph-based features of the completion model.
	\item We use the joint PoS-tagging with two-best tags for each token.
	\item We use a beam of 40.
	\item We use 25 iterations of training.
	\item We do \textit{not} change the sentence order of the training data during training.
\end{enumerate}

\section{Out-of-domain Parsing}\label{section:domain}
The release of the large manually annotated Penn Treebank (PTB) \cite{marcus93} and the development of the supervised learning techniques enable researchers to work on the supervised learning based parsing systems. Over the last two decades, the parsing accuracy has been significantly improved. A number of strong parsing systems for both constituency and dependency families have been developed \cite{klein03,petrov07,bohnet2013joint,martins2013turning,weiss2015neural,dozat2017deep}. The parsers based on supervised learning techniques capture statistics from labelled corpora to enable the systems to correctly predict parse trees when input the corresponding sentences. Since the PTB corpus contains mainly texts from news domain, the supervised learning based parsers trained on PTB corpus are sensitive to domain shifting. Those systems are able to achieve high accuracies when tested on the PTB test set (i.e. in-domain parsing). However, when applying them on data from different sources (i.e. out-of-domain parsing), such as web domain \cite{petrov2012overview} and chemical text\cite{nivre07conll}, the accuracy drops significantly. Table \ref{table:cross-domain-accuracy} shows a comparison of the in-domain and out-of-domain parsing performance of three parsers that have been frequently used by researchers (i.e. MST \cite{mcdonald2006online}, Malt \cite{nivre2009non}, and Mate parser \cite{bohnet2013joint}). Those parsers are trained on the training data from the major shared task on dependency parsing (i.e. CoNLL 2009 \cite{hajic09conll}). The training set contains mainly the news domain data from the Penn Treebank. In our evaluation, we first test them on the CoNLL test set which denotes our in-domain examples; for our out-of-domain examples we test the parsers on a number of different domains from the OntoNotes v5.0\footnote{https://catalog.ldc.upenn.edu/LDC2013T19} corpus.  As we can see from the results, the accuracies on out-of-domain texts are much lower than that of in-domain texts, with the largest accuracy difference of more than 15\% (i.e. Mate parser has an accuracy of 90.1\% on in-domain texts and an accuracy of 74.4\% on texts from broadcast conversations). How can we reduce the accuracy gap between the in-domain and the out-of-domain parsing? The most straightforward way would be annotating more text for the target domain, however, this approach is very expensive and time-consuming. There are only very limited manually annotated corpora available, which confirms the high costs of the annotation process. Domain adaptation is a task focused on solving the out-of-domain problems but without the need for manual annotation. There are a number of directions to work on the domain adaptation task, each of them focusing on a different aspect. These directions include semi-supervised techniques, domain specific training data selection, external lexicon resources and parser ensembles. Each direction has its own advantages and disadvantages, we briefly discuss in Section \ref{adaptaion-approaches}. In this thesis, we mainly focus on one direction that improves the out-of-domain accuracy by using unlabelled data (Semi-supervised approaches). Similar to other domain adaptation approaches, semi-supervised approaches do not require to manually annotate new data, but instead, they use the widely available unlabelled data. Some semi-supervised approaches focus on boosting the training data by unlabelled data that is automatically annotated by the base models, others aid the parsers by incorporating features extracted from the large unlabelled data. In Section \ref{semi-supervised} we discuss both approaches in detail.

\begin{table}[t]
\begin{center}
\begin{tabular}{|l|c|c|c|}
\hline \bf Domain & \bf MST & \bf Malt & \bf Mate \\ \hline
Newswire & 84.8 & 81.7 & 87.1 \\
Pivot Texts & 84.9 & 83.0 & 86.6 \\
Broadcast News & 79.4 & 78.1 & 81.2 \\
Magazines & 77.1 & 74.7 & 79.3 \\
Broadcast Conversation & 73.4 & 70.5 & 74.4 \\ \hline
CoNLL & 86.9 & 84.7 & 90.1 \\
\hline
\end{tabular}
\end{center}
\caption{\label{table:cross-domain-accuracy} Labelled attachment scores achieved by the MST, Malt, and Mate parsers trained on the \textsc{Conll} training set and tested on different domains. }
\end{table}

\subsection{Approaches to Out-of-Domain Parsing}\label{adaptaion-approaches}
As stated above, the domain adaptation techniques are designed to fill the accuracy gaps between the source domain and the target domain.
Previous work on domain adaptation tasks is mainly focused on four directions: semi-supervised techniques \cite{sarkar01,mcclosky2006reranking,reichart2007self,sagae07,koo08,sagae2010self,zhou2011exploiting,zhang12hit}, target domain training data selection \cite{plank2011effective,sogaard12sancl,khan13towards}, external lexicon resources \cite{szolovits2003adding,pyysalo2006,pekar2014exploring} and parser ensembles \cite{nivre07conll,le2012dcu,zhang12hit,petrov2012overview}. 

The semi-supervised techniques focus on exploring the largely available unlabelled data. There are two major ways to use the unlabelled data. The first family aims to boost the training data. Data that has been automatically annotated by the base models are used directly in re-training as the additional training set, up-training, self-training and co-training are techniques of this family.  The other family uses the features extracted from unlabelled data to aid the base model, this type of techniques include word embeddings, word clusters and dependency language models. In this thesis, we use semi-supervised techniques from both families and we will discuss them in detail in Section \ref{semi-supervised}.

Domain specific training data selection is a technique based on the assumption that similarity methods are able to derive a subset of the source domain training data that fits an individual test domain.  \newcite{plank2011effective} investigated several similarity methods to automatically select sentences from training data for the target domain, which gain significant improvements when comparing with random selection. Positive impacts are also found by \newcite{khan13towards} when they experimented with training data selection on parsing five sub-genres of web data. The advantage of this technique is that it does not need any extra data, however, it is also restricted to learn only from the source domain training set. 

Lack of the knowledge of the unknown words is one of the well-known problems faced by domain adaptation tasks, i.e. target domain test sets usually contain more unknown words (vocabularies which did not appear in the training data) than source domain test sets \cite{nivre07conll,petrov2012overview} . One way to solve this problem is to use the external lexicon resources created by the linguistics. External lexicons provide additional information for tokens, such as word lemma, part-of-speech tags, morphological information and so on. This information can be used by parsers directly to help making the decision. Previously, lexicons have been used by \newcite{szolovits2003adding} and \newcite{pyysalo2006} to improve the link grammar parser on the medical domain. Both approaches showed large improvements on parsing accuracy. Recently, \newcite{pekar2014exploring} extracted a lexicon from a crowd-sourced online dictionary (Wiktionary) and applied it to a strong dependency parser. Unfortunately, in their approach, the dictionary achieved a moderate improvement only.

The fourth direction of domain adaptation is parser ensembles, it becomes more noticeable, due to its good performance in shared tasks. For example, in the first workshop on syntactic analysis of non-canonical language (SANCL), the ensemble-based systems on average produced much better results than that of single parsers \cite{le2012dcu,zhang12hit,petrov2012overview}. However, those ensemble-based systems are not used in real-world tasks, due to the complex architectures and high running time.

\subsection{Semi-Supervised Approaches}\label{semi-supervised}
Semi-supervised approaches use unlabelled data to bridge the accuracy gap between in-domain and out-of-domain. In recent years, unlabelled data has gained large popularity in syntactic parsing tasks, as it can easily and inexpensively be obtained, cf. \cite{sarkar01,steedman2003semi,mcclosky06naacl,koo08,sogaard2010,petrov2012overview,chen2013feature,weiss2015neural}. This is in stark contrast to the high costs of manually labelling new data. Some techniques such as self-training \cite{mcclosky06naacl} and co-training \cite{sarkar01} use auto-parsed data as additional training data. This enables the parser to learn from its own or other parsers' annotations. Other techniques include word clustering \cite{koo08} and word embedding \cite{bengio03a} which are generated from a large amount of unlabelled data. The outputs can be used as features or inputs for parsers. Both groups of techniques have been shown effective on syntactic parsing tasks \cite{zhou2005tri,reichart2007self,sagae2010self,sogaard2010,yu2015iwpt,weiss2015neural}.

\subsubsection{Boosting the Training Set}

The first group uses unlabelled data (usually parsed data) directly in the training process as additional training data. The most common approaches in this group are co-training and self-training.

\textbf{Co-training} is a technique, that has been frequently used by domain adaptation for parsers \cite{sarkar01,sagae07,zhang12hit,petrov2012overview}. The early version of co-training uses two different 'views' of the classifier, each 'view' has a distinct feature set. Two 'views' are used to annotate unlabelled set after trained on the same training set. Then both classifiers are retrained on the newly annotated data and the initial training set \cite{blum98}. \newcite{blum98} first applied a multi-iteration co-training on classifying web pages. Then it was extended by \newcite{collins99} to investigate named entity classification. At that stage, co-training strongly depended on the splitting of features \cite{zhu2005semi}. One year after, \newcite{goldman00} introduced a new variant of co-training which used two different learners, but both of them took the whole feature sets. One learner's high confidence data are used to teach the other learner. After that, \newcite{zhou2005tri} proposed another variant of co-training (tri-training). Tri-training used three learners, each learner is designed to learn from data on which the other two learners have agreed.

In terms of the use of co-training in the syntactic analysis area, \newcite{sarkar01} first applied the co-training to a phrase structure parser. He used a subset (9695 sentences) of labelled Wall Street Journal data as initial training set and a larger pool of unlabelled data (about 30K sentences). In each iteration of co-training, the most probable $n$ sentences from two views are added to the training set of the next iteration. In their experiments, the parser achieved significant improvements in both precision and recall (7.79\% and 10.52\% respectively) after 12 iterations of co-training. 

The work most close to ours was presented by \cite{sagae07} in the shared task of the conference on computational natural language learning (CoNLL). They used two different settings of a shift-reduce parser to complete a one iteration co-training, and their approach successfully achieved improvements of approximately 2-3\%. Their outputs have also scored the best in the out-of-domain track \cite{nivre07conll}. The two settings they used in their experiments are distinguished from each other in three ways. Firstly, they parse the sentences in reverse directions (forward vs backward). Secondly, the search strategies are also \textit{not} the same (best-first vs deterministic). Finally, they use different learners (maximum entropy classifier vs support vector machine). The maximum entropy classifier learns a conditional model $p(y|x)$ by maximising the conditional entropy ($H(p) = - \sum_{x,y} \tilde{p}(x)p(y|x) \log p(y|x) $)\footnote{$\tilde{p}(x)$ is the empirical distribution of $x$ in the training data.}, while the support vector machines (SVMs) are linear classifiers trained to maximise the margin between different classes. In order to enable the multi-class classification, they used the all-versus-all strategy to train multiple SVMs for predicting the next transition. In addition, a polynomial kernel with degree 2 is used to make the data linearly separable. \newcite{sagae07} proved their assumptions in their experiments. Firstly, the two settings they used are different enough to produce distinct results. Secondly, the perfect agreement between two learners is an indication of correctness. They reported that the labelled attachment score could be above 90\% when the two views agreed. By contrast, the labelled attachment scores of the individual view were only between 78\% and 79\%. 

Tri-training is a variant of co-training. A tri-training approach uses three learners, in which the source learner is retrained on the data produced by the other two learners. This allows the source learner to explore additional annotations that are not predicted by its own, thus it has a potential to be more effective than the co-training. Tri-training is used by \cite{zhang12hit} in the first workshop on syntactic analysis of non-canonical language (SANCL)\cite{petrov2012overview}. They add the sentences which the two parsers agreed on into the third parser's training set, then retrain the third parser on the new training set. However, in their experiments, tri-training did \textit{not} significantly affect their results. 

Recently, \newcite{weiss2015neural} used normal agreement based co-training and tri-training in their evaluation of a state-of-the-art neural network parser. Their evaluation is similar to the Chapter \ref{chapter:cotrain} of this thesis, although they used different parsers. Please note their paper is published after our evaluation on co-training \cite{pekar2014exploring}.  In their work, the annotations agreed by a conventional transition-based parser (zPar) \cite{zhang11} and the Berkeley constituency parser \cite{petrov07} have been used as additional training data. They retrained their neural network parser and the zPar parser on the extended training data. The neural network parser gained around 0.3\% from the tri-training, and it outperforms the state-of-the-art accuracy by a large 1\%. By contrast, their co-training evaluation on the zPar parser found only negative effects.

\textbf{Self-training} is another semi-supervised technique that only involves one learner. In a typical self-training iteration, a learner is firstly trained on the labelled data, and then the trained learner is used to label some unlabelled data. After that, the unlabelled data with the predictions (usually the high confident predictions of the model) are added to the training data to re-train the learner. The self-training iteration can also be repeated to do a multi-iteration self-training.   When compared with co-training, self-training has a number of advantages. Firstly unlike the co-training that requires two to three learners, the self-training only requires one learner, thus it is more likely we can use the self-training than co-training in an under resourced scenario. Secondly, to generate the additional training data, co-training requires the unlabelled data to be double annotated by different learners, this is more time-consuming than self-training's single annotation requirement. In term of the previous work on parsing via self-training, \newcite{charniak1997statistical} first applied self-training to a PCFG parser, but this first attempt of using self-training for parsing failed. \newcite{steedman2003semi} implemented self-training and evaluated it using several settings. 
They used a 500 sentences training data and parsed only 30 sentences in each self-training iteration. After multiple self-training iterations, it only achieved moderate improvements. This is caused probably by the small number of additional sentences used for self-training.
  
\newcite{mcclosky06naacl} reported strong self-training results with an improvement of 1.1\% f-score by using the Charniak-parser, cf. \cite{charniak2005}. 
The Charniak-parser is a two stage parser that contains a lexicalized context-free parser and a discriminative reranker. They evaluated on two different settings. In the first setting, they add the data annotated by both stages and retrain the first stage parser on the new training set, this results in a large improvement of 1.1\%. In the second setting, they retrain the first stage parser on its own annotations, the result shows no improvement. Their first setting is similar to the co-training as the first stage parser is retrained on the annotation co-selected by the second stage reranker, in which the additional training data is more accurate than the predictions of first stage parser.  \newcite{mcclosky2006reranking} applied the same method later on out-of-domain texts which show
good accuracy gains too. 

\newcite{reichart2007self} showed that self-training can improve the performance of a constituency parser without a reranker for the in-domain parsing. However, their approach used only a rather small training set when compared to that of \newcite{mcclosky06naacl}.

\newcite{sagae2010self} investigated the contribution of the reranker for a constituency parser in a domain adaptation setting.
Their results suggest that constituency parsers without a reranker can achieve statistically significant improvements in the out-of-domain parsing, 
but the improvement is still larger when the reranker is used.

In the workshop on syntactic analysis of non-canonical language (SANCL) 2012 shared task, self-training was used by most of the constituency-based systems, cf. \cite{petrov2012overview}.
The top ranked system is also enhanced by self-training, this indicates that self-training is probably an established technique to improve
the accuracy of constituency parsing on out-of-domain data, cf. \cite{le2012dcu}. 
However, none of the dependency-based systems used self-training in the SANCL 2012 shared task.

One of the few successful approaches to self-training for dependency parsing was introduced by \newcite{chen2008learning}. They improved the unlabelled attachment score by about one percentage point for Chinese.\newcite{chen2008learning} added parsed sentences that have a high ratio of dependency edges that span only a short distance, i.e. the head and dependent are close together. The rationale for this procedure is the observation that short dependency edges show a higher accuracy than longer edges.

\newcite{kawahara2008learning} used a separately trained binary classifier to select reliable sentences as additional training data. 
Their approach improved the unlabelled accuracy of texts from a chemical domain by about 0.5\%. 

\newcite{goutam2011exploring} applied a multi-iteration self-training approach on Hindi to improve parsing accuracy within the training domain. In each iteration, they add a small number (1,000) of additional sentences to a small initial training set of 2,972 sentences, the additional sentences were selected due to their parse scores. They improved upon the baseline by up to 0.7\% and 0.4\% for labelled and unlabelled attachment scores after 23 self-training iterations. 

While many other evaluations on self-training for dependency parsing are found unhelpful or even have negative effects on results. \newcite{bplank2011phd} applied self-training with single and multiple iterations for parsing of Dutch using the Alpino parser \cite{malouf04wide}, which was modified to produce dependency trees. She found self-training produces only a slight improvement in some cases but worsened when more unlabelled data is added. 

\newcite{plank2013experiments} used self-training in conjunction with dependency triplets statistics and the similarity-based sentence selection for Italian out-of-domain parsing. They found the effects of self-training are unstable and does not lead to an improvement.

\newcite{cerisara2014spmrl} and \newcite{bjorkelund2014spmrl} applied self-training to dependency parsing on nine languages. \newcite{cerisara2014spmrl} could only report negative results in their self-training evaluations for dependency parsing. Similarly, \newcite{bjorkelund2014spmrl} could observe only on Swedish a positive effect.

\subsubsection{Integrating with Features Learned from Unlabelled Data}
The second group uses the unlabelled data indirectly. Instead of using the unlabelled data as training data, they incorporate the information extracted from large unlabelled data as features to the parser. Word clusters \cite{koo08,cerisara2014spmrl} and word embeddings \cite{chen2014neural,weiss2015neural} are most well-known approaches of this family. However, other attempts have also been evaluated, such as dependency language models (DLM) \cite{chen2012utilizing}.

\textbf{Word Clustering} is an unsupervised algorithm that is able to group the similar words into the same classes by analysing the co-occurrence of the words in a large unlabelled corpus. The popular clustering algorithm includes Brown \cite{brown1992class,liang05master} and the Latent dirichlet allocation (LDA) \cite{chrupala2011lda} clusters.

\newcite{koo08} first employed a set of features based on brown clusters to a second-order graph-based dependency parser. They evaluated on two languages (English and Czech) and yield about one percentage improvements for both languages. The similar features have been adapted to a transition-based parser of \newcite{bohnet2012emnlp}. The LDA clusters have been used by \newcite{cerisara2014spmrl} in the workshop on statistical parsing of morphologically rich languages (SPMRL) 2014 shared tasks \cite{seddah2014introducing} on parsing nine different languages, their system achieved the best average results across all non-ensemble parsers.

\textbf{Word embeddings} is another approach that relies on the co-occurrence of the words. Instead of assigning the words into clusters, word embedding represent words as a  low dimensional vector (such as 50 or 300 dimensional vector), popular word embedding algorithms include word2vec \cite{mikolov2013distributed} and global vectors for word representation (GloVe) \cite{pennington2014glove}. Due to the nature of the neural networks, word embeddings can be effectively used in the parsers based on neural networks. By using pre-trained word embeddings the neural network-based parsers can usually achieve a higher accuracy compared with those who used randomly initialised embeddings \cite{chen2014neural,weiss2015neural,dozat2017deep}.

\textbf{Other Approaches} that use different ways to extract features from unlabelled data have also been reported. 

\newcite{mirroshandel12} used lexical affinities to rescore the n-best parses. 
They extract the lexical affinities from parsed French corpora by calculating the relative frequencies of head-dependent pairs for nine manually selected patterns. Their approach gained a labelled improvement of 0.8\% over the baseline.

\newcite{chen2012utilizing} applied high-order DLMs to a second-order graph-based parser. This approach is most close to the Chapter \ref{chapter:dlm} of this thesis. The DLMs allow the new parser to explore higher-order features without increasing the time complexity. The DLMs are extracted from a 43 million words English corpus \cite{charniak00} and a 311 million words corpus of Chinese \cite{huang09} parsed by the baseline parser. Features based on the DLMs are used in the parser. 
They gained 0.66\% UAS for English and an impressive 2.93\% for Chinese.

\newcite{chen2013feature} combined the basic first- and second-order features with meta features based on frequencies. 
The meta features are extracted from auto-parsed annotations by counting the frequencies of basic feature representations in a large corpus. 
With the help of meta features, the parser achieved the state-of-the-art accuracy on Chinese.

\section{Corpora}
As mentioned previously, one contribution of this thesis is evaluating major semi-supervised techniques in a unified framework. For our main evaluation, we used English data from the conference on computational natural language learning (\textsc{Conll}) 2009 shared task \cite{hajic09conll} as our source of in-domain evaluation. For out-of-domain evaluation, we used weblogs portion of OntoNotes v5.0\footnote{https://catalog.ldc.upenn.edu/LDC2013T19} corpus (\textsc{Weblogs}) and the first workshop on syntactic analysis of non-canonical language shared task data (\textsc{Newsgroups,Reviews,Answers}) \cite{petrov2012overview}. Section \ref{section:main_corpora} introduces our main evaluation corpora in detail. For comparison and multi-lingual evaluation, we also evaluated some of our approaches in various additional corpora. Our self-training approach has been evaluated on chemical domain data (\textsc{Chemical}) from the conference on computational natural language learning 2007 shared task \cite{nivre07conll} and  nine languages datasets from the workshop on statistical parsing of morphologically rich languages (\textsc{Spmrl}) 2014 shared task\cite{seddah2014introducing}. Our dependency language models approach has been evaluated in addition on Wall Street Journal portion of Penn English Treebank 3 (\textsc{Wsj}) \cite{marcus93} and Chinese Treebank 5 (\textsc{Ctb}) \cite{xue05}. As both treebanks do not contain unlabelled data, we used the data of \newcite{chelba13onebillion} and the Xinhua portion of Chinese Gigaword
Version 5.0 \footnote{https://catalog.ldc.upenn.edu/LDC2011T13} for our English and Chinese tests respectively. We introduce those corpora in the experiment set-up section of the relevant chapters. 

\subsection{The Main Evaluation Corpora}\label{section:main_corpora}

\begin{table}[t]
	\begin{center}
		\begin{tabular}{|l|r|r|}
			\cline{2-3}
			\multicolumn{1}{c|}{}&  \multicolumn{1}{|c|}{\bf train}   &  \multicolumn{1}{|c|}{\bf test}   \\
			\cline{2-3}
			\multicolumn{1}{c|}{}& \multicolumn{1}{|c|}{\sc \ \ \ Conll\ \ \ } & \multicolumn{1}{|c|}{\sc \ \ \ Conll\ \ \ }\\ \hline
			\bf Sentences  &39,279  &2,399 \\
			\bf Tokens     &958,167  &57,676\\
			\bf Avg. Length&24.39&24.04\\
			\hline
		\end{tabular}
	\end{center}
	\caption{\label{table:main_conll09_corpus_stats} The size of the source domain (\textsc{Conll}) training and test sets for our main evaluation corpora.}
\end{table}

\begin{table}[t]
	\begin{center}
		\begin{tabular}{|l|r|rrrr|}
			\cline{2-6}
			\multicolumn{1}{c|}{}& \multicolumn{1}{|c|}{\bf dev}& \multicolumn{4}{|c|}{\bf test}\\ \cline{2-6}
			\multicolumn{1}{c|}{} & \sc Weblogs  & \sc Weblogs & \sc Newsgroups  &\sc Reviews & \sc Answers  \\ \hline
			\bf Source &OntoNotes&OntoNotes&SANCL&SANCL&SANCL\\
			\bf Sentences  &2,150   &2,141  & 1,195  & 1,906 & 1,744 \\
			\bf Tokens     &42,144  &40,733   & 20,651  & 28,086&28,823\\
			\bf Avg. Length&19.6&19.03&17.28&14.74&16.53\\
			\hline
		\end{tabular}
	\end{center}
	\caption{\label{table:main_corpora_test_stats} The size of the target domain test datasets for our main evaluation corpora.}
\end{table}

\begin{table}[t]
	\begin{center}
		\begin{tabular}{|l|r|r|r|r|}
			\cline{2-5}
			\multicolumn{1}{c|}{}&\multicolumn{4}{|c|}{\bf unlabelled}\\\cline{2-5}
			\multicolumn{1}{c|}{}& \sc Weblogs & \sc Newsgroups& \sc Reviews& \sc Answers\\ \hline
			\bf Sentences  &  513,687& 512,000   &512,000 &27,274\\
			\bf Tokens     &   9,882,352  &9,373,212  &7,622,891&424,299  \\
			\bf Avg. Length& 19.24 &18.31 &14.89 &15.55\\
			\hline
		\end{tabular}
		
	\end{center}
	\caption{\label{table:main_corpora_unlabelled_stats} The size of unlabelled datasets for our main evaluation corpora.}
\end{table}

In this section, we introduce our main evaluation corpora that have been used in all of the semi-supervised approaches evaluated in this thesis. 

The \textsc{Conll} English corpus built on the Penn English Treebank 3 \cite{marcus93} which contains mainly Wall Street Journals but also included a small portion of Brown corpus\cite{browncorpus1979}. The training set contains only Wall Street Journals, the small subset of the Brown corpus has been included in the test set. The constituency trees from Penn English Treebank are converted to dependency representation by the LTH constituent-to-dependency conversion tool, cf. \cite{johansson2007extended}. A basic statistic of the corpus can be found in Table \ref{table:main_conll09_corpus_stats}.

For our \textsc{Weblogs} domain test we used the Ontonotes v5.0\footnote{https://catalog.ldc.upenn.edu/LDC2013T19} corpus. The Ontonotes corpus contains various domains of text such as weblogs, broadcasts, talk shows and pivot texts. We used the last 20\% of the weblogs portion of the Ontonotes v5.0 corpus as our target domain development set and the main test set. The selected subset allows us to build sufficient sized datasets similar to the source domain test set. More precisely, the first half of the selected corpus is used as a test set while the second half is used as the development set. Table \ref{table:main_corpora_test_stats} shows some basic statistic of those datasets. 

\textsc{Newsgroups, Reviews} and \textsc{Answers} domain data are used as additional test sets for our evaluation. Those additional test domains are provided by the first workshop on syntactic analysis of non-canonical language (SANCL) shared task \cite{petrov2012overview}. The shared task is focused on the parsing English web text, in total, they prepared five web domain datasets, two of them are development datasets (Email, Weblogs) and the other three (Newsgroups, Reviews and Answers) are used as test sets. For each of the domains, a small labelled set and a large unlabelled set are provided. In this thesis, we used all three test datasets (both labelled and unlabelled data). In addition, one of the unlabelled texts (Weblogs) from the development portion of the shared task is also used. We used for each domain a similar sized unlabelled dataset to make the evaluation more unified. The only exception is the answers domain, as its unlabelled dataset is much smaller than the other three domains, thus we used all of the data provided. A basic statistic of the labelled test sets and unlabelled data can be found in Table \ref{table:main_corpora_test_stats} and \ref{table:main_corpora_unlabelled_stats} respectively.

In term of the dependency representation, we used the LTH conversion for our main evaluation corpora. Same as the CoNLL 2009 shared task we converted all the labelled data from constituent trees to dependency representation by the LTH constituent-to-dependency conversion tool \cite{johansson2007extended} when needed.

\section{Evaluation Methods}
To measure the parser's performance, we report labelled attachment scores (LAS) and unlabelled attachment scores (UAS).  For our evaluation on the main corpora, we use the official evaluation script of the CoNLL 2009 shared task, in which all punctuation marks are included in the evaluation. The LAS and UAS are the standard ways to evaluate the accuracy of a dependency parser. Due to the single-head property of the dependency trees, the dependency parsing can be seen as a tagging task, thus the single accuracy metric is well suited for the evaluation.  Both LAS and UAS measure the accuracy by calculating the percentage of the dependency edges that have been correctly attached. The UAS considers an edge is correct if the attachment is correct, it does not take the label into account, while the LAS counts only the edges that are both correctly attached and the correct label also assigned. The LAS is more strict than UAS thus we mainly focus on LAS in our evaluation. Let $C_a$ be the number of edges that are correctly attached, $C_{a+l}$ be the number of edges that are both correctly attached and have the correct label, $C_t$ be the total number of edges, we compute:

\begin{equation}
Unlabelled\ attachment\ score\ (UAS)\ =\ C_a\ /\ C_t
\end{equation}
\begin{equation}
Labelled\ attachment\ score\ (LAS)\ =\ C_{a+l}\ /\ C_t
\end{equation}

For significance testing, we use the randomised parsing evaluation comparator from a major shared task on dependency parsing \cite{nivre07conll} . The script takes predictions annotated by two different models of the same dataset. Let the first input be the one which has a higher overall accuracy. The null hypothesis of the script is that the accuracy difference between the first input and the second input is not statistically significant. And the p-values represent the probability that the null hypothesis is correct. We use the script's default setting of 10,000 iterations ($i_{total}$), for each iteration, the comparator randomly selects one sentence from the dataset and compares the accuracies of the sentence predicted in the two different inputs. Let $i_{less}$ be the number of randomly selected instances that are predicted less accurately in the first input when compared to the predictions in the second input. The p-value is calculated by:
 $$p = \frac{i_{less}}{i_{total}}$$
We mark the significance levels based on their p-values, * for $p < 0.05$, ** for $p < 0.01$. 

\section{Analysis Techniques}
To understand the behaviour of our methods, we assess our results on a number of tests. We analyse the results on both token level and sentences level. For token level, we focus on the accuracies of individual syntactic labels and the known/unknown words accuracies. For sentence level, we used the methods from \newcite{mcclosky06naacl} to evaluate sentences in four factors. We used all four factors from their analysis, i.e. sentence length, the number of unknown words, the number of prepositions and number of conjunctions. 

\textbf{Token Level Analysis.}
Our token level analysis consists of two tests, the first test assesses the accuracy changes for individual labels. The goal of this test is to find out the effects of our semi-supervised methods on different labels. For an individual label, we calculate the recall, precision and the f-score. Let $P_L$ be the number of the label $L$ predicted by the parser, $G_L$ be the count of label $L$ presented in the gold data and $PG_L$ be the number of the label predicted correctly. The precision ($Pre_L$), recall ($Rec_L$) and the f-score ($F_L$) are calculated as follows: 

\begin{equation}
	Pre_L\ =\ PG_L\ /\ P_L
\end{equation}
\begin{equation}
	Rec_L\ =\ PG_L\ /\ G_L
\end{equation}
\begin{equation}
	F_L\ =\ 2\ *\ \frac{Pre_L\ *\ Rec_L}{Pre_L\ +\ Rec_L}
\end{equation}

%label evaluation
\begin{figure}[t]
	\begin{center}
		\begin{tikzpicture}
		\begin{axis}[
		ymin=-0.5, ymax=1.8,
		width=12cm,height=7cm,
		ybar,ybar interval =0.7,
		ylabel=Accuracy Change (\%),
		enlargelimits=0.01,
		legend style={at={(0.5,-0.15)},anchor=north,legend columns=-1},
		symbolic x coords={NMOD,P,PMOD,SBJ,OBJ,ROOT,ADV,NAME,VC,COORD,TMP,DEP,CONJ,LOC,AMOD},
		xtick=data,
		x tick label style={font=\tiny,rotate=45,anchor=east}
		]
		\addplot coordinates{(NMOD,0.3) (P,0.1) (PMOD,0.2) (SBJ,0.1) (OBJ,0.5) (ROOT,0.3) (ADV,1.0) (NAME,0.5) (VC,-0.1) (COORD,0.5) (TMP,0.9) (DEP,-0.3) (CONJ,1.1) (LOC,0.7) (AMOD,-0.3) }; 
		\addlegendentry{\tiny Recall}
		
		\addplot coordinates{(NMOD,0.3) (P,0.0) (PMOD,0.0) (SBJ,0.4) (OBJ,0.3) (ROOT,0.3) (ADV,0.4) (NAME,-0.4) (VC,0.4) (COORD,0.6) (TMP,-0.2) (DEP,-0.1) (CONJ,1.2) (LOC,1.2) (AMOD,1.4) }; 
		\addlegendentry{\tiny Precision}
		
		\addplot coordinates{(NMOD,0.3) (P,0.1) (PMOD,0.1) (SBJ,0.2) (OBJ,0.4) (ROOT,0.3) (ADV,0.7) (NAME,0.0) (VC,0.2) (COORD,0.5) (TMP,0.3) (DEP,-0.2) (CONJ,1.2) (LOC,1.0) (AMOD,0.5) }; 
		\addlegendentry{\tiny F-score}
		
		\end{axis}
		%NMOD cnt:15286 R/P/F95.7/95.6/95.7 Base R/P/F95.4/95.3/95.4 diff: F:0.3 R:0.3 P:0.3
		%P cnt:6831 R/P/F99.9/99.7/99.8 Base R/P/F99.8/99.7/99.7 diff: F:0.1 R:0.1 P:0.0
		%PMOD cnt:5469 R/P/F95.5/95.5/95.5 Base R/P/F95.3/95.5/95.4 diff: F:0.1 R:0.2 P:0.0
		%SBJ cnt:4332 R/P/F96.8/97.5/97.1 Base R/P/F96.7/97.1/96.9 diff: F:0.2 R:0.1 P:0.4
		%OBJ cnt:3073 R/P/F95.1/93.7/94.4 Base R/P/F94.6/93.4/94.0 diff: F:0.4 R:0.5 P:0.3
		%ROOT cnt:2399 R/P/F96.3/96.3/96.3 Base R/P/F96.0/96.0/96.0 diff: F:0.3 R:0.3 P:0.3
		%ADV cnt:2091 R/P/F84.1/75.5/79.8 Base R/P/F83.1/75.1/79.1 diff: F:0.7 R:1.0 P:0.4
		%NAME cnt:2002 R/P/F94.7/94.0/94.3 Base R/P/F94.2/94.4/94.3 diff: F:0.0 R:0.5 P:-0.4
		%VC cnt:1804 R/P/F97.4/97.4/97.4 Base R/P/F97.5/97.0/97.2 diff: F:0.2 R:-0.1 P:0.4
		%COORD cnt:1374 R/P/F92.8/94.2/93.5 Base R/P/F92.3/93.6/93.0 diff: F:0.5 R:0.5 P:0.6
		%TMP cnt:1341 R/P/F84.6/86.8/85.7 Base R/P/F83.7/87.0/85.4 diff: F:0.3 R:0.9 P:-0.2
		%DEP cnt:1302 R/P/F86.0/85.4/85.7 Base R/P/F86.3/85.5/85.9 diff: F:-0.2 R:-0.3 P:-0.1
		%CONJ cnt:1103 R/P/F91.6/91.2/91.4 Base R/P/F90.5/90.0/90.2 diff: F:1.2 R:1.1 P:1.2
		%LOC cnt:955 R/P/F84.1/87.5/85.8 Base R/P/F83.4/86.3/84.8 diff: F:1.0 R:0.7 P:1.2
		%AMOD cnt:913 R/P/F77.0/84.5/80.7 Base R/P/F77.3/83.1/80.2 diff: F:0.5 R:-0.3 P:1.4
		
		\end{tikzpicture}
	\end{center}
	\caption{\label{figure:eg-analysis-label} The bar chart used to visualise our analysis on individual labels.}
\end{figure}
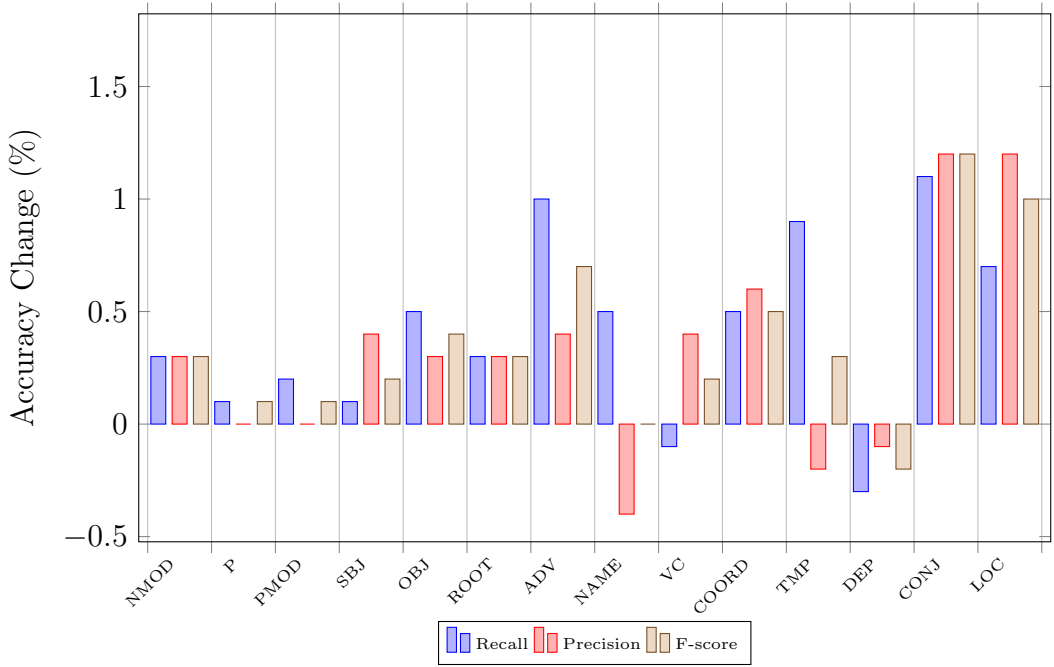

We compute for each label, the score differences between our enhanced model and the base model. The results for the most frequent labels are visualised by the bar chart. Figure \ref{figure:eg-analysis-label} is an example of the bar chart we used, the x-axis shows the names of the relevant label, the y-axis shows the accuracy changes in percentage. For each of the labels, we report the accuracy changes of all three scores (recall, precision and f-score), the left (blue) bar represents the recall, the middle (red) bar represents the precision and the right (brown) bar is for f-score.

The second test assesses the overall accuracy of known words and unknown words. The unknown words are defined as the words that are not presented in the initial training set. The initial training set is the one we used to train the base model. To compute the accuracy for known and unknown words, we first assign all the tokens in the dataset into two groups (known and unknown) and then we calculate the labelled and unlabelled accuracies for each of the groups separately. We compare the improvements achieved by our enhanced model on known and unknown words to understand the ability of our model on handling unknown words.

%number of unknown words
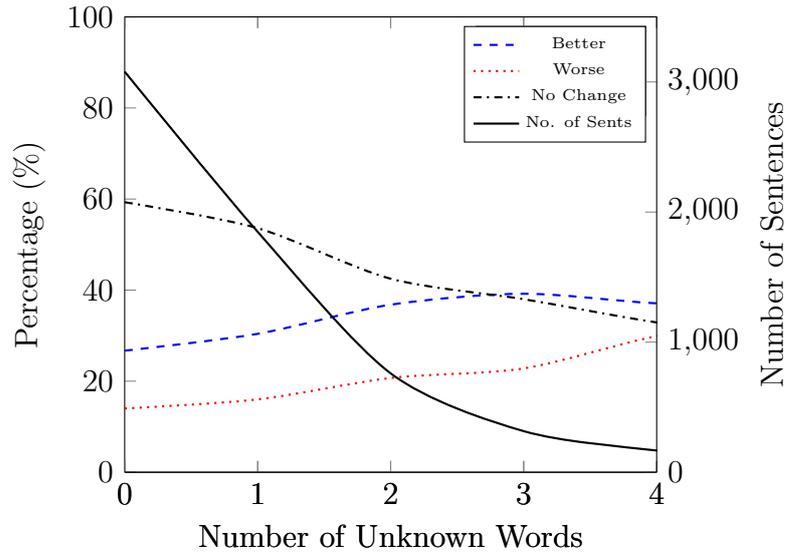
\begin{figure}[t]
	\begin{center}
		\begin{tikzpicture}
		\pgfplotsset{
			xmin=0,xmax=4,
			xtick={0,1,...,4},
			xlabel=Number of Unknown Words}
		\begin{axis}[
		axis y line*=left,
		ymin=0,ymax=100,
		ylabel=Percentage (\%)]
		%Better
		\addplot[smooth,thick,dashed,color=blue] coordinates {(0,26.7) (1,30.4) (2,36.8) (3,39.2) (4,37.1) (5,36.6) (6,33.3) };\label{better}
		\addlegendentry{\tiny Better}
		
		%Worse
		\addplot[smooth,thick,dotted,color=red] coordinates{(0,14.0) (1,16.0) (2,20.7) (3,22.8) (4,29.9) (5,34.1) (6,33.3) };\label{worse}
		\addlegendentry{\tiny Worse}
		
		%No Change
		\addplot[smooth,thick,dashdotted] coordinates{(0,59.3) (1,53.6) (2,42.5) (3,38.0) (4,32.9) (5,29.3) (6,33.3) };\label{nochange}
		\addlegendentry{\tiny No Change}
		
		\end{axis}
		\begin{axis}[
		axis y line*=right,
		ymin=0,ymax=3500,
		ylabel=Number of Sentences]
		\addlegendimage{/pgfplots/refstyle=better}\addlegendentry{\tiny Better}
		\addlegendimage{/pgfplots/refstyle=worse}\addlegendentry{\tiny Worse}
		\addlegendimage{/pgfplots/refstyle=nochange}\addlegendentry{\tiny No Change}
		%All Sent
		\addplot[smooth,thick,solid] coordinates{(0,3079.0) (1,1848.0) (2,760.0) (3,316.0) (4,167.0) (5,82.0) (6,42.0) };\addlegendentry{\tiny No. of Sents}
		
		\end{axis}
		%0 Total:3079 LAS better/worse/nochange:821/432/1826 26.7/14.0/59.3 12.7
		%1 Total:1848 LAS better/worse/nochange:562/296/990 30.4/16.0/53.6 14.4
		%2 Total:760 LAS better/worse/nochange:280/157/323 36.8/20.7/42.5 16.1
		%3 Total:316 LAS better/worse/nochange:124/72/120 39.2/22.8/38.0 16.4
		%4 Total:167 LAS better/worse/nochange:62/50/55 37.1/29.9/32.9 7.2
		%5 Total:82 LAS better/worse/nochange:30/28/24 36.6/34.1/29.3 2.5
		%6 Total:42 LAS better/worse/nochange:14/14/14 33.3/33.3/33.3 0.0
		
		\end{tikzpicture}
	\end{center}
	\caption{\label{figure:eg-analysis-sentunk} An example of our sentence level analysis on different number of unknown words per sentence. }
\end{figure}

\textbf{Sentence Level Analysis.} For our sentence level analysis, we evaluate on four factors (sentence length, the number of unknown words, the number of prepositions and the number of conjunctions) that are known to be problematic in parsing. We use a method similar to \newcite{mcclosky06naacl} in our analysis. For each of the factors, we assign sentences to different classes according to their property, sentences that have the same property are assigned to the same class. Take unknown words factor as an example, sentences which contain the same number of unknown words are grouped together. For each group, we calculate the percentage of sentences that are improved, worsened or  unchanged in accuracy by our enhanced model. The reason for using the percentage instead of the number of sentences that were used by \newcite{mcclosky06naacl} is mainly because the absolute numbers vary greatly both within the factor and between factors, thus is not suitable for comparison. The percentage, on the other hand, can be easily compared. In addition to the above values,  we also report the number of the sentences in each class. Figure \ref{figure:eg-analysis-sentunk} shows an example of our sentence level analysis on the different number of unknown words per sentence. The x-axis shows the conditions of the classes. In this example, it represents the different number of unknown words in a single sentence. The y-axis to the left is the percentage and the y-axis to the right is the number of sentences. The blue dashed line represents the percentage of the sentences that are parsed better by our enhanced model, the red dotted line represent the portion that is parsed less accurate, the black dash-dotted line shows the portion of sentences whose accuracy are unchanged. The black solid line is the number of sentences in the individual classes. 

\section{Chapter Summary}
This chapter introduced the background and the experiment set-up. The first part focused on dependency parsers, it introduced three major types of dependency parsers and gave a detailed introduction of the base parser used in this thesis. The second part discussed the problem caused by parsing out-of-domain text and the techniques that have been used by previous work to solve the problem. The third part introduced the corpora we used. The last two parts showed our evaluation methods and analysis techniques.

\chapter{Co-training}\label{chapter:cotrain}
In this chapter, we introduce our co-training approach. Co-training is one of the popular semi-supervised techniques that has been applied to many natural language processing tasks, such as named entity recognition \cite{collins99}, constituency parsing \cite{sarkar01} and dependency parsing \cite{sagae07,petrov2012overview}. Although co-training approaches are popular, they do not always bring positive effects \cite{zhang12hit,weiss2015neural}. Improvements on results are usually reported by learners that are carefully designed to be as different as possible. Such as in \newcite{sagae07}'s approach, they form the co-training with parsers consisting of different learning algorithms and search strategies. However, off-the-shelf parsers use many similar features, the output of these parsers are more likely to agree with each other. Thus it is unclear whether the off-the-shelf parsers are suitable for co-training.  

In this work we evaluate co-training with a number of off-the-shelf parsers that are freely available to the research community, namely Malt parser \cite{nivre2009non}, MST parser \cite{mcdonald2006online}, Mate parser \cite{bohnet2013joint}, and Turbo parser \cite{martins2010turbo}. We evaluate those parsers on agreement based co-training algorithms. The evaluation learner is retrained on the training set that is boosted by automatically annotated sentences agreed by two source learners. We investigate both normal agreement based co-training and a variant called tri-training. In a normal co-training setting the evaluation learner is used as one of the source learners, and in a tri-training scenario, the source learners are different from the evaluation learner.

In the following sections we introduce our approaches in Section \ref{section:co-our-approach}. We then introduce our experiment settings and results in Section \ref{section:co-experiment-setup} and Section \ref{section:co-results} respectively. After that, in Section \ref{section:co-analysis} we analyse the results and trying to understand how co-training helps. In the last section (Section \ref{section:co-conclusion}), we summarise our finding.

\section{Agreement Based Co-training}\label{section:co-our-approach}
In this work, we apply an agreement based co-training to out-of-domain dependency parsing.  Our agreement based co-training is inspired by the observation from \newcite{sagae07} in which the two parsers agreeing on an annotation is an indication of a higher accuracy. We proposed two types of agreement based approaches: one uses parser pairs (normal co-training), the other uses three parsers which is also known as tri-training.

Two approaches use a similar algorithm, which involves two source learners and one evaluation learner. Two source learners are used to produce additional training data for retraining the evaluation learner. More precisely, our algorithm is as follows:

\begin{enumerate}
	\item Two source learners are trained separately on the source domain training set to generate two base models.
	\item Both models are used to parse a large number of target domain unlabelled data.
	\item \label{label:co-newtrainingset}After that, we compare two automatically labelled predictions for each of the sentences, the first N (such as 10k, 20k) predictions that both models agreed are added to the end of the source domain training set.
	\item Finally, we retrain the evaluation learner on the boosted training set generated in step \ref{label:co-newtrainingset}.
\end{enumerate}

Although both approaches share the similar algorithm, the major differences between them are: both parsers involved by normal co-training are used as the source learners, in which one of them is also used as the evaluation learner; by contrast, tri-training uses three parsers in total, in which two of them are used as the source learners and the third one is used as the evaluation learner. 

In terms of parsers selection, we selected four public available dependency parsers, which include two benchmark parsers (Malt parser \cite{nivre2009non} and MST parser \cite{mcdonald2006online}), one transition-based Mate parser \cite{bohnet2013joint}, and one graph-based Turbo parser \cite{martins2010turbo}. These parsers have been widely used by researchers. A  more detailed discussion of the dependency parser can be found in section \ref{section:depparsing}.

\begin{table}[t]
\begin{center}
\begin{tabular}{|l|r|r|r|r|}
\hline
%&\bf LAS (Single)&\bf LAS (Identical)&\bf Identical rate&\bf Avg. Length\\\hline
%\bf Malt&72.63&89.32&19.81&8.92\\%426
%\bf MST&75.35&89.08&20.32&9.03\\%437
%\bf Turbo&74.85&90.48&22.28&8.96\\%479
%\bf Mate &77.54&77.54&100&19.6\\%2150

&\bf Malt&\bf MST &\bf Turbo &\bf Mate\\ \hline
\bf LAS (Single)&72.63&75.35&74.85&77.54\\
\bf LAS (Identical)&89.32&89.08&90.48&-\\
\bf Identical rate&19.81&20.32&22.28&-\\
\bf Avg. Length&8.92&9.03&8.96&-\\
%&426&437&479&2150

\hline
\end{tabular}
\end{center}
\caption{\label{table:co-training_identical_las} The analysis of identical annotations on \textsc{Weblogs} development set.}
\end{table}

The agreement based co-training depends on the assumption that identical annotations between two learners indicate the correctness. To confirm the suitability of selected parsers, in the preliminary evaluation we assessed the accuracy of identical analysis generated by parser pairs. Because we intend to use the Mate parser as our evaluation parser, we paired each of the other three parsers with Mate parser to create three co-training pairs. We assess our assumption by annotating our \textsc{Weblogs} development set, the development set is parsed by all four parsers. We then extract the identical annotations (whole sentence) from parser pairs. We show the accuracy of individual parsers and the accuracy of identical annotations in Table \ref{table:co-training_identical_las}. The second row shows the labelled accuracy of each parser on the \textsc{Weblogs} development set. The third row shows the labelled accuracy of the identical annotations between the named parser and Mate parser. The fourth row shows the agreement rate of the parser pairs. The last row shows the average sentence length of the identical annotations. As we can see from the table, our assumption is correct on all the parser pairs. Actually, when they agreed on the annotations, the accuracies can be 16\% higher than that of individual parsers. However, we also noticed that the average sentence length of the identical annotations is in stark contrast with that of the entire development set (19.6 tokens/sentence). We will discuss this potential conflict in the later section.

\section{Experiment Set-up} \label{section:co-experiment-setup}
In our evaluation on co-training we use our main evaluation corpora that consists of a source domain training set (\textsc{Conll}), a \textsc{Weblogs} domain development set, a in-domain test set (\textsc{Conll}) and four out-of-domain test sets (\textsc{Weblogs, Newsgroups, Reviews} and \textsc{Answers}). For each target domains, we used in addition a large unlabelled dataset to supply the additional training set. We evaluate various different settings on the development set to tune the best configuration, after that, we apply the best setting to all the test domains.

As mentioned before, we used four parsers in our experiments. cf. the Malt parser \cite{nivre2009non}, MST parser \cite{mcdonald2006online}, Mate parser \cite{bohnet2013joint}, and the Turbo parser \cite{martins2010turbo}. We use the default settings for all the parsers.  The part-of-speech tags is annotated by Mate parser's internal tagger. 

To create the additional training corpus, the unlabelled datasets are annotated by all the parsers which are trained on the \textsc{Conll} source domain training set. The Mate parser is used as our evaluation learner, the baseline for all the domains are generated by Mate parser trained on the same \textsc{Conll} training set and applied directly to target domains.

We mainly report the labelled attachment scores (LAS), but also include the unlabelled attachment scores (UAS) for our evaluations on test sets. We mark the significance levels according to the p-values, * indicates significance at the p \textless\ 0.05 level, ** for the p \textless\ 0.01 level.

\section{Empirical Results}\label{section:co-results}
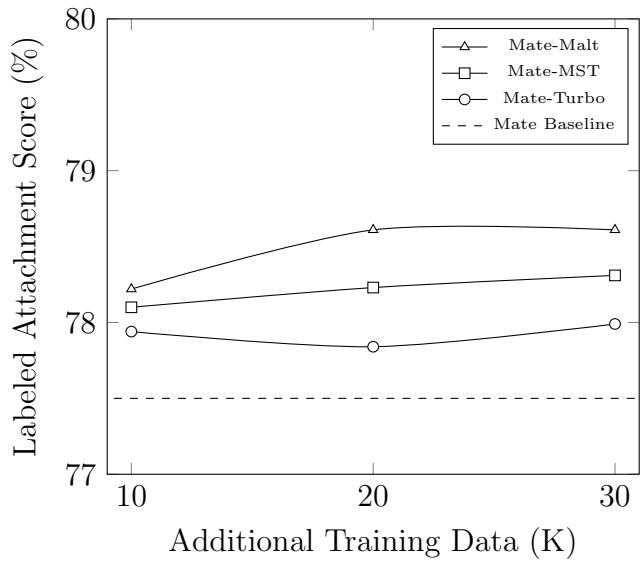
\begin{figure}[t]
	\begin{center}
		\begin{tikzpicture}
		\pgfplotsset{
			xmin=9,xmax=31,
			xtick={10,20,...,30},
			ymin=77,ymax=80,
			xlabel=Additional Training Data (K),
			ylabel=Labeled Attachment Score (\%)
		}
		\begin{axis}

		%Mate+Malt &  78.22& \bf 78.61&\bf 78.61 \\
		\addplot[smooth,mark=triangle*,mark options={fill=white}]
		coordinates {(10,78.22) (20,78.61) (30,78.61)};
		\addlegendentry{\tiny Mate-Malt}
		
		%Mate+MST &78.10 &78.23&78.31 \\
		\addplot[smooth,mark=square*, mark options={fill=white}]
		coordinates{(10,78.10) (20,78.23) (30,78.31)};
		\addlegendentry{\tiny Mate-MST}
		
		%Mate+Turbo & 77.94 &77.84 &77.99
		\addplot[smooth,mark=*, mark options={fill=white}] 
		coordinates {(10,77.94) (20,77.84) (30,77.99)};
		\addlegendentry{\tiny Mate-Turbo}
		
		%baseline
		\addplot[smooth,dashed] 
		coordinates{(0,77.5) (40,77.5)};
		\addlegendentry{\tiny Mate Baseline}
		
		\end{axis}
		
		\end{tikzpicture}
	
	\end{center}
	\caption{\label{figure:co-normal-cotraining} The results of our normal agreement-based co-training with three different parser pairs. }
\end{figure}

\textbf{Agreement based co-training.} We first evaluate the parser pairs on the normal agreement based co-training. Each of the other three parsers is paired with Mate parser to be the source learners of our co-training. For each pairwise parser combinations, the unlabelled \textsc{Weblogs} text is double parsed by the parser pairs. The sentences that are annotated identically by both parsers are used as candidates for the additional training set. We take different amount of additional training sentences from the candidates pool to retrain the Mate parser. Figure \ref{figure:co-normal-cotraining} shows the co-training results of adding 10k to 30k additional training data for all three parser pairs. As we can see from the figure, all the co-training results achieved improvements when compared with the Mate baseline. The largest improvement of one percentage point is achieved by Mate-Malt parser pair when adding 20k or 30k additional training data. We also notice a negative correlation between the improvement and the identical rate mentioned previously in Table \ref{table:co-training_identical_las}. The Turbo parser has the highest identical rate, in which it annotated 479 out of 2150 sentences (22.28\%) exactly the same as Mate parser when evaluated on the development set. This is 2\% higher than that of MST parser and 2.5\% higher than the Malt parser. However, the improvements achieved by the pairs are shown to be negatively correlated, i.e. the Mate-Malt pair gains the largest improvement, the Mate-Turbo pair achieved the lowest gain. This finding is in-line with the fundamental of co-training that requires the learners to be as different as possible. 

%\begin{table}[t]
%\begin{center}
%\begin{tabular}{|l|l|l|l|}
%\hline  & \sc Conll +10k & \sc Conll +20k &\sc Conll +30k \\ \hline
%Mate+Malt &  78.22& \bf 78.61&\bf 78.61 \\
%Mate+MST &78.10 &78.23&78.31 \\
%Mate+Turbo & 77.94 &77.84 &77.99 \\
%%Mate+All&77.89&78.07&78.3\\
%\hline
%Baseline & \multicolumn{3}{c|}{ 77.54}\\
%\hline
%\end{tabular}
%\end{center}
%\caption{\label{table:co-normal-cotraining} Agreement-based co-training using two parsers. }
%\end{table}

\begin{figure}[t]
	\begin{center}
		\begin{tikzpicture}
		\pgfplotsset{
			xmin=3.9,xmax=6.1,
			xtick={4,5,...,6},
			ymin=78,ymax=80,
			xlabel=Mininum Sentence Length,
			ylabel=Labeled Attachment Score (\%)
		}
		\begin{axis}

		%Exclude Short Sentences &78.67&78.61*78.88
		\addplot[smooth,mark=triangle*,mark options={fill=white}]
		coordinates {(4,78.67) (5,78.61)(6,78.88)};
		\addlegendentry{\tiny Exclude Short Sentences}

		%baseline
		\addplot[smooth,dashed] 
		coordinates{(0,78.61) (10,78.61)};
		\addlegendentry{\tiny All Sentences}
		
		\end{axis}
		
		\end{tikzpicture}
					
	\end{center}
	\caption{\label{figure:co-gt-cotraining} The effect of omitting short sentences from additional training data. }
\end{figure}
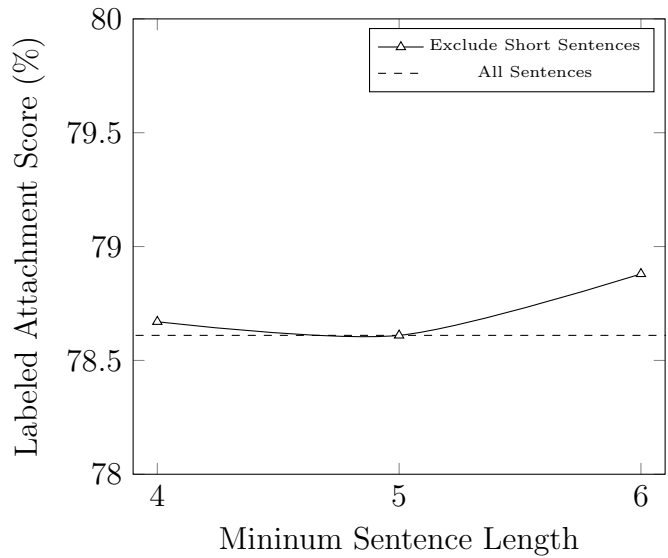

\textbf{Removing short sentences from identical data.} The identical annotations between the parsers are like a double-edged sword, they consist of a higher accuracy but in the same time shorter in average sentence length. Take our Mate-Malt pair as an example, the average sentence length of the identical annotations is only 8 tokens, this is much lower than the development set's 19.6 tokens/sentence and the \textsc{Conll} training set's 24.4 tokens/sentence. To make the additional training data more similar to the manually annotated data, we exclude the extremely short sentences from the pool. More precisely we set three minimal sentence length thresholds (4, 5 and 6 tokens), sentences shorter than the thresholds are removed from the pool.  We then take 30k sentences from the remaining pool as the additional training data. By taking out the short sentences the average sentence length of the selected sentences is closer to that of the development set. As shown in Table \ref{table:co-gt-eval}, the average sentence length reached 13 tokens/sentence. One of the major concerns when we exclude the short sentences from the pool is that the accuracy of the remaining pool might drop. The short sentences are easier to parse, thus they usually have a higher accuracy. However, an evaluation on the development set shows that there is almost no effect on the accuracies (see Table \ref{table:co-gt-eval}). In term of the results, we gained a 0.27\% additional improvement when discarding short sentences (Figure \ref{figure:co-gt-cotraining}).

\begin{table}[t]
	\begin{center}
		\begin{tabular}{|l|r|r|r|}
			\hline &\bf LAS (Identical) &\bf Avg. Length&\bf Identical Sentences\\ \hline
			\textgreater 6 tokens & 89.44 &13.1&248\\
			\textgreater 5 tokens &89.29 &12.67&278\\
			\textgreater 4 tokens &89.19&11.94&311 \\
			All sentences & 89.32&8.35&426\\
			\hline
		\end{tabular}
	\end{center}
	\caption{\label{table:co-gt-eval} The quantity and quality (LAS) of identical (Mate-Malt) development set sentences when omitting the short sentences.}
\end{table}

%\begin{table}[t]
%\begin{center}
%\begin{tabular}{|l|r|}
%\hline & \sc Conll+30k \\ \hline
%\textgreater 6 tokens & \bf 78.88\\
%\textgreater 5 tokens & 78.61\\
%\textgreater 4 tokens & 78.67 \\
%All sentences & 78.61\\
%\hline
%\end{tabular}
%\end{center}
%\caption{\label{table:co-gt-cotraining} The effect of removing short sentences from generated training data.}
%\end{table}

\begin{figure}[t]
	\begin{center}
		\begin{tikzpicture}
		\pgfplotsset{
			xmin=8,xmax=42,
			xtick={10,20,...,40},
			ymin=77,ymax=80,
			xlabel=Additional Training Data (K),
			ylabel=Labeled Attachment Score (\%)
		}
		\begin{axis}

		%Mate+Malt+MST & 78.70* & \textbf{79.12*} & 78.95 \\
		\addplot[smooth,mark=triangle*,mark options={fill=white}]
		coordinates {(10,78.70) (20,79.12) (30,78.95)};
		\addlegendentry{\tiny Tri-training}
		
		%Mate+Malt & 78.43 & 78.70 & 78.88 &78.61\\
		\addplot[smooth,mark=square*, mark options={fill=white}]
		coordinates{(10,78.43) (20,78.70) (30,78.88) (40,78.61)};
		\addlegendentry{\tiny Normal Co-training}

		%baseline
		\addplot[smooth,dashed] 
		coordinates{(0,77.5) (50,77.5)};
		\addlegendentry{\tiny Mate Baseline}
		
		\end{axis}
		
		\end{tikzpicture}
	\end{center}
	\caption{\label{figure:co-tritraining} The results of our tri-training compared with normal co-training. }
\end{figure}
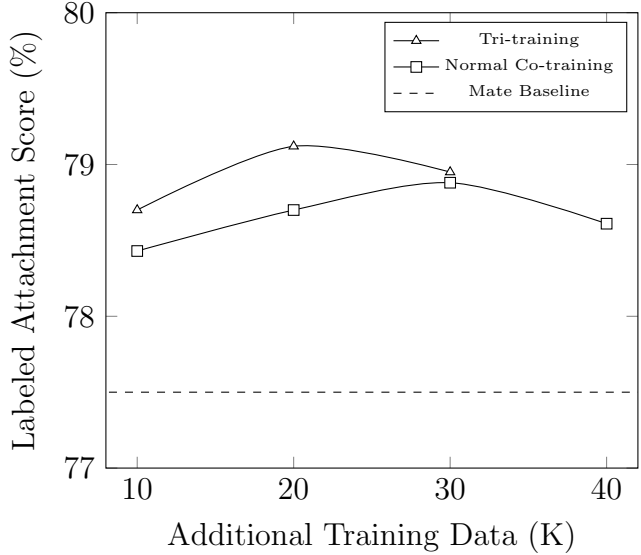

\textbf{Three learners co-training.} In the normal co-training setting, the Mate parser is used as one of the source learners to provide additional training data for retraining itself. Based on this setting the Mate parser can learn only from the annotations it has already known. The tri-training algorithm is on the other hand designed to allow the evaluation learner to learn from sources other than itself. This gives the Mate parser the potential to explore novel examples from other parsers. In our tri-training experiments, we used the Malt parser and the MST parser as our source learners. The sentences that are annotated identically by these parsers are added to the pool for retraining the Mate parser. To assess the quality of the identical annotations between Malt and MST parsers we apply them to our development set. We also assessed the sentences that are annotated identically by Malt and MST parsers but different to Mate parser's annotation, this allows us to know the scale of the novel examples. As shown in Table \ref{table:co-tri-nomate-eval}, the accuracy of the sentences agreed by Malt and MST parsers is even slightly higher than that of Mate and Malt parsers, this is surprising as MST parser is less accurate than Mate parser. The analysis also showed that half of the identical annotations from Malt and MST parsers are actually novel to Mate parser. We compared our tri-training and co-training results in Figure \ref{figure:co-tritraining}, the tri-training results constantly outperform the normal co-training. The best result of 79.12\% is achieved by retraining the Mate parser with 20k additional training data agreed by Malt-MST parsers (tri-training). The best tri-training result is 0.24\% higher than that of co-training and nearly 1.6\% higher than the Mate baseline.  

%\begin{table}[h]
%\begin{center}
%\begin{tabular}{|l|l|l|l|}
%\hline  & \bf +10k &\bf +20k& \bf +30k \\ \hline
%Mate+Malt+MST & 78.70* & \textbf{79.12*} & 78.95 \\
%Mate+Malt & 78.43 & 78.70 & 78.88 \\
%\hline
%\end{tabular}
%\end{center}
%\caption{\label{table:tritraining} Accuracy scores for tri-training (Mate+Malt+MST) and the best two-parser co-training algorithm (Mate+Malt).}
%\end{table}

\begin{table}[t]
	\begin{center}
		\begin{tabular}{|l|r|r|}
			\hline &\bf LAS (Identical) &\bf Identical Sentences\\ \hline
			Mate-Malt &89.44 &248\\
			Malt-MST &90.20 &300\\
			Malt-MST excl. Mate &89.28&147 \\
			\hline
		\end{tabular}
	\end{center}
	\caption{\label{table:co-tri-nomate-eval} The quantity and quality (LAS) of identical development set sentences agreed by different parser pairs.}
\end{table}

\textbf{Evaluating on test domains.} We then evaluated our best configuration (tri-training) on our four test domains. Under the tri-training setting, the unlabelled datasets of each domain are double parsed by Malt-MST pairs, the first 20k identical annotations are used as additional training data to retrain the Mate parser. The only exception is for answers domain. Due to the lack of unlabelled data the additional training data is much smaller, we used all 3k identical sentences for retraining. Table \ref{table:co-test-domain} shows our tri-training results accompanied by the baselines. The tri-training setting achieved large labelled improvements up to 1.8 percentage points. For unlabelled attachment scores, the models gained up to 0.59\% absolute improvements. We also tested the retrained \textsc{Weblogs} domain model on the in-domain test set. The results show the tri-trained model does not affect the in-domain accuracy.

\begin{table}[t]
\begin{center}
\begin{tabular}{|l|ll|ll|}
\cline{2-5}
\multicolumn{1}{c|}{} &\multicolumn{2}{|c|}{\bf Tri-training} &\multicolumn{2}{|c|}{\bf Baseline}\\ \cline{2-5}
\multicolumn{1}{c|}{}&\bf LAS&\bf UAS&\bf LAS&\bf UAS\\\hline
\sc Weblogs &80.59**&85.61**&78.99&85.1\\
\sc Newsgroups&76.44**&83.13&75.3&82.88\\
\sc Reviews&76.87**&83.27**&75.07&82.68\\
\sc Answers&74.59**&81.58&73.08&81.15\\\hline
\sc Conll&90.16&92.47&90.07&92.4\\\hline
\end{tabular}
\end{center}
\caption{\label{table:co-test-domain} The effect of applying the best configuration (tri-training) to our test domains. }
\end{table}

\section{Analysis}\label{section:co-analysis}
From the above experiments, we demonstrated the effect of co-/tri-training on parsing out-of-domain text with the off-the-shelf parsers. It remains unclear how the additional training data helps the target domain parsing. To understand where the improvements come from, in this section we give a detailed study on the results. We compare the annotations produced by our tri-training approach and the baseline and evaluate the changes on both token level and sentence level. For our analysis, we treat all the target domain as the same, the \textsc{Weblogs, Newsgroups, Reviews} and \textsc{Answers} domain test sets are used as a single set. 

\subsection{Token Level Analysis}

%label evaluation
\begin{figure}[t]
	\begin{center}
		\begin{tikzpicture}
		\pgfplotsset{
			width=12cm,height=7cm,
			enlargelimits=0.01,
			legend style={at={(0.5,-0.15)},anchor=north,legend columns=-1},
		}
		\begin{axis}[
		ymin=-0.7,ymax=4,
		ybar,ybar interval=0.7,
		axis y line*=left,
		ylabel=Accuracy Change (\%),
		symbolic x coords={NMOD,P,SBJ,PMOD,ROOT,ADV,COORD,VC,CONJ,DEP,AMOD,TMP,OBJ,PRD,IM},
		xtick=data,
		x tick label style={font=\tiny,rotate=45,anchor=east}
		]
		\addplot coordinates{(NMOD,0.3) (P,-0.5) (SBJ,0.8) (PMOD,0.6) (ROOT,0.4)  (ADV,0.5) (COORD,0.0) (VC,3.5) (CONJ,0.4) (DEP,-0.4)  (AMOD,0.9) (TMP,0.4)(OBJ,0) (PRD,0)(IM,0)}; 
		\addlegendentry{\tiny Recall}
		
		\addplot coordinates{(NMOD,0.5) (P,0.4) (SBJ,0.3) (PMOD,0.5) (ROOT,0.5)  (ADV,2.1) (COORD,1.2) (VC,-0.6) (CONJ,0.1) (DEP,2.0)  (AMOD,0.4) (TMP,3.0)(OBJ,0) (PRD,0)(IM,0)}; 
		\addlegendentry{\tiny Precision}
		
		\addplot coordinates{(NMOD,0.4) (P,-0.1) (SBJ,0.6) (PMOD,0.6) (ROOT,0.4) (ADV,1.3) (COORD,0.5) (VC,1.5) (CONJ,0.3) (DEP,0.8)  (AMOD,0.7) (TMP,1.7) (OBJ,0) (PRD,0)(IM,0) }; 
		\addlegendentry{\tiny F-score}
		
		\end{axis}
		\begin{axis}[
		ybar,ybar interval=0.7,
		ylabel=Accuracy Change (\%),
		axis y line*=right,
		ymin=-7,ymax=40,
		symbolic x coords={NMOD,P,SBJ,PMOD,ROOT,ADV,COORD,VC,CONJ,DEP,AMOD,TMP,OBJ,PRD,IM},
		xtick=data,
		xticklabels={\empty}
		]
		\addplot coordinates{(NMOD,0) (P,0) (SBJ,0) (PMOD,0) (ROOT,0)  (ADV,0) (COORD,0) (VC,0) (CONJ,0) (DEP,0)  (AMOD,0) (TMP,0)(OBJ,1.2) (PRD,34.3) (IM,0)}; 
		\addlegendentry{\tiny Recall}
		
		\addplot coordinates{(NMOD,0) (P,0) (SBJ,0) (PMOD,0) (ROOT,0)  (ADV,0) (COORD,0) (VC,0) (CONJ,0) (DEP,0)  (AMOD,0) (TMP,0)(OBJ,10.6) (PRD,-3.8) (IM,0)}; 
		\addlegendentry{\tiny Precision}
		
		\addplot coordinates{(NMOD,0) (P,0) (SBJ,0) (PMOD,0) (ROOT,0)  (ADV,0) (COORD,0) (VC,0) (CONJ,0) (DEP,0)  (AMOD,0) (TMP,0)(OBJ,5.9) (PRD,15.2)(IM,0) }; 
		\addlegendentry{\tiny F-score}
		
		\draw [very thick,dashed,color=blue] (120,-100) -- (120,1000);
		
		\end{axis}

		%NMOD cnt:24656 R/P/F89.6/87.9/88.7 Base R/P/F89.3/87.4/88.3 diff: F:0.4 R:0.3 P:0.5
		%P cnt:11977 R/P/F92.2/99.1/95.6 Base R/P/F92.7/98.7/95.7 diff: F:-0.1 R:-0.5 P:0.4
		%SBJ cnt:9199 R/P/F92.8/88.2/90.5 Base R/P/F92.0/87.9/89.9 diff: F:0.6 R:0.8 P:0.3
		%PMOD cnt:9195 R/P/F90.2/90.4/90.3 Base R/P/F89.6/89.9/89.7 diff: F:0.6 R:0.6 P:0.5
		%ROOT cnt:6345 R/P/F86.2/86.3/86.2 Base R/P/F85.8/85.8/85.8 diff: F:0.4 R:0.4 P:0.5
		%OBJ cnt:6291 R/P/F88.3/80.1/84.2 Base R/P/F87.1/69.5/78.3 diff: F:5.9 R:1.2 P:10.6
		%ADV cnt:6088 R/P/F67.6/71.4/69.5 Base R/P/F67.1/69.3/68.2 diff: F:1.3 R:0.5 P:2.1
		%COORD cnt:4751 R/P/F81.4/90.1/85.7 Base R/P/F81.4/88.9/85.2 diff: F:0.5 R:0.0 P:1.2
		%VC cnt:4028 R/P/F92.6/93.0/92.8 Base R/P/F89.1/93.6/91.3 diff: F:1.5 R:3.5 P:-0.6
		%CONJ cnt:3455 R/P/F86.0/86.0/86.0 Base R/P/F85.6/85.9/85.7 diff: F:0.3 R:0.4 P:0.1
		%DEP cnt:2866 R/P/F24.1/61.3/42.7 Base R/P/F24.5/59.3/41.9 diff: F:0.8 R:-0.4 P:2.0
		%PRD cnt:2827 R/P/F77.9/85.6/81.7 Base R/P/F43.6/89.4/66.5 diff: F:15.2 R:34.3 P:-3.8
		%AMOD cnt:2511 R/P/F58.2/72.3/65.3 Base R/P/F57.3/71.9/64.6 diff: F:0.7 R:0.9 P:0.4
		%TMP cnt:2387 R/P/F65.4/70.0/67.7 Base R/P/F65.0/67.0/66.0 diff: F:1.7 R:0.4 P:3.0
		%IM cnt:1692 R/P/F98.3/96.9/97.6 Base R/P/F98.1/97.2/97.6 diff: F:0.0 R:0.2 P:-0.3
		
		\end{tikzpicture}
	\end{center}
	\caption[The performance comparison between the tri-training approach and the baseline on major labels.]{\label{figure:analysis-label-tri-training} The performance comparison between the tri-training approach and the baseline on major labels. The x-axis shows the labels, the y-axis to the left shows the accuracy changes for labels from start to TMP, the y-axis to the right-hand side is for label OBJ and PRD only.}
\end{figure}
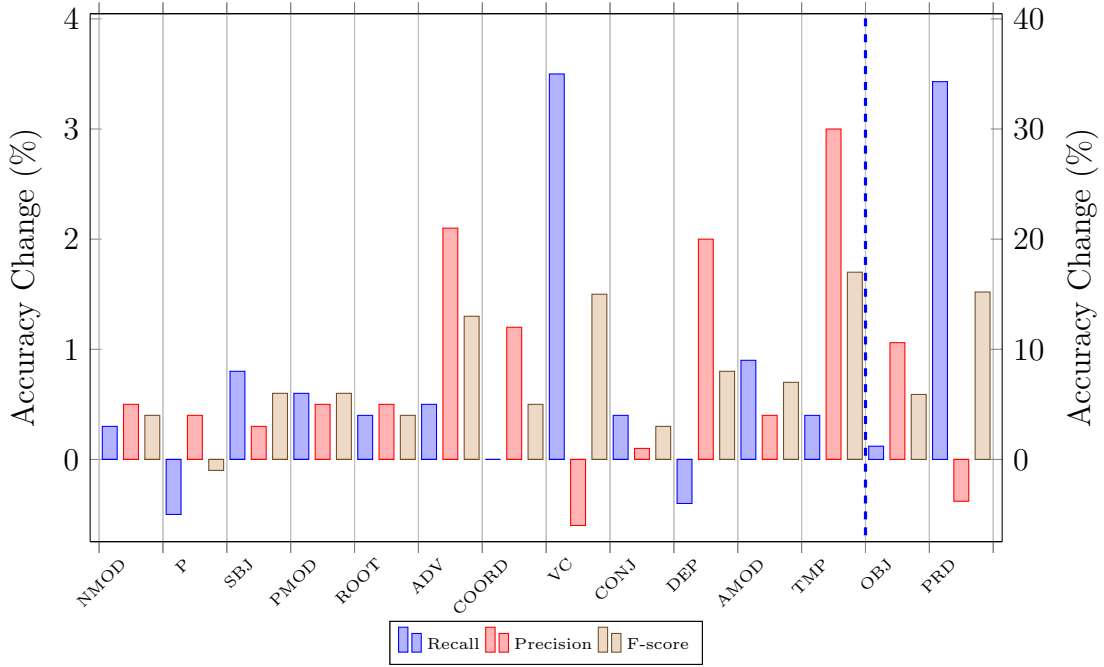

%Confution matric
\begin{table}
\begin{center}
\begin{tabular}{l|r|r}
\hline
\bf Confusion & \bf Baseline & \bf tri-training\\\hline
\footnotesize NMOD $\rightarrow$ \footnotesize ADV & 235 & 229\\
\footnotesize NMOD $\rightarrow$ \footnotesize LOC & 162 & 164\\
\footnotesize NMOD $\rightarrow$ \footnotesize HYPH & 198 & 196\\
\footnotesize NMOD $\rightarrow$ \footnotesize NAME & 569 & 583\\
\footnotesize NMOD $\rightarrow$ \footnotesize PMOD & 187 & 179\\
\footnotesize NMOD $\rightarrow$ \footnotesize HMOD & 217 & 213\\
\footnotesize NMOD $\rightarrow$ \footnotesize ROOT,OBJ,SBJ,DEP & 491 & 442\\\hline
\footnotesize P $\rightarrow$ \footnotesize HYPH & 162 & 173\\
\footnotesize P $\rightarrow$ \footnotesize NAME,NMOD & 233 & 245\\\hline
\footnotesize SBJ $\rightarrow$ \footnotesize NMOD & 169 & 157\\
\footnotesize SBJ $\rightarrow$ \footnotesize OBJ & 132 & 72\\\hline
\footnotesize OBJ $\rightarrow$ \footnotesize NMOD & 218 & 202\\
\footnotesize OBJ $\rightarrow$ \footnotesize SBJ & 117 & 89\\\hline
\footnotesize PMOD $\rightarrow$ \footnotesize NMOD & 290 & 275\\
\footnotesize PMOD $\rightarrow$ \footnotesize OBJ & 122 & 92\\\hline
\footnotesize ROOT $\rightarrow$ \footnotesize NMOD & 235 & 235\\
\footnotesize ROOT $\rightarrow$ \footnotesize OBJ,SBJ & 256 & 216\\\hline
\footnotesize ADV $\rightarrow$ \footnotesize MNR & 150 & 129\\
\footnotesize ADV $\rightarrow$ \footnotesize AMOD & 152 & 134\\
\footnotesize ADV $\rightarrow$ \footnotesize LOC & 227 & 214\\
\footnotesize ADV $\rightarrow$ \footnotesize NMOD & 382 & 373\\
\footnotesize ADV $\rightarrow$ \footnotesize TMP & 182 & 195\\
\footnotesize ADV $\rightarrow$ \footnotesize DIR & 118 & 113\\\hline
\footnotesize COORD $\rightarrow$ \footnotesize NMOD & 164 & 140\\
\footnotesize COORD $\rightarrow$ \footnotesize ROOT & 102 & 94\\\hline
\footnotesize VC $\rightarrow$ \footnotesize OPRD & 114 & 23\\\hline
\footnotesize CONJ $\rightarrow$ \footnotesize NMOD & 132 & 130\\\hline
\footnotesize DEP $\rightarrow$ \footnotesize ROOT & 190 & 199\\
\footnotesize DEP $\rightarrow$ \footnotesize OBJ & 267 & 241\\
\footnotesize DEP $\rightarrow$ \footnotesize SBJ & 403 & 420\\
\footnotesize DEP $\rightarrow$ \footnotesize NMOD & 382 & 396\\
\footnotesize DEP $\rightarrow$ \footnotesize TMP & 176 & 165\\
\footnotesize DEP $\rightarrow$ \footnotesize ADV & 142 & 133\\\hline
\footnotesize AMOD $\rightarrow$ \footnotesize ADV & 169 & 157\\
\footnotesize AMOD $\rightarrow$ \footnotesize NMOD & 265 & 273\\
\footnotesize AMOD $\rightarrow$ \footnotesize HYPH & 104 & 105\\\hline
\footnotesize TMP $\rightarrow$ \footnotesize ADV & 280 & 268\\
\footnotesize TMP $\rightarrow$ \footnotesize NMOD & 133 & 128\\\hline
\footnotesize PRD $\rightarrow$ \footnotesize OBJ & 854 & 97\\
\footnotesize PRD $\rightarrow$ \footnotesize ADV,VC & 255 & 184\\\hline
\end{tabular}
\end{center}
\caption{\label{table:analysis-confusion-tri-training} The confusion matrix of dependency labels, compared between the tri-training approach and the baseline.}
\end{table}

\textbf{Individual Label Accuracy.} We first compared the individual label accuracies of the tri-trained model and the baseline. For each of the label we calculate recalls, precisions and f-scores, we then compute the score differences between the tri-trained model and the baseline model.    Table \ref{figure:analysis-label-tri-training} shows the score changes of the most frequent labels. All the f-scores of our tri-trained model outperform the baseline, the only exception is the P (punctuations) which drops slightly by 0.1\%. Eight labels achieved around 0.5\% improvements which include ROOT (root of the sentence), SBJ (subject), COORD (coordination), CONJ (conjunct), modifiers (NMOD (modifier of nominal), PMOD (modifier of preposition), AMOD (modifier of adjective or adverbial)) and DEP (unclassified relations). ADV  (adverbial), VC (verb chain) and TMP (temporal adverbial or nominal modifier) are labels that have improvements between 1\% and 2\%. The accuracy changes are much larger for label OBJ and PRD, thus we used a secondary y-axis for them. More precisely, an improvement of 5.9\% is found on OBJ (object), a much better precision of 10\% suggests this improvement is mainly contributed by the reduced false positive. The largest improvement of 15\% comes from label PRD (predicative complement), the improvement is as a result of significant recall change. The baseline parser can only recall 43\% of the label, it has been improved significantly (34\%) by the tri-trained model. Table \ref{table:analysis-confusion-tri-training} shows the confusion matrix of dependency labels. As we can see from the table, the PRD has been frequently labeled as OBJ by the baseline, but this has been largely corrected by our tri-training model.

%corpus UNK
\begin{table}[t]
	\begin{center}
		\begin{tabular}{|l|r|rr|rr|}
			\cline{3-6}
			\multicolumn{2}{c|}{}& \multicolumn{2}{|c|}{\bf Tri-training} &\multicolumn{2}{|c|}{\bf Baseline}\\
			\cline{2-6}
			\multicolumn{1}{c|}{}&\bf Tokens &\bf LAS&\bf UAS&\bf LAS&\bf UAS\\
			\hline
			\bf Known &101616&78.7&84.5&77.1&84.1\\%diff: LAS:1.6 UAS:0.4
			\bf Unknown &6055&63.2&72.6&61.4&71.9\\%diff: LAS:1.8 UAS:0.7
			\hline
			\bf All &107671&77.8&83.8&76.3&83.4\\%diff: LAS:1.5 UAS:0.4
			\hline
		\end{tabular}
	\end{center}
	\caption{\label{table:analysis-corpusunk-tri-training} The accuracy comparison between the tri-training approach and the baseline on unknown words.}
\end{table}

\textbf{Unknown Words Accuracy.} We then evaluate unknown words at the token level, by comparing the labelled and unlabelled accuracy scores between words that presented in the source domain training data (Known) and words that are unseen from training sets (Unknown). We present the accuracy comparison of known/unknown words together with that of all tokens in Table \ref{table:analysis-corpusunk-tri-training}. The tri-trained model achieved better gains on unknown words for both labelled and unlabelled accuracies. The labelled gains of the tri-trained model on unknown words are 1.8\%, which is 0.2\% higher than that of known words (1.6\%). The unlabelled improvements on unknown words (0.7\%) is 0.3\% higher than known words (0.4\%). Although the absolute gains for unknown words are larger, the performance of known words is still better in terms of the error reduction rate. For known words, tri-trained model reduced 7\% errors on labelled accuracy and this is 2.4\% better than that of unknown words. The error reduction for unlabelled accuracy is the same (2.5\%) for both unknown and known words.

\subsection{Sentence Level Analysis}
We then carry out our sentence level analysis, the sentence level analysis use sentences as a whole, all the tokens in the same sentences are always put into the same class. In total, we analysis four different sentences factors, our goal is to have a more clear picture about the improvements of different type of sentences.

%number of tokens
\begin{figure}[t]
	\begin{center}
		\begin{tikzpicture}
		\pgfplotsset{
			xmin=0,xmax=40,
			xlabel=Number of Tokens}
		\begin{axis}[
		axis y line*=left,
		ymin=0,ymax=100,
		ylabel=Percentage (\%)]
		%Better
		\addplot[smooth,thick,dashed,color=blue] coordinates {(2,8.1) (6,18.5) (10,27.1) (14,30.9) (18,35.7) (22,40.5) (26,42.0) (30,48.7) (34,43.8) (38,38.2) (42,41.4) (46,47.3) (50,46.9) (54,48.0) (58,52.4) (62,50.0) };\label{better}
		\addlegendentry{\tiny Better}
		
		%Worse
		\addplot[smooth,thick,dotted,color=red] coordinates{(2,5.2) (6,10.7) (10,13.5) (14,17.3) (18,16.3) (22,22.7) (26,23.1) (30,24.7) (34,28.6) (38,31.9) (42,37.9) (46,27.3) (50,30.6) (54,36.0) (58,28.6) (62,25.0) };\label{worse}
		\addlegendentry{\tiny Worse}
		
		%No Change
		\addplot[smooth,thick,dashdotted] coordinates{(2,86.7) (6,70.8) (10,59.4) (14,51.8) (18,48.0) (22,36.8) (26,34.9) (30,26.6) (34,27.6) (38,29.9) (42,20.7) (46,25.5) (50,22.4) (54,16.0) (58,19.0) (62,25.0) };\label{nochange}
		\addlegendentry{\tiny No Change}
		
		\end{axis}
		\begin{axis}[
		axis y line*=right,
		ymin=0,ymax=300,
		ylabel=Number of Sentences]
		\addlegendimage{/pgfplots/refstyle=better}\addlegendentry{\tiny Better}
		\addlegendimage{/pgfplots/refstyle=worse}\addlegendentry{\tiny Worse}
		\addlegendimage{/pgfplots/refstyle=nochange}\addlegendentry{\tiny No Change}
		%All Sent
		\addplot[smooth,thick,solid] coordinates{(2,178.0) (6,249.0) (10,252.0) (14,226.0) (18,189.0) (22,154.0) (26,101.0) (30,79.0) (34,46.0) (38,36.0) (42,21.0) (46,13.0) (50,12.0) (54,6.0) (58,5.0) (62,3.0) };\addlegendentry{\tiny No. of Sents}
		
		\end{axis}
		%2 Total:713 LAS better/worse/nochange:58/37/618 8.1/5.2/86.7 2.9
		%6 Total:996 LAS better/worse/nochange:184/107/705 18.5/10.7/70.8 7.8
		%10 Total:1010 LAS better/worse/nochange:274/136/600 27.1/13.5/59.4 13.6
		%14 Total:906 LAS better/worse/nochange:280/157/469 30.9/17.3/51.8 13.6
		%18 Total:759 LAS better/worse/nochange:271/124/364 35.7/16.3/48.0 19.4
		%22 Total:617 LAS better/worse/nochange:250/140/227 40.5/22.7/36.8 17.8
		%26 Total:407 LAS better/worse/nochange:171/94/142 42.0/23.1/34.9 18.9
		%30 Total:316 LAS better/worse/nochange:154/78/84 48.7/24.7/26.6 24.0
		%34 Total:185 LAS better/worse/nochange:81/53/51 43.8/28.6/27.6 15.2
		%38 Total:144 LAS better/worse/nochange:55/46/43 38.2/31.9/29.9 6.3
		%42 Total:87 LAS better/worse/nochange:36/33/18 41.4/37.9/20.7 3.5
		%46 Total:55 LAS better/worse/nochange:26/15/14 47.3/27.3/25.5 20.0
		%50 Total:49 LAS better/worse/nochange:23/15/11 46.9/30.6/22.4 16.3
		%54 Total:25 LAS better/worse/nochange:12/9/4 48.0/36.0/16.0 12.0
		%58 Total:21 LAS better/worse/nochange:11/6/4 52.4/28.6/19.0 23.8
		%62 Total:12 LAS better/worse/nochange:6/3/3 50.0/25.0/25.0 25.0
		
		\end{tikzpicture}
	\end{center}
	\caption{\label{figure:analysis-sentlength-tri-training} The comparison between the tri-training approach and the baseline on different number of tokens per sentence.}
\end{figure}
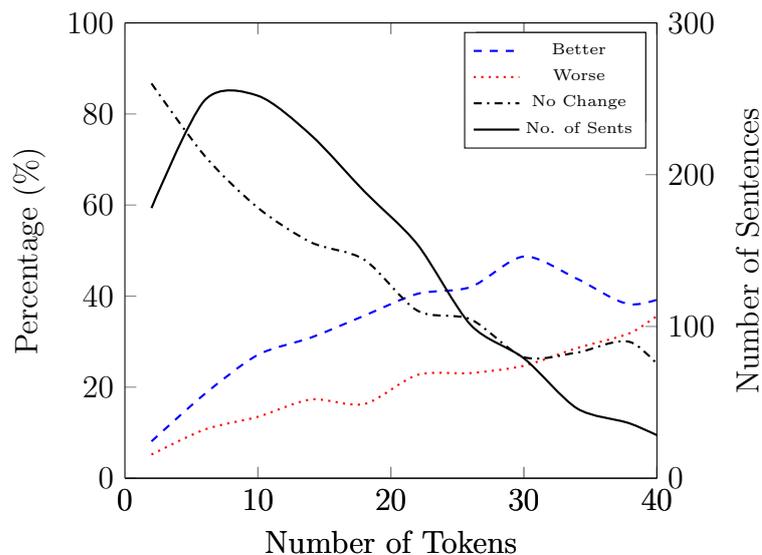

\textbf{Sentence Length.} Figure \ref{figure:analysis-sentlength-tri-training} shows the performance changes for sentences of different length, the results of the tri-trained model is compared with the baseline. As we can see from the figure, the percentage of sentences that remain the same accuracies continuously decrease when the sentence length increases. We suggest this is mainly because longer sentences are harder to parse, thus are less likely to have the same accuracy. The rate of sentences parsed better is constantly larger than that of parsed worse. The gaps widened when the sentence length increases until reached the widest point at a length of 30, after that the gap narrowed and become very close at 40 tokens. However, there are only less than 200 sentences in the classes which have a sentence length of more than 35, thus the results of those classes become less reliable. Overall, the analysis suggests the major improvements are contributed by sentences that have a length between 15 and 30 tokens.

%number of unknown words
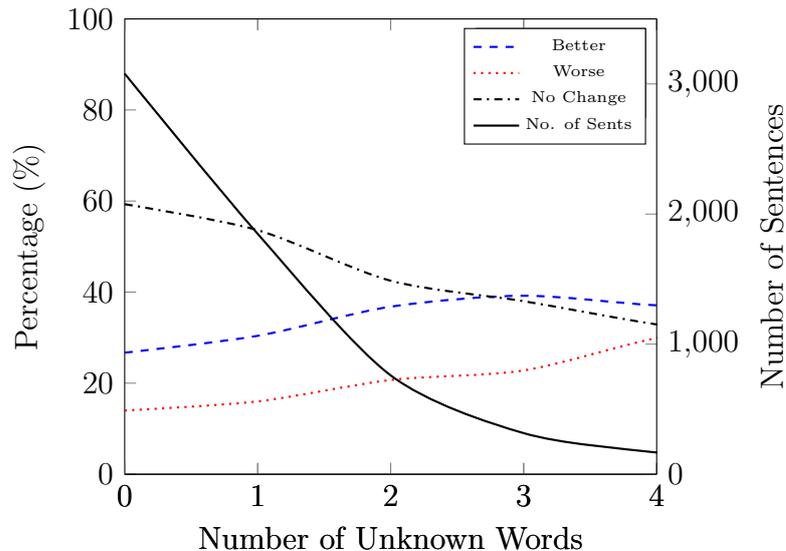
\begin{figure}[t]
	\begin{center}
		\begin{tikzpicture}
		\pgfplotsset{
			xmin=0,xmax=4,
			xtick={0,1,...,4},
			xlabel=Number of Unknown Words}
		\begin{axis}[
		axis y line*=left,
		ymin=0,ymax=100,
		ylabel=Percentage (\%)]
		%Better
		\addplot[smooth,thick,dashed,color=blue] coordinates {(0,26.7) (1,30.4) (2,36.8) (3,39.2) (4,37.1) (5,36.6) (6,33.3) };\label{better}
		\addlegendentry{\tiny Better}
		
		%Worse
		\addplot[smooth,thick,dotted,color=red] coordinates{(0,14.0) (1,16.0) (2,20.7) (3,22.8) (4,29.9) (5,34.1) (6,33.3) };\label{worse}
		\addlegendentry{\tiny Worse}
		
		%No Change
		\addplot[smooth,thick,dashdotted] coordinates{(0,59.3) (1,53.6) (2,42.5) (3,38.0) (4,32.9) (5,29.3) (6,33.3) };\label{nochange}
		\addlegendentry{\tiny No Change}
		
		\end{axis}
		\begin{axis}[
		axis y line*=right,
		ymin=0,ymax=3500,
		ylabel=Number of Sentences]
		\addlegendimage{/pgfplots/refstyle=better}\addlegendentry{\tiny Better}
		\addlegendimage{/pgfplots/refstyle=worse}\addlegendentry{\tiny Worse}
		\addlegendimage{/pgfplots/refstyle=nochange}\addlegendentry{\tiny No Change}
		%All Sent
		\addplot[smooth,thick,solid] coordinates{(0,3079.0) (1,1848.0) (2,760.0) (3,316.0) (4,167.0) (5,82.0) (6,42.0) };\addlegendentry{\tiny No. of Sents}
		
		\end{axis}
		%0 Total:3079 LAS better/worse/nochange:821/432/1826 26.7/14.0/59.3 12.7
		%1 Total:1848 LAS better/worse/nochange:562/296/990 30.4/16.0/53.6 14.4
		%2 Total:760 LAS better/worse/nochange:280/157/323 36.8/20.7/42.5 16.1
		%3 Total:316 LAS better/worse/nochange:124/72/120 39.2/22.8/38.0 16.4
		%4 Total:167 LAS better/worse/nochange:62/50/55 37.1/29.9/32.9 7.2
		%5 Total:82 LAS better/worse/nochange:30/28/24 36.6/34.1/29.3 2.5
		%6 Total:42 LAS better/worse/nochange:14/14/14 33.3/33.3/33.3 0.0
		
		\end{tikzpicture}
	\end{center}
	\caption{\label{figure:analysis-sentunk-tri-training} The comparison between the tri-training approach and the baseline on different number of unknown words per sentence. }
\end{figure}

\textbf{Unknown Words.} Unknown words are hard to parse as the model trained on training data do not have sufficient information to annotate those words. Thus a large number of unknown words in a sentence usually results in a poor accuracy. We group sentences that have the same number of unknown words and then apply our analysis method to each class. We noted that 50\% of the sentences do not contain unknown words, 30\% of them contain one unseen word, 12\% of which contain 2 such words, the rest 8\% contain 3 or 4 unknown words.  For the sentences that do not contain unknown words, about 60\% of them remain the same accuracy, 25\% of them have a higher accuracy and 15\% of them are pared worse. This gap widened slowly until 3 unknown words per sentence, after that the gap narrowed for sentences have 4 unknown words. Overall, the gains on sentences with unknown words are slightly better than that of sentences contain only known words. This is in line with our finding in the token level analysis.

%number of prepositions
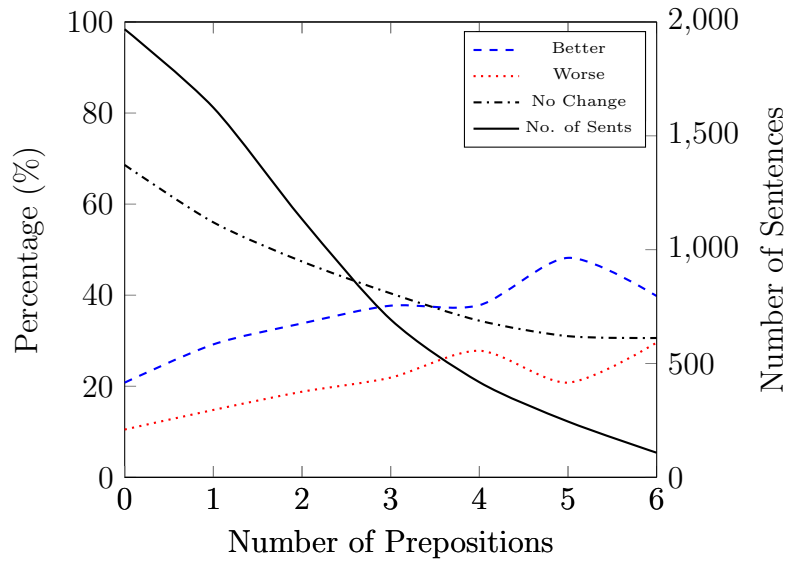
\begin{figure}[t]
	\begin{center}
		\begin{tikzpicture}
		\pgfplotsset{
			xmin=0,xmax=6,
			xlabel=Number of Prepositions}
		\begin{axis}[
		axis y line*=left,
		ymin=0,ymax=100,
		ylabel=Percentage (\%)]
		%Better
		\addplot[smooth,thick,dashed,color=blue] coordinates {(0,20.8) (1,29.2) (2,33.8) (3,37.7) (4,37.8) (5,48.2) (6,39.8) };\label{better}
		\addlegendentry{\tiny Better}
		
		%Worse
		\addplot[smooth,thick,dotted,color=red] coordinates{(0,10.5) (1,14.8) (2,18.8) (3,21.9) (4,27.8) (5,20.8) (6,29.6) };\label{worse}
		\addlegendentry{\tiny Worse}
		
		%No Change
		\addplot[smooth,thick,dashdotted] coordinates{(0,68.6) (1,56.0) (2,47.4) (3,40.4) (4,34.4) (5,31.0) (6,30.6) };\label{nochange}
		\addlegendentry{\tiny No Change}
		
		\end{axis}
		\begin{axis}[
		axis y line*=right,
		ymin=0,ymax=2000,
		ylabel=Number of Sentences]
		\addlegendimage{/pgfplots/refstyle=better}\addlegendentry{\tiny Better}
		\addlegendimage{/pgfplots/refstyle=worse}\addlegendentry{\tiny Worse}
		\addlegendimage{/pgfplots/refstyle=nochange}\addlegendentry{\tiny No Change}
		%All Sent
		\addplot[smooth,thick,solid] coordinates{(0,1968.0) (1,1624.0) (2,1134.0) (3,693.0) (4,418.0) (5,245.0) (6,108.0) };\addlegendentry{\tiny No. of Sents}
		
		\end{axis}
		%0 Total:1968 LAS better/worse/nochange:410/207/1351 20.8/10.5/68.6 Diff:10.3
		%1 Total:1624 LAS better/worse/nochange:474/241/909 29.2/14.8/56.0 Diff:14.4
		%2 Total:1134 LAS better/worse/nochange:383/213/538 33.8/18.8/47.4 Diff:15.0
		%3 Total:693 LAS better/worse/nochange:261/152/280 37.7/21.9/40.4 Diff:15.8
		%4 Total:418 LAS better/worse/nochange:158/116/144 37.8/27.8/34.4 Diff:10.0
		%5 Total:245 LAS better/worse/nochange:118/51/76 48.2/20.8/31.0 Diff:27.4
		%6 Total:108 LAS better/worse/nochange:43/32/33 39.8/29.6/30.6 Diff:10.2
		
		\end{tikzpicture}
	\end{center}
	\caption{\label{figure:analysis-sentin-tri-training} The comparison between the tri-training approach and the baseline on different number of prepositions per sentence. }
\end{figure}

\textbf{Prepositions.} The attachment of prepositions is one of the complex problems that are difficult for parsing. It can be found even harder when going out-of-domain, as their behaviour might change. To address those changes we looked at the labels assigned to the prepositions. For both source and target domain we find NMOD (Modifier of nominal), ADV (General adverbial), LOC (Locative adverbial or nominal modifier) and TMP (Temporal adverbial or nominal modifier) are the most frequently assigned labels, those labels covering 80\% of the total prepositions. However, the percentages for the source domain and the target domain are very different. In the source domain 35\% of the prepositions are labelled as NMOD and 19\% of them are labelled as ADV, while, in the target domain, the rate for NMOD and ADV are very close, both labels contribute around 28\%. In terms of our sentence level analysis on the number of prepositions, Figure \ref{figure:analysis-sentin-tri-training} illustrates the performance changes when the number of prepositions increases in sentences. The percentages of sentences parsed better and worse increased smoothly when the number of preposition increases, the tri-training gains at least 10\% for all the cases. Generally speaking, tri-training works better for sentences that have prepositions, the average gain for sentences that have prepositions is 15\% and this is 5\% more than that of sentences that do not have a proposition.

%number of conjunctions
\begin{figure}[t]
	\begin{center}
		\begin{tikzpicture}
		\pgfplotsset{
			xmin=0,xmax=3,
			xtick={0,1,...,3},
			xlabel=Number of Conjunctions}
			\begin{axis}[
			axis y line*=left,
			ymin=0,ymax=100,
			ylabel=Percentage (\%)]
			%Better
			\addplot[smooth,thick,dashed,color=blue] coordinates {(0,25.3) (1,34.1) (2,40.0) (3,50.7) };\label{better}
			\addlegendentry{\tiny Better}
			
			%Worse
			\addplot[smooth,thick,dotted,color=red] coordinates{(0,13.8) (1,18.8) (2,25.2) (3,23.3) };\label{worse}
			\addlegendentry{\tiny Worse}
			
			%No Change
			\addplot[smooth,thick,dashdotted] coordinates{(0,60.9) (1,47.1) (2,34.8) (3,26.0) };\label{nochange}
			\addlegendentry{\tiny No Change}
			
			\end{axis}
			\begin{axis}[
			axis y line*=right,
			ymin=0,ymax=4000,
			ylabel=Number of Sentences]
			\addlegendimage{/pgfplots/refstyle=better}\addlegendentry{\tiny Better}
			\addlegendimage{/pgfplots/refstyle=worse}\addlegendentry{\tiny Worse}
			\addlegendimage{/pgfplots/refstyle=nochange}\addlegendentry{\tiny No Change}
			%All Sent
			\addplot[smooth,thick,solid] coordinates{(0,3640.0) (1,1878.0) (2,615.0) (3,150.0) };\addlegendentry{\tiny No. of Sents}
			
			\end{axis}
			%0 Total:3640 LAS better/worse/nochange:922/501/2217 25.3/13.8/60.9 Diff:11.5
			%1 Total:1878 LAS better/worse/nochange:640/353/885 34.1/18.8/47.1 Diff:15.3
			%2 Total:615 LAS better/worse/nochange:246/155/214 40.0/25.2/34.8 Diff:14.8
			%3 Total:150 LAS better/worse/nochange:76/35/39 50.7/23.3/26.0 Diff:27.4
			
			\end{tikzpicture}
	\end{center}
	\caption{\label{figure:analysis-sentcc-tri-training} The comparison between the tri-training approach and the baseline on different number of conjunctions per sentence. }
\end{figure}

%examples
\begin{table}
\begin{center}
\begin{tabular}{p{\linewidth}}
\hline
\hline
\colorbox{blue!30}{\footnotesize If}$_{11_{\textsc{adv}}}^{1 \colorbox{red!30}{\tiny p}}$ \footnotesize you$_{3_{\textsc{sbj}}}^{2}$ \footnotesize come$_{1_{\textsc{sub}}}^{3}$ \footnotesize upon$_{3_{\textsc{adv}}}^{4 \colorbox{red!30}{\tiny p}}$ \footnotesize something$_{4_{\textsc{pmod}}}^{5}$ \footnotesize important$_{5_{\textsc{appo}}}^{6}$ \colorbox{blue!30}{\footnotesize ,}$_{11_{\textsc{p}}}^{7}$ \colorbox{blue!30}{\footnotesize by}$_{11_{\textsc{adv}}}^{8 \colorbox{red!30}{\tiny p}}$ \footnotesize all$_{10_{\textsc{nmod}}}^{9}$ \colorbox{blue!30}{\footnotesize means}$_{8_{\textsc{pmod}}}^{10}$ \colorbox{blue!30}{\footnotesize make}$_{0_{\textsc{root}}}^{11}$ \footnotesize a$_{13_{\textsc{nmod}}}^{12}$ \footnotesize note$_{11_{\textsc{obj}}}^{13}$ \footnotesize of$_{13_{\textsc{nmod}}}^{14 \colorbox{red!30}{\tiny p}}$ \footnotesize it$_{14_{\textsc{pmod}}}^{15}$ \colorbox{blue!30}{\footnotesize ,}$_{11_{\textsc{p}}}^{16}$ \colorbox{blue!30}{\footnotesize and}$_{11_{\textsc{coord}}}^{17 \colorbox{red!30}{\tiny c}}$ \footnotesize so$_{11_{\textsc{adv}}}^{18}$ \footnotesize on$_{18_{\textsc{amod}}}^{19}$ \colorbox{blue!30}{\footnotesize .}$_{11_{\textsc{p}}}^{20}$\\\hline

\colorbox{blue!30}{\footnotesize But}$_{31_{\textsc{dep}}}^{1 \colorbox{red!30}{\tiny c}}$ \colorbox{blue!30}{\footnotesize creating}$_{31_{\textsc{sbj}}}^{2}$ \footnotesize a$_{5_{\textsc{nmod}}}^{3}$ \footnotesize balanced$_{5_{\textsc{nmod}}}^{4}$ \footnotesize community$_{2_{\textsc{obj}}}^{5}$ \footnotesize with$_{5_{\textsc{nmod}}}^{6 \colorbox{red!30}{\tiny p}}$ \footnotesize a$_{8_{\textsc{nmod}}}^{7}$ \footnotesize mix$_{6_{\textsc{pmod}}}^{8}$ \footnotesize of$_{8_{\textsc{nmod}}}^{9 \colorbox{red!30}{\tiny p}}$ \footnotesize housing$_{9_{\textsc{pmod}}}^{10}$ \footnotesize ,$_{10_{\textsc{p}}}^{11}$ \footnotesize offices$_{10_{\textsc{coord}}}^{12}$ \footnotesize ,$_{12_{\textsc{p}}}^{13}$ \footnotesize shopping$_{12_{\textsc{coord}}}^{14}$ \footnotesize and$_{14_{\textsc{coord}}}^{15 \colorbox{red!30}{\tiny c}}$ \footnotesize other$_{17_{\textsc{nmod}}}^{16}$ \footnotesize amenities$_{15_{\textsc{conj}}}^{17}$ \footnotesize --$_{5_{\textsc{p}}}^{18}$ \footnotesize allowing$_{5_{\textsc{appo}}}^{19}$ \footnotesize people$_{19_{\textsc{obj}}}^{20}$ \footnotesize to$_{19_{\textsc{oprd}}}^{21}$ \footnotesize live$_{21_{\textsc{im}}}^{22}$ \footnotesize close$_{22_{\textsc{loc}}}^{23}$ \footnotesize to$_{23_{\textsc{amod}}}^{24 \colorbox{red!30}{\tiny p}}$ \colorbox{blue!30}{\footnotesize where}$_{27_{\textsc{loc}}}^{25}$ \footnotesize they$_{27_{\textsc{sbj}}}^{26}$ \footnotesize work$_{24_{\textsc{pmod}}}^{27}$ \footnotesize and$_{27_{\textsc{coord}}}^{28 \colorbox{red!30}{\tiny c}}$ \footnotesize play$_{28_{\textsc{conj}}}^{29}$ \footnotesize --$_{5_{\textsc{p}}}^{30}$ \colorbox{blue!30}{\footnotesize is}$_{0_{\textsc{root}}}^{31}$ \footnotesize an$_{36_{\textsc{nmod}}}^{32}$ \colorbox{blue!30}{\footnotesize even}$_{35_{\textsc{amod}}}^{33}$ \colorbox{blue!30}{\footnotesize more}$_{35_{\textsc{amod}}}^{34}$ \colorbox{blue!30}{\footnotesize worthy}$_{36_{\textsc{nmod}}}^{35}$ \colorbox{green!30}{\footnotesize goal}$_{31_{\textsc{prd}}}^{36}$ \colorbox{blue!30}{\footnotesize .}$_{31_{\textsc{p}}}^{37}$\\\hline

\footnotesize In$_{5_{\textsc{adv}}}^{1 \colorbox{red!30}{\tiny p}}$ \footnotesize some$_{3_{\textsc{nmod}}}^{2}$ \footnotesize respects$_{1_{\textsc{pmod}}}^{3}$ \footnotesize ,$_{5_{\textsc{p}}}^{4}$ \footnotesize is$_{0_{\textsc{root}}}^{5}$ \footnotesize n't$_{5_{\textsc{adv}}}^{6}$ \footnotesize that$_{5_{\textsc{sbj}}}^{7}$ \footnotesize essentially$_{5_{\textsc{adv}}}^{8}$ \footnotesize what$_{14_{\textsc{obj}}}^{9}$ \colorbox{blue!30}{\footnotesize No}$_{11_{\textsc{name}}}^{10}$ \footnotesize Va$_{12_{\textsc{nmod}}}^{11}$ \footnotesize jursidictions$_{13_{\textsc{sbj}}}^{12 \colorbox{red!30}{\tiny u}}$ \colorbox{blue!30}{\footnotesize are}$_{5_{\textsc{prd}}}^{13}$ \colorbox{blue!30}{\footnotesize doing}$_{13_{\textsc{vc}}}^{14}$ \footnotesize -$_{14_{\textsc{p}}}^{15}$ \colorbox{blue!30}{\footnotesize favoring}$_{14_{\textsc{adv}}}^{16}$ \footnotesize non-residential$_{18_{\textsc{nmod}}}^{17}$ \colorbox{blue!30}{\footnotesize development}$_{16_{\textsc{obj}}}^{18}$ \colorbox{blue!30}{\footnotesize and}$_{16_{\textsc{coord}}}^{19 \colorbox{red!30}{\tiny c}}$ \footnotesize letting$_{19_{\textsc{conj}}}^{20}$ \footnotesize other$_{22_{\textsc{nmod}}}^{21}$ \colorbox{blue!30}{\footnotesize jursidictions}$_{20_{\textsc{obj}}}^{22 \colorbox{red!30}{\tiny u}}$ \colorbox{green!30}{\footnotesize handle}$_{20_{\textsc{oprd}}}^{23}$ \footnotesize the$_{25_{\textsc{nmod}}}^{24}$ \footnotesize residential$_{23_{\textsc{obj}}}^{25}$ \footnotesize ?$_{5_{\textsc{p}}}^{26}$\\\hline

\colorbox{blue!30}{\footnotesize Her}$_{4_{\textsc{nmod}}}^{1}$ \footnotesize ``$_{4_{\textsc{p}}}^{2}$ \footnotesize Rubble$_{4_{\textsc{name}}}^{3}$ \colorbox{green!30}{\footnotesize Division}$_{6_{\textsc{sbj}}}^{4}$ \footnotesize ''$_{4_{\textsc{p}}}^{5}$ \colorbox{blue!30}{\footnotesize mixes}$_{0_{\textsc{root}}}^{6}$ \footnotesize such$_{9_{\textsc{nmod}}}^{7}$ \footnotesize disparate$_{9_{\textsc{nmod}}}^{8}$ \colorbox{blue!30}{\footnotesize materials}$_{6_{\textsc{obj}}}^{9}$ \footnotesize as$_{9_{\textsc{nmod}}}^{10 \colorbox{red!30}{\tiny p}}$ \footnotesize ink$_{13_{\textsc{nmod}}}^{11}$ \footnotesize -$_{13_{\textsc{nmod}}}^{12}$ \footnotesize jet$_{14_{\textsc{nmod}}}^{13}$ \colorbox{blue!30}{\footnotesize prints}$_{10_{\textsc{pmod}}}^{14}$ \colorbox{blue!30}{\footnotesize pasted}$_{14_{\textsc{appo}}}^{15 \colorbox{red!30}{\tiny u}}$ \colorbox{blue!30}{\footnotesize on}$_{15_{\textsc{loc}}}^{16 \colorbox{red!30}{\tiny p}}$ \footnotesize board$_{16_{\textsc{pmod}}}^{17}$ \footnotesize ,$_{14_{\textsc{p}}}^{18}$ \footnotesize foam$_{20_{\textsc{nmod}}}^{19}$ \footnotesize rubber$_{14_{\textsc{coord}}}^{20}$ \footnotesize ,$_{20_{\textsc{p}}}^{21}$ \footnotesize galvanized$_{23_{\textsc{nmod}}}^{22}$ \footnotesize steel$_{20_{\textsc{coord}}}^{23}$ \footnotesize ,$_{23_{\textsc{p}}}^{24}$ \footnotesize concrete$_{23_{\textsc{coord}}}^{25}$ \footnotesize ,$_{25_{\textsc{p}}}^{26}$ \footnotesize steel$_{28_{\textsc{nmod}}}^{27}$ \footnotesize rebar$_{25_{\textsc{coord}}}^{28 \colorbox{red!30}{\tiny u}}$ \footnotesize and$_{28_{\textsc{coord}}}^{29 \colorbox{red!30}{\tiny c}}$ \footnotesize bungee$_{31_{\textsc{nmod}}}^{30 \colorbox{red!30}{\tiny u}}$ \footnotesize cords$_{29_{\textsc{conj}}}^{31 \colorbox{red!30}{\tiny u}}$ \colorbox{blue!30}{\footnotesize .}$_{6_{\textsc{p}}}^{32}$\\\hline

\footnotesize they$_{2_{\textsc{sbj}}}^{1}$ \footnotesize were$_{0_{\textsc{root}}}^{2}$ \colorbox{green!30}{\footnotesize convinced}$_{2_{\textsc{prd}}}^{3}$ \colorbox{green!30}{\footnotesize that}$_{3_{\textsc{amod}}}^{4 \colorbox{red!30}{\tiny p}}$ \colorbox{blue!30}{\footnotesize if}$_{20_{\textsc{adv}}}^{5 \colorbox{red!30}{\tiny p}}$ \footnotesize only$_{8_{\textsc{adv}}}^{6}$ \footnotesize they$_{8_{\textsc{sbj}}}^{7}$ \colorbox{blue!30}{\footnotesize could}$_{5_{\textsc{sub}}}^{8}$ \footnotesize speak$_{8_{\textsc{vc}}}^{9}$ \footnotesize to$_{9_{\textsc{adv}}}^{10 \colorbox{red!30}{\tiny p}}$ \footnotesize an$_{12_{\textsc{nmod}}}^{11}$ \footnotesize American$_{10_{\textsc{pmod}}}^{12}$ \colorbox{blue!30}{\footnotesize ,}$_{20_{\textsc{p}}}^{13}$ \footnotesize Abather$_{19_{\textsc{nmod}}}^{14 \colorbox{red!30}{\tiny u}}$ \footnotesize 's$_{14_{\textsc{suffix}}}^{15}$ \footnotesize charred$_{19_{\textsc{nmod}}}^{16}$ \footnotesize and$_{16_{\textsc{coord}}}^{17 \colorbox{red!30}{\tiny c}}$ \footnotesize mangled$_{17_{\textsc{conj}}}^{18 \colorbox{red!30}{\tiny u}}$ \footnotesize flesh$_{20_{\textsc{sbj}}}^{19}$ \colorbox{blue!30}{\footnotesize would}$_{4_{\textsc{sub}}}^{20}$ \footnotesize make$_{20_{\textsc{vc}}}^{21}$ \footnotesize their$_{23_{\textsc{nmod}}}^{22}$ \footnotesize case$_{21_{\textsc{obj}}}^{23}$ \colorbox{blue!30}{\footnotesize ,}$_{2_{\textsc{p}}}^{24}$ \colorbox{blue!30}{\footnotesize but}$_{2_{\textsc{coord}}}^{25 \colorbox{red!30}{\tiny c}}$ \footnotesize they$_{27_{\textsc{sbj}}}^{26}$ \footnotesize had$_{25_{\textsc{conj}}}^{27}$ \footnotesize never$_{27_{\textsc{tmp}}}^{28}$ \footnotesize gotten$_{27_{\textsc{vc}}}^{29}$ \footnotesize past$_{29_{\textsc{adv}}}^{30 \colorbox{red!30}{\tiny p}}$ \footnotesize the$_{34_{\textsc{nmod}}}^{31}$ \footnotesize Jordanian$_{34_{\textsc{nmod}}}^{32 \colorbox{red!30}{\tiny u}}$ \footnotesize security$_{34_{\textsc{nmod}}}^{33}$ \footnotesize guards$_{30_{\textsc{pmod}}}^{34}$ \footnotesize .$_{2_{\textsc{p}}}^{35}$\\\hline

\colorbox{blue!30}{\footnotesize and}$_{3_{\textsc{dep}}}^{1 \colorbox{red!30}{\tiny c}}$ \footnotesize i$_{3_{\textsc{sbj}}}^{2}$ \colorbox{blue!30}{\footnotesize promise}$_{0_{\textsc{root}}}^{3}$ \colorbox{green!30}{\footnotesize to}$_{3_{\textsc{oprd}}}^{4}$ \footnotesize fess$_{4_{\textsc{im}}}^{5 \colorbox{red!30}{\tiny u}}$ \footnotesize up$_{5_{\textsc{prt}}}^{6}$ \footnotesize eventually$_{5_{\textsc{tmp}}}^{7}$ \footnotesize and$_{5_{\textsc{coord}}}^{8 \colorbox{red!30}{\tiny c}}$ \footnotesize tell$_{8_{\textsc{conj}}}^{9}$ \footnotesize of$_{9_{\textsc{adv}}}^{10 \colorbox{red!30}{\tiny p}}$ \footnotesize at$_{13_{\textsc{dep}}}^{11 \colorbox{red!30}{\tiny p}}$ \footnotesize least$_{11_{\textsc{amod}}}^{12}$ \colorbox{blue!30}{\footnotesize one}$_{15_{\textsc{nmod}}}^{13}$ \colorbox{blue!30}{\footnotesize such}$_{15_{\textsc{nmod}}}^{14}$ \colorbox{blue!30}{\footnotesize epic}$_{10_{\textsc{pmod}}}^{15}$ \colorbox{blue!30}{\footnotesize i}$_{17_{\textsc{sbj}}}^{16}$ \colorbox{blue!30}{\footnotesize survived}$_{15_{\textsc{nmod}}}^{17}$ \colorbox{blue!30}{\footnotesize --}$_{3_{\textsc{p}}}^{18}$\\\hline

\colorbox{yellow!30}{\footnotesize -}$_{3_{\textsc{p}}}^{1}$ \footnotesize Dr.$_{3_{\textsc{title}}}^{2}$ \colorbox{blue!30}{\footnotesize Seuss}$_{0_{\textsc{root}}}^{3 \colorbox{red!30}{\tiny u}}$ \colorbox{blue!30}{\footnotesize ,}$_{3_{\textsc{p}}}^{4}$ \footnotesize ``$_{3_{\textsc{p}}}^{5}$ \footnotesize One$_{7_{\textsc{nmod}}}^{6}$ \colorbox{blue!30}{\footnotesize Fish}$_{3_{\textsc{coord}}}^{7}$ \footnotesize ,$_{7_{\textsc{p}}}^{8}$ \colorbox{green!30}{\footnotesize Two}$_{10_{\textsc{nmod}}}^{9}$ \colorbox{green!30}{\footnotesize Fish}$_{7_{\textsc{coord}}}^{10}$ \footnotesize ,$_{10_{\textsc{p}}}^{11}$ \footnotesize Red$_{13_{\textsc{nmod}}}^{12}$ \colorbox{blue!30}{\footnotesize Fish}$_{10_{\textsc{coord}}}^{13}$ \colorbox{blue!30}{\footnotesize ,}$_{13_{\textsc{p}}}^{14}$ \footnotesize Blue$_{16_{\textsc{nmod}}}^{15}$ \colorbox{blue!30}{\footnotesize Fish}$_{13_{\textsc{coord}}}^{16}$ \colorbox{blue!30}{\footnotesize ''}$_{3_{\textsc{p}}}^{17}$\\\hline

\colorbox{blue!30}{\footnotesize or}$_{19_{\textsc{dep}}}^{1 \colorbox{red!30}{\tiny c}}$ \colorbox{blue!30}{\footnotesize ,}$_{19_{\textsc{p}}}^{2}$ \colorbox{blue!30}{\footnotesize as}$_{19_{\textsc{adv}}}^{3 \colorbox{red!30}{\tiny p}}$ \colorbox{yellow!30}{\footnotesize warren}$_{5_{\textsc{name}}}^{4}$ \colorbox{blue!30}{\footnotesize harding}$_{7_{\textsc{sbj}}}^{5 \colorbox{red!30}{\tiny u}}$ \colorbox{blue!30}{\footnotesize once}$_{7_{\textsc{tmp}}}^{6}$ \colorbox{blue!30}{\footnotesize said}$_{3_{\textsc{sub}}}^{7}$ \footnotesize :$_{19_{\textsc{p}}}^{8}$ \footnotesize ``$_{19_{\textsc{p}}}^{9}$ \footnotesize At$_{19_{\textsc{loc}}}^{10 \colorbox{red!30}{\tiny p}}$ \footnotesize either$_{12_{\textsc{nmod}}}^{11}$ \footnotesize end$_{10_{\textsc{pmod}}}^{12}$ \footnotesize of$_{12_{\textsc{nmod}}}^{13 \colorbox{red!30}{\tiny p}}$ \footnotesize the$_{16_{\textsc{nmod}}}^{14}$ \footnotesize social$_{16_{\textsc{nmod}}}^{15}$ \footnotesize spectrum$_{13_{\textsc{pmod}}}^{16}$ \footnotesize ,$_{19_{\textsc{p}}}^{17}$ \footnotesize there$_{19_{\textsc{loc}}}^{18}$ \colorbox{blue!30}{\footnotesize lies}$_{0_{\textsc{root}}}^{19}$ \footnotesize a$_{22_{\textsc{nmod}}}^{20}$ \footnotesize leisure$_{22_{\textsc{nmod}}}^{21}$ \footnotesize class$_{19_{\textsc{sbj}}}^{22}$ \colorbox{blue!30}{\footnotesize .}$_{19_{\textsc{p}}}^{23}$ \colorbox{blue!30}{\footnotesize ''}$_{19_{\textsc{p}}}^{24}$\\\hline

\colorbox{blue!30}{\footnotesize when}$_{12_{\textsc{tmp}}}^{1}$ \footnotesize the$_{3_{\textsc{nmod}}}^{2}$ \colorbox{blue!30}{\footnotesize guy}$_{12_{\textsc{sbj}}}^{3}$ \colorbox{green!30}{\footnotesize (}$_{9_{\textsc{p}}}^{4}$ \footnotesize the$_{6_{\textsc{nmod}}}^{5}$ \colorbox{blue!30}{\footnotesize owner}$_{9_{\textsc{dep}}}^{6}$ \footnotesize ,$_{9_{\textsc{p}}}^{7}$ \footnotesize it$_{9_{\textsc{sbj}}}^{8}$ \colorbox{blue!30}{\footnotesize turned}$_{3_{\textsc{prn}}}^{9}$ \footnotesize out$_{9_{\textsc{prt}}}^{10}$ \footnotesize )$_{9_{\textsc{p}}}^{11}$ \colorbox{blue!30}{\footnotesize arrived}$_{20_{\textsc{tmp}}}^{12}$ \colorbox{yellow!30}{\footnotesize to}$_{12_{\textsc{prp}}}^{13}$ \footnotesize open$_{13_{\textsc{im}}}^{14}$ \footnotesize the$_{17_{\textsc{nmod}}}^{15}$ \footnotesize gas$_{17_{\textsc{nmod}}}^{16}$ \footnotesize station$_{14_{\textsc{obj}}}^{17}$ \colorbox{blue!30}{\footnotesize ,}$_{20_{\textsc{p}}}^{18}$ \footnotesize he$_{20_{\textsc{sbj}}}^{19}$ \colorbox{blue!30}{\footnotesize took}$_{0_{\textsc{root}}}^{20}$ \footnotesize one$_{22_{\textsc{nmod}}}^{21}$ \footnotesize look$_{20_{\textsc{obj}}}^{22}$ \colorbox{blue!30}{\footnotesize at}$_{20_{\textsc{adv}}}^{23 \colorbox{red!30}{\tiny p}}$ \footnotesize our$_{26_{\textsc{nmod}}}^{24}$ \footnotesize cow$_{26_{\textsc{nmod}}}^{25 \colorbox{red!30}{\tiny u}}$ \footnotesize pie$_{23_{\textsc{pmod}}}^{26}$ \footnotesize with$_{26_{\textsc{nmod}}}^{27 \colorbox{red!30}{\tiny p}}$ \footnotesize wheels$_{27_{\textsc{pmod}}}^{28}$ \footnotesize and$_{20_{\textsc{coord}}}^{29 \colorbox{red!30}{\tiny c}}$ \footnotesize said$_{29_{\textsc{conj}}}^{30}$ \footnotesize ``$_{30_{\textsc{p}}}^{31}$ \colorbox{green!30}{\footnotesize what}$_{34_{\textsc{nmod}}}^{32}$ \footnotesize the$_{34_{\textsc{nmod}}}^{33}$ \colorbox{green!30}{\footnotesize fook}$_{30_{\textsc{obj}}}^{34 \colorbox{red!30}{\tiny u}}$ \colorbox{blue!30}{\footnotesize ?}$_{20_{\textsc{p}}}^{35}$ \colorbox{blue!30}{\footnotesize ''}$_{20_{\textsc{p}}}^{36}$\\\hline

\colorbox{blue!30}{\footnotesize SO}$_{14_{\textsc{adv}}}^{1}$ \colorbox{blue!30}{\footnotesize ,}$_{14_{\textsc{p}}}^{2}$ \colorbox{blue!30}{\footnotesize IF}$_{14_{\textsc{adv}}}^{3 \colorbox{red!30}{\tiny p}}$ \footnotesize YOU$_{5_{\textsc{sbj}}}^{4}$ \colorbox{blue!30}{\footnotesize WANT}$_{3_{\textsc{sub}}}^{5}$ \footnotesize A$_{7_{\textsc{nmod}}}^{6}$ \footnotesize BURGER$_{5_{\textsc{obj}}}^{7}$ \footnotesize AND$_{7_{\textsc{coord}}}^{8 \colorbox{red!30}{\tiny c}}$ \footnotesize FRIES$_{8_{\textsc{conj}}}^{9 \colorbox{red!30}{\tiny u}}$ \colorbox{blue!30}{\footnotesize ,}$_{14_{\textsc{p}}}^{10}$ \colorbox{yellow!30}{\footnotesize WELL}$_{14_{\textsc{dep}}}^{11}$ \colorbox{blue!30}{\footnotesize ,}$_{14_{\textsc{p}}}^{12}$ \colorbox{blue!30}{\footnotesize IT}$_{14_{\textsc{sbj}}}^{13}$ \colorbox{blue!30}{\footnotesize IS}$_{0_{\textsc{root}}}^{14}$ \colorbox{blue!30}{\footnotesize OK}$_{14_{\textsc{prd}}}^{15}$ \colorbox{blue!30}{\footnotesize .}$_{14_{\textsc{p}}}^{16}$\\\hline

\hline
\end{tabular}
\end{center}
\caption[The example sentences that have been improved by the tri-training approach when compared to the baseline.]{\label{table:analysis-examples-tri-training} The example sentences that have been improved by the tri-training approach when compared to the baseline. In which the dependency head/relation of a token are marked as the subscript, while the superscript is the index of token. The unknown words, prepositions and conjunctions are highlighted with \colorbox{red!30}{u}, \colorbox{red!30}{p} and \colorbox{red!30}{c} respectively. We highlight the different levels of the improvements achieved by our tri-training model on the dependency edges by different colours. In which the \colorbox{blue!30}{blue} colour means both head and label are corrected, the \colorbox{yellow!30}{yellow} colour means only the head is corrected and the \colorbox{green!30}{green} colour means only the label is corrected.}
\end{table}

\textbf{Conjunctions.} The annotation of conjunctions is another well-known problem for parsing. More conjunction usually results in a longer sentence and are more complex as well. Figure \ref{figure:analysis-sentcc-tri-training} shows the analysis on conjunctions. The figure is similar to that of prepositions, the tri-training model gained more than 11\% for all the classes and have higher gains for sentences containing conjunctions.

\textbf{Example Sentences.} Table \ref{table:analysis-examples-tri-training} shows some example sentences that have been improved largely by our tri-training approach. 

\section{Chapter Summary}\label{section:co-conclusion}
In this chapter we present our evaluations on two co-training approaches (co-training and tri-training). The main contribution of our evaluation on co-training is to assess the suitability of using the off-the-shelf parsers to form co-training.  We first evaluated on the normal agreement based co-training with four off-the-shelf parsers. Three of them are paired with the Mate parser to generate additional training data for retraining the Mate parser. We evaluated the parser pairs by adding different number of sentences into the training data. We also evaluated the pairs with additional training data that excluded the short annotations. The results show co-training is able to improve largely on target domain and additional gains are achieved when excluding the short sentences. We then evaluated the second approach (tri-training) that retrains the Mate parser on additional training data annotated identically by MST-Malt parsers. Benefit from the novel annotations that not predicted by the Mate parser,  tri-training outperforms our best co-training setting. The further evaluation on tri-training shows large improvements on all four test domains. The method achieved the largest improvement of 1.8\% and 0.6\% for labelled and unlabelled accuracies. We then applied both token level and sentence level analysis to find out where the improvement comes from. The analysis suggests tri-training gained particularly large improvement on label OBJ (objects) and PRD (predicative complement). The analysis of unknown words on both token level and sentence level shows only a slightly larger improvement on unknown words when compared with known words. The analysis on sentence length suggests tri-training helped mainly on sentences with a length between 15 and 30 tokens. The analysis on prepositions and conjunctions shows larger gains are achieved on sentences containing prepositions or conjunctions. Overall we demonstrated that co-/tri-training are powerful techniques for out-of-domain parsing when the off-the-shelf parsers are used.

\chapter{Self-training}\label{chapter:selftrain}
In this chapter, we introduce our self-training approach for English out-of-domain text. Self-training is one of the semi-supervised techniques that improves the learner's performance by its own annotations. Taking parsing as an example, a basic self-training iteration usually consists of three steps: firstly a base model is trained on the original manually annotated training data, then the base model is used to annotate unlabelled sentences (usually much larger than the original training set), finally the parser is retrained on the new training set, which consists of both manually and automatically annotated data. The self-training iteration can also be repeated to conduct a multi-iteration approach. Self-training has been adapted first to constituency parsers and achieved reasonably good gains for both in- and out-of-domain parsing \cite{mcclosky2006reranking,mcclosky06naacl,reichart2007self,sagae2010self,petrov2012overview}. While self-training approaches for dependency parsing are less successful, the evaluations usually found no impact or even negative effects on accuracy \cite{plank2011effective,plank2013experiments,cerisara2014spmrl,bjorkelund2014spmrl}. There are only a few successful self-training approaches reported on the dependency parsing, but those approaches are usually more complex than the basic self-training iterations.  \newcite{kawahara2008learning}'s approach needs a separately trained classifier to select additional training data,  \newcite{chen2008learning} used only partial parse trees and \newcite{goutam2011exploring}'s approach conditions on a small initial training set.     

In this work, we introduce a novel confidence-based self-training approach to out-of-domain dependency parsing. Our approach uses confidence-based methods to select training sentences for self-training. The confidence scores are generated during the parsing thus we do not need to train a separate classifier. Our self-training approach employs a single basic self-training iteration, except for the second step we add only sentences that have higher confidence scores to the training set. Overall, we present a simple but effective confidence-based self-training approach for English out-of-domain dependency parsing. We compare two confidence-based methods to select training data for our self-training. We evaluate our approaches on the main evaluation corpora as well as the \textsc{Chemical} domain text from the domain adaptation track of CoNLL 2007 shared task. 

The remaining parts of this chapter are organised as follows. Section \ref{section:self-en-appraoch} shows the detail of our self-training approaches. Section \ref{section:self-en-setup} introduces the experiment set-up of our evaluation. We then discuss and analyse the results in Section \ref{section:self-en-results} and \ref{section:self-en-analysis} respectively. The last section (Section \ref{section:self-en-conclusion}) summarises the chapter.

\section{Confidence-based Self-training}\label{section:self-en-appraoch}
The confidence-based self-training approach is inspired by the successful use of the high-quality dependency trees in our agreement based co-training and the correlation between the prediction quality and the confidence-based methods \cite{dredze2008confidence,crammer2009adaptive,mejer2012}. The confidence-besed methods were previously used by \newcite{mejer2012} to assess the parsing quality of a graph-based parser, but they haven't been used in self-training or transition-based parser before this work.  Based on our experience on co-training and the results of the previous work on self-training, we believe the selection of high-quality dependency trees is a crucial precondition for the successful application of self-training to dependency parsing. Therefore, we explore two confidence-based methods to select such dependency trees from newly parsed sentences. More precisely, our self-training approach consists of the following steps:

\begin{enumerate}
	\item A parser is trained on the source domain training set in order to generate a base model.
	\item A large number of unlabelled sentences from a target domain is annotated by the base model.
	\item The newly parsed sentences which have a high confidence score are added to the source domain training set as additional training data.
	\item The parser is then retrained on the new training set in order to produce a self-trained model.
	\item Finally, we evaluate the target domain data by the self-trained model.
\end{enumerate}

We test two methods to gain confidence scores for a dependency tree. 
The first method uses the parse scores, which is based on the observation that a higher parse score is correlated with a higher parsing quality.   
The second method uses the method of \newcite{mejer2012} to compute the Delta score.
\newcite{mejer2012} compute a confidence score for each edge. The algorithm attaches each edge to an alternative head.
The Delta is the score difference between the original dependency tree and the tree with the changed edge. 
This method provides a per-edge confidence score. 
Note that the scores are real numbers and might be greater than 1. 
We changed the Delta-approach in two aspects from that of \newcite{mejer2012}. 
We request that the new parse tree contains a node that has either a different head or might have a different edge label or both, since we use labelled dependency trees in contrast to \newcite{mejer2012}. To obtain a single score for a tree, we use the averaged score of scores computed for the individual edge by the Delta function.

% for our approach
\begin{figure}[t]
	\begin{center}
		\begin{tikzpicture}
		\pgfplotsset{
			xmin=0,xmax=100,
			ymin=75,ymax=95,
			xlabel=Percentage of Sentences,
			ylabel=Labeled Attachment Score (\%)
		}
		\begin{axis}

		%parse score
		\addplot[smooth,mark=triangle*,mark options={fill=white}]
		coordinates {(10,81.0) (20,80.1) (30,80.1) (40,79.6) (50,79.4)(60,79.0)(70,78.6)(80,78.1)(90,77.7)(100,77.5)};
		\addlegendentry{\tiny Parse Score}
		
		%adjusted parse score
		\addplot[smooth,mark=*, mark options={fill=white}] 
		coordinates {(10,88.3) (20,85.9) (30,85.7) (40,84.5) (50,83.6)(60,82.2)(70,81.3)(80,80.1)(90,79.0)(100,77.5)};
		\addlegendentry{\tiny Adjusted Parse Score}
		
		%delta
		\addplot[smooth,mark=square*, mark options={fill=white}]
		coordinates{(10,91.2) (20,88.0) (30,86.7) (40,86.0) (50,84.3)(60,83.1)(70,82.2)(80,81.1)(90,79.9)(100,77.5)};
		\addlegendentry{\tiny Delta}

		%random
		\addplot[smooth,dashed] 
		coordinates{(0,77.5) (100,77.5)};
		\addlegendentry{\tiny Average Accuracy of Entire Set}
		
		\end{axis}
		
		\end{tikzpicture}
	\end{center}
	\caption{\label{figure:self-en-accuracy-assess} 
	The accuracies when inspecting 10-100\% sentences of the \textsc{Weblogs} development set ranked by the confidence-based methods.}
\end{figure}
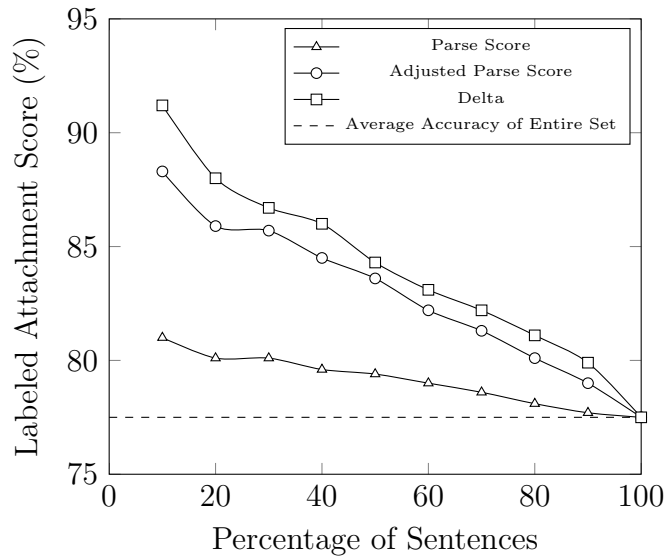

\begin{figure}[t]
\begin{center}
\begin{tikzpicture}
\begin{axis}[
    xlabel={\small Sentence Length},
    ylabel={\small Parse Score},
    zlabel={\small Labelled Attachment Score(\%)}
]

%history
\addplot3 [scatter, only marks] file{self-training-3d.txt};

\end{axis}
\end{tikzpicture}
\end{center}
\caption{\label{figure:self-en-accuracy-lenght-parse-socres} The accuracies, sentence lengths and the parse scores of individual sentences in \textsc{Weblogs} development set.}
\end{figure}
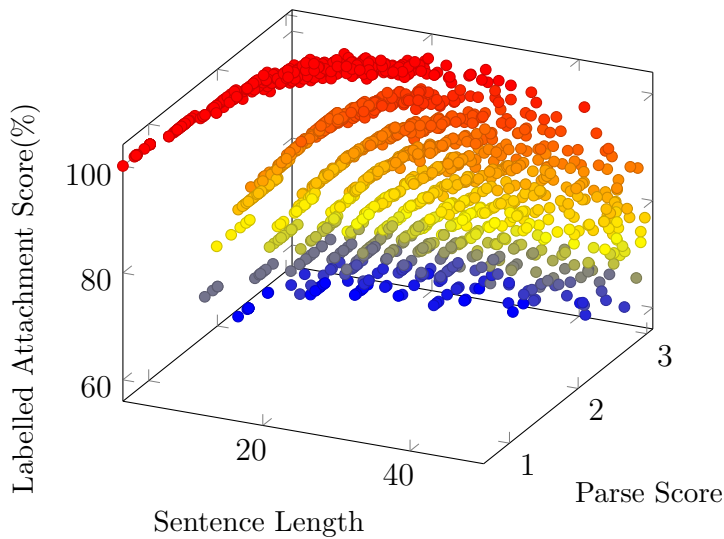

We use our main evaluation parser (Mate parser \cite{bohnet2013joint}) to implement our self-training approach. Mate is an arc-standard transition-based parser which employs beam search and a graph-based rescoring model. This parser computes a score for each dependency tree by summing up the scores for each transition and dividing the score by the total number of transitions. Due to the swap-operation (used for non-projective parsing), the number of transitions can vary, cf. \cite{kahane98acl,nivre07naacl}. 

Our second confidence-based method requires the computation of the score differences between the best tree and alternative trees. 
To compute the smallest difference (Delta), we modified the parser to derive the highest scoring alternative parse tree that replaces a given edge with an alternative one. This means either that the dependent is attached to another node or the edge label is changed, or both the dependent
is attached to another node and the edge is relabelled. More precisely, during the parsing for alternative trees, beam candidates that contain the specified labelled edge will be removed from the beam at the end of each transition.  Let $Score_{best}$ be the score of the best tree, $Score_{i}$ be the score of the alternative tree for the $i_{th}$ labelled edge and $L$ be the length of the sentence, the Delta ($Score_{Delta}$) for a parse tree is then calculated as follows:

\begin{equation}
Score_{Delta} = \frac{\sum\limits_{i=1}^L |Score_{best}-Score_{i}|}{L}
\vspace{0.3cm}
\end{equation}

To obtain high-accuracy dependency trees is crucial for our self-training approaches, 
thus we first assess the performance of the confidence-based methods on the development set for selecting high-quality dependency trees. 
We rank the parsed sentences by their confidence scores in a descending order. 
Figure \ref{figure:self-en-accuracy-assess} shows the accuracy scores when selecting 10-100\% of sentences with an increment of 10\%. 
The Delta method shows the best performance for detecting high-quality parse trees. 
We observed that when inspecting 10\% of sentences, the accuracy score difference between the Delta method and the average score of the entire set is nearly 14\%. 
The method using the parse score does not show such a high accuracy difference. The accuracy of the 10\% top ranked sentences are lower. 

We observed that despite that the parse score is the averaged value of the transitions, long sentences generally exhibit a higher score. Thus, short sentences tend to be ranked at the bottom, regardless of the accuracy. 
To give a more clear view, we plot the relations between the sentence lengths, parse scores and the accuracies in figure \ref{figure:self-en-accuracy-lenght-parse-socres}. The sentences of the \textsc{Weblogs} development set are represented by dots in the figure based on their properties. 
To soften the sentences proportional to their length, we  penalise the original parser score according to the sentence length, i.e. longer sentences are penalised more. The penalisation is done assuming a subtractive relationship between the original score and the length of the sentences ($L$) weighted by a constant  ($d$) which we fit on the development set.
%To reduce the dependence of the score on the sentence length and to maximise the correlation of the score
%and the accuracy, we adjust the scores for each parse tree by subtracting from them a constant $d$ multiplied by the sentence length ($L$).
The new parse scores are calculated as follows:

\begin{equation}
Score_{adjusted} = Score_{original}-L \times d
\vspace{0.3cm}
\end{equation}

To obtain the constant $d$, we apply the defined equation to all sentences of the development set and rank the sentences according to their adjusted scores in a descending order. 
The value of $d$ is selected to minimise the root mean square-error ($f_{r}$) of the ranked sentences. Following \newcite{mejer2012} we compute the $f_{r}$ by: 

\begin{equation}
f_{r} = \sqrt{\sum_{i} n_i (c_i-a_i)^2/(\sum_{i} n_i)}
\vspace{0.3cm}
\end{equation}

\begin{figure}[t]
\begin{center}
	\begin{tikzpicture}
	\pgfplotsset{
		xmin=0,xmax=0.05,
		ymin=0,ymax=0.5,
		xlabel=Value of $d$,
		ylabel=Root Mean Square-Error ($f_{r}$)
	}
	\begin{axis}

	%adjusted parse score
	\addplot[smooth,mark=*, mark options={fill=white}] 
	coordinates {(0,0.150)(0.005,0.101)(0.01,0.064)(0.015,0.062)(0.02,0.097)(0.025,0.142)(0.03,0.193)(0.035,0.247)(0.04,0.297)(0.045,0.341)(0.05,0.381)};
	\addlegendentry{\tiny Adjusted Parse Score}

	%random
	\addplot[smooth,dashed] 
	coordinates{(0,0.382)(0.05,0.382)};
	\addlegendentry{\tiny Unranked}
	
	\end{axis}
	
	\end{tikzpicture}
\end{center}
\caption{\label{figure:self-en-dvalue} The root mean square-error ($f_{r}$) of \textsc{Weblogs} development set after ranked by adjusted parse scores with different values of $d$.}
\end{figure}
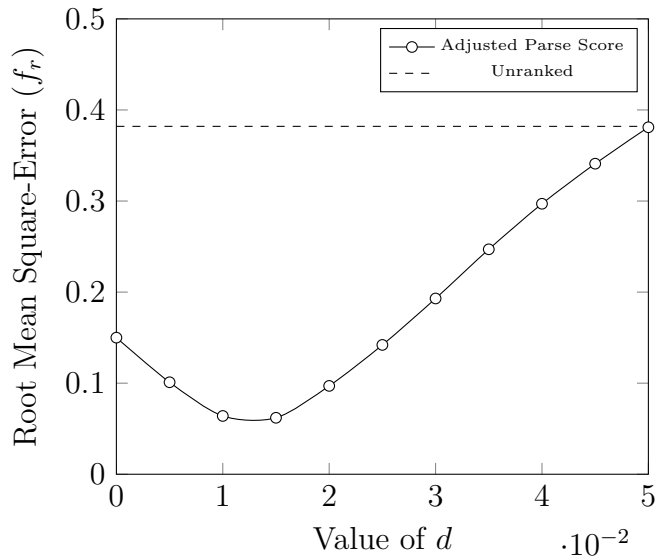

We use 100 bins to divide the accuracy into ranges of one percent. As the parse scores computed by the parser are generally in the range of [0,3], the parse scores in the range of $[\frac{(i-1)\times3}{100},\frac{i\times 3}{100}]$ are assigned to the $i_{th}$ 
bin. 
Let $n_i$ be the number of sentences in $i{th}$ bin, $c_i$ be the estimated accuracy of the bin calculated by $\frac{i-0.5}{100}$ and $a_i$ be the actual accuracy of the bin. We calculate $f_{r}$ by iterating stepwise over $d$ from 0 to 0.05 with an increment of 0.005. Figure \ref{figure:self-en-dvalue} shows the $f_{r}$ for the adjusted parse scores with different values of $d$. The lowest $f_{r}$ is achieved when $d = 0.015$, this reduces the $f_{r}$ from 0.15 to 0.06 when compared to the parse score method without adjustment ($d=0$). In contrast to the $f_{r}=0.06$ calculated when $d$ is set to 0.015, the unranked sentences have a $f_{r}$ of 0.38, which is six times larger than that of the adjusted one. The reduction on $f_{r}$ achieved by our adjustment indicates that the adjusted parse scores have a higher correlation to the accuracy when compared to the ones without the adjustment.

% for our approach

Figure \ref{figure:self-en-accuracy-assess} shows the performance of the adjusted parse scores for finding high accuracy parse trees in relation to the original parse score and the Delta-based method. The adjusted parse score-based method performs significantly better than that of the original score with a performance similar to the Delta method.
The method based on the parse scores is faster as we do not need to apply the parser to find alternatives for each edge of a dependency tree.

\section{Experiment Set-up}\label{section:self-en-setup}
For our evaluation on self-training, we used our main evaluation corpora and the \textsc{Chemical} domain text from the domain adaptation track of CoNLL 2007 shared task. We mainly evaluated on our main evaluation corpora and the best setting is also tuned on the development set of the main evaluation corpora. The \textsc{Chemical} domain evaluation is only used for comparison with previous work, we do not optimise our approaches specifically for this domain. 

For the main evaluation corpora, we used the \textsc{Conll} source domain training set, the \textsc{Weblogs} domain development set, the \textsc{Conll} source domain test set and \textsc{Weblogs, Newsgroups, Reviews} domain test sets. We do not evaluate our approach on the \textsc{Answers} domain as the unlabelled data for this domain is not large enough for our self-training.

The evaluation corpus for \textsc{Chemical} domain is taken from the domain adaptation track of the CoNLL 2007 shared task \cite{nivre07conll}. The shared task is the second year running for the dependency parsing task. Besides the multi-lingual parsing track introduced from the previous year, the 2007 shared task also included a track on domain adaptation task. The domain adaptation track provided mainly two domains (Biomedical and Chemical), in which the biomedical domain is used as development set and the chemical domain is used as evaluation set. The source domain training set consists of sections 2-11 of the Wall Street Journal section of the Penn Treebank \cite{marcus93}. A sufficient size of unlabelled data are also provided by the organiser, we used the first 256k sentences in our work. The labelled data are converted to dependency relations by the LTH constituent-to-dependency conversion tool \cite{johansson2007extended}. Table \ref{table:chem_datasets_stats} shows the basic statistics of the training, development and the test set. For the \textsc{Chemical} domain test we used only the data from the CoNLL 2007 shared task to make a fair comparison with \newcite{kawahara2008learning}'s results.

We use the Mate transition-based parser in our experiments. The parser is modified to output the confidence scores, other than that we used its default settings. For part-of-speech tagging, we use predicted tags from Mate's internal tagger for all the evaluated domains. For \textsc{Chemical} domain we evaluated additionally on gold tags as they are used by previous work. The baselines are trained only on the respective source domain training data. 

For the evaluation of the parser's accuracy, we report both labelled (LAS) and unlabelled (UAS) attachment scores, but mainly focus on the labelled version. We included all punctuation marks in the evaluation. The significance levels are marked according to the p-values, * and ** are used to represent the p-value of 0.05 and 0.01 levels respectively.

\begin{table}[t]
	\begin{center}
		\begin{tabular}{|l|r|r|r|}
			\cline{2-4}
			\multicolumn{1}{c|}{}& \multicolumn{1}{|c|}{\bf train} & \multicolumn{1}{|c|}{\bf \ \ test\ \ }& \multicolumn{1}{|c|}{\bf unlabelled}\\ \hline
			\bf Sentences  & 18,577 & 195  &256,000 \\
			\bf Tokens     & 446,573  &5,001 &6,848,072 \\
			\bf Avg. Length &24.04&25.65&26.75\\
			\hline
		\end{tabular}
		
	\end{center}
	\caption{\label{table:chem_datasets_stats} The size of datasets for \textsc{Chemical} domain evaluation.}
\end{table}

\section{Empirical Results}\label{section:self-en-results}

\begin{figure}[t]
	\begin{center}
		\begin{tikzpicture}
		\pgfplotsset{
			xmin=40,xmax=310,
			ymin=76.9,ymax=79.1,
			xlabel=Additional Training Data (K),
			ylabel=Labeled Attachment Score (\%)
		}
		\begin{axis}

		%Random
		\addplot[smooth,mark=triangle*,mark options={fill=white}]
		coordinates {(50,77.7) (100,77.78) (150,77.7) (200,77.68)};
		\addlegendentry{\tiny Random}

		%adjusted parse score
		\addplot[smooth,mark=*, mark options={fill=white}] 
		coordinates {(50,78.06) (100,78.09) (150,78.11) (200,78.13) (250,78.2) (300,77.89)};
		\addlegendentry{\tiny Adjusted Parse Score}
		
		%Delta
		\addplot[smooth,mark=square*, mark options={fill=white}]
		coordinates{(50,77.02) (100,77.33) (150,77.83) (200,78.14) (250,78.27) (300,78.13)};
		\addlegendentry{\tiny Delta}

		%baseline
		\addplot[smooth,dashed] 
		coordinates{(0,77.54) (350,77.54)};
		\addlegendentry{\tiny Baseline}
		
		\end{axis}
		
		\end{tikzpicture}
	\end{center}
	\caption{\label{figure:self-en-dev} The effect of our self-training approaches on the \textsc{Weblogs} development set. }
\end{figure}
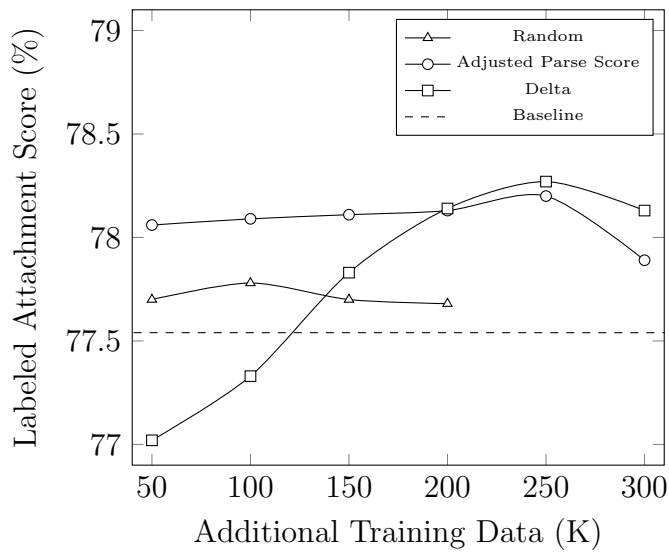

\textbf{Random Selection-based Self-training.} To have an idea of the performance of basic self-training, we first evaluated with randomly selected additional training data. The triangle marked curve in Figure \ref{figure:self-en-dev} shows the accuracy of the random selection-based self-training. We used from 50k to 200k randomly selected additional training data to retrain the Mate parser. The retrained models obtain some small improvements when compared with the baseline. The improvements achieved by the different number of additional training data are very similar: they all around 0.2\%. Those small improvements obtained by the basic self-training are not statistically significant. This finding is in line with previous work of applying non-confidence-based self-training approaches to dependency parsing, cf. \cite{cerisara2014spmrl,bjorkelund2014spmrl}.

%for evaluating on the dev set

\textbf{Parse Score-based Self-training.} We then evaluate with our first confidence-based method, that uses parse scores. As proposed the automatically annotated sentences are ranked in descending order by the adjusted parse scores before they are used as additional training data. As shown in Figure \ref{figure:self-en-dev}, we add between 50k to 300k top ranked sentences from the \textsc{Weblogs} auto-annotated dataset. The method achieved 0.52\% improvement when we use 50k additional training data and the improvement increased to 0.66\% when 250k sentences are used. After that, the improvement decreased.  We use an auto-labelled dataset of 500k sentences. After we rank the sentences by our confidence-based methods, the first half is expected to have an accuracy higher than the average, and the second half is expected to have one lower than average. Thus we should avoid using sentences from the second half of the ranked dataset.

\textbf{Delta-based self-training.} For our Delta-based approach, we select additional training data with the Delta method. We train the parser by adding between 50k to 300k sentences from the target domain. Same as the parse score-based method, we gain the largest improvement when 250k sentences are used, which improves the baseline by 0.73\% (cf. Figure \ref{figure:self-en-dev}). Although this improvement is slightly higher than that of the parse score-based method, the accuracies are lower than the baseline when we use 50k and 100k ranked sentences from Delta based method. Our error analysis shows that these parse trees are mainly short sentences consisting of only three words. These sentences contribute probably no additional information that the parser can exploit.

\textbf{Evaluating on test domains.} 
We adapt our best settings of 250k additional sentences for both approaches and apply them to three test sets (\textsc{Weblogs, Newsgroups} and \textsc{Reviews}). 
As illustrated in Table \ref{table:self-en-main-test}, nearly all the results produced by both approaches are statistically significant improvements when compared to the baselines. The only exception is the unlabelled improvement of the parse score approach on \textsc{Reviews} domain which has a p-value of 0.08. Both approaches achieved the largest improvements on \textsc{Weblogs} domain. The largest labelled improvement of 0.81\% is achieved by the parse score-based method, while the largest unlabelled improvement of 0.77\% is achieved by the Delta method. For \textsc{Newsgroups} domain both approaches gained the similar labelled and unlabelled improvements of 0.6\%. For \textsc{Reviews} domain the Delta method achieved 0.4 - 0.5\% improvements on labelled and unlabelled accuracies. The parse score-based approach achieved lower improvements of 0.3\%. In terms of the in-domain evaluation, the accuracies of both approaches are lower than the baseline.

\begin{table}[t]
	\begin{center}
		\begin{tabular}{|l|llll|ll|}
			\cline{2-7}
			\multicolumn{1}{c|}{}&\multicolumn{2}{|c}{\bf Parse Score}&\multicolumn{2}{c|}{\bf Delta}&\multicolumn{2}{|c|}{\bf Baseline}\\ \cline{2-7}
			\multicolumn{1}{c|}{}&\bf LAS&\bf UAS&\bf LAS&\bf UAS&\bf LAS&\bf UAS\\\hline
			\sc Weblogs&79.8**&85.82**&79.68**&85.88**&78.99&85.1\\
			\sc Newsgroups&75.88**&83.41*&75.87*&83.49**&75.3&82.88\\
			\sc Reviews&75.43*&82.99&75.6**&83.09*&75.07&82.68\\ \hline
			\sc Conll&89.4&91.88&89.67&92.13&90.07&92.4\\ \hline
		\end{tabular}
	\end{center}
	\caption{\label{table:self-en-main-test} The effect of the adjusted parse score-based and the Delta-based self-training approaches on our main test sets.}
\end{table} 

\begin{table}[t]
	\begin{center}
		\begin{tabular}{|l|ll|ll|}
			\cline{2-5}
			\multicolumn{1}{c|}{}&\multicolumn{2}{|c|}{\bf PPOS}&\multicolumn{2}{|c|}{\bf GPOS}\\\cline{2-5}
			\multicolumn{1}{c|}{} & \bf LAS&\bf UAS & \bf LAS&\bf UAS \\ \hline
			\bf Parse Score&80.8*&83.62*&83.44**&85.74**\\
			\bf Delta &81.1*&83.71*&83.58**&85.8**\\
			\bf Baseline &79.68&82.5&81.96&84.28\\ \hline
			\bf Kawahara (Self-trained)&\multicolumn{1}{|c}{-}&\multicolumn{1}{c|}{-}&\multicolumn{1}{|c}{-}&84.12\\
			\bf Kawahara (Baseline) &\multicolumn{1}{|c}{-}&\multicolumn{1}{c|}{-}&\multicolumn{1}{|c}{-}&83.58\\
			\bf Sagae (Co-training)&\multicolumn{1}{|c}{-}&\multicolumn{1}{c|}{-}&81.06&83.42\\
			\hline
		\end{tabular}
	\end{center}
	\caption[The results of the adjusted parse score-based and the Delta-based self-training approaches on the \textsc{Chemical} test set compared with previous work.]{\label{table:self-en-chem-test} The results of the adjusted parse score-based and the Delta-based self-training approaches on the \textsc{Chemical} test set compared with the best-reported self-training gain \cite{kawahara2008learning} and the best results of CoNLL 2007 shared task, cf. \newcite{sagae07}. (PPOS: results based on predicted tags, GPOS: results based on gold tags, Self-trained: results of self-training experiments, Co-trained: results of co-training experiments.)}
\end{table}

We further evaluate our best settings on \textsc{Chemical} texts provided by the CoNLL 2007 shared task. 
We adapt the best settings of the main evaluation corpora and apply both confidence-based approaches to the \textsc{Chemical} domain. For the constant $d$, we use 0.015 and we use 125k additional training data out of the 256k from the unlabelled data of the \textsc{Chemical} domain. 
We evaluate our confidence-based methods on both predicted and gold part-of-speech tags. After retraining, both confidence-based methods achieve
significant improvements in all experiments. 
Table \ref{table:self-en-chem-test} shows the results for the \textsc{Chemical} domain. 
When we use predicted part-of-speech tags, the Delta-based method gains a labelled improvement of 1.42\%, while the parse score-based approach gains 1.12\%. 
For the experiments based on gold tags, we achieved larger labelled improvements of 1.62\% for the Delta-based and 1.48\% for the parse score-based methods. For all experiments, the unlabelled improvements are similar to that of labelled ones.

Table \ref{table:self-en-chem-test} compares our results with that of \newcite{kawahara2008learning}. We added also the results of \newcite{sagae07} but those are not directly comparable since they were gained with co-training.
\newcite{sagae07} gained additional training data by parsing the unlabelled data with two parsers and then they select those sentences where the parsers agree on.

\newcite{kawahara2008learning} reported positive results for self-training.
They used a separately trained binary classifier to select additional training data and are evaluated only on gold tags. 
Our baseline is higher than \newcite{kawahara2008learning}'s self-training result. Starting from this strong baseline, we could improve by 1.62\% LAS 
and 1.52\% UAS which is an error reduction of 9.6\% on the UAS (cf. Table \ref{table:self-en-chem-test}). 
The largest improvement of 1.52\% compared to that of \newcite{kawahara2008learning} (0.54\% UAS) is substantially larger. 
We obtained the result by a simple method, and we do not need a separately trained classifier.

\section{Analysis}\label{section:self-en-analysis}
\begin{figure}[t]
	\begin{center}
		\begin{tikzpicture}
		\pgfplotsset{
			xmin=0,xmax=100,
			ymin=0,ymax=100,
			xlabel=Percentage of Sentences,
			ylabel=Identical Rate (\%)
		}
		\begin{axis}

		\addplot[smooth,mark=*, mark options={fill=white}] 
		coordinates {(10,4.65) (20,18.14) (30,33.02) (40,46.28) (50,56.19) (60,65.04) (70,75.42) (80,84.53) (90,91.63) };

		\end{axis}
		
		\end{tikzpicture}
	\end{center}
	\caption{\label{figure:analysis-agree-self-training} The identical rate between the adjusted parse score-based and the Delta-based methods, when top ranked $n$ percent is concerned.}
\end{figure}
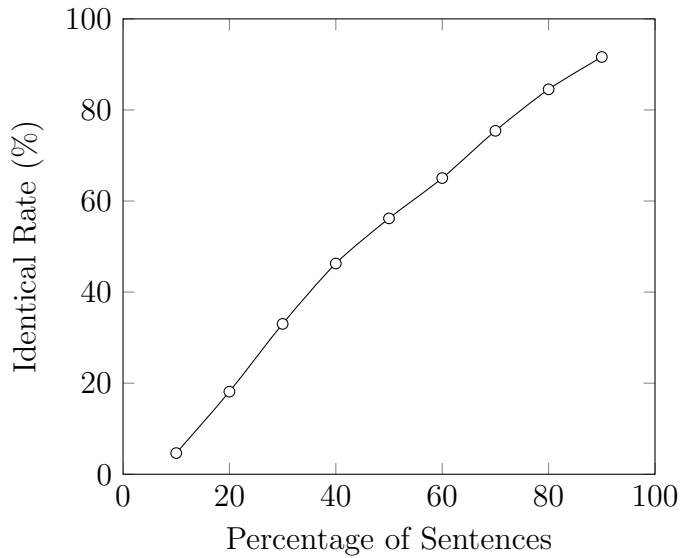

Our self-training approaches demonstrated their merit in the above experiments, two confidence-based methods work equally well on most of the domains. This suggests self-training can be used for out-of-domain dependency parsing when there is a reasonably good confidence-based method available. As two confidence-based methods showed similar performances on our tested domains, the first guess would be they might consist of a large portion of identical additional training data. We assess our assumption on the development set. We first rank the dataset by different methods. Let $Delta_n$ and $PS_n$ be the top ranked $n$\% sentences of the development set by their Delta and adjusted parse scores. The identical rate is defined as the percentage of sentences that are presented in both $Delta_n$ and $PS_n$. Figure \ref{figure:analysis-agree-self-training} shows the identical rate of our methods. The identical rates are lower than we expected, for top ranked 10\% sentences only 5\% of them are identical, and the identical rate is 56\% for the first half of the ranked list. As the additional training data from Delta and adjusted parse scores can consist of more than 40 percent different sentences, we suspect there might be some behaviour difference between two methods. In order to have a more clear picture about the behaviours of our confidence-based methods, we applied both token level and sentence level analysis to those methods. This allows us to have an in-depth comparison between our confidence-based methods. In the same way as we did in our analysis for co-training, we plot the accuracy changes of major syntactic labels and compute improvements different on unknown/known words in our token level analysis. For sentence level analysis, we evaluate all four factors on both confidence-based methods, cf. sentence length, the number of unknown words, the number of prepositions and the number of conjunctions. For our analysis, three target domain test sets are used as a single set.

\subsection{Token Level Analysis}

%label evaluation
\begin{figure}[t]
	\begin{center}
		\begin{tikzpicture}
		\begin{axis}[
		title=\tiny a) Parse Score Method,
		width=12cm,height=5cm,
		ymin=-1.4, ymax=4.7,
		ybar,ybar interval =0.7,
		ylabel=Accuracy Change (\%),
		enlargelimits=0.01,
		legend style={at={(0.5,-0.2)},anchor=north,legend columns=-1},
		symbolic x coords={NMOD,P,PMOD,SBJ,ROOT,OBJ,ADV,COORD,VC,CONJ,PRD,DEP,AMOD,TMP,IM},
		xtick=data,
		x tick label style={font=\tiny,rotate=45,anchor=east}
		]
		\addplot coordinates{(NMOD,0.3) (P,-0.5) (PMOD,0.4) (SBJ,0.4) (ROOT,1.0) (OBJ,1.1) (ADV,1.4) (COORD,0.7) (VC,0.9) (CONJ,0.2) (PRD,0.4) (DEP,-1.3) (AMOD,0.4) (TMP,-0.6) (IM,-0.3) }; 
		%\addlegendentry{\tiny Recall}
		
		\addplot coordinates{(NMOD,0.4) (P,0.4) (PMOD,0.6) (SBJ,0.6) (ROOT,1.0) (OBJ,0.5) (ADV,-0.2) (COORD,0.8) (VC,-0.1) (CONJ,-0.2) (PRD,0.0) (DEP,4.6) (AMOD,1.7) (TMP,1.2) (IM,0.0) }; 
		%\addlegendentry{\tiny Precision}
		
		\addplot coordinates{(NMOD,0.3) (P,-0.1) (PMOD,0.5) (SBJ,0.5) (ROOT,1.0) (OBJ,0.7) (ADV,0.6) (COORD,0.7) (VC,0.4) (CONJ,0.0) (PRD,0.2) (DEP,1.7) (AMOD,1.0) (TMP,0.4) (IM,-0.2) }; 
	%	\addlegendentry{\tiny F-score}
		
		\end{axis}
		%NMOD cnt:21231 R/P/F89.5/88.5/89.0 Base R/P/F89.2/88.1/88.7 diff: F:0.3 R:0.3 P:0.4
		%P cnt:10100 R/P/F92.3/99.1/95.7 Base R/P/F92.8/98.7/95.8 diff: F:-0.1 R:-0.5 P:0.4
		%PMOD cnt:7906 R/P/F90.3/91.1/90.7 Base R/P/F89.9/90.5/90.2 diff: F:0.5 R:0.4 P:0.6
		%SBJ cnt:7438 R/P/F92.5/88.7/90.6 Base R/P/F92.1/88.1/90.1 diff: F:0.5 R:0.4 P:0.6
		%ROOT cnt:5242 R/P/F87.0/87.0/87.0 Base R/P/F86.0/86.0/86.0 diff: F:1.0 R:1.0 P:1.0
		%OBJ cnt:5058 R/P/F88.8/71.1/79.9 Base R/P/F87.7/70.6/79.2 diff: F:0.7 R:1.1 P:0.5
		%ADV cnt:4808 R/P/F69.1/68.9/69.0 Base R/P/F67.7/69.1/68.4 diff: F:0.6 R:1.4 P:-0.2
		%COORD cnt:3857 R/P/F82.6/90.4/86.5 Base R/P/F81.9/89.6/85.8 diff: F:0.7 R:0.7 P:0.8
		%VC cnt:3182 R/P/F90.2/93.4/91.8 Base R/P/F89.3/93.5/91.4 diff: F:0.4 R:0.9 P:-0.1
		%CONJ cnt:2826 R/P/F86.3/86.8/86.6 Base R/P/F86.1/87.0/86.6 diff: F:0.0 R:0.2 P:-0.2
		%PRD cnt:2277 R/P/F43.6/90.4/67.0 Base R/P/F43.2/90.4/66.8 diff: F:0.2 R:0.4 P:0.0
		%DEP cnt:2173 R/P/F25.8/66.2/46.0 Base R/P/F27.1/61.6/44.3 diff: F:1.7 R:-1.3 P:4.6
		%AMOD cnt:2075 R/P/F57.1/74.6/65.8 Base R/P/F56.7/72.9/64.8 diff: F:1.0 R:0.4 P:1.7
		%TMP cnt:2032 R/P/F65.4/71.1/68.3 Base R/P/F66.0/69.9/67.9 diff: F:0.4 R:-0.6 P:1.2
		%IM cnt:1363 R/P/F97.7/97.3/97.5 Base R/P/F98.0/97.3/97.7 diff: F:-0.2 R:-0.3 P:0.0
		
		\end{tikzpicture}
		\begin{tikzpicture}
		\begin{axis}[
		title=\tiny b) Delta Based Method,
		width=12cm,height=5cm,
		ymin=-1.4, ymax=4.7,
		ybar,ybar interval =0.7,
		ylabel=Accuracy Change (\%),
		enlargelimits=0.01,
		legend style={at={(0.5,-0.2)},anchor=north,legend columns=-1},
		symbolic x coords={NMOD,P,PMOD,SBJ,ROOT,OBJ,ADV,COORD,VC,CONJ,PRD,DEP,AMOD,TMP,IM},
		xtick=data,
		x tick label style={font=\tiny,rotate=45,anchor=east}
		]
		\addplot coordinates{(NMOD,0.4) (P,-0.2) (PMOD,0.9) (SBJ,0.4) (ROOT,0.7) (OBJ,1.1) (ADV,1.0) (COORD,0.5) (VC,1.7) (CONJ,0.1) (PRD,0.6) (DEP,-0.9) (AMOD,0.4) (TMP,-1.2) (IM,-0.1) }; 
		\addlegendentry{\tiny Recall}
		
		\addplot coordinates{(NMOD,0.3) (P,0.3) (PMOD,0.7) (SBJ,0.2) (ROOT,0.7) (OBJ,-0.1) (ADV,-0.7) (COORD,1.3) (VC,0.0) (CONJ,0.0) (PRD,-0.7) (DEP,3.0) (AMOD,1.5) (TMP,1.2) (IM,0.2) }; 
		\addlegendentry{\tiny Precision}
		
		\addplot coordinates{(NMOD,0.3) (P,0.0) (PMOD,0.8) (SBJ,0.3) (ROOT,0.7) (OBJ,0.4) (ADV,0.1) (COORD,0.9) (VC,0.9) (CONJ,0.0) (PRD,0.0) (DEP,1.1) (AMOD,0.9) (TMP,0.1) (IM,0.0) }; 
		\addlegendentry{\tiny F-score}
		
		\end{axis}
		%NMOD cnt:21231 R/P/F89.6/88.4/89.0 Base R/P/F89.2/88.1/88.7 diff: F:0.3 R:0.4 P:0.3
		%P cnt:10100 R/P/F92.6/99.0/95.8 Base R/P/F92.8/98.7/95.8 diff: F:0.0 R:-0.2 P:0.3
		%PMOD cnt:7906 R/P/F90.8/91.2/91.0 Base R/P/F89.9/90.5/90.2 diff: F:0.8 R:0.9 P:0.7
		%SBJ cnt:7438 R/P/F92.5/88.3/90.4 Base R/P/F92.1/88.1/90.1 diff: F:0.3 R:0.4 P:0.2
		%ROOT cnt:5242 R/P/F86.7/86.7/86.7 Base R/P/F86.0/86.0/86.0 diff: F:0.7 R:0.7 P:0.7
		%OBJ cnt:5058 R/P/F88.8/70.5/79.6 Base R/P/F87.7/70.6/79.2 diff: F:0.4 R:1.1 P:-0.1
		%ADV cnt:4808 R/P/F68.7/68.4/68.5 Base R/P/F67.7/69.1/68.4 diff: F:0.1 R:1.0 P:-0.7
		%COORD cnt:3857 R/P/F82.4/90.9/86.7 Base R/P/F81.9/89.6/85.8 diff: F:0.9 R:0.5 P:1.3
		%VC cnt:3182 R/P/F91.0/93.5/92.3 Base R/P/F89.3/93.5/91.4 diff: F:0.9 R:1.7 P:0.0
		%CONJ cnt:2826 R/P/F86.2/87.0/86.6 Base R/P/F86.1/87.0/86.6 diff: F:0.0 R:0.1 P:0.0
		%PRD cnt:2277 R/P/F43.8/89.7/66.8 Base R/P/F43.2/90.4/66.8 diff: F:0.0 R:0.6 P:-0.7
		%DEP cnt:2173 R/P/F26.2/64.6/45.4 Base R/P/F27.1/61.6/44.3 diff: F:1.1 R:-0.9 P:3.0
		%AMOD cnt:2075 R/P/F57.1/74.4/65.7 Base R/P/F56.7/72.9/64.8 diff: F:0.9 R:0.4 P:1.5
		%TMP cnt:2032 R/P/F64.8/71.1/68.0 Base R/P/F66.0/69.9/67.9 diff: F:0.1 R:-1.2 P:1.2
		%IM cnt:1363 R/P/F97.9/97.5/97.7 Base R/P/F98.0/97.3/97.7 diff: F:0.0 R:-0.1 P:0.2
		
		\end{tikzpicture}
	\end{center}
	\caption{\label{figure:analysis-label-self-training} The performance comparison between the self-training approach and the baseline on major labels.}
\end{figure}

%Confution matric
\begin{table}
\begin{center}
\begin{tabular}{l|r|r|r}
\hline
\bf Confusion & \bf Baseline & \bf Parse Score& \bf Delta\\\hline
\footnotesize NMOD $\rightarrow$ \footnotesize ADV & 200 &192 & 226\\
\footnotesize NMOD $\rightarrow$ \footnotesize HYPH & 190 &190 & 193\\
\footnotesize NMOD $\rightarrow$ \footnotesize NAME & 530 &565 & 510\\
\footnotesize NMOD $\rightarrow$ \footnotesize PMOD & 151 &128 & 128\\
\footnotesize NMOD $\rightarrow$ \footnotesize HMOD & 206 &211 & 212\\
\footnotesize NMOD $\rightarrow$ \footnotesize OBJ,SBJ,LOC & 361 &333 & 337\\\hline
\footnotesize P $\rightarrow$ \footnotesize HYPH,NMOD & 245 &263 & 254\\\hline
\footnotesize SBJ $\rightarrow$ \footnotesize NMOD,OBJ & 238 &221 & 224\\\hline
\footnotesize OBJ $\rightarrow$ \footnotesize NMOD & 174 &150 & 170\\\hline
\footnotesize PMOD $\rightarrow$ \footnotesize NMOD & 228 &225 & 215\\
\footnotesize PMOD $\rightarrow$ \footnotesize OBJ & 104 &88 & 91\\\hline
\footnotesize ROOT $\rightarrow$ \footnotesize NMOD & 201 &192 & 198\\
\footnotesize ROOT $\rightarrow$ \footnotesize OBJ & 105 &88 & 101\\\hline
\footnotesize ADV $\rightarrow$ \footnotesize LOC & 195 &180 & 186\\
\footnotesize ADV $\rightarrow$ \footnotesize NMOD & 305 &285 & 300\\
\footnotesize ADV $\rightarrow$ \footnotesize MNR,AMOD,DIR,TMP & 457 &416 & 411\\\hline
\footnotesize COORD $\rightarrow$ \footnotesize NMOD & 125 &109 & 109\\\hline
\footnotesize VC $\rightarrow$ \footnotesize OPRD & 95 & 90& 75\\\hline
\footnotesize CONJ $\rightarrow$ \footnotesize NMOD & 101 &99 & 100\\\hline
\footnotesize DEP $\rightarrow$ \footnotesize ROOT & 152 &153 & 155\\
\footnotesize DEP $\rightarrow$ \footnotesize OBJ & 173 &161 & 166\\
\footnotesize DEP $\rightarrow$ \footnotesize SBJ & 305 &303 & 311\\
\footnotesize DEP $\rightarrow$ \footnotesize NMOD & 294 &322 & 302\\
\footnotesize DEP $\rightarrow$ \footnotesize ADV,TMP & 229 &241 & 234\\\hline
\footnotesize AMOD $\rightarrow$ \footnotesize NMOD & 208 &223 & 227\\
\footnotesize AMOD $\rightarrow$ \footnotesize ADV,HYPH & 236 &242 & 235\\\hline
\footnotesize TMP $\rightarrow$ \footnotesize ADV & 225 &250 & 259\\
\footnotesize TMP $\rightarrow$ \footnotesize NMOD & 112 &117 & 115\\\hline
\footnotesize PRD $\rightarrow$ \footnotesize OBJ & 700 &724 & 722\\
\footnotesize PRD $\rightarrow$ \footnotesize ADV,VC & 218 &220 & 213\\\hline
\end{tabular}
\end{center}
\caption{\label{table:analysis-confusion-self-training} The confusion matrix of dependency labels, compared between the self-training approaches and the baseline.}
\end{table}

\textbf{Individual Label Accuracy.} Figure \ref{figure:analysis-label-self-training} shows the comparison of accuracy changes between our adjusted parse score-based approach and the Delta-based approach. Two approaches show similar patterns on the individual labels, both of them show no effect on labels such as P (punctuations), CONJ (conjunct) and PRD (predicative complement). They both gained more than 0.5\% f-score on ROOT (root of the sentence), COORD (coordination), some modifiers (PMOD, AMOD) and unclassified relations (DEP). In addition to the common improvements between two methods, the Delta method also gains a 0.9\% improvement on VC (Verb chain), and the parse score method achieved 0.5\% improvement on SBJ (subject). Figure \ref{table:analysis-confusion-self-training} shows the confusion matrix of your self-training methods compared with the baseline. 

%corpus UNK
\begin{table}[t]
	\begin{center}
		\begin{tabular}{|l|r|rrrr|rr|}
			\cline{3-8}
			\multicolumn{2}{c|}{}& \multicolumn{2}{|c}{\bf Parse Score} &\multicolumn{2}{c|}{\bf Delta} &\multicolumn{2}{|c|}{\bf Baseline}\\
			\cline{2-8}
			\multicolumn{1}{c|}{}&\bf Tokens &\bf LAS&\bf UAS&\bf LAS&\bf UAS&\bf LAS&\bf UAS\\
			\hline
			\bf Known &84421&78.4&85.0&78.4&85.1&77.8&84.5\\
			%diff: LAS:0.6 UAS:0.5 LAS:0.6 UAS:0.6
			\bf Unknown &5049&62.4&73.5&62.5&73.8&61.6&72.5\\
			%diff: LAS:0.8 UAS:1.0 LAS:0.9 UAS:1.3
			\hline
			\bf All &89470&77.5&84.4&77.5&84.5&76.9&83.8\\
			%diff: LAS:0.6 UAS:0.6 LAS:0.6 UAS:0.7
			\hline
		\end{tabular}
	\end{center}
	\caption{\label{table:analysis-corpusunk-self-training} The accuracy comparison between the self-training approach and the baseline on unknown words.}
\end{table}

\textbf{Unknown Words Accuracy.} For unlabelled improvements, both methods showed a large gap between known words and unknown words. Improvements on unknown words are at least doubled in value when compared to that of known words. The improvement differences are smaller on the labelled accuracies. The value for unknown words is only 0.2\% higher than that of known words. This is an indication that self-training is able to improve unknown words attachment but still does not have sufficient information to make label decisions. The improvements of the entire set are same as that of known words and are not affected largely by the unknown words. This is due to the unknown words only occupying 5\% of the dataset.

\subsection{Sentence Level Analysis}

%number of tokens
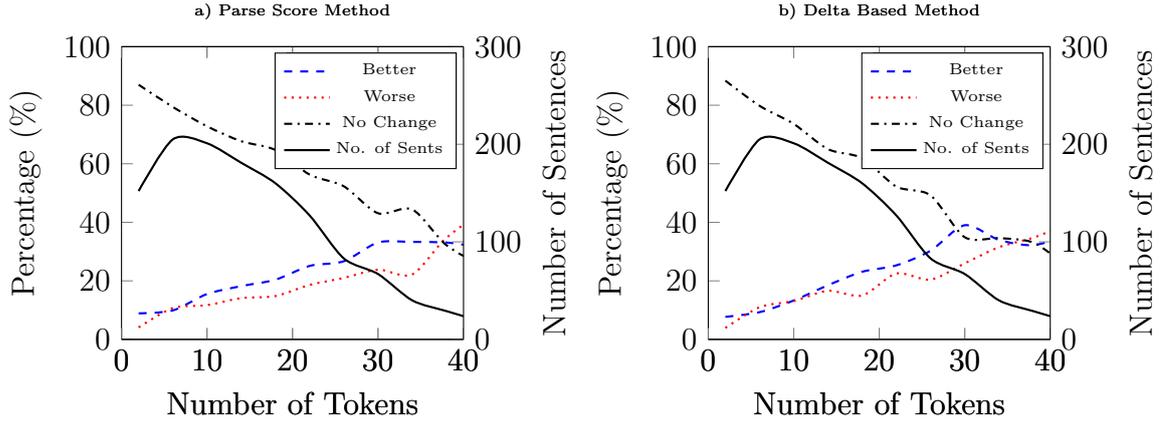
\begin{figure}[t]
	\begin{center}
		\begin{tikzpicture}
		\pgfplotsset{
			width=4.5cm,
			title=\tiny a) Parse Score Method,
			xmin=0,xmax=40,
			xlabel=Number of Tokens}
		\begin{axis}[
		axis y line*=left,
		ymin=0,ymax=100,
		ylabel=Percentage (\%)]
		%Better
		\addplot[smooth,thick,dashed,color=blue] coordinates {(2,8.9) (6,9.9) (10,15.5) (14,18.2) (18,20.4) (22,25.1) (26,26.7) (30,33.1) (34,33.3) (38,33.1) (42,32.1) (46,37.0) (50,51.2) (54,43.5) (58,29.4) (62,50.0) };\label{better}
		\addlegendentry{\tiny Better}
		
		%Worse
		\addplot[smooth,thick,dotted,color=red] coordinates{(2,4.1) (6,10.7) (10,11.7) (14,14.1) (18,14.8) (22,18.6) (26,21.0) (30,23.8) (34,22.2) (38,34.7) (42,41.0) (46,28.3) (50,24.4) (54,43.5) (58,52.9) (62,20.0) };\label{worse}
		\addlegendentry{\tiny Worse}
		
		%No Change
		\addplot[smooth,thick,dashdotted] coordinates{(2,87.0) (6,79.4) (10,72.8) (14,67.7) (18,64.7) (22,56.3) (26,52.3) (30,43.1) (34,44.4) (38,32.2) (42,26.9) (46,34.8) (50,24.4) (54,13.0) (58,17.6) (62,30.0) };\label{nochange}
		\addlegendentry{\tiny No Change}
		
		\end{axis}
		\begin{axis}[
		axis y line*=right,
		ymin=0,ymax=300,
		ylabel=Number of Sentences]
		\addlegendimage{/pgfplots/refstyle=better}\addlegendentry{\tiny Better}
		\addlegendimage{/pgfplots/refstyle=worse}\addlegendentry{\tiny Worse}
		\addlegendimage{/pgfplots/refstyle=nochange}\addlegendentry{\tiny No Change}
		%All Sent
		\addplot[smooth,thick,solid] coordinates{(2,152.0) (6,205.0) (10,201.0) (14,181.0) (18,160.0) (22,127.0) (26,83.0) (30,67.0) (34,40.0) (38,29.0) (42,19.0) (46,11.0) (50,10.0) (54,5.0) (58,4.0) (62,2.0) };\addlegendentry{\tiny No. of Sents}
		
		\end{axis}
		%2 Total:610 LAS better/worse/nochange:54/25/531 8.9/4.1/87.0 4.8
		%6 Total:820 LAS better/worse/nochange:81/88/651 9.9/10.7/79.4 -0.8
		%10 Total:804 LAS better/worse/nochange:125/94/585 15.5/11.7/72.8 3.8
		%14 Total:725 LAS better/worse/nochange:132/102/491 18.2/14.1/67.7 4.1
		%18 Total:641 LAS better/worse/nochange:131/95/415 20.4/14.8/64.7 5.6
		%22 Total:510 LAS better/worse/nochange:128/95/287 25.1/18.6/56.3 6.5
		%26 Total:333 LAS better/worse/nochange:89/70/174 26.7/21.0/52.3 5.7
		%30 Total:269 LAS better/worse/nochange:89/64/116 33.1/23.8/43.1 9.3
		%34 Total:162 LAS better/worse/nochange:54/36/72 33.3/22.2/44.4 11.1
		%38 Total:118 LAS better/worse/nochange:39/41/38 33.1/34.7/32.2 -1.6
		%42 Total:78 LAS better/worse/nochange:25/32/21 32.1/41.0/26.9 -8.9
		%46 Total:46 LAS better/worse/nochange:17/13/16 37.0/28.3/34.8 8.7
		%50 Total:41 LAS better/worse/nochange:21/10/10 51.2/24.4/24.4 26.8
		%54 Total:23 LAS better/worse/nochange:10/10/3 43.5/43.5/13.0 0.0
		%58 Total:17 LAS better/worse/nochange:5/9/3 29.4/52.9/17.6 -23.5
		%62 Total:10 LAS better/worse/nochange:5/2/3 50.0/20.0/30.0 30.0
		
		\end{tikzpicture}
		\begin{tikzpicture}
		\pgfplotsset{
			width=4.5cm,
			title=\tiny b) Delta Based Method,
			xmin=0,xmax=40,
			xlabel=Number of Tokens}
		\begin{axis}[
		axis y line*=left,
		ymin=0,ymax=100,
		ylabel=Percentage (\%)]
		%Better
		\addplot[smooth,thick,dashed,color=blue] coordinates {(2,7.7) (6,9.3) (10,13.3) (14,18.5) (18,23.1) (22,25.3) (26,30.3) (30,39.0) (34,34.0) (38,32.2) (42,35.9) (46,43.5) (50,46.3) (54,34.8) (58,35.3) (62,40.0) };\label{better}
		\addlegendentry{\tiny Better}
		
		%Worse
		\addplot[smooth,thick,dotted,color=red] coordinates{(2,3.9) (6,11.0) (10,13.1) (14,16.6) (18,15.0) (22,22.5) (26,20.4) (30,26.0) (34,31.5) (38,34.7) (42,38.5) (46,32.6) (50,19.5) (54,43.5) (58,29.4) (62,30.0) };\label{worse}
		\addlegendentry{\tiny Worse}
		
		%No Change
		\addplot[smooth,thick,dashdotted] coordinates{(2,88.4) (6,79.8) (10,73.6) (14,65.0) (18,61.9) (22,52.2) (26,49.2) (30,34.9) (34,34.6) (38,33.1) (42,25.6) (46,23.9) (50,34.1) (54,21.7) (58,35.3) (62,30.0) };\label{nochange}
		\addlegendentry{\tiny No Change}
		
		\end{axis}
		\begin{axis}[
		axis y line*=right,
		ymin=0,ymax=300,
		ylabel=Number of Sentences]
		\addlegendimage{/pgfplots/refstyle=better}\addlegendentry{\tiny Better}
		\addlegendimage{/pgfplots/refstyle=worse}\addlegendentry{\tiny Worse}
		\addlegendimage{/pgfplots/refstyle=nochange}\addlegendentry{\tiny No Change}
		%All Sent
		\addplot[smooth,thick,solid] coordinates{(2,152.0) (6,205.0) (10,201.0) (14,181.0) (18,160.0) (22,127.0) (26,83.0) (30,67.0) (34,40.0) (38,29.0) (42,19.0) (46,11.0) (50,10.0) (54,5.0) (58,4.0) (62,2.0) };\addlegendentry{\tiny No. of Sents}
		
		\end{axis}
		%2 Total:610 LAS better/worse/nochange:47/24/539 7.7/3.9/88.4 3.8
		%6 Total:820 LAS better/worse/nochange:76/90/654 9.3/11.0/79.8 -1.7
		%10 Total:804 LAS better/worse/nochange:107/105/592 13.3/13.1/73.6 0.2
		%14 Total:725 LAS better/worse/nochange:134/120/471 18.5/16.6/65.0 1.9
		%18 Total:641 LAS better/worse/nochange:148/96/397 23.1/15.0/61.9 8.1
		%22 Total:510 LAS better/worse/nochange:129/115/266 25.3/22.5/52.2 2.8
		%26 Total:333 LAS better/worse/nochange:101/68/164 30.3/20.4/49.2 9.9
		%30 Total:269 LAS better/worse/nochange:105/70/94 39.0/26.0/34.9 13.0
		%34 Total:162 LAS better/worse/nochange:55/51/56 34.0/31.5/34.6 2.5
		%38 Total:118 LAS better/worse/nochange:38/41/39 32.2/34.7/33.1 -2.5
		%42 Total:78 LAS better/worse/nochange:28/30/20 35.9/38.5/25.6 -2.6
		%46 Total:46 LAS better/worse/nochange:20/15/11 43.5/32.6/23.9 10.9
		%50 Total:41 LAS better/worse/nochange:19/8/14 46.3/19.5/34.1 26.8
		%54 Total:23 LAS better/worse/nochange:8/10/5 34.8/43.5/21.7 -8.7
		%58 Total:17 LAS better/worse/nochange:6/5/6 35.3/29.4/35.3 5.9
		%62 Total:10 LAS better/worse/nochange:4/3/3 40.0/30.0/30.0 10.0
		
		\end{tikzpicture}			
	\end{center}
	\caption{\label{figure:analysis-sentlength-self-training} The comparison between the self-training approach and the baseline on different number of tokens per sentence. }
\end{figure}

\textbf{Sentence Length.} For the sentence level analysis we first evaluate the performance of our self-training approaches on the different sentence lengths. The sentences that have the same length are grouped into classes. For each class, the sentences are further classified into three subclasses (better, worse and no change) according to their accuracies when compared with the baseline.  We plot them together with the number of sentences in individual classes in Figure \ref{figure:analysis-sentlength-self-training}. The left-hand side is the figure for the parse score-based method, while the right-hand side is that of the Delta-based method. At a first glance, both methods show similar behaviours, they both do not help the very short sentences. The percentages for sentences longer than 30 tokens are varied. More precisely, the parse score-based method helps most on the sentences containing between 10 and 35 tokens, and the Delta-based method is most productive on sentences which have a length between 15 and 30 tokens.

%number of unknown words
\begin{figure}[t]
	\begin{center}
		\begin{tikzpicture}
		\pgfplotsset{
			width=4.5cm,
			title=\tiny a) Parse Score Method (LAS),
			xmin=0,xmax=4,
			xtick={0,1,...,4},
			xlabel=Number of Unknown Words}
		\begin{axis}[
		axis y line*=left,
		ymin=0,ymax=100,
		ylabel=Percentage (\%)]
		%Better
		\addplot[smooth,thick,dashed,color=blue] coordinates {(0,15.5) (1,19.0) (2,24.6) (3,29.9) (4,33.1) (5,50.0) (6,33.3) };\label{better}
		\addlegendentry{\tiny Better}
		
		%Worse
		\addplot[smooth,thick,dotted,color=red] coordinates{(0,11.9) (1,14.7) (2,20.9) (3,26.1) (4,25.0) (5,23.4) (6,38.9) };\label{worse}
		\addlegendentry{\tiny Worse}
		
		%No Change
		\addplot[smooth,thick,dashdotted] coordinates{(0,72.7) (1,66.3) (2,54.4) (3,43.9) (4,41.9) (5,26.6) (6,27.8) };\label{nochange}
		\addlegendentry{\tiny No Change}
		
		\end{axis}
		\begin{axis}[
		axis y line*=right,
		ymin=0,ymax=3000,
		ylabel=Number of Sentences]
		\addlegendimage{/pgfplots/refstyle=better}\addlegendentry{\tiny Better}
		\addlegendimage{/pgfplots/refstyle=worse}\addlegendentry{\tiny Worse}
		\addlegendimage{/pgfplots/refstyle=nochange}\addlegendentry{\tiny No Change}
		%All Sent
		\addplot[smooth,thick,solid] coordinates{(0,2540.0) (1,1523.0) (2,621.0) (3,264.0) (4,148.0) (5,64.0) (6,36.0) };\addlegendentry{\tiny No. of Sents}
		
		\end{axis}
		%0 Total:2540 LAS better/worse/nochange:393/301/1846 15.5/11.9/72.7 3.6
		%1 Total:1523 LAS better/worse/nochange:289/224/1010 19.0/14.7/66.3 4.3
		%2 Total:621 LAS better/worse/nochange:153/130/338 24.6/20.9/54.4 3.7
		%3 Total:264 LAS better/worse/nochange:79/69/116 29.9/26.1/43.9 3.8
		%4 Total:148 LAS better/worse/nochange:49/37/62 33.1/25.0/41.9 8.1
		%5 Total:64 LAS better/worse/nochange:32/15/17 50.0/23.4/26.6 26.6
		%6 Total:36 LAS better/worse/nochange:12/14/10 33.3/38.9/27.8 -5.6
		
		\end{tikzpicture}
		\begin{tikzpicture}
		\pgfplotsset{
			width=4.5cm,
			title=\tiny b) Delta Based Method (LAS),
			xmin=0,xmax=4,
			xtick={0,1,...,4},
			xlabel=Number of Unknown Words}
		\begin{axis}[
		axis y line*=left,
		ymin=0,ymax=100,
		ylabel=Percentage (\%)]
		%Better
		\addplot[smooth,thick,dashed,color=blue] coordinates {(0,15.9) (1,20.4) (2,23.2) (3,31.1) (4,31.8) (5,42.2) (6,36.1) };\label{better}
		\addlegendentry{\tiny Better}
		
		%Worse
		\addplot[smooth,thick,dotted,color=red] coordinates{(0,13.0) (1,16.7) (2,21.6) (3,24.6) (4,27.0) (5,26.6) (6,36.1) };\label{worse}
		\addlegendentry{\tiny Worse}
		
		%No Change
		\addplot[smooth,thick,dashdotted] coordinates{(0,71.1) (1,63.0) (2,55.2) (3,44.3) (4,41.2) (5,31.3) (6,27.8) };\label{nochange}
		\addlegendentry{\tiny No Change}
		
		\end{axis}
		\begin{axis}[
		axis y line*=right,
		ymin=0,ymax=3000,
		ylabel=Number of Sentences]
		\addlegendimage{/pgfplots/refstyle=better}\addlegendentry{\tiny Better}
		\addlegendimage{/pgfplots/refstyle=worse}\addlegendentry{\tiny Worse}
		\addlegendimage{/pgfplots/refstyle=nochange}\addlegendentry{\tiny No Change}
		%All Sent
		\addplot[smooth,thick,solid] coordinates{(0,2540.0) (1,1523.0) (2,621.0) (3,264.0) (4,148.0) (5,64.0) (6,36.0) };\addlegendentry{\tiny No. of Sents}
		
		\end{axis}
		%0 Total:2540 LAS better/worse/nochange:403/330/1807 15.9/13.0/71.1 2.9
		%1 Total:1523 LAS better/worse/nochange:310/254/959 20.4/16.7/63.0 3.7
		%2 Total:621 LAS better/worse/nochange:144/134/343 23.2/21.6/55.2 1.6
		%3 Total:264 LAS better/worse/nochange:82/65/117 31.1/24.6/44.3 6.5
		%4 Total:148 LAS better/worse/nochange:47/40/61 31.8/27.0/41.2 4.8
		%5 Total:64 LAS better/worse/nochange:27/17/20 42.2/26.6/31.3 15.6
		%6 Total:36 LAS better/worse/nochange:13/13/10 36.1/36.1/27.8 0.0
		
		\end{tikzpicture}
		\begin{tikzpicture}
		\pgfplotsset{
			width=4.5cm,
			title=\tiny c) Parse Score Method (UAS),
			xmin=0,xmax=4,
			xtick={0,1,...,4},
			xlabel=Number of Unknown Words}
		\begin{axis}[
		axis y line*=left,
		ymin=0,ymax=100,
		ylabel=Percentage (\%)]
		%Better
		\addplot[smooth,thick,dashed,color=blue] coordinates {(0,13.0) (1,16.3) (2,20.9) (3,27.7) (4,32.4) (5,40.6) (6,33.3) };\label{better}
		\addlegendentry{\tiny Better}
		
		%Worse
		\addplot[smooth,thick,dotted,color=red] coordinates{(0,9.6) (1,13.2) (2,19.2) (3,22.7) (4,24.3) (5,21.9) (6,50.0) };\label{worse}
		\addlegendentry{\tiny Worse}
		
		%No Change
		\addplot[smooth,thick,dashdotted] coordinates{(0,77.4) (1,70.5) (2,59.9) (3,49.6) (4,43.2) (5,37.5) (6,16.7) };\label{nochange}
		\addlegendentry{\tiny No Change}
		
		\end{axis}
		\begin{axis}[
		axis y line*=right,
		ymin=0,ymax=3000,
		ylabel=Number of Sentences]
		\addlegendimage{/pgfplots/refstyle=better}\addlegendentry{\tiny Better}
		\addlegendimage{/pgfplots/refstyle=worse}\addlegendentry{\tiny Worse}
		\addlegendimage{/pgfplots/refstyle=nochange}\addlegendentry{\tiny No Change}
		%All Sent
		\addplot[smooth,thick,solid] coordinates{(0,2540.0) (1,1523.0) (2,621.0) (3,264.0) (4,148.0) (5,64.0) (6,36.0) };\addlegendentry{\tiny No. of Sents}
		
		\end{axis}
		%0 Total:2540 UAS better/worse/nochange:329/244/1967 13.0/9.6/77.4 Diff:3.4
		%1 Total:1523 UAS better/worse/nochange:249/201/1073 16.3/13.2/70.5 Diff:3.1
		%2 Total:621 UAS better/worse/nochange:130/119/372 20.9/19.2/59.9 Diff:1.7
		%3 Total:264 UAS better/worse/nochange:73/60/131 27.7/22.7/49.6 Diff:5.0
		%4 Total:148 UAS better/worse/nochange:48/36/64 32.4/24.3/43.2 Diff:8.1
		%5 Total:64 UAS better/worse/nochange:26/14/24 40.6/21.9/37.5 Diff:18.7
		%6 Total:36 UAS better/worse/nochange:12/18/6 33.3/50.0/16.7 Diff:-16.7
		
		\end{tikzpicture}
		\begin{tikzpicture}
		\pgfplotsset{
			width=4.5cm,
			title=\tiny d) Delta Based Method (UAS),
			xmin=0,xmax=4,
			xtick={0,1,...,4},
			xlabel=Number of Unknown Words}
		\begin{axis}[
		axis y line*=left,
		ymin=0,ymax=100,
		ylabel=Percentage (\%)]
		%Better
		\addplot[smooth,thick,dashed,color=blue] coordinates {(0,13.3) (1,16.7) (2,19.6) (3,29.2) (4,32.4) (5,37.5) (6,36.1) };\label{better}
		\addlegendentry{\tiny Better}
		
		%Worse
		\addplot[smooth,thick,dotted,color=red] coordinates{(0,11.0) (1,14.1) (2,18.7) (3,20.1) (4,23.6) (5,26.6) (6,38.9) };\label{worse}
		\addlegendentry{\tiny Worse}
		
		%No Change
		\addplot[smooth,thick,dashdotted] coordinates{(0,75.7) (1,69.1) (2,61.7) (3,50.8) (4,43.9) (5,35.9) (6,25.0) };\label{nochange}
		\addlegendentry{\tiny No Change}
		
		\end{axis}
		\begin{axis}[
		axis y line*=right,
		ymin=0,ymax=3000,
		ylabel=Number of Sentences]
		\addlegendimage{/pgfplots/refstyle=better}\addlegendentry{\tiny Better}
		\addlegendimage{/pgfplots/refstyle=worse}\addlegendentry{\tiny Worse}
		\addlegendimage{/pgfplots/refstyle=nochange}\addlegendentry{\tiny No Change}
		%All Sent
		\addplot[smooth,thick,solid] coordinates{(0,2540.0) (1,1523.0) (2,621.0) (3,264.0) (4,148.0) (5,64.0) (6,36.0) };\addlegendentry{\tiny No. of Sents}
		
		\end{axis}
		%0 Total:2540 UAS better/worse/nochange:339/279/1922 13.3/11.0/75.7 Diff:2.3
		%1 Total:1523 UAS better/worse/nochange:255/215/1053 16.7/14.1/69.1 Diff:2.6
		%2 Total:621 UAS better/worse/nochange:122/116/383 19.6/18.7/61.7 Diff:0.9
		%3 Total:264 UAS better/worse/nochange:77/53/134 29.2/20.1/50.8 Diff:9.1
		%4 Total:148 UAS better/worse/nochange:48/35/65 32.4/23.6/43.9 Diff:8.8
		%5 Total:64 UAS better/worse/nochange:24/17/23 37.5/26.6/35.9 Diff:10.9
		%6 Total:36 UAS better/worse/nochange:13/14/9 36.1/38.9/25.0 Diff:-2.8
		
		\end{tikzpicture}
	\end{center}
	\caption{\label{figure:analysis-sentunk-self-training} The comparison between the self-training approach and the baseline on different number of unknown words per sentence. }
\end{figure}
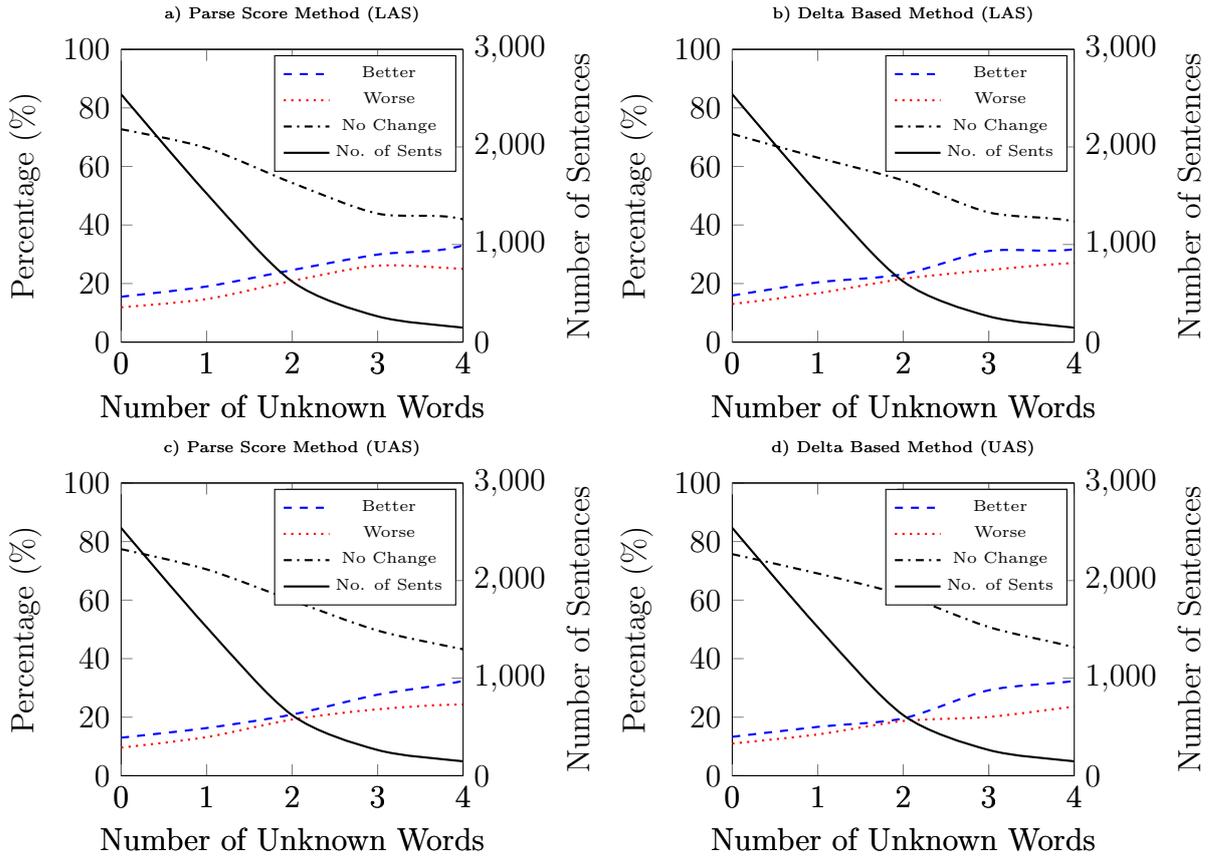

\textbf{Unknown Words.} For the sentence level analysis of unknown words, we evaluate on both labelled and unlabelled accuracy scores. This is mainly because according to our token level analysis our self-training gained much larger unlabelled improvements on the unknown words than that of known words. Figure \ref{figure:analysis-sentunk-self-training} shows our analysis of unknown words, the upper figures are the analysis of labelled accuracies and the lower two are that of unlabelled accuracies. As we can see from the above two figures, the gap between sentences that have a better labelled accuracy and sentences worsened in accuracy are not affected by the increasing number of unknown words in sentences. The gap on unlabelled accuracies shows a clear increasement when more than two unknown words are found in the sentence. This is in line with our finding in the token level analysis that self-training could improve more on unknown words attachment.

%number of prepositions
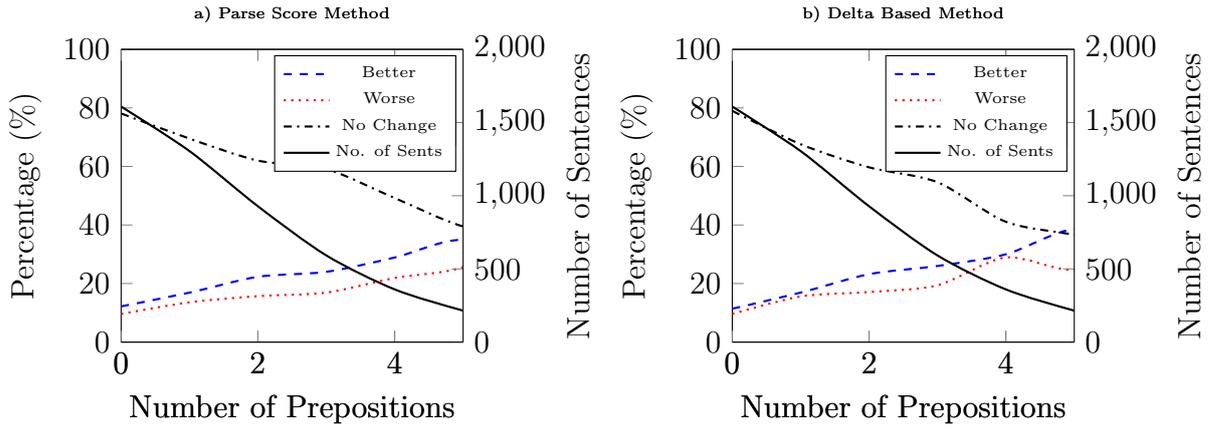
\begin{figure}[t]
	\begin{center}
		\begin{tikzpicture}
		\pgfplotsset{
			width=4.5cm,
			title=\tiny a) Parse Score Method,
			xmin=0,xmax=5,
			xlabel=Number of Prepositions}
		\begin{axis}[
		axis y line*=left,
		ymin=0,ymax=100,
		ylabel=Percentage (\%)]
		%Better
		\addplot[smooth,thick,dashed,color=blue] coordinates {(0,12.2) (1,16.9) (2,22.3) (3,24.0) (4,28.9) (5,34.9) (6,25.8) };\label{better}
		\addlegendentry{\tiny Better}
		
		%Worse
		\addplot[smooth,thick,dotted,color=red] coordinates{(0,9.6) (1,13.6) (2,15.7) (3,16.9) (4,21.9) (5,25.6) (6,38.1) };\label{worse}
		\addlegendentry{\tiny Worse}
		
		%No Change
		\addplot[smooth,thick,dashdotted] coordinates{(0,78.1) (1,69.5) (2,62.0) (3,59.1) (4,49.2) (5,39.5) (6,36.1) };\label{nochange}
		\addlegendentry{\tiny No Change}
		
		\end{axis}
		\begin{axis}[
		axis y line*=right,
		ymin=0,ymax=2000,
		ylabel=Number of Sentences]
		\addlegendimage{/pgfplots/refstyle=better}\addlegendentry{\tiny Better}
		\addlegendimage{/pgfplots/refstyle=worse}\addlegendentry{\tiny Worse}
		\addlegendimage{/pgfplots/refstyle=nochange}\addlegendentry{\tiny No Change}
		%All Sent
		\addplot[smooth,thick,solid] coordinates{(0,1609.0) (1,1305.0) (2,928.0) (3,592.0) (4,360.0) (5,215.0) (6,97.0) };\addlegendentry{\tiny No. of Sents}
		
		\end{axis}
		%0 Total:1609 LAS better/worse/nochange:197/155/1257 12.2/9.6/78.1 Diff:2.6
		%1 Total:1305 LAS better/worse/nochange:220/178/907 16.9/13.6/69.5 Diff:3.3
		%2 Total:928 LAS better/worse/nochange:207/146/575 22.3/15.7/62.0 Diff:6.6
		%3 Total:592 LAS better/worse/nochange:142/100/350 24.0/16.9/59.1 Diff:7.1
		%4 Total:360 LAS better/worse/nochange:104/79/177 28.9/21.9/49.2 Diff:7.0
		%5 Total:215 LAS better/worse/nochange:75/55/85 34.9/25.6/39.5 Diff:9.3
		%6 Total:97 LAS better/worse/nochange:25/37/35 25.8/38.1/36.1 Diff:-12.3
		
		\end{tikzpicture}
		\begin{tikzpicture}
		\pgfplotsset{
			width=4.5cm,
			title=\tiny b) Delta Based Method,
			xmin=0,xmax=5,
			xlabel=Number of Prepositions}
		\begin{axis}[
		axis y line*=left,
		ymin=0,ymax=100,
		ylabel=Percentage (\%)]
		%Better
		\addplot[smooth,thick,dashed,color=blue] coordinates {(0,11.4) (1,16.9) (2,23.2) (3,26.0) (4,30.0) (5,38.6) (6,34.0) };\label{better}
		\addlegendentry{\tiny Better}
		
		%Worse
		\addplot[smooth,thick,dotted,color=red] coordinates{(0,9.6) (1,15.6) (2,17.1) (3,19.4) (4,28.9) (5,24.7) (6,33.0) };\label{worse}
		\addlegendentry{\tiny Worse}
		
		%No Change
		\addplot[smooth,thick,dashdotted] coordinates{(0,79.0) (1,67.6) (2,59.7) (3,54.6) (4,41.1) (5,36.7) (6,33.0) };\label{nochange}
		\addlegendentry{\tiny No Change}
		
		\end{axis}
		\begin{axis}[
		axis y line*=right,
		ymin=0,ymax=2000,
		ylabel=Number of Sentences]
		\addlegendimage{/pgfplots/refstyle=better}\addlegendentry{\tiny Better}
		\addlegendimage{/pgfplots/refstyle=worse}\addlegendentry{\tiny Worse}
		\addlegendimage{/pgfplots/refstyle=nochange}\addlegendentry{\tiny No Change}
		%All Sent
		\addplot[smooth,thick,solid] coordinates{(0,1609.0) (1,1305.0) (2,928.0) (3,592.0) (4,360.0) (5,215.0) (6,97.0) };\addlegendentry{\tiny No. of Sents}
		
		\end{axis}
		%0 Total:1609 LAS better/worse/nochange:183/155/1271 11.4/9.6/79.0 Diff:1.8
		%1 Total:1305 LAS better/worse/nochange:220/203/882 16.9/15.6/67.6 Diff:1.3
		%2 Total:928 LAS better/worse/nochange:215/159/554 23.2/17.1/59.7 Diff:6.1
		%3 Total:592 LAS better/worse/nochange:154/115/323 26.0/19.4/54.6 Diff:6.6
		%4 Total:360 LAS better/worse/nochange:108/104/148 30.0/28.9/41.1 Diff:1.1
		%5 Total:215 LAS better/worse/nochange:83/53/79 38.6/24.7/36.7 Diff:13.9
		%6 Total:97 LAS better/worse/nochange:33/32/32 34.0/33.0/33.0 Diff:1.0
		
		\end{tikzpicture}
	\end{center}
	\caption{\label{figure:analysis-sentin-self-training} The comparison between the self-training approach and the baseline on different number of prepositions per sentence. }
\end{figure}

\textbf{Prepositions.} The preposition analysis of our confidence-based self-training is shown in Figure \ref{figure:analysis-sentin-self-training}. Both methods show very similar curves, they gain small improvements around 1\% on sentences that have up to one preposition, but they achieved larger improvements on sentences that have at least 2 prepositions. Although the differences between sentences that are parsed better and those parsed worse varies for the different number of prepositions, most of the gains are larger than 6\% and the largest gain is around 14\%. Overall, the confidence-based self-training methods show clear better performances on sentences that have multiple prepositions.

%number of conjunctions
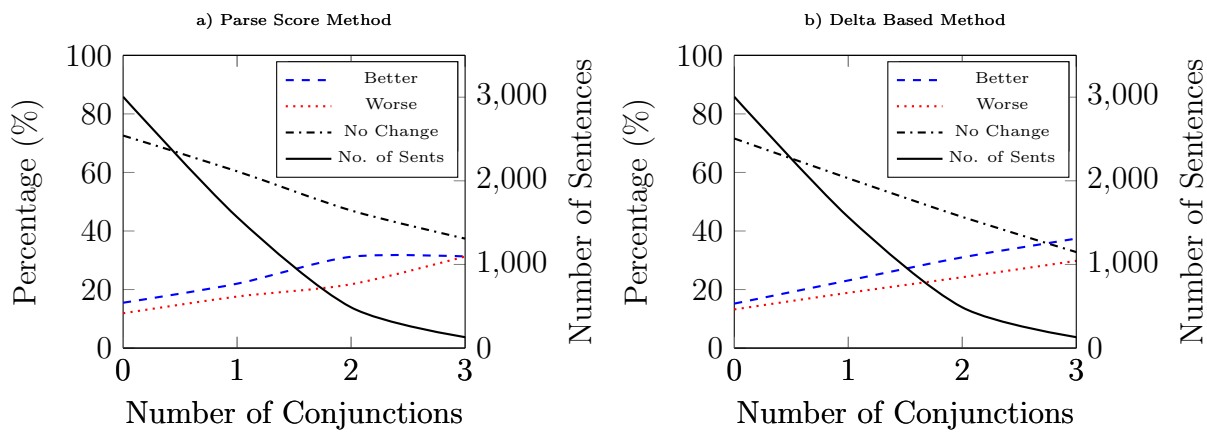
\begin{figure}[t]
	\begin{center}
		\begin{tikzpicture}
		\pgfplotsset{
			width=4.5cm,
			title=\tiny a) Parse Score Method,
			xmin=0,xmax=3,
			xtick={0,1,...,3},
			xlabel=Number of Conjunctions}
		\begin{axis}[
		axis y line*=left,
		ymin=0,ymax=100,
		ylabel=Percentage (\%)]
		%Better
		\addplot[smooth,thick,dashed,color=blue] coordinates {(0,15.5) (1,22.0) (2,31.2) (3,31.3) };\label{better}
		\addlegendentry{\tiny Better}
		
		%Worse
		\addplot[smooth,thick,dotted,color=red] coordinates{(0,11.9) (1,17.6) (2,21.8) (3,31.3) };\label{worse}
		\addlegendentry{\tiny Worse}
		
		%No Change
		\addplot[smooth,thick,dashdotted] coordinates{(0,72.6) (1,60.4) (2,47.0) (3,37.4) };\label{nochange}
		\addlegendentry{\tiny No Change}
		
		\end{axis}
		\begin{axis}[
		axis y line*=right,
		ymin=0,ymax=3500,
		ylabel=Number of Sentences]
		\addlegendimage{/pgfplots/refstyle=better}\addlegendentry{\tiny Better}
		\addlegendimage{/pgfplots/refstyle=worse}\addlegendentry{\tiny Worse}
		\addlegendimage{/pgfplots/refstyle=nochange}\addlegendentry{\tiny No Change}
		%All Sent
		\addplot[smooth,thick,solid] coordinates{(0,3005.0) (1,1566.0) (2,487.0) (3,131.0) };\addlegendentry{\tiny No. of Sents}
		
		\end{axis}
		%0 Total:3005 LAS better/worse/nochange:465/358/2182 15.5/11.9/72.6 Diff:3.6
		%1 Total:1566 LAS better/worse/nochange:345/275/946 22.0/17.6/60.4 Diff:4.4
		%2 Total:487 LAS better/worse/nochange:152/106/229 31.2/21.8/47.0 Diff:9.4
		%3 Total:131 LAS better/worse/nochange:41/41/49 31.3/31.3/37.4 Diff:0.0
		
		\end{tikzpicture}
		\begin{tikzpicture}
		\pgfplotsset{
			width=4.5cm,
			title=\tiny b) Delta Based Method,
			xmin=0,xmax=3,
			xtick={0,1,...,3},
			xlabel=Number of Conjunctions}
		\begin{axis}[
		axis y line*=left,
		ymin=0,ymax=100,
		ylabel=Percentage (\%)]
		%Better
		\addplot[smooth,thick,dashed,color=blue] coordinates {(0,15.2) (1,23.1) (2,31.0) (3,37.4) };\label{better}
		\addlegendentry{\tiny Better}
		
		%Worse
		\addplot[smooth,thick,dotted,color=red] coordinates{(0,13.2) (1,18.9) (2,24.2) (3,29.8) };\label{worse}
		\addlegendentry{\tiny Worse}
		
		%No Change
		\addplot[smooth,thick,dashdotted] coordinates{(0,71.6) (1,58.0) (2,44.8) (3,32.8) };\label{nochange}
		\addlegendentry{\tiny No Change}
		
		\end{axis}
		\begin{axis}[
		axis y line*=right,
		ymin=0,ymax=3500,
		ylabel=Number of Sentences]
		\addlegendimage{/pgfplots/refstyle=better}\addlegendentry{\tiny Better}
		\addlegendimage{/pgfplots/refstyle=worse}\addlegendentry{\tiny Worse}
		\addlegendimage{/pgfplots/refstyle=nochange}\addlegendentry{\tiny No Change}
		%All Sent
		\addplot[smooth,thick,solid] coordinates{(0,3005.0) (1,1566.0) (2,487.0) (3,131.0) };\addlegendentry{\tiny No. of Sents}
		
		\end{axis}
		%0 Total:3005 LAS better/worse/nochange:456/396/2153 15.2/13.2/71.6 Diff:2.0
		%1 Total:1566 LAS better/worse/nochange:362/296/908 23.1/18.9/58.0 Diff:4.2
		%2 Total:487 LAS better/worse/nochange:151/118/218 31.0/24.2/44.8 Diff:6.8
		%3 Total:131 LAS better/worse/nochange:49/39/43 37.4/29.8/32.8 Diff:7.6
		
		\end{tikzpicture}
	\end{center}
	\caption{\label{figure:analysis-sentcc-self-training} The comparison between the self-training approach and the baseline on different number of conjunctions per sentence. }
\end{figure}

\textbf{Conjunctions.} In terms of conjunctions, both methods show similar figures, cf. Figure \ref{figure:analysis-sentcc-self-training}. They both show gains for most of the cases, except that the parse score-based method shows no effect on sentences that have 3 conjunctions. They both start with a small gain of 2-3\% when there is no conjunction in the sentence and the improvement widened to 7-10\% for sentences have more conjunctions. There are only 100 sentences in the class of 3 conjunctions, thus the numbers of this class are less reliable. Generally speaking, the self-training approaches work slightly better on the sentences that have more conjunctions.

%examples
\begin{table}
\begin{center}
\begin{tabular}{p{\linewidth}}
\hline
\hline
\colorbox{blue!30}{\footnotesize But}$_{31_{\textsc{dep}}}^{1 \colorbox{red!30}{\tiny c}}$ \colorbox{blue!30}{\footnotesize creating}$_{31_{\textsc{sbj}}}^{2}$ \footnotesize a$_{5_{\textsc{nmod}}}^{3}$ \footnotesize balanced$_{5_{\textsc{nmod}}}^{4}$ \footnotesize community$_{2_{\textsc{obj}}}^{5}$ \footnotesize with$_{5_{\textsc{nmod}}}^{6 \colorbox{red!30}{\tiny p}}$ \footnotesize a$_{8_{\textsc{nmod}}}^{7}$ \footnotesize mix$_{6_{\textsc{pmod}}}^{8}$ \footnotesize of$_{8_{\textsc{nmod}}}^{9 \colorbox{red!30}{\tiny p}}$ \footnotesize housing$_{9_{\textsc{pmod}}}^{10}$ \footnotesize ,$_{10_{\textsc{p}}}^{11}$ \footnotesize offices$_{10_{\textsc{coord}}}^{12}$ \footnotesize ,$_{12_{\textsc{p}}}^{13}$ \footnotesize shopping$_{12_{\textsc{coord}}}^{14}$ \footnotesize and$_{14_{\textsc{coord}}}^{15 \colorbox{red!30}{\tiny c}}$ \footnotesize other$_{17_{\textsc{nmod}}}^{16}$ \footnotesize amenities$_{15_{\textsc{conj}}}^{17}$ \footnotesize --$_{5_{\textsc{p}}}^{18}$ \footnotesize allowing$_{5_{\textsc{appo}}}^{19}$ \footnotesize people$_{19_{\textsc{obj}}}^{20}$ \footnotesize to$_{19_{\textsc{oprd}}}^{21}$ \footnotesize live$_{21_{\textsc{im}}}^{22}$ \footnotesize close$_{22_{\textsc{loc}}}^{23}$ \footnotesize to$_{23_{\textsc{amod}}}^{24 \colorbox{red!30}{\tiny p}}$ \colorbox{blue!30}{\footnotesize where}$_{27_{\textsc{loc}}}^{25}$ \footnotesize they$_{27_{\textsc{sbj}}}^{26}$ \footnotesize work$_{24_{\textsc{pmod}}}^{27}$ \footnotesize and$_{27_{\textsc{coord}}}^{28 \colorbox{red!30}{\tiny c}}$ \footnotesize play$_{28_{\textsc{conj}}}^{29}$ \footnotesize --$_{5_{\textsc{p}}}^{30}$ \colorbox{blue!30}{\footnotesize is}$_{0_{\textsc{root}}}^{31}$ \footnotesize an$_{36_{\textsc{nmod}}}^{32}$ \colorbox{blue!30}{\footnotesize even}$_{35_{\textsc{amod}}}^{33}$ \colorbox{blue!30}{\footnotesize more}$_{35_{\textsc{amod}}}^{34}$ \colorbox{blue!30}{\footnotesize worthy}$_{36_{\textsc{nmod}}}^{35}$ \colorbox{green!30}{\footnotesize goal}$_{31_{\textsc{prd}}}^{36}$ \colorbox{blue!30}{\footnotesize .}$_{31_{\textsc{p}}}^{37}$\\\hline

\colorbox{blue!30}{\footnotesize Her}$_{4_{\textsc{nmod}}}^{1}$ \footnotesize ``$_{4_{\textsc{p}}}^{2}$ \footnotesize Rubble$_{4_{\textsc{name}}}^{3}$ \colorbox{green!30}{\footnotesize Division}$_{6_{\textsc{sbj}}}^{4}$ \footnotesize ''$_{4_{\textsc{p}}}^{5}$ \colorbox{blue!30}{\footnotesize mixes}$_{0_{\textsc{root}}}^{6}$ \footnotesize such$_{9_{\textsc{nmod}}}^{7}$ \footnotesize disparate$_{9_{\textsc{nmod}}}^{8}$ \colorbox{blue!30}{\footnotesize materials}$_{6_{\textsc{obj}}}^{9}$ \footnotesize as$_{9_{\textsc{nmod}}}^{10 \colorbox{red!30}{\tiny p}}$ \footnotesize ink$_{13_{\textsc{nmod}}}^{11}$ \footnotesize -$_{13_{\textsc{nmod}}}^{12}$ \footnotesize jet$_{14_{\textsc{nmod}}}^{13}$ \colorbox{blue!30}{\footnotesize prints}$_{10_{\textsc{pmod}}}^{14}$ \colorbox{blue!30}{\footnotesize pasted}$_{14_{\textsc{appo}}}^{15 \colorbox{red!30}{\tiny u}}$ \colorbox{blue!30}{\footnotesize on}$_{15_{\textsc{loc}}}^{16 \colorbox{red!30}{\tiny p}}$ \footnotesize board$_{16_{\textsc{pmod}}}^{17}$ \footnotesize ,$_{14_{\textsc{p}}}^{18}$ \footnotesize foam$_{20_{\textsc{nmod}}}^{19}$ \footnotesize rubber$_{14_{\textsc{coord}}}^{20}$ \footnotesize ,$_{20_{\textsc{p}}}^{21}$ \footnotesize galvanized$_{23_{\textsc{nmod}}}^{22}$ \footnotesize steel$_{20_{\textsc{coord}}}^{23}$ \footnotesize ,$_{23_{\textsc{p}}}^{24}$ \footnotesize concrete$_{23_{\textsc{coord}}}^{25}$ \footnotesize ,$_{25_{\textsc{p}}}^{26}$ \footnotesize steel$_{28_{\textsc{nmod}}}^{27}$ \footnotesize rebar$_{25_{\textsc{coord}}}^{28 \colorbox{red!30}{\tiny u}}$ \footnotesize and$_{28_{\textsc{coord}}}^{29 \colorbox{red!30}{\tiny c}}$ \footnotesize bungee$_{31_{\textsc{nmod}}}^{30 \colorbox{red!30}{\tiny u}}$ \footnotesize cords$_{29_{\textsc{conj}}}^{31 \colorbox{red!30}{\tiny u}}$ \colorbox{blue!30}{\footnotesize .}$_{6_{\textsc{p}}}^{32}$\\\hline

\footnotesize Go$_{0_{\textsc{root}}}^{1}$ \footnotesize look$_{1_{\textsc{oprd}}}^{2}$ \footnotesize at$_{2_{\textsc{adv}}}^{3 \colorbox{red!30}{\tiny p}}$ \footnotesize DHRM$_{3_{\textsc{pmod}}}^{4 \colorbox{red!30}{\tiny u}}$ \colorbox{blue!30}{\footnotesize and}$_{4_{\textsc{coord}}}^{5 \colorbox{red!30}{\tiny c}}$ \footnotesize the$_{9_{\textsc{nmod}}}^{6}$ \footnotesize state$_{9_{\textsc{nmod}}}^{7}$ \footnotesize courts$_{9_{\textsc{nmod}}}^{8}$ \footnotesize system$_{5_{\textsc{conj}}}^{9}$ \footnotesize ($_{9_{\textsc{p}}}^{10}$ \colorbox{blue!30}{\footnotesize separate}$_{13_{\textsc{nmod}}}^{11}$ \footnotesize HR$_{13_{\textsc{nmod}}}^{12 \colorbox{red!30}{\tiny u}}$ \colorbox{blue!30}{\footnotesize dept}$_{9_{\textsc{appo}}}^{13 \colorbox{red!30}{\tiny u}}$ \footnotesize )$_{9_{\textsc{p}}}^{14}$ \footnotesize and$_{2_{\textsc{coord}}}^{15 \colorbox{red!30}{\tiny c}}$ \footnotesize see$_{15_{\textsc{conj}}}^{16}$ \footnotesize what$_{21_{\textsc{obj}}}^{17}$ \colorbox{blue!30}{\footnotesize the}$_{20_{\textsc{nmod}}}^{18}$ \colorbox{green!30}{\footnotesize state}$_{20_{\textsc{nmod}}}^{19}$ \colorbox{blue!30}{\footnotesize folks}$_{21_{\textsc{sbj}}}^{20}$ \colorbox{blue!30}{\footnotesize do}$_{16_{\textsc{obj}}}^{21}$ \colorbox{blue!30}{\footnotesize ,}$_{21_{\textsc{p}}}^{22}$ \colorbox{blue!30}{\footnotesize and}$_{21_{\textsc{coord}}}^{23 \colorbox{red!30}{\tiny c}}$ \footnotesize who$_{30_{\textsc{obj}}}^{24}$ \footnotesize all$_{26_{\textsc{nmod}}}^{25}$ \footnotesize you$_{27_{\textsc{sbj}}}^{26}$ \footnotesize 're$_{23_{\textsc{conj}}}^{27}$ \footnotesize talking$_{27_{\textsc{vc}}}^{28}$ \footnotesize about$_{28_{\textsc{adv}}}^{29 \colorbox{red!30}{\tiny p}}$ \footnotesize furloughing$_{29_{\textsc{pmod}}}^{30 \colorbox{red!30}{\tiny u}}$ \footnotesize .$_{1_{\textsc{p}}}^{31}$\\\hline

\colorbox{blue!30}{\footnotesize and}$_{3_{\textsc{dep}}}^{1 \colorbox{red!30}{\tiny c}}$ \colorbox{green!30}{\footnotesize i}$_{3_{\textsc{sbj}}}^{2}$ \colorbox{blue!30}{\footnotesize promise}$_{0_{\textsc{root}}}^{3}$ \colorbox{green!30}{\footnotesize to}$_{3_{\textsc{oprd}}}^{4}$ \footnotesize fess$_{4_{\textsc{im}}}^{5 \colorbox{red!30}{\tiny u}}$ \footnotesize up$_{5_{\textsc{prt}}}^{6}$ \footnotesize eventually$_{5_{\textsc{tmp}}}^{7}$ \footnotesize and$_{5_{\textsc{coord}}}^{8 \colorbox{red!30}{\tiny c}}$ \footnotesize tell$_{8_{\textsc{conj}}}^{9}$ \footnotesize of$_{9_{\textsc{adv}}}^{10 \colorbox{red!30}{\tiny p}}$ \footnotesize at$_{13_{\textsc{dep}}}^{11 \colorbox{red!30}{\tiny p}}$ \footnotesize least$_{11_{\textsc{amod}}}^{12}$ \colorbox{blue!30}{\footnotesize one}$_{15_{\textsc{nmod}}}^{13}$ \colorbox{blue!30}{\footnotesize such}$_{15_{\textsc{nmod}}}^{14}$ \colorbox{blue!30}{\footnotesize epic}$_{10_{\textsc{pmod}}}^{15}$ \colorbox{blue!30}{\footnotesize i}$_{17_{\textsc{sbj}}}^{16}$ \colorbox{blue!30}{\footnotesize survived}$_{15_{\textsc{nmod}}}^{17}$ \colorbox{blue!30}{\footnotesize --}$_{3_{\textsc{p}}}^{18}$\\\hline

\colorbox{blue!30}{\footnotesize If}$_{11_{\textsc{adv}}}^{1 \colorbox{red!30}{\tiny p}}$ \footnotesize you$_{3_{\textsc{sbj}}}^{2}$ \footnotesize come$_{1_{\textsc{sub}}}^{3}$ \footnotesize upon$_{3_{\textsc{adv}}}^{4 \colorbox{red!30}{\tiny p}}$ \footnotesize something$_{4_{\textsc{pmod}}}^{5}$ \colorbox{green!30}{\footnotesize important}$_{5_{\textsc{appo}}}^{6}$ \colorbox{blue!30}{\footnotesize ,}$_{11_{\textsc{p}}}^{7}$ \colorbox{blue!30}{\footnotesize by}$_{11_{\textsc{adv}}}^{8 \colorbox{red!30}{\tiny p}}$ \footnotesize all$_{10_{\textsc{nmod}}}^{9}$ \colorbox{blue!30}{\footnotesize means}$_{8_{\textsc{pmod}}}^{10}$ \colorbox{blue!30}{\footnotesize make}$_{0_{\textsc{root}}}^{11}$ \footnotesize a$_{13_{\textsc{nmod}}}^{12}$ \footnotesize note$_{11_{\textsc{obj}}}^{13}$ \footnotesize of$_{13_{\textsc{nmod}}}^{14 \colorbox{red!30}{\tiny p}}$ \footnotesize it$_{14_{\textsc{pmod}}}^{15}$ \colorbox{blue!30}{\footnotesize ,}$_{11_{\textsc{p}}}^{16}$ \colorbox{blue!30}{\footnotesize and}$_{11_{\textsc{coord}}}^{17 \colorbox{red!30}{\tiny c}}$ \footnotesize so$_{11_{\textsc{adv}}}^{18}$ \footnotesize on$_{18_{\textsc{amod}}}^{19}$ \colorbox{blue!30}{\footnotesize .}$_{11_{\textsc{p}}}^{20}$\\\hline

\footnotesize [jingzhe19]$_{12_{\textsc{dep}}}^{1 \colorbox{red!30}{\tiny u}}$ \colorbox{blue!30}{\footnotesize However}$_{12_{\textsc{adv}}}^{2}$ \colorbox{blue!30}{\footnotesize ,}$_{12_{\textsc{p}}}^{3}$ \footnotesize the$_{5_{\textsc{nmod}}}^{4}$ \colorbox{blue!30}{\footnotesize post}$_{12_{\textsc{sbj}}}^{5}$ \footnotesize of$_{5_{\textsc{nmod}}}^{6 \colorbox{red!30}{\tiny p}}$ \footnotesize driving$_{6_{\textsc{pmod}}}^{7}$ \footnotesize on$_{7_{\textsc{tmp}}}^{8 \colorbox{red!30}{\tiny p}}$ \footnotesize a$_{11_{\textsc{nmod}}}^{9}$ \footnotesize snowy$_{11_{\textsc{nmod}}}^{10 \colorbox{red!30}{\tiny u}}$ \footnotesize day$_{8_{\textsc{pmod}}}^{11}$ \footnotesize reminds$_{0_{\textsc{root}}}^{12}$ \footnotesize me$_{12_{\textsc{obj}}}^{13}$ \footnotesize again$_{12_{\textsc{adv}}}^{14}$ \colorbox{blue!30}{\footnotesize of}$_{12_{\textsc{adv}}}^{15 \colorbox{red!30}{\tiny p}}$ \footnotesize the$_{17_{\textsc{nmod}}}^{16}$ \footnotesize story$_{15_{\textsc{pmod}}}^{17}$ \footnotesize of$_{17_{\textsc{nmod}}}^{18 \colorbox{red!30}{\tiny p}}$ \footnotesize Hua$_{20_{\textsc{name}}}^{19 \colorbox{red!30}{\tiny u}}$ \footnotesize Xin$_{18_{\textsc{pmod}}}^{20 \colorbox{red!30}{\tiny u}}$ \footnotesize and$_{20_{\textsc{coord}}}^{21 \colorbox{red!30}{\tiny c}}$ \footnotesize Wang$_{23_{\textsc{name}}}^{22}$ \footnotesize Lang$_{21_{\textsc{conj}}}^{23}$ \footnotesize in$_{17_{\textsc{loc}}}^{24 \colorbox{red!30}{\tiny p}}$ \colorbox{blue!30}{\footnotesize the}$_{27_{\textsc{nmod}}}^{25}$ \colorbox{blue!30}{\footnotesize New}$_{27_{\textsc{name}}}^{26}$ \colorbox{blue!30}{\footnotesize Anecdotes}$_{24_{\textsc{pmod}}}^{27 \colorbox{red!30}{\tiny u}}$ \colorbox{blue!30}{\footnotesize of}$_{27_{\textsc{nmod}}}^{28 \colorbox{red!30}{\tiny p}}$ \footnotesize Social$_{30_{\textsc{name}}}^{29}$ \colorbox{blue!30}{\footnotesize Talk}$_{28_{\textsc{pmod}}}^{30}$ \footnotesize .$_{12_{\textsc{p}}}^{31}$\\\hline

\colorbox{blue!30}{\footnotesize But}$_{32_{\textsc{dep}}}^{1 \colorbox{red!30}{\tiny c}}$ \colorbox{yellow!30}{\footnotesize to}$_{32_{\textsc{sbj}}}^{2}$ \footnotesize stand$_{2_{\textsc{im}}}^{3}$ \colorbox{blue!30}{\footnotesize ,}$_{3_{\textsc{p}}}^{4}$ \footnotesize day$_{3_{\textsc{tmp}}}^{5}$ \footnotesize after$_{5_{\textsc{nmod}}}^{6 \colorbox{red!30}{\tiny p}}$ \footnotesize day$_{6_{\textsc{pmod}}}^{7}$ \footnotesize ,$_{2_{\textsc{p}}}^{8}$ \footnotesize and$_{2_{\textsc{coord}}}^{9 \colorbox{red!30}{\tiny c}}$ \footnotesize to$_{9_{\textsc{conj}}}^{10}$ \footnotesize make$_{10_{\textsc{im}}}^{11}$ \footnotesize such$_{14_{\textsc{nmod}}}^{12}$ \footnotesize preposterous$_{14_{\textsc{nmod}}}^{13}$ \footnotesize statements$_{11_{\textsc{obj}}}^{14}$ \colorbox{blue!30}{\footnotesize ,}$_{14_{\textsc{p}}}^{15}$ \colorbox{blue!30}{\footnotesize known}$_{14_{\textsc{appo}}}^{16}$ \footnotesize to$_{16_{\textsc{adv}}}^{17 \colorbox{red!30}{\tiny p}}$ \footnotesize everybody$_{17_{\textsc{pmod}}}^{18}$ \colorbox{blue!30}{\footnotesize to}$_{16_{\textsc{oprd}}}^{19}$ \footnotesize be$_{19_{\textsc{im}}}^{20}$ \colorbox{blue!30}{\footnotesize lies}$_{20_{\textsc{prd}}}^{21}$ \footnotesize ,$_{2_{\textsc{p}}}^{22}$ \footnotesize without$_{2_{\textsc{mnr}}}^{23 \colorbox{red!30}{\tiny p}}$ \footnotesize even$_{25_{\textsc{adv}}}^{24}$ \footnotesize being$_{23_{\textsc{pmod}}}^{25}$ \footnotesize ridiculed$_{25_{\textsc{vc}}}^{26}$ \footnotesize in$_{26_{\textsc{loc}}}^{27 \colorbox{red!30}{\tiny p}}$ \footnotesize your$_{30_{\textsc{nmod}}}^{28}$ \footnotesize own$_{30_{\textsc{nmod}}}^{29}$ \footnotesize milieu$_{27_{\textsc{pmod}}}^{30 \colorbox{red!30}{\tiny u}}$ \footnotesize ,$_{32_{\textsc{p}}}^{31}$ \colorbox{blue!30}{\footnotesize can}$_{0_{\textsc{root}}}^{32}$ \footnotesize only$_{32_{\textsc{adv}}}^{33}$ \footnotesize happen$_{32_{\textsc{vc}}}^{34}$ \footnotesize in$_{34_{\textsc{loc}}}^{35 \colorbox{red!30}{\tiny p}}$ \footnotesize this$_{37_{\textsc{nmod}}}^{36}$ \footnotesize region$_{35_{\textsc{pmod}}}^{37}$ \colorbox{blue!30}{\footnotesize .}$_{32_{\textsc{p}}}^{38}$\\\hline

\footnotesize The$_{3_{\textsc{nmod}}}^{1}$ \footnotesize only$_{3_{\textsc{nmod}}}^{2}$ \colorbox{blue!30}{\footnotesize thing}$_{7_{\textsc{sbj}}}^{3}$ \footnotesize that$_{5_{\textsc{dep}}}^{4}$ \footnotesize was$_{3_{\textsc{nmod}}}^{5}$ \colorbox{blue!30}{\footnotesize edible}$_{5_{\textsc{prd}}}^{6}$ \colorbox{blue!30}{\footnotesize was}$_{0_{\textsc{root}}}^{7}$ \footnotesize the$_{10_{\textsc{nmod}}}^{8}$ \footnotesize steamed$_{10_{\textsc{nmod}}}^{9 \colorbox{red!30}{\tiny u}}$ \footnotesize rice$_{7_{\textsc{prd}}}^{10}$ \colorbox{blue!30}{\footnotesize and}$_{7_{\textsc{coord}}}^{11 \colorbox{red!30}{\tiny c}}$ \footnotesize the$_{15_{\textsc{nmod}}}^{12}$ \colorbox{blue!30}{\footnotesize vegetable}$_{15_{\textsc{nmod}}}^{13}$ \footnotesize lo$_{15_{\textsc{nmod}}}^{14 \colorbox{red!30}{\tiny u}}$ \colorbox{blue!30}{\footnotesize mein}$_{16_{\textsc{sbj}}}^{15 \colorbox{red!30}{\tiny u}}$ \colorbox{blue!30}{\footnotesize was}$_{11_{\textsc{conj}}}^{16}$ \footnotesize barely$_{18_{\textsc{amod}}}^{17}$ \footnotesize tolerable$_{16_{\textsc{prd}}}^{18}$ \colorbox{blue!30}{\footnotesize .}$_{7_{\textsc{p}}}^{19}$\\\hline

\footnotesize The$_{2_{\textsc{nmod}}}^{1}$ \colorbox{blue!30}{\footnotesize asparagus}$_{10_{\textsc{sbj}}}^{2 \colorbox{red!30}{\tiny u}}$ \colorbox{blue!30}{\footnotesize ,}$_{2_{\textsc{p}}}^{3}$ \colorbox{blue!30}{\footnotesize seared}$_{5_{\textsc{nmod}}}^{4 \colorbox{red!30}{\tiny u}}$ \colorbox{blue!30}{\footnotesize tuna}$_{2_{\textsc{coord}}}^{5}$ \colorbox{blue!30}{\footnotesize ,}$_{5_{\textsc{p}}}^{6}$ \colorbox{blue!30}{\footnotesize and}$_{5_{\textsc{coord}}}^{7 \colorbox{red!30}{\tiny c}}$ \footnotesize lobster$_{9_{\textsc{nmod}}}^{8}$ \colorbox{blue!30}{\footnotesize tail}$_{7_{\textsc{conj}}}^{9}$ \colorbox{blue!30}{\footnotesize were}$_{0_{\textsc{root}}}^{10}$ \footnotesize the$_{12_{\textsc{nmod}}}^{11}$ \footnotesize best$_{10_{\textsc{prd}}}^{12}$ \footnotesize we$_{15_{\textsc{sbj}}}^{13}$ \footnotesize ever$_{15_{\textsc{tmp}}}^{14}$ \footnotesize had$_{12_{\textsc{nmod}}}^{15}$ \colorbox{blue!30}{\footnotesize .}$_{10_{\textsc{p}}}^{16}$\\\hline

\hline
\end{tabular}
\end{center}
\caption[The example sentences that have been improved by the parse score-based self-training approach when compared to the baseline.]{\label{table:analysis-examples-ps-self-training} The example sentences that have been improved by the parse score-based self-training approach when compared to the baseline. In which the dependency head/relation of a token are marked as the subscript, while the superscript is the index of token. The unknown words, prepositions and conjunctions are highlighted with \colorbox{red!30}{u}, \colorbox{red!30}{p} and \colorbox{red!30}{c} respectively. We highlight the different levels of the improvements achieved by our parse score-based self-training model on the dependency edges by different colours. In which the \colorbox{blue!30}{blue} colour means both head and label are corrected, the \colorbox{yellow!30}{yellow} colour means only the head is corrected and the \colorbox{green!30}{green} colour means only the label is corrected.}
\end{table}

%examples
\begin{table}
\begin{center}
\begin{tabular}{p{\linewidth}}
\hline
\hline
\colorbox{blue!30}{\footnotesize But}$_{31_{\textsc{dep}}}^{1 \colorbox{red!30}{\tiny c}}$ \colorbox{blue!30}{\footnotesize creating}$_{31_{\textsc{sbj}}}^{2}$ \footnotesize a$_{5_{\textsc{nmod}}}^{3}$ \footnotesize balanced$_{5_{\textsc{nmod}}}^{4}$ \footnotesize community$_{2_{\textsc{obj}}}^{5}$ \footnotesize with$_{5_{\textsc{nmod}}}^{6 \colorbox{red!30}{\tiny p}}$ \footnotesize a$_{8_{\textsc{nmod}}}^{7}$ \footnotesize mix$_{6_{\textsc{pmod}}}^{8}$ \footnotesize of$_{8_{\textsc{nmod}}}^{9 \colorbox{red!30}{\tiny p}}$ \footnotesize housing$_{9_{\textsc{pmod}}}^{10}$ \footnotesize ,$_{10_{\textsc{p}}}^{11}$ \footnotesize offices$_{10_{\textsc{coord}}}^{12}$ \footnotesize ,$_{12_{\textsc{p}}}^{13}$ \footnotesize shopping$_{12_{\textsc{coord}}}^{14}$ \footnotesize and$_{14_{\textsc{coord}}}^{15 \colorbox{red!30}{\tiny c}}$ \footnotesize other$_{17_{\textsc{nmod}}}^{16}$ \footnotesize amenities$_{15_{\textsc{conj}}}^{17}$ \footnotesize --$_{5_{\textsc{p}}}^{18}$ \footnotesize allowing$_{5_{\textsc{appo}}}^{19}$ \footnotesize people$_{19_{\textsc{obj}}}^{20}$ \footnotesize to$_{19_{\textsc{oprd}}}^{21}$ \footnotesize live$_{21_{\textsc{im}}}^{22}$ \footnotesize close$_{22_{\textsc{loc}}}^{23}$ \footnotesize to$_{23_{\textsc{amod}}}^{24 \colorbox{red!30}{\tiny p}}$ \colorbox{blue!30}{\footnotesize where}$_{27_{\textsc{loc}}}^{25}$ \footnotesize they$_{27_{\textsc{sbj}}}^{26}$ \footnotesize work$_{24_{\textsc{pmod}}}^{27}$ \footnotesize and$_{27_{\textsc{coord}}}^{28 \colorbox{red!30}{\tiny c}}$ \footnotesize play$_{28_{\textsc{conj}}}^{29}$ \footnotesize --$_{5_{\textsc{p}}}^{30}$ \colorbox{blue!30}{\footnotesize is}$_{0_{\textsc{root}}}^{31}$ \footnotesize an$_{36_{\textsc{nmod}}}^{32}$ \colorbox{blue!30}{\footnotesize even}$_{35_{\textsc{amod}}}^{33}$ \colorbox{blue!30}{\footnotesize more}$_{35_{\textsc{amod}}}^{34}$ \colorbox{blue!30}{\footnotesize worthy}$_{36_{\textsc{nmod}}}^{35}$ \footnotesize goal$_{31_{\textsc{prd}}}^{36}$ \colorbox{blue!30}{\footnotesize .}$_{31_{\textsc{p}}}^{37}$\\\hline

\colorbox{blue!30}{\footnotesize Her}$_{4_{\textsc{nmod}}}^{1}$ \footnotesize ``$_{4_{\textsc{p}}}^{2}$ \footnotesize Rubble$_{4_{\textsc{name}}}^{3}$ \colorbox{green!30}{\footnotesize Division}$_{6_{\textsc{sbj}}}^{4}$ \footnotesize ''$_{4_{\textsc{p}}}^{5}$ \colorbox{blue!30}{\footnotesize mixes}$_{0_{\textsc{root}}}^{6}$ \footnotesize such$_{9_{\textsc{nmod}}}^{7}$ \footnotesize disparate$_{9_{\textsc{nmod}}}^{8}$ \colorbox{blue!30}{\footnotesize materials}$_{6_{\textsc{obj}}}^{9}$ \footnotesize as$_{9_{\textsc{nmod}}}^{10 \colorbox{red!30}{\tiny p}}$ \footnotesize ink$_{13_{\textsc{nmod}}}^{11}$ \footnotesize -$_{13_{\textsc{nmod}}}^{12}$ \footnotesize jet$_{14_{\textsc{nmod}}}^{13}$ \colorbox{blue!30}{\footnotesize prints}$_{10_{\textsc{pmod}}}^{14}$ \colorbox{blue!30}{\footnotesize pasted}$_{14_{\textsc{appo}}}^{15 \colorbox{red!30}{\tiny u}}$ \colorbox{blue!30}{\footnotesize on}$_{15_{\textsc{loc}}}^{16 \colorbox{red!30}{\tiny p}}$ \footnotesize board$_{16_{\textsc{pmod}}}^{17}$ \footnotesize ,$_{14_{\textsc{p}}}^{18}$ \footnotesize foam$_{20_{\textsc{nmod}}}^{19}$ \footnotesize rubber$_{14_{\textsc{coord}}}^{20}$ \footnotesize ,$_{20_{\textsc{p}}}^{21}$ \footnotesize galvanized$_{23_{\textsc{nmod}}}^{22}$ \footnotesize steel$_{20_{\textsc{coord}}}^{23}$ \footnotesize ,$_{23_{\textsc{p}}}^{24}$ \footnotesize concrete$_{23_{\textsc{coord}}}^{25}$ \footnotesize ,$_{25_{\textsc{p}}}^{26}$ \footnotesize steel$_{28_{\textsc{nmod}}}^{27}$ \footnotesize rebar$_{25_{\textsc{coord}}}^{28 \colorbox{red!30}{\tiny u}}$ \footnotesize and$_{28_{\textsc{coord}}}^{29 \colorbox{red!30}{\tiny c}}$ \footnotesize bungee$_{31_{\textsc{nmod}}}^{30 \colorbox{red!30}{\tiny u}}$ \footnotesize cords$_{29_{\textsc{conj}}}^{31 \colorbox{red!30}{\tiny u}}$ \colorbox{blue!30}{\footnotesize .}$_{6_{\textsc{p}}}^{32}$\\\hline

\footnotesize Go$_{0_{\textsc{root}}}^{1}$ \footnotesize look$_{1_{\textsc{oprd}}}^{2}$ \footnotesize at$_{2_{\textsc{adv}}}^{3 \colorbox{red!30}{\tiny p}}$ \footnotesize DHRM$_{3_{\textsc{pmod}}}^{4 \colorbox{red!30}{\tiny u}}$ \footnotesize and$_{4_{\textsc{coord}}}^{5 \colorbox{red!30}{\tiny c}}$ \footnotesize the$_{9_{\textsc{nmod}}}^{6}$ \footnotesize state$_{9_{\textsc{nmod}}}^{7}$ \footnotesize courts$_{9_{\textsc{nmod}}}^{8}$ \footnotesize system$_{5_{\textsc{conj}}}^{9}$ \footnotesize ($_{9_{\textsc{p}}}^{10}$ \colorbox{blue!30}{\footnotesize separate}$_{13_{\textsc{nmod}}}^{11}$ \footnotesize HR$_{13_{\textsc{nmod}}}^{12 \colorbox{red!30}{\tiny u}}$ \colorbox{blue!30}{\footnotesize dept}$_{9_{\textsc{appo}}}^{13 \colorbox{red!30}{\tiny u}}$ \footnotesize )$_{9_{\textsc{p}}}^{14}$ \footnotesize and$_{2_{\textsc{coord}}}^{15 \colorbox{red!30}{\tiny c}}$ \footnotesize see$_{15_{\textsc{conj}}}^{16}$ \footnotesize what$_{21_{\textsc{obj}}}^{17}$ \colorbox{blue!30}{\footnotesize the}$_{20_{\textsc{nmod}}}^{18}$ \colorbox{green!30}{\footnotesize state}$_{20_{\textsc{nmod}}}^{19}$ \colorbox{blue!30}{\footnotesize folks}$_{21_{\textsc{sbj}}}^{20}$ \colorbox{blue!30}{\footnotesize do}$_{16_{\textsc{obj}}}^{21}$ \colorbox{blue!30}{\footnotesize ,}$_{21_{\textsc{p}}}^{22}$ \colorbox{blue!30}{\footnotesize and}$_{21_{\textsc{coord}}}^{23 \colorbox{red!30}{\tiny c}}$ \footnotesize who$_{30_{\textsc{obj}}}^{24}$ \footnotesize all$_{26_{\textsc{nmod}}}^{25}$ \footnotesize you$_{27_{\textsc{sbj}}}^{26}$ \footnotesize 're$_{23_{\textsc{conj}}}^{27}$ \footnotesize talking$_{27_{\textsc{vc}}}^{28}$ \footnotesize about$_{28_{\textsc{adv}}}^{29 \colorbox{red!30}{\tiny p}}$ \footnotesize furloughing$_{29_{\textsc{pmod}}}^{30 \colorbox{red!30}{\tiny u}}$ \footnotesize .$_{1_{\textsc{p}}}^{31}$\\\hline

\colorbox{blue!30}{\footnotesize and}$_{3_{\textsc{dep}}}^{1 \colorbox{red!30}{\tiny c}}$ \colorbox{green!30}{\footnotesize i}$_{3_{\textsc{sbj}}}^{2}$ \colorbox{blue!30}{\footnotesize promise}$_{0_{\textsc{root}}}^{3}$ \colorbox{green!30}{\footnotesize to}$_{3_{\textsc{oprd}}}^{4}$ \footnotesize fess$_{4_{\textsc{im}}}^{5 \colorbox{red!30}{\tiny u}}$ \footnotesize up$_{5_{\textsc{prt}}}^{6}$ \footnotesize eventually$_{5_{\textsc{tmp}}}^{7}$ \footnotesize and$_{5_{\textsc{coord}}}^{8 \colorbox{red!30}{\tiny c}}$ \footnotesize tell$_{8_{\textsc{conj}}}^{9}$ \footnotesize of$_{9_{\textsc{adv}}}^{10 \colorbox{red!30}{\tiny p}}$ \footnotesize at$_{13_{\textsc{dep}}}^{11 \colorbox{red!30}{\tiny p}}$ \footnotesize least$_{11_{\textsc{amod}}}^{12}$ \colorbox{blue!30}{\footnotesize one}$_{15_{\textsc{nmod}}}^{13}$ \colorbox{blue!30}{\footnotesize such}$_{15_{\textsc{nmod}}}^{14}$ \colorbox{blue!30}{\footnotesize epic}$_{10_{\textsc{pmod}}}^{15}$ \colorbox{blue!30}{\footnotesize i}$_{17_{\textsc{sbj}}}^{16}$ \colorbox{blue!30}{\footnotesize survived}$_{15_{\textsc{nmod}}}^{17}$ \colorbox{blue!30}{\footnotesize --}$_{3_{\textsc{p}}}^{18}$\\\hline

\colorbox{blue!30}{\footnotesize In}$_{11_{\textsc{adv}}}^{1 \colorbox{red!30}{\tiny p}}$ \footnotesize fact$_{1_{\textsc{pmod}}}^{2}$ \colorbox{blue!30}{\footnotesize ,}$_{11_{\textsc{p}}}^{3}$ \footnotesize the$_{6_{\textsc{nmod}}}^{4}$ \footnotesize MINI$_{6_{\textsc{name}}}^{5 \colorbox{red!30}{\tiny u}}$ \colorbox{blue!30}{\footnotesize COOPER}$_{11_{\textsc{sbj}}}^{6}$ \colorbox{blue!30}{\footnotesize she}$_{8_{\textsc{sbj}}}^{7}$ \colorbox{blue!30}{\footnotesize was}$_{6_{\textsc{nmod}}}^{8}$ \colorbox{blue!30}{\footnotesize riding}$_{8_{\textsc{vc}}}^{9}$ \colorbox{green!30}{\footnotesize in}$_{9_{\textsc{adv}}}^{10 \colorbox{red!30}{\tiny p}}$ \colorbox{blue!30}{\footnotesize is}$_{0_{\textsc{root}}}^{11}$ \footnotesize not$_{11_{\textsc{adv}}}^{12}$ \footnotesize what$_{16_{\textsc{obj}}}^{13}$ \footnotesize the$_{15_{\textsc{nmod}}}^{14}$ \footnotesize reports$_{16_{\textsc{sbj}}}^{15}$ \footnotesize said$_{11_{\textsc{prd}}}^{16}$ \colorbox{blue!30}{\footnotesize .}$_{11_{\textsc{p}}}^{17}$\\\hline

\footnotesize But$_{3_{\textsc{dep}}}^{1 \colorbox{red!30}{\tiny c}}$ \footnotesize everyone$_{3_{\textsc{sbj}}}^{2}$ \footnotesize needs$_{0_{\textsc{root}}}^{3}$ \footnotesize to$_{3_{\textsc{oprd}}}^{4}$ \footnotesize recognize$_{4_{\textsc{im}}}^{5}$ \footnotesize that$_{5_{\textsc{obj}}}^{6 \colorbox{red!30}{\tiny p}}$ \footnotesize Arlington$_{9_{\textsc{nmod}}}^{7}$ \footnotesize 's$_{7_{\textsc{suffix}}}^{8}$ \colorbox{blue!30}{\footnotesize decision}$_{16_{\textsc{sbj}}}^{9}$ \footnotesize not$_{11_{\textsc{adv}}}^{10}$ \footnotesize to$_{9_{\textsc{nmod}}}^{11}$ \footnotesize pursue$_{11_{\textsc{im}}}^{12}$ \colorbox{blue!30}{\footnotesize a}$_{15_{\textsc{nmod}}}^{13}$ \colorbox{blue!30}{\footnotesize balanced}$_{15_{\textsc{nmod}}}^{14}$ \colorbox{blue!30}{\footnotesize community}$_{12_{\textsc{obj}}}^{15}$ \colorbox{blue!30}{\footnotesize means}$_{6_{\textsc{sub}}}^{16}$ \colorbox{blue!30}{\footnotesize that}$_{16_{\textsc{obj}}}^{17 \colorbox{red!30}{\tiny p}}$ \footnotesize housing$_{19_{\textsc{sbj}}}^{18}$ \colorbox{blue!30}{\footnotesize will}$_{17_{\textsc{sub}}}^{19}$ \footnotesize end$_{19_{\textsc{vc}}}^{20}$ \footnotesize up$_{20_{\textsc{prt}}}^{21}$ \footnotesize somewhere$_{20_{\textsc{loc}}}^{22}$ \footnotesize else$_{22_{\textsc{amod}}}^{23}$ \footnotesize ,$_{22_{\textsc{p}}}^{24}$ \footnotesize presumably$_{26_{\textsc{pmod}}}^{25}$ \footnotesize in$_{22_{\textsc{amod}}}^{26 \colorbox{red!30}{\tiny p}}$ \colorbox{blue!30}{\footnotesize outlying}$_{28_{\textsc{nmod}}}^{27}$ \colorbox{blue!30}{\footnotesize counties}$_{26_{\textsc{pmod}}}^{28}$ \footnotesize .$_{3_{\textsc{p}}}^{29}$\\\hline

\footnotesize Jim$_{0_{\textsc{root}}}^{1}$ \colorbox{blue!30}{\footnotesize ,}$_{1_{\textsc{p}}}^{2}$ \colorbox{blue!30}{\footnotesize on}$_{1_{\textsc{adv}}}^{3 \colorbox{red!30}{\tiny p}}$ \footnotesize behalf$_{3_{\textsc{pmod}}}^{4}$ \footnotesize of$_{4_{\textsc{nmod}}}^{5 \colorbox{red!30}{\tiny p}}$ \footnotesize the$_{11_{\textsc{nmod}}}^{6}$ \footnotesize Virginia$_{11_{\textsc{name}}}^{7}$ \footnotesize Recycling$_{9_{\textsc{name}}}^{8}$ \footnotesize Markets$_{10_{\textsc{nmod}}}^{9}$ \footnotesize Development$_{11_{\textsc{nmod}}}^{10}$ \footnotesize Council$_{5_{\textsc{pmod}}}^{11}$ \footnotesize and$_{11_{\textsc{coord}}}^{12 \colorbox{red!30}{\tiny c}}$ \colorbox{blue!30}{\footnotesize the}$_{15_{\textsc{nmod}}}^{13}$ \colorbox{blue!30}{\footnotesize Mid-Atlantic}$_{15_{\textsc{name}}}^{14 \colorbox{red!30}{\tiny u}}$ \colorbox{blue!30}{\footnotesize Consortium}$_{12_{\textsc{conj}}}^{15}$ \colorbox{blue!30}{\footnotesize of}$_{15_{\textsc{nmod}}}^{16 \colorbox{red!30}{\tiny p}}$ \footnotesize Recycling$_{21_{\textsc{nmod}}}^{17}$ \colorbox{blue!30}{\footnotesize and}$_{17_{\textsc{coord}}}^{18 \colorbox{red!30}{\tiny c}}$ \footnotesize Economic$_{20_{\textsc{name}}}^{19}$ \footnotesize Development$_{18_{\textsc{conj}}}^{20}$ \colorbox{blue!30}{\footnotesize Officials}$_{16_{\textsc{pmod}}}^{21}$ \footnotesize ($_{15_{\textsc{p}}}^{22}$ \footnotesize MACREDO$_{15_{\textsc{appo}}}^{23 \colorbox{red!30}{\tiny u}}$ \footnotesize )$_{15_{\textsc{p}}}^{24}$ \colorbox{blue!30}{\footnotesize ,}$_{1_{\textsc{p}}}^{25}$ \footnotesize thanks$_{1_{\textsc{dep}}}^{26}$ \footnotesize for$_{26_{\textsc{nmod}}}^{27 \colorbox{red!30}{\tiny p}}$ \colorbox{blue!30}{\footnotesize plugging}$_{27_{\textsc{pmod}}}^{28}$ \colorbox{blue!30}{\footnotesize e-cycling}$_{28_{\textsc{obj}}}^{29 \colorbox{red!30}{\tiny u}}$ \footnotesize .$_{1_{\textsc{p}}}^{30}$\\\hline

\colorbox{blue!30}{\footnotesize when}$_{12_{\textsc{tmp}}}^{1}$ \footnotesize the$_{3_{\textsc{nmod}}}^{2}$ \colorbox{blue!30}{\footnotesize guy}$_{12_{\textsc{sbj}}}^{3}$ \colorbox{green!30}{\footnotesize (}$_{9_{\textsc{p}}}^{4}$ \footnotesize the$_{6_{\textsc{nmod}}}^{5}$ \colorbox{blue!30}{\footnotesize owner}$_{9_{\textsc{dep}}}^{6}$ \footnotesize ,$_{9_{\textsc{p}}}^{7}$ \footnotesize it$_{9_{\textsc{sbj}}}^{8}$ \colorbox{blue!30}{\footnotesize turned}$_{3_{\textsc{prn}}}^{9}$ \footnotesize out$_{9_{\textsc{prt}}}^{10}$ \footnotesize )$_{9_{\textsc{p}}}^{11}$ \colorbox{blue!30}{\footnotesize arrived}$_{20_{\textsc{tmp}}}^{12}$ \colorbox{yellow!30}{\footnotesize to}$_{12_{\textsc{prp}}}^{13}$ \footnotesize open$_{13_{\textsc{im}}}^{14}$ \footnotesize the$_{17_{\textsc{nmod}}}^{15}$ \footnotesize gas$_{17_{\textsc{nmod}}}^{16}$ \footnotesize station$_{14_{\textsc{obj}}}^{17}$ \colorbox{blue!30}{\footnotesize ,}$_{20_{\textsc{p}}}^{18}$ \footnotesize he$_{20_{\textsc{sbj}}}^{19}$ \colorbox{blue!30}{\footnotesize took}$_{0_{\textsc{root}}}^{20}$ \footnotesize one$_{22_{\textsc{nmod}}}^{21}$ \footnotesize look$_{20_{\textsc{obj}}}^{22}$ \footnotesize at$_{20_{\textsc{adv}}}^{23 \colorbox{red!30}{\tiny p}}$ \footnotesize our$_{26_{\textsc{nmod}}}^{24}$ \footnotesize cow$_{26_{\textsc{nmod}}}^{25 \colorbox{red!30}{\tiny u}}$ \footnotesize pie$_{23_{\textsc{pmod}}}^{26}$ \footnotesize with$_{26_{\textsc{nmod}}}^{27 \colorbox{red!30}{\tiny p}}$ \footnotesize wheels$_{27_{\textsc{pmod}}}^{28}$ \footnotesize and$_{20_{\textsc{coord}}}^{29 \colorbox{red!30}{\tiny c}}$ \footnotesize said$_{29_{\textsc{conj}}}^{30}$ \footnotesize ``$_{30_{\textsc{p}}}^{31}$ \footnotesize what$_{34_{\textsc{nmod}}}^{32}$ \footnotesize the$_{34_{\textsc{nmod}}}^{33}$ \footnotesize fook$_{30_{\textsc{obj}}}^{34 \colorbox{red!30}{\tiny u}}$ \colorbox{blue!30}{\footnotesize ?}$_{20_{\textsc{p}}}^{35}$ \colorbox{blue!30}{\footnotesize ''}$_{20_{\textsc{p}}}^{36}$\\\hline

\colorbox{blue!30}{\footnotesize In}$_{26_{\textsc{loc}}}^{1 \colorbox{red!30}{\tiny p}}$ \footnotesize Pakistan$_{4_{\textsc{nmod}}}^{2}$ \footnotesize national$_{4_{\textsc{nmod}}}^{3}$ \footnotesize chart$_{1_{\textsc{pmod}}}^{4}$ \footnotesize besides$_{26_{\textsc{adv}}}^{5 \colorbox{red!30}{\tiny p}}$ \footnotesize the$_{8_{\textsc{nmod}}}^{6}$ \footnotesize transit$_{8_{\textsc{nmod}}}^{7}$ \colorbox{blue!30}{\footnotesize affliction}$_{5_{\textsc{pmod}}}^{8}$ \footnotesize to$_{8_{\textsc{nmod}}}^{9 \colorbox{red!30}{\tiny p}}$ \footnotesize transit$_{11_{\textsc{nmod}}}^{10}$ \footnotesize Venus$_{9_{\textsc{pmod}}}^{11}$ \footnotesize in$_{8_{\textsc{loc}}}^{12 \colorbox{red!30}{\tiny p}}$ \footnotesize the$_{15_{\textsc{nmod}}}^{13}$ \footnotesize fourth$_{15_{\textsc{nmod}}}^{14}$ \footnotesize house$_{12_{\textsc{pmod}}}^{15}$ \footnotesize by$_{8_{\textsc{nmod}}}^{16 \colorbox{red!30}{\tiny p}}$ \footnotesize FMs$_{18_{\textsc{nmod}}}^{17 \colorbox{red!30}{\tiny u}}$ \footnotesize Rahu$_{16_{\textsc{pmod}}}^{18 \colorbox{red!30}{\tiny u}}$ \footnotesize and$_{18_{\textsc{coord}}}^{19 \colorbox{red!30}{\tiny c}}$ \footnotesize Mercury$_{19_{\textsc{conj}}}^{20}$ \colorbox{blue!30}{\footnotesize ,}$_{26_{\textsc{p}}}^{21}$ \colorbox{green!30}{\footnotesize natal}$_{23_{\textsc{nmod}}}^{22 \colorbox{red!30}{\tiny u}}$ \colorbox{blue!30}{\footnotesize Saturn}$_{26_{\textsc{sbj}}}^{23}$ \footnotesize and$_{23_{\textsc{coord}}}^{24 \colorbox{red!30}{\tiny c}}$ \footnotesize Venus$_{24_{\textsc{conj}}}^{25}$ \colorbox{blue!30}{\footnotesize are}$_{0_{\textsc{root}}}^{26}$ \footnotesize also$_{26_{\textsc{adv}}}^{27}$ \footnotesize under$_{26_{\textsc{prd}}}^{28 \colorbox{red!30}{\tiny p}}$ \footnotesize the$_{31_{\textsc{nmod}}}^{29}$ \footnotesize close$_{31_{\textsc{nmod}}}^{30}$ \footnotesize affliction$_{28_{\textsc{pmod}}}^{31}$ \footnotesize of$_{31_{\textsc{nmod}}}^{32 \colorbox{red!30}{\tiny p}}$ \colorbox{yellow!30}{\footnotesize transit}$_{34_{\textsc{nmod}}}^{33}$ \colorbox{blue!30}{\footnotesize Rahu}$_{32_{\textsc{pmod}}}^{34 \colorbox{red!30}{\tiny u}}$ \colorbox{blue!30}{\footnotesize .}$_{26_{\textsc{p}}}^{35}$\\\hline

\hline
\end{tabular}
\end{center}
\caption[The example sentences that have been improved by the Delta-based self-training approach when compared to the baseline.]{\label{table:analysis-examples-delta-self-training} The example sentences that have been improved by the Delta-based self-training approach when compared to the baseline. In which the dependency head/relation of a token are marked as the subscript, while the superscript is the index of token. The unknown words, prepositions and conjunctions are highlighted with \colorbox{red!30}{u}, \colorbox{red!30}{p} and \colorbox{red!30}{c} respectively. We highlight the different levels of the improvements achieved by our Delta-based self-training model on the dependency edges by different colours. In which the \colorbox{blue!30}{blue} colour means both head and label are corrected, the \colorbox{yellow!30}{yellow} colour means only the head is corrected and the \colorbox{green!30}{green} colour means only the label is corrected.}
\end{table}

\textbf{Example Sentences.} Table \ref{table:analysis-examples-ps-self-training} and table \ref{table:analysis-examples-delta-self-training} present example sentences that have been improved by the parse score-based and the Delta-based self-training approaches respectively. We choose four sentences (the first four sentences) that have been largely improved by both approaches, as we can see from table the improvements achieved by both models are very similar, some are even identical.

\section{Chapter Summary}\label{section:self-en-conclusion}
In this chapter, we introduced two novel confidence-based self-training approaches to domain adaptation for dependency parsing. 
We compared a self-training approach that uses random selection and two confidence-based approaches. 
The random selection-based self-training method did \textit{not} improve the accuracy which is in line with previously published negative results, 
both confidence-based methods achieved statistically significant improvements and showed relatively high accuracy gains. 

We tested both confidence-based approaches on three web related domains of our main evaluation corpora (\textsc{Weblogs, Newsgroups, Reviews}) and the \textsc{Chemical} domain. Our confidence-based approaches achieved statistically significant improvements in all tested domains. For web domains, we gained up to 0.8 percentage points for both labelled and unlabelled accuracies. On average the Delta-based approach improved the accuracy by 0.6\% for both labelled and unlabelled accuracies. Similarly, the parse score-based method improved labelled accuracy scores by 0.6\% and unlabelled accuracy scores by 0.5\%.
In terms of the \textsc{Chemical} domain, the Delta-based and the parse score-based approaches gained 1.42\% and 1.12\% labelled accuracies respectively when using predicted PoS tags. 
When we used gold PoS tags, a larger labelled improvement of 1.62\% is achieved by the Delta method and 1.48\% is gained by the parse score method. The unlabelled improvements for both methods are similar to their labelled improvements for all the experiments. In total, our approaches achieved significantly better accuracy for all four domains. 

We conclude from the experiments that self-training based on confidence is worth applying in a domain adaptation scenario and that a confidence-based self-training approach seems to be crucial for the successful application of self-training in dependency parsing. Our evaluation underlines the finding that the pre-selection of parse trees is probably a precondition that self-training becomes effective in the case of dependency parsing and to reach a significant accuracy gain.

The further analysis compared the behaviour of two approaches and gave a clearer picture of in which part self-training helps most. As a preliminary analysis, we assessed the overlap between the top ranked sentences of two methods. When we compared the top ranked 50\% of the development set by different methods, 56\% of them are identical. As there are more than 40\% sentences which are selected differently by different methods, we expect some clear differences in our in-depth analysis on token and sentence level. Surprisingly, the further analysis suggested that both methods played similar roles on most of the analysis, the behaviour differences are rather small. In our token level analysis, both methods gained large improvements on the root, coordination, modifiers and unclassified relations. We also found much larger unlabelled improvements for unknown words. For sentence level analysis, we noticed that our approaches helped most the medium length sentences (10-30 tokens/sentence). Generally speaking, they also have a better performance on sentences that have certain levels of complexity, such as sentences that have more than 2 unknown words or at least 2 prepositions. This might also because of the simpler sentences have already a reasonably good accuracy when baseline model is used, thus are harder to improve.

\chapter{Multi-lingual Self-training}\label{chapter:multiselftrain}
Self-training approaches have previously been used mainly for English parsing \cite{mcclosky06naacl,mcclosky2006reranking,reichart2007self,kawahara2008learning,sagae2010self,petrov2012overview}. The few successful attempts of using self-training for languages other than English were limited only to a single language \cite{chen2008learning,goutam2011exploring}. The evaluations of using self-training for multiple languages are still found no improvements on accuracies \cite{cerisara2014spmrl,bjorkelund2014spmrl}. 

In the previous chapter we demonstrated the power of the confidence-based self-training on English out-of-domain parsing, the evaluation on four different domains showed large gains. We wonder if the self-training methods could be adapted to other languages. The first problem with going beyond English is the lack of resources. To the best of our knowledge, there is no out-of-domain corpus available for languages other than English. In fact, even for English, the out-of-domain dataset is very limited. Thus, we are not able to evaluate on the same domain adaptation scenario as we did for English. In English evaluation, we do not use any target domain manually annotated data for training, which is a typical domain adaptation scenario that assume no target domain training data is annotated. The other common domain adaptation scenario assumes that there is a small number of target domain training data available. In this chapter, we use a small training set (5,000 sentences) to simulate the latter scenario. The same domain unlabelled set is annotated by the base model to enlarge the training data. Strictly speaking, this is an under-resourced in-domain parsing setting as in the 2014 shared task at the workshop on statistical parsing of morphologically rich language (SPMRL) \cite{seddah2014introducing}.  
More precisely, in this chapter, we evaluate with the adjusted parse score-based method, as both methods have very similar performances and the adjusted parse scores are fast to compute.
We evaluate this method on nine languages (\textsc{Arabic, Basque, French, German, Hebrew, Hungarian, Korean, Polish, Swedish}) corpora of the SPMRL shared task \cite{seddah2014introducing}.

The rest of the chapter are organized as follows: We introduce our approach and experiment settings in Section \ref{section:self-multi-appraoch} and \ref{section:self-multi-setup} respectively. Section \ref{section:self-multi-results} and \ref{section:self-multi-analysis} discusses and analyses the results. We summarise the chapter in Section \ref{section:self-multi-conclusion}.

\section{Multi-lingual Confidence-based Self-training}\label{section:self-multi-appraoch}

Our goal for the multi-lingual experiments is to evaluate the performance of our confidence-based method on more languages. Our previous evaluations on multiple web domains and the \textsc{Chemical} domain showed that our configuration is robust and can be directly used across domains. Thus, in our multi-lingual evaluation we again directly adapt our best configuration from our English evaluation, in which the first half of the ranked auto-annotated dataset is used as additional training data for all the languages. We also do not tune different configurations for individual language, as we want to evaluate the confidence-based self-training in a unified framework. More precisely, our multi-lingual self-training approach consists of a single iteration with the following steps:

\begin{enumerate}
	\item A parser is trained on a (small) initial training set to generate a base model.
	\item We analyse a large number of unlabelled sentences with the base model.
	\item We build a new training set consisting of the initial training set and 50\% newly analysed sentences parsed with a high confidence.
	\item We retrain the parser on the new training set to produce a self-trained model.
	\item Finally, the self-trained model is used to annotate the test set.
\end{enumerate}

%for our approach
\begin{figure}[t]
	\begin{center}
		\begin{tikzpicture}
		\pgfplotsset{
			xmin=-0.002,xmax=0.052,
			ymin=89.5,ymax=91.5,
			xlabel=Value of $d$,
			ylabel=Labeled Attachment Score (\%)
		}
		\begin{axis}
		
		%Adjusted parse score
		\addplot[smooth,mark=triangle*,mark options={fill=white}]
		coordinates {(0,89.8)(0.005,90.8	)(0.01,91.1)(0.015,91.1)(0.02,90.9)(0.025,90.7)(0.03,90.4)(0.035,90.1)(0.04,90.1)(0.045,90.0)(0.05,90.1)};
		\addlegendentry{\tiny Adjusted Parse Score}

		\end{axis}
		
		\end{tikzpicture}
	
	\end{center}
	\caption{\label{figure:self-multi-dvalue} Accuracies of sentences which have a position number within the top 50\% after ranking the auto-parsed sentences of \textsc{German} development set by the adjusted parse scores with different values of $d$.}
\end{figure}
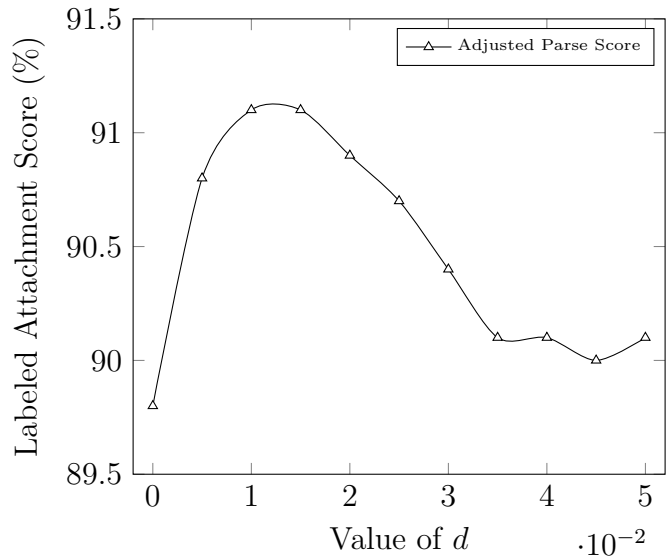

%for our approach
\begin{figure}[t]
	\begin{center}
		\begin{tikzpicture}
		\pgfplotsset{
			xmin=0,xmax=100,
			ymin=85,ymax=95,
			xlabel=Percentage of Sentences (\%),
			ylabel=Labeled Attachment Score (\%)
		}
		\begin{axis}

		%parse score
		\addplot[smooth,mark=triangle*,mark options={fill=white}]
		coordinates {(10,92.55)(20,91.87)(30,91.23)(40,90.31)(50,89.83)(60,88.89)(70,87.94)(80,87.09)(90,86.5)(100,86.49)};
		\addlegendentry{\tiny Original Parse Score}
		
		%adjusted parse score
		\addplot[smooth,mark=square*, mark options={fill=white}]
		coordinates{(10,93.88)(20,93.4)(30,92.53)(40,91.86)(50,91.14)(60,90.3)(70,89.46)(80,88.44)(90,87.3)(100,86.49)};
		\addlegendentry{\tiny Adjusted Parse Score}
		
		%baseline
		\addplot[smooth,dashed] 
		coordinates{(0,86.49) (100,86.49)};
		\addlegendentry{\tiny Average Accuracy}
		
		\end{axis}
		
		\end{tikzpicture}
	\end{center}
	\caption{\label{figure:self-multi-accuracy-assess} The accuracies when inspecting 10-100\% sentences of the \textsc{German} development set ranked by the confidence-based methods.}
\end{figure}
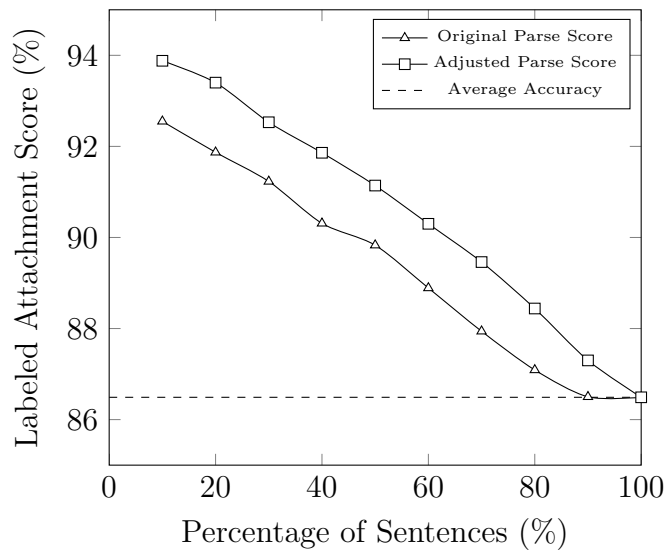

Here we give a recap of our adjusted parse score method and confirm the correlation between accuracy and the adjusted parse scores on the multi-lingual development set. The adjusted parse score method which we proposed in the previous chapter is mainly based on the observation that the parse scores of sentences are correlated with their accuracies. However, the original parse scores are sensitive to sentence length, in which longer sentences usually have higher scores. To tackle this problem, we introduce a simple but effective adjustment on the scores. The original parse score of an auto-parsed sentence ($Score_{original}$) is subtracted by its sentence length ($L$) multiplied by a fixed number $d$. More precisely, the adjusted parse scores are calculated as follows:

\begin{equation}
Score_{adjusted} = Score_{original}-L \times d
\vspace{0.3cm}
\end{equation}

To obtain the constant $d$, we apply the defined equation with different values of $d$ to all sentences of the development set and rank the sentences by their adjusted scores in a descending order.  Let $No(i)$ be the position number of the $i_{th}$ sentence after ranking them by the adjusted scores. The value of $d$ is selected to maximize the accuracy of sentences that have a $No(i)$ within the top 50\%. We evaluate stepwise different values of $d$ from 0 to 0.05 with an increment of 0.005. The highest accuracy of the top ranked sentences is achieved when $d=0.015$ (see Figure \ref{figure:self-multi-dvalue}), thus $d$ is set to 0.015 in our experiments. The $d$ value used in our English evaluations is the same 0.015, this shows a stability of our equation. Figure \ref{figure:self-multi-accuracy-assess} shows the accuracies when inspecting 10 -100\% of sentences ranked by adjusted and original parse scores. We found that adjusted parse scores lead to a higher correlation with accuracies compared to original parse scores. This is in line with our finding in previous evaluation on English out-of-domain data.

\section{Experiment Set-up}\label{section:self-multi-setup}

\begin{table}
	\begin{center}
		\begin{tabular}{l|rrrrr}
			\hline &\sc Arabic & \sc Basque&\sc French&\sc German&\sc Hebrew \\ \hline\hline
			\bf train: &&&&&\\ \hline 
			\bf Sentences &5,000&5,000&5,000&5,000&5,000\\ 
			\bf Tokens &224,907& 61,905& 150,984& 87,841& 128,046\\
			\bf Avg. Length&44.98& 12.38 &30.19& 17.56& 25.60\\\hline\hline
			\bf test: &&&&&\\ \hline 
			\bf Sentences &1,959& 946& 2,541& 5,000& 716\\ 
			\bf Tokens &73,878& 11,457& 75,216& 92,004& 16,998\\
			\bf Avg. Length&37.71& 12.11& 29.60 &18.40 &23.74\\\hline\hline
			\bf unlabelled: &&&&&\\ \hline 
			\bf Sentences &100,000&100,000&100,000&100,000&100,000\\ 
			\bf Tokens &4,340,695&1,785,474&1,618,324&1,962,248&2,776,500\\
			\bf Avg. Length&43.41&17.85&16.18&19.62&27.77\\\hline
			\multicolumn{6}{l}{}\\
			\hline
			
			&\sc Hungarian & \sc Korean&\sc Polish&\sc Swedish &\\ \hline\hline
			\bf train: &&&&&\\ \hline 
			\bf Sentences &5,000&5,000&5,000&5,000&\\ 
			\bf Tokens &109,987 &68,336& 52,123& 76,357&\\
			\bf Avg. Length&21.99& 13.66& 10.42 &15.27&\\\hline\hline
			\bf test: &&&&&\\ \hline 
			\bf Sentences &1,009 &2,287& 822& 666&\\ 
			\bf Tokens &19,908& 33,766& 8,545 &10,690&\\
			\bf Avg. Length&19.73& 14.76& 10.39 &16.05&\\\hline\hline
			\bf unlabelled: &&&&&\\ \hline 
			\bf Sentences &100,000&100,000&100,000&100,000&\\ 
			\bf Tokens &1,913,154&2,147,605&2,024,323&1,575,868&\\
			\bf Avg. Length&19.13&21.48&20.24&15.76&\\\hline
		\end{tabular}
	\end{center}
	\caption{\label{table:spmrl_datasets_stats} Statistics about the \textsc{Spmrl} multi-lingual corpora}
\end{table}

We evaluate our adjusted parse score-based self-training approach with the \textsc{Spmrl} multi-lingual corpora.
The \textsc{Spmrl} multi-lingual corpora consist of nine languages (\textsc{Arabic, Basque, French, German, Hebrew, Hungarian, Korean, Polish, Swedish}) in-domain datasets available from 2014 Shared Task at the workshop on statistical parsing of morphologically rich languages (SPMRL), 
cf. \cite{seddah2014introducing}. We have chosen the datasets as there are no multi-lingual out-of-domain corpora available. Actually, even the in-domain corpora for many languages are rather small. We used the 5k smaller training set from the shared task, to make the scenario similar to the domain adaptation task that assumes a small number of target domain data is available. This setting is also a good basis for exploration for improving parsing accuracy of under-resourced languages. For each language, the shared task also provided a sufficient unlabelled data which is required by our evaluation. We evaluate nine languages in a unified setting, in which the 5k training set and a 100k unlabelled dataset are used for all the languages. 
For additional training set, we parse all 100k sentences for each of the languages and use 50k of them as the additional training set. For tuning the $d$ value of our adjusted parse score-based method, we used only the \textsc{German} development set, as we intend to use a unified setting for all languages and the \textsc{German} development set is the largest in size.
Table \ref{table:spmrl_datasets_stats} shows statistics about the corpora that we used in our experiments.

We evaluate all nine languages on the Mate parser \cite{bohnet2013joint}, the default settings are used in all the experiments. To output the confidence scores we slightly modified the parser, however, this does not affect the parser's accuracy. For part-of-speech tagging, we use the Mate parser's internal tagger for all the evaluations. The baselines are obtained from models trained only on the 5k initial training data. 

We report both labelled (LAS) and unlabelled (UAS) attachment scores, and mainly focus on the labelled accuracy. In line with the shared task official evaluation method, we include all the punctuations in our evaluation. The statistically significance levels are marked according to their p-values, (*) p-value \textless\ 0.05, (**) p-value \textless\ 0.01.

\section{Empirical Results}\label{section:self-multi-results}

\begin{table}[t]
\begin{center}
\begin{tabular}{|l|ll|ll|ll|}
\cline{2-7}
\multicolumn{1}{c|}{}& \multicolumn{2}{|c|}{\bf Baseline}&\multicolumn{2}{|c|}{\bf Self-train}&\multicolumn{2}{|c|}{\bf LORIA} \\\cline{2-7}
\multicolumn{1}{c|}{}&\bf LAS&\bf UAS&\bf LAS&\bf UAS&\bf LAS&\bf UAS\\ \hline
\sc Arabic		&82.09&85.17	&82.22	&85.21		&81.65&84.56  \\
\sc Basque  	&78.35&84.8 	&79.22**&85.61**	&81.39&86.86 \\
\sc French  	&81.91&86.03	&81.48	&85.63		&81.74&85.89\\ 
\sc German  	&81.54&84.72	&81.87**&85.18**	&83.35&86.37  \\
\sc Hebrew  	&78.86&85.08	&79.04	&85.26		&75.55&82.79  \\
\sc Hungarian   &83.13&87.48	&83.56*	&87.65		&82.88&87.26\\
\sc Korean  	&73.31&77.75	&75.45**&79.54**	&74.15&78.53  \\
\sc Polish  	&81.97&87.8 	&81.35	&87.25		&79.95&87.98 \\
\sc Swedish 	&79.67&86.1 	&80.26	&86.78*		&80.04&86.3 \\ \hline
 Average		&80.09&84.99	&80.49	&85.35		&80.08&85.17\\\hline
\end{tabular}
\end{center}
\caption{\label{table:self-multi-results} Comparing our self-trained results with the best non-ensemble system in the SPMRL Shared Task (LORIA).}
\end{table}

In this section, we report our results of the adjusted parse score-based self-training approach on the test sets of nine languages. To obtain the increased training data for our self-trained model, the unlabelled data is parsed and ranked by their confidence scores. The 50\% (50k) top ranked sentences are added to the initial training set. We retrain the Mate parser on the new training set.

The empirical results on nine languages show that our approach worked for five languages which are \textsc{Basque, German, Hungarian, Korean} and \textsc{Swedish}. Moreover, the self-trained model achieved on average (nine languages) 0.4\% gains for both labelled and unlabelled accuracies. These improvements are achieved only by a unified experiment setting, we do not tune parameters for individual language. Our self-training approach has the potential to achieve even better performances if we treat each of the languages separately, however, this is beyond the scope of this work.

More precisely, our self-training method achieved the largest labelled and unlabelled improvements on \textsc{Korean} with absolute gains of 2.14 and 1.79 percentage points respectively. Other than \textsc{Korean}, we also gain statistically significant improvements on \textsc{Basque, German, Hungarian} and \textsc{Swedish}. For \textsc{Basque}, the method achieved 0.87\% gain for labelled accuracy and the improvement for unlabelled accuracy is 0.81\%. For \textsc{German}, improvements of 0.33\% and 0.46\%  are gained by our self-trained model for labelled and unlabelled scores respectively. For \textsc{Hungarian}, we achieved a 0.42\% gain on labelled accuracy, the unlabelled improvement is smaller (0.17\%) thus not statistically significant. For \textsc{Swedish}, improvements of 0.59\% and 0.68\% are achieved for labelled and unlabelled accuracies. The unlabelled gain is statistically significant, while the labelled gain is not a statistically significant improvement which has a p-value of 0.067. As the improvements on \textsc{Swedish} are large but the test set is small (only contains 666 sentences), we decided to enlarge the test set by the \textsc{Swedish} development set. The \textsc{Swedish} development set contains 494 sentences and is not used for tuning in our experiments. The evaluation on the combined set showed 0.7\% and 0.6\% statistically significant (p \textless 0.01) improvements for labelled and unlabelled scores. This confirms the effectiveness of our self-training method on \textsc{Swedish}. In terms of the effects of our method on other languages, our method gains moderate improvements on \textsc{Arabic} and \textsc{Hebrew} but these are statistically insignificant accuracy gains.
We find negative results for \textsc{French} and \textsc{Polish}. Table \ref{table:self-multi-results} shows detailed results of our self-training experiments.

We compare our self-training results with the best non-ensemble parsing system of the SPMRL shared tasks \cite{seddah2013overview,seddah2014introducing}. The best results of the non-ensemble system are achieved by \newcite{cerisara2014spmrl}. Their system is also based on the semi-supervised learning, the LDA clusters \cite{chrupala2011lda} are used to explore the unlabelled data. The average labelled accuracy of our baseline on nine languages is same as the one achieved by \newcite{cerisara2014spmrl} and our self-trained results are 0.41\% higher than their results. The average unlabelled accuracy of our self-trained model also surpasses that of \newcite{cerisara2014spmrl} but with a smaller margin of 0.18\%. Overall, our self-trained models perform better in six languages (\textsc{Arabic, Hebrew, Hungarian, Korean, Polish} and \textsc{Swedish}) compared to the best non-ensemble system of \newcite{cerisara2014spmrl}.

\section{Analysis}\label{section:self-multi-analysis}
In this section, we analyse the results achieved by our self-training approach. Our approach achieved improvements on most of the languages, but also showed negative effects on two languages.  Thus, we analyse both positive and negative effects introduced by our self-training approach.  

For the analysis on positive effects, we choose the \textsc{Korean} dataset, as our self-training method achieved the largest improvement on it. The goal for our analysis on \textsc{Korean} is to find out where the improvement comes from. We apply our token and sentence level analysis to \textsc{Korean}. We evaluate for the token level the accuracy changes of individual labels and compare the improvements of unknown and known words. For our sentence level evaluation, we evaluate the performances on different sentence length and the number of unknown words per sentence. We do not evaluate on the number of subjects, the number of prepositions and number of conjunctions as those factors are language specific, thus they might not suitable for \textsc{Korean}.

For the analysis of negative effects, we analyse the \textsc{French} dataset as the \textsc{French} test set is larger than that of \textsc{Polish}. We aim to have an idea why our self-training approach has a negative effect on results. Our analysis focuses on two directions, firstly, we check the correlation between the quality of \textsc{French} data and our confidence scores, as the correlation is the pre-condition of the successful use of our self-training approach; secondly, we check the similarity between the test set and the unlabelled set to assess the suitability of unlabelled data.

\subsection{Positive Effects Analysis}
\subsubsection{Token Level Analysis}

%Confution matric
\begin{table}[!h]
\begin{center}
\begin{tabular}{l|r|r}
\hline
\bf Confusion & \bf Baseline & \bf Self-training\\\hline
\footnotesize adn $\rightarrow$ \footnotesize nmod & 99 & 113\\
\footnotesize adn $\rightarrow$ \footnotesize sub,root & 88 & 84\\\hline
\footnotesize adv $\rightarrow$ \footnotesize adn & 52 & 35\\
\footnotesize adv $\rightarrow$ \footnotesize sub,nmod,vmod & 126 & 130\\\hline
\footnotesize p $\rightarrow$ \footnotesize conj & 55 & 28\\
\footnotesize p $\rightarrow$ \footnotesize nmod & 126 & 136\\
\footnotesize p $\rightarrow$ \footnotesize adn & 103 & 91\\
\footnotesize p $\rightarrow$ \footnotesize vmod & 57 & 50\\\hline
\footnotesize nmod $\rightarrow$ \footnotesize conj & 62 & 39\\
\footnotesize nmod $\rightarrow$ \footnotesize adn & 209 & 166\\
\footnotesize nmod $\rightarrow$ \footnotesize adv & 99 & 88\\
\footnotesize nmod $\rightarrow$ \footnotesize vmod & 215 & 179\\
\footnotesize nmod $\rightarrow$ \footnotesize sub & 41 & 38\\\hline
\footnotesize root $\rightarrow$ \footnotesize aux & 103 & 116\\
\footnotesize root $\rightarrow$ \footnotesize adn & 41 & 15\\\hline
\footnotesize tpc $\rightarrow$ \footnotesize adn & 107 & 74\\
\footnotesize tpc $\rightarrow$ \footnotesize nmod & 30 & 29\\\hline
\footnotesize sub $\rightarrow$ \footnotesize conj & 75 & 69\\
\footnotesize sub $\rightarrow$ \footnotesize adn & 74 & 57\\
\footnotesize sub $\rightarrow$ \footnotesize adv & 40 & 50\\\hline
\footnotesize sbj $\rightarrow$ \footnotesize comp & 35 & 36\\\hline
\footnotesize aux $\rightarrow$ \footnotesize root & 66 & 59\\
\footnotesize aux $\rightarrow$ \footnotesize sub,adn & 68 & 57\\\hline
\footnotesize conj $\rightarrow$ \footnotesize sub & 88 & 86\\
\footnotesize conj $\rightarrow$ \footnotesize adn & 56 & 42\\
\footnotesize conj $\rightarrow$ \footnotesize nmod & 48 & 54\\\hline
\footnotesize vmod $\rightarrow$ \footnotesize nmod & 187 & 195\\
\footnotesize vmod $\rightarrow$ \footnotesize adv & 77 & 78\\
\footnotesize vmod $\rightarrow$ \footnotesize sub,adn,amod & 116 & 108\\\hline
\end{tabular}
\end{center}
\caption{\label{table:analysis-confusion-multi-lingual-self-training} The confusion matrix of dependency labels, compared between the multi-lingual self-training approach and the baseline.}
\end{table}

%label evaluation
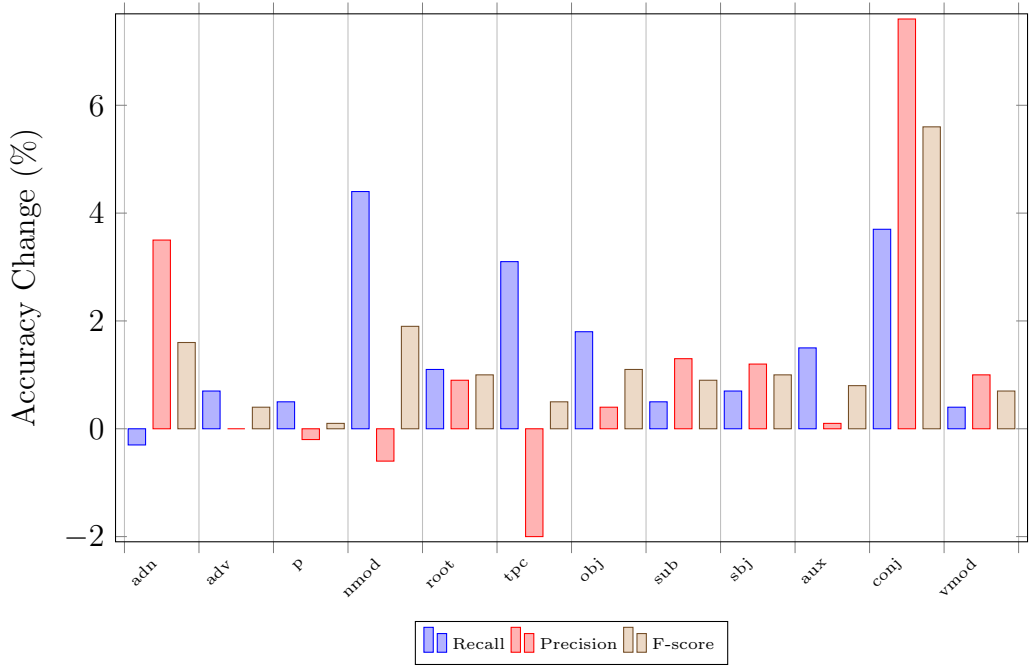
\begin{figure}[t]
	\begin{center}
		\begin{tikzpicture}
		\begin{axis}[
		width=12cm,height=7cm,
		ybar,ybar interval =0.7,
		ylabel=Accuracy Change (\%),
		enlargelimits=0.01,
		legend style={at={(0.5,-0.15)},anchor=north,legend columns=-1},
		symbolic x coords={adn,adv,p,nmod,root,tpc,obj,sub,sbj,aux,conj,vmod,cc,amod,ejx},
		xtick=data,
		x tick label style={font=\tiny,rotate=45,anchor=east}
		]
		\addplot coordinates{(adn,-0.3) (adv,0.7) (p,0.5) (nmod,4.4) (root,1.1) (tpc,3.1) (obj,1.8) (sub,0.5) (sbj,0.7) (aux,1.5) (conj,3.7) (vmod,0.4) (cc,0.3)  }; 
		\addlegendentry{\tiny Recall}
		
		\addplot coordinates{(adn,3.5) (adv,0.0) (p,-0.2) (nmod,-0.6) (root,0.9) (tpc,-2.0) (obj,0.4) (sub,1.3) (sbj,1.2) (aux,0.1) (conj,7.6) (vmod,1.0) (cc,-0.4)  }; 
		\addlegendentry{\tiny Precision}
		
		\addplot coordinates{(adn,1.6) (adv,0.4) (p,0.1) (nmod,1.9) (root,1.0) (tpc,0.5) (obj,1.1) (sub,0.9) (sbj,1.0) (aux,0.8) (conj,5.6) (vmod,0.7) (cc,-0.1)}; 
		\addlegendentry{\tiny F-score}
		
		\end{axis}
		%adn cnt:4686 R/P/F92.9/86.9/89.9 Base R/P/F93.2/83.4/88.3 diff: F:1.6 R:-0.3 P:3.5
		%adv cnt:3548 R/P/F93.5/89.0/91.3 Base R/P/F92.8/89.0/90.9 diff: F:0.4 R:0.7 P:0.0
		%p cnt:3227 R/P/F87.0/99.3/93.1 Base R/P/F86.5/99.5/93.0 diff: F:0.1 R:0.5 P:-0.2
		%nmod cnt:3019 R/P/F80.8/75.7/78.3 Base R/P/F76.4/76.3/76.4 diff: F:1.9 R:4.4 P:-0.6
		%root cnt:2287 R/P/F92.1/92.1/92.1 Base R/P/F91.0/91.2/91.1 diff: F:1.0 R:1.1 P:0.9
		%tpc cnt:1885 R/P/F91.1/93.0/92.0 Base R/P/F88.0/95.0/91.5 diff: F:0.5 R:3.1 P:-2.0
		%obj cnt:1642 R/P/F96.6/93.9/95.2 Base R/P/F94.8/93.5/94.1 diff: F:1.1 R:1.8 P:0.4
		%sub cnt:1575 R/P/F79.9/77.9/78.9 Base R/P/F79.4/76.6/78.0 diff: F:0.9 R:0.5 P:1.3
		%sbj cnt:1569 R/P/F93.6/93.8/93.7 Base R/P/F92.9/92.6/92.7 diff: F:1.0 R:0.7 P:1.2
		%aux cnt:1475 R/P/F90.0/88.9/89.5 Base R/P/F88.5/88.8/88.7 diff: F:0.8 R:1.5 P:0.1
		%conj cnt:986 R/P/F76.7/76.9/76.8 Base R/P/F73.0/69.3/71.2 diff: F:5.6 R:3.7 P:7.6
		%vmod cnt:976 R/P/F54.5/52.1/53.3 Base R/P/F54.1/51.1/52.6 diff: F:0.7 R:0.4 P:1.0

		\end{tikzpicture}
	\end{center}
	\caption{\label{figure:analysis-label-multi-self-training} The performance comparison between the multi-lingual self-training approach and the baseline on major labels.}
\end{figure}

\textbf{Individual Label Accuracy.} The \textsc{Korean} syntactic labels set used in the shared task contains 22 labels \cite{seddah2014introducing}. We listed the 12 most frequently used labels in our analysis. Those labels are presented in the \textsc{Korean} test set for at least 1,000 times. As we can see from the Figure \ref{figure:analysis-label-multi-self-training}, the largest f-score improvement of 5.6\% is achieved on conjuncts (conj). Large gains of more than 0.4\% are achieved on nearly all the labels, the only exception is punctuations (p), for punctuations our self-training approach only achieved a moderate improvement of 0.1\%. The adverbial modifier (adv), topic (tpc), subordination (sub), auxiliary verb (aux) and modifier of predicate (vmod) have improvements between 0.4\% and 0.9\%. The other five labels, adnominal modifier (adn), modifier of nominal (nmod), root of the sentence (root), object (obj), subject (sbj) are improved by more than 1\%. Table \ref{table:analysis-confusion-multi-lingual-self-training} shows the confusion matrix of the dependency labels.

%corpus UNK
\begin{table}[t]
	\begin{center}
		\begin{tabular}{|l|r|rr|rr|}
			\cline{3-6}
			\multicolumn{2}{c|}{}& \multicolumn{2}{|c|}{\bf Self-training} &\multicolumn{2}{|c|}{\bf Baseline}\\
			\cline{2-6}
			\multicolumn{1}{c|}{}&\bf Tokens &\bf LAS&\bf UAS&\bf LAS&\bf UAS\\
			\hline
			\bf Known &15567&81.6&84.0&79.7&82.2\\%diff: LAS:1.9 UAS:1.8
			\bf Unknown &12799&67.9&74.1&65.5&72.3\\%diff: LAS:2.4 UAS:1.8
			\hline
			\bf All &28366&75.5&79.5&73.3&77.7\\%diff: LAS:2.2 UAS:1.8
			\hline
		\end{tabular}
	\end{center}
	\caption{\label{table:analysis-corpusunk-multi-self-training} The accuracy comparison between the multi-lingual self-training approach and the baseline on unknown words.}
\end{table}

\textbf{Unknown Words Accuracy.} Table \ref{table:analysis-corpusunk-multi-self-training} shows our analysis of the unknown words. The unknown words rate for the \textsc{Korean} test is surprisingly higher than expected, more than 45\% of the words in the test set are not presented in the training set. This might due to two reasons: firstly the training set is very small only contains 5k sentences thus have a less coverage of vocabulary; secondly and the main reason is the \textsc{Korean} tokens used in the shared task are combinations of the word form and the grammatical affixes. The latter creates much more unique tokens. The vocabulary of the training set is 29,715, but the total number of tokens is only 68,336, which means each token only shows less than 2.3 times on average. Despite the high unknown words rate, our self-training approach showed a better labelled improvement (2.4\%) on unknown words than that of known words (1.9\%). While the unlabelled improvement (1.8\%) is exactly the same for both known and unknown words.

\subsubsection{Sentence Level Analysis}

%number of tokens
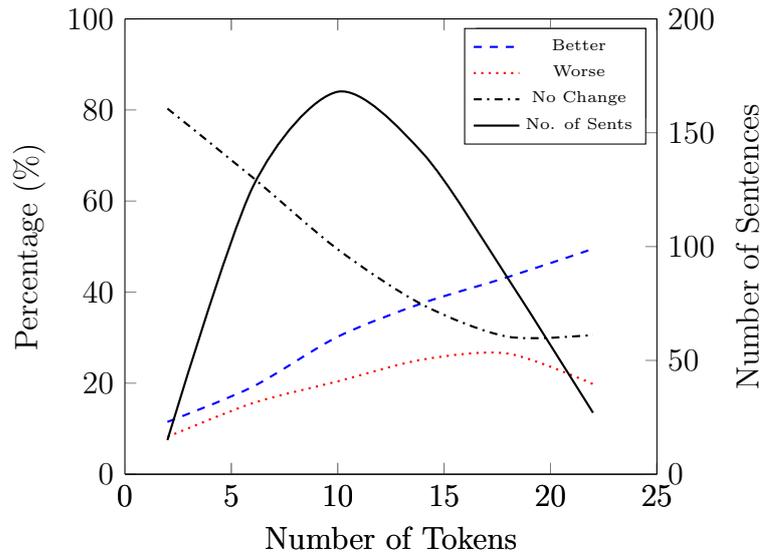
\begin{figure}[t]
	\begin{center}
		\begin{tikzpicture}
		\pgfplotsset{
			xmin=0,xmax=25,
			xlabel=Number of Tokens}
		\begin{axis}[
		axis y line*=left,
		ymin=0,ymax=100,
		ylabel=Percentage (\%)]
		%Better
		\addplot[smooth,thick,dashed,color=blue] coordinates {(2,11.5) (6,19.2) (10,30.2) (14,37.6) (18,43.3) (22,49.5)  };\label{better}
		\addlegendentry{\tiny Better}
		
		%Worse
		\addplot[smooth,thick,dotted,color=red] coordinates{(2,8.2) (6,15.6) (10,20.4) (14,25.2) (18,26.5) (22,19.8)  };\label{worse}
		\addlegendentry{\tiny Worse}
		
		%No Change
		\addplot[smooth,thick,dashdotted] coordinates{(2,80.3) (6,65.2) (10,49.4) (14,37.2) (18,30.2) (22,30.6)  };\label{nochange}
		\addlegendentry{\tiny No Change}
		
		\end{axis}
		\begin{axis}[
		axis y line*=right,
		ymin=0,ymax=200,
		ylabel=Number of Sentences]
		\addlegendimage{/pgfplots/refstyle=better}\addlegendentry{\tiny Better}
		\addlegendimage{/pgfplots/refstyle=worse}\addlegendentry{\tiny Worse}
		\addlegendimage{/pgfplots/refstyle=nochange}\addlegendentry{\tiny No Change}
		%All Sent
		\addplot[smooth,thick,solid] coordinates{(2,15.0) (6,126.0) (10,168.0) (14,141.0) (18,86.0) (22,27.0)  };\addlegendentry{\tiny No. of Sents}
		
		\end{axis}
		%2 Total:61 LAS better/worse/nochange:7/5/49 11.5/8.2/80.3 Diff:3.3
		%6 Total:506 LAS better/worse/nochange:97/79/330 19.2/15.6/65.2 Diff:3.6
		%10 Total:672 LAS better/worse/nochange:203/137/332 30.2/20.4/49.4 Diff:9.8
		%14 Total:564 LAS better/worse/nochange:212/142/210 37.6/25.2/37.2 Diff:12.4
		%18 Total:344 LAS better/worse/nochange:149/91/104 43.3/26.5/30.2 Diff:16.8
		%22 Total:111 LAS better/worse/nochange:55/22/34 49.5/19.8/30.6 Diff:29.7
	
		\end{tikzpicture}
	\end{center}
	\caption{\label{figure:analysis-sentlength-multi-self-training} The comparison between the multi-lingual self-training approach and the baseline on different number of tokens per sentence. }
\end{figure}

\textbf{Sentence Length.} We then apply the sentence level analysis for \textsc{Korean} test set. We first evaluate on the different sentence length, sentences that have the same length are assigned into the same group. We then calculate the percentage of sentences that are improved, decreased or unchanged in accuracy for each group. We plot the results along with the number of sentences in each of the groups in Figure \ref{figure:analysis-sentlength-multi-self-training}. As we can see from the figure, the gap between the improved and decreased sentences are smaller (about 3\%) on short sentences that contain less than 10 tokens. The gap significantly widens when the sentence length grows. The gap increased to 30\% for sentences containing more than 20 tokens. This is a clear indication that our self-training yielded stronger enhancements on longer sentences.

%number of unknown words
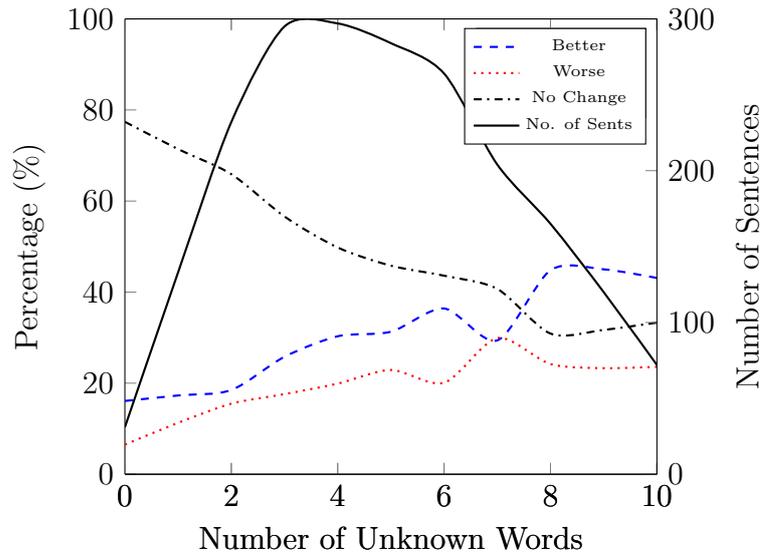
\begin{figure}[t]
	\begin{center}
		\begin{tikzpicture}
		\pgfplotsset{
			xmin=0,xmax=10,
			xlabel=Number of Unknown Words}
		\begin{axis}[
		axis y line*=left,
		ymin=0,ymax=100,
		ylabel=Percentage (\%)]
		%Better
		\addplot[smooth,thick,dashed,color=blue] coordinates {(0,16.1) (1,17.3) (2,18.5) (3,25.8) (4,30.3) (5,31.3) (6,36.4) (7,29.4) (8,44.8) (9,45.0) (10,43.1) };\label{better}
		\addlegendentry{\tiny Better}
		
		%Worse
		\addplot[smooth,thick,dotted,color=red] coordinates{(0,6.5) (1,11.3) (2,15.5) (3,17.6) (4,19.9) (5,22.9) (6,20.1) (7,29.9) (8,24.2) (9,23.3) (10,23.6) };\label{worse}
		\addlegendentry{\tiny Worse}
		
		%No Change
		\addplot[smooth,thick,dashdotted] coordinates{(0,77.4) (1,71.4) (2,65.9) (3,56.6) (4,49.8) (5,45.8) (6,43.6) (7,40.7) (8,30.9) (9,31.7) (10,33.3) };\label{nochange}
		\addlegendentry{\tiny No Change}
		
		\end{axis}
		\begin{axis}[
		axis y line*=right,
		ymin=0,ymax=300,
		ylabel=Number of Sentences]
		\addlegendimage{/pgfplots/refstyle=better}\addlegendentry{\tiny Better}
		\addlegendimage{/pgfplots/refstyle=worse}\addlegendentry{\tiny Worse}
		\addlegendimage{/pgfplots/refstyle=nochange}\addlegendentry{\tiny No Change}
		%All Sent
		\addplot[smooth,thick,solid] coordinates{(0,31.0) (1,133.0) (2,232.0) (3,295.0) (4,297.0) (5,284.0) (6,264.0) (7,204.0) (8,165.0) (9,120.0) (10,72.0) };\addlegendentry{\tiny No. of Sents}
		
		\end{axis}
		%0 Total:31 LAS better/worse/nochange:5/2/24 16.1/6.5/77.4 Diff:9.6
		%1 Total:133 LAS better/worse/nochange:23/15/95 17.3/11.3/71.4 Diff:6.0
		%2 Total:232 LAS better/worse/nochange:43/36/153 18.5/15.5/65.9 Diff:3.0
		%3 Total:295 LAS better/worse/nochange:76/52/167 25.8/17.6/56.6 Diff:8.2
		%4 Total:297 LAS better/worse/nochange:90/59/148 30.3/19.9/49.8 Diff:10.4
		%5 Total:284 LAS better/worse/nochange:89/65/130 31.3/22.9/45.8 Diff:8.4
		%6 Total:264 LAS better/worse/nochange:96/53/115 36.4/20.1/43.6 Diff:16.3
		%7 Total:204 LAS better/worse/nochange:60/61/83 29.4/29.9/40.7 Diff:-0.5
		%8 Total:165 LAS better/worse/nochange:74/40/51 44.8/24.2/30.9 Diff:20.6
		%9 Total:120 LAS better/worse/nochange:54/28/38 45.0/23.3/31.7 Diff:21.7
		%10 Total:72 LAS better/worse/nochange:31/17/24 43.1/23.6/33.3 Diff:19.5
		
		\end{tikzpicture}
	\end{center}
	\caption{\label{figure:analysis-sentunk-multi-self-training} The comparison between the multi-lingual self-training approach and the baseline on different number of unknown words per sentence. }
\end{figure}

\textbf{Unknown Words.} As we found in the token level analysis, the unknown words rate is very high for \textsc{Korean} test set. In the extreme case, there could be more than 20 unknown words in a single sentence. The curve shows an overall increased gap between the sentences improved by the self-trained model and those worsened when the number of unknown words per sentence increases. However, the gains sometimes drop, the most notable group is the one for sentences containing 7 unknown words. The percentage of worsened sentences are even 0.5\% higher than that of improved ones. It is unclear the reason why the behaviour changes, but due to the group size is small (only 200 sentences) we suggest this might caused by chance.

\subsection{Negative Effects Analysis}

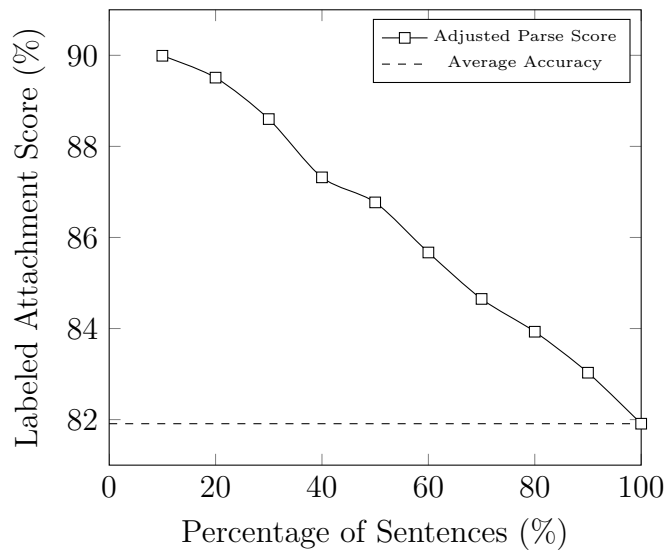
\begin{figure}[t]
	\begin{center}
		\begin{tikzpicture}
		\pgfplotsset{
			xmin=0,xmax=100,
			ymin=81,ymax=91,
			xlabel=Percentage of Sentences (\%),
			ylabel=Labeled Attachment Score (\%)
		}
		\begin{axis}

		%adjusted parse score
		\addplot[smooth,mark=square*, mark options={fill=white}]
		coordinates{(10,89.99) (20,89.51) (30,88.6) (40,87.32) (50,86.77) (60,85.67) (70,84.65) (80,83.93) (90,83.03) (100,81.91)};
		\addlegendentry{\tiny Adjusted Parse Score}
		
		%baseline
		\addplot[smooth,dashed] 
		coordinates{(0,81.91) (100,81.91)};
		\addlegendentry{\tiny Average Accuracy}
		
		\end{axis}
		
		\end{tikzpicture}
	\end{center}
	\caption{\label{figure:analysis-negative-accuracy-assess-multi-self-training} The accuracies when inspecting 10-100\% sentences of the \textsc{French} test set ranked by the confidence-based methods.}
\end{figure}

\subsubsection{Confidence Score Analysis}
As our confidence-based self-training is based on the hypothesis that the confidence scores are able to indicate the quality of the annotations. Thus when our self-training approach showed a negative effect on the accuracy, the first thing comes to our mind is to check the correlation between confidence scores and accuracies. We analyse the correlation on the \textsc{French} test set by ranking the sentences in the dataset according to their confidence scores. We assess the accuracy of the top ranked $n$ percent sentences. We set $n$ to 10\% and increase it by 10\% in each step until all the sentences are included. We show the analysis in Figure \ref{figure:analysis-negative-accuracy-assess-multi-self-training}. The analysis suggests that there is a reasonably high correlation between the quality of the sentences and our confidence-based method. The top ranked 10\% sentences have an accuracy of 89.99\% which is 8\% higher than the average. The accuracy for top ranked 50\% sentences is 86.77\% which surpasses the average by 5\%.

\begin{table}[t]
	\begin{center}
		\begin{tabular}{|l|r|r|r|}
			\cline{2-4}
			\multicolumn{1}{c|}{}& \multicolumn{1}{|c|}{\bf train} & \multicolumn{1}{|c|}{\bf \ \ test\ \ }& \multicolumn{1}{|c|}{\bf unlabelled}\\ \hline
			\bf Sentences  & 5,000 & 2,541 &100,000 \\
			\bf Tokens     & 150,984  &75,216 &1,618,324 \\
			\bf Avg. Length &30.19&29.60&16.18\\
			\bf UNK (\%) &-&5.91&16.82\\
			\bf Similarity (\%)&-&99.74&95.47\\
			\hline
		\end{tabular}
		
	\end{center}
	\caption{\label{table:analysis-negative-datasets-stats-multi-self-training} The basic statistic of datasets for \textsc{French} evaluation.}
\end{table}
\subsubsection{Unlabelled Data Analysis}

The quality of unlabelled data is another issue that might affect the results. We first compute the basic statistics of the training, test and unlabelled dataset to have a surface level comparison. As shown in Table \ref{table:analysis-negative-datasets-stats-multi-self-training} the unlabelled data is very different from the training and test set. More precisely, the average sentence length of the unlabelled data is much shorter. The unknown words rate of the unlabelled dataset (16.82\%) is three times higher than that of the test set (5.91\%). We further calculate the cosine similarity between the training set and the test/unlabelled dataset. The test set is highly similar to the training set with a similarity of 99.74\%. The similarity score of the unlabelled data is more than 4\% lower, which suggests the unlabelled data is more different.

\section{Chapter Summary}\label{section:self-multi-conclusion}
In this chapter, we evaluated an effective confidence-based self-training approach on nine languages. Due to the lack of out-of-domain resources, we used an under-resourced in-domain setting instead. We used for all languages a unified setting, the parser is retrained on the new training set boosted by the top 50k ranked parse trees selected from a 100k auto-parsed dataset. 

Our approach successfully improved accuracies of five languages (\textsc{Basque, German, Hungarian, Korean} and \textsc{Swedish}) without tuning variables for the individual language. 
We can report the largest labelled and unlabelled accuracy gain of 2.14\% and 1.79\% on \textsc{Korean}, on average we improved the baselines of five languages by 0.87\% (LAS) and 0.78\% (UAS). 

We further did an in-depth analysis on \textsc{Korean} and \textsc{French}. For \textsc{Korean}, we did a number of analysis on both token level and sentence level to understand where the improvement comes from. The analysis on the individual label showed that the self-trained model achieved large improvement on all the major labels, and it achieved the largest gain on conjuncts (conj). The analysis of unknown words showed that the self-trained model gained a larger labelled improvement for unknown words. The analysis on sentence length suggested the self-training approach achieved larger improvements on longer sentences. For \textsc{French}, we aim to understand why self-training did not work. The analysis showed the confidence scores have a reasonably high correlation with the annotation quality, hence it is less likely be the reason of self-training's negative effect. While the large difference between unlabelled data and the training/test sets is more likely a major contributor to the accuracy drop. 

\chapter{Dependency Language Models}\label{chapter:dlm}
In this chapter, we introduce our dependency language models (DLM) approach for both in-domain and out-of-domain dependency parsing. The co-training and self-training approaches evaluated in the previous chapters have demonstrated their effectiveness on the out-of-domain parsing, however, neither approaches gained large improvements on the source domain accuracy. In fact, sometimes they even have a negative effect on the in-domain results. Another disadvantage of co-/self-training is that they can use only a relatively small additional training dataset, as training parsers on a large corpus might be time-consuming or even intractable on a corpus of millions of sentences. The goal of our DLM approach is to create a robust model that is able to improve both in-domain and out-of-domain accuracies. Unlike the co-/self-training, the DLM approach does not use the unlabelled data directly for retraining. Instead, a small number of features based on DLMs are integrated into the parser, thus we could explore much larger unlabelled datasets. Other semi-supervised techniques that use the unlabelled data indirectly include word clustering \cite{brown1992class,chrupala2011lda} and word embedding \cite{bengio03a,mikolov2013distributed,pennington2014glove}. However, both word clustering and word embedding are generated from unannotated data, thus do not consider the syntactic structures. The DLMs used in this work are generated from the automatically annotated dataset, which could benefit additionally from the syntactic annotations. 

Dependency language models are variants of language models based on dependency structures. An N-gram DLM is able to predict the next child when given N-1 immediate previous children and their head. DLMs were first introduced by \newcite{shen2008new} and were later adapted to dependency parsing by \newcite{chen2012utilizing}. \newcite{chen2012utilizing} integrated DLMs extracted from large auto-parsed corpora into a second-order graph-based parser. DLMs allow the parser to explore higher order features but without increasing the time complexity.  We use a similar approach as \newcite{chen2012utilizing}, but our approach is different in six important aspects: 

\begin{enumerate}
	\item We apply DLMs to a transition-based dependency parser.
	\item We additionally use syntactic labels in the DLM-based features as our parser produces the labelled annotations.
	\item The DLM-based features are integrated into a strong parser that is able to achieve competitive baselines.
	\item We use not only single DLM but also multiple DLMs in our experiments.
	\item We evaluate our approach on both in-domain and out-of-domain parsing.
	\item Inspired by our co-training approach, we also investigate the parser with DLMs generated from high-quality auto-parsed data.
\end{enumerate}

In the rest of this chapter, we introduce our approaches in Section \ref{section:dlm-appraoch}, we present our experiment set-up in Section \ref{section:dlm-setup}. In Section \ref{section:dlm-results} and \ref{section:dlm-analysis} we discuss and analyse the results. In the final section (Section \ref{section:dlm-conclusion}) we summarise the chapter.

\section{Dependency Language Models for Transition-based System}\label{section:dlm-appraoch}
Dependency language models were introduced by \newcite{shen2008new} to capture long distance relations in syntactic structures. 
An N-gram DLM predicts the next child based on N-1 immediate previous children and their head. 
We integrate DLMs extracted from a large parsed corpus into the Mate parser \cite{bohnet2013joint}. We first train a base model with the manually annotated training set. The base model is then used to annotate a large number of unlabelled sentences. After that, we extract DLMs from the auto-annotated corpus. Finally, we retrain the parser with additional DLM-based features.

Further, we experimented with techniques to improve the quality of the syntactic annotations which we use to build the DLMs. 
We parse the unlabelled data with two different parsers and then select the annotations on which both parsers agree on. 
The method is similar to co-training except that we do not train the parser directly on these auto-labelled sentences.

We build the DLMs with the method of \newcite{chen2012utilizing}. 
For each child $x_{ch}$, we gain the probability distribution $P_u(x_{ch}|HIS)$, where $HIS$ refers to $N-1$ immediate previous children and their head $x_h$. The previous children for $x_{ch}$ are those who share the same head with $x_{ch}$ but are closer to the head word according to the word sequence in the sentence.
Consider the left side child $x_{Lk}$ in the dependency relations $(x_{Lk}...x_{L1}, x_h, x_{R1}...x_{Rm})$ as an example; the N-1 immediate previous children for $x_{Lk}$ are $x_{Lk-1}..x_{Lk-N+1}$.  
In our approach, we estimate $P_u(x_{ch}|HIS)$ by the relative frequency: %$P_u(x_{ch}|HIS) = \frac{count(x_{ch},HIS)}{\sum_{x'_{ch}} count(x'_{ch},HIS)}$
\begin{equation}
P_u(x_{ch}|HIS) = \frac{count(x_{ch},HIS)}{\sum_{x'_{ch}} count(x'_{ch},HIS)}
\vspace{0.3cm}
\end{equation}

By their probabilities, the N-grams are sorted in a descending order. We then used the thresholds of \newcite{chen2012utilizing} to replace the probabilities with one of the three classes ($PH,PM,PL$) according to their position in the sorted list, i.e. the probabilities having an index in the first 10\% of the sorted list are replaced
with $PH$, $PM$ refers to probabilities ranked between 10\% and 30\%, probabilities that are ranked below 30\% are replaced with $PL$. 
During parsing, we use an additional class $PO$ for relations not presented in DLMs. We use the classes instead of the probability is because our baseline parser uses the binary feature representations, classes are required to map the features into the binary feature representations.  As a result, the real number features are hard to be integrated into the existing system. In the preliminary experiments, the $PH$ class is mainly filled by unusual relations that only appeared a few times in the parsed text. To avoid this we configured the DLMs to only use elements which have a minimum frequency of three, i.e. $count(x_{ch},HIS) \geq 3$.
Table \ref{table:dlm-feature} shows our feature templates, where $NO_{DLM}$ is an index which allows DLMs to be distinguished from each other, $s_0$, $s_1$ are the top and the second top of the stack, $\phi(P_u(s_0/s_1))$ refers the coarse label of probabilities $P_u(x_{s_0/s_1}|HIS)$ (one of the $PH,PM,PL,PO$), 
$s_0/s_1\_pos, s_0/s_1\_word$ refer to part-of-speech tags, word forms of $s_0/s_1$, and $label$ is the dependency label between $s_0$ and $s_1$.

\begin{table}[t]
\begin{center}
\begin{tabular}{l}
\hline  \( <NO_{DLM}, \phi(P_u(s_0)), \phi(P_u(s_1)), label> \) \\
\( <NO_{DLM}, \phi(P_u(s_0)), \phi(P_u(s_1)), label, s_0\_pos> \) \\
\( <NO_{DLM}, \phi(P_u(s_0)), \phi(P_u(s_1)), label, s_0\_word> \) \\
\( <NO_{DLM}, \phi(P_u(s_0)), \phi(P_u(s_1)), label, s_1\_pos> \) \\
\( <NO_{DLM}, \phi(P_u(s_0)), \phi(P_u(s_1)), label, s_1\_word> \) \\
\( <NO_{DLM}, \phi(P_u(s_0)), \phi(P_u(s_1)), label, s_0\_pos, s_1\_pos> \) \\
\( <NO_{DLM}, \phi(P_u(s_0)), \phi(P_u(s_1)), label, s_0\_word, s_1\_word> \) \\
 \hline

\end{tabular}
\end{center}
\caption{\label{table:dlm-feature} DLM-based feature templates which we used in the parser.}
\end{table} 

\section{Experiment Set-up} \label{section:dlm-setup}
\begin{table}[t]
	\begin{center}
		\begin{tabular}{|l|r|r|r|r|}
			\cline{2-5}
			\multicolumn{1}{c|}{}& \multicolumn{1}{|c|}{\bf train} & \multicolumn{1}{|c|}{\bf dev}&\multicolumn{1}{|c|}{\bf test}& \multicolumn{1}{|c|}{\bf unlabelled}\\ \hline
			\bf Section&2-21&22&23&-\\ \hline
			\bf Sentences &39,832 &1,700 &2,416  & 30,546,808 \\
			\bf Tokens  &950,028  &40,117&56,684 & 771,306,902 \\
			\bf Avg. Length &23.85&23.60&23.46&25.25\\
			\hline
		\end{tabular}
		
	\end{center}
	\caption{\label{table:wsj_datasets_stats} The size of datasets for the \textsc{Wsj} Stanford conversion evaluation.}
\end{table}

For our experiments on English in-domain text, we used the Wall Street Journal portion (\textsc{Wsj}) of the Penn English Treebank \cite{marcus93}. The constituency trees are converted to the Stanford style dependency relations. The Stanford conversion attracts more attention during the recent years, it has been used in the SANCL 2012 shared tasks \cite{petrov2012overview} and many state-of-the-art results were also reported using this conversion \cite{weiss2015neural,andor2016globally,dozat2017deep}. We follow the standard splits of the corpus, 
section 2-21 are used for training, section 22 and 23 are used as the development set and the test set respectively.
We used the Stanford parser
\footnote{http://nlp.stanford.edu/software/lex-parser.shtml} 
v3.3.0 to convert the constituency trees into Stanford style dependencies \cite{demarneffe06}. For unlabelled data, we used the data of \newcite{chelba13onebillion} which contains around 30 million sentences (800 million words) from the news domain. Table \ref{table:wsj_datasets_stats} shows the basic statistics about the corpus;

In addition to the \textsc{Wsj} corpus, we also evaluate our approach on the main evaluation corpus of this thesis. Our main evaluation corpus consists of a \textsc{Conll} source domain training set, a source domain test set and four target domain test sets (\textsc{Weblogs, Newsgroups, Reviews} and \textsc{Answers}). Unlike our \textsc{Wsj} corpus that uses Stanford dependencies, the main evaluation corpus is based on the LTH conversion \cite{johansson2007extended}. Experimenting on different conversions and domains allow us to evaluate our method's robustness. For unlabelled data, we use the same dataset as in our \textsc{Wsj} evaluation. 

\begin{table}[t]
	\begin{center}
		\begin{tabular}{|l|r|r|r|r|}
			\cline{2-5}
			\multicolumn{1}{c|}{}& \multicolumn{1}{|c|}{\bf train} & \multicolumn{1}{|c|}{\bf dev}&\multicolumn{1}{|c|}{\bf test}& \multicolumn{1}{|c|}{\bf unlabelled}\\ \hline
			\bf Section&001-815,&886-931,&816-885, &-\\
			&1001-1136& 1148-1151&1137-1147&\\ \hline
			\bf Sentences  &16,118 &805 & 1,915  & 19,806,808 \\
			\bf Tokens     & 437,860 &20,454&50,319 &467,242,601 \\
			\bf Avg. Length &27.16&25.41&26.27&23.59\\
			\hline
		\end{tabular}
		
	\end{center}
	\caption{\label{table:ctb_datasets_stats} The size of datasets for the Chinese Treebank 5 (\textsc{Ctb}) evaluation.}
\end{table}

For Chinese, we evaluate our approach only on the in-domain scenario, this is due to the lack of out-of-domain corpus. We use Chinese Treebank 5 (CTB5) \cite{xue05} as the source of our gold standard data. The Chinese Treebank 5 corpus mainly consists of articles from Xinhua news agency but also contains some articles from Sinorama magazine and information services department of HKSAR.
We follow the splits of \newcite{zhang11}, the constituency trees are converted to dependency relations by the Penn2Malt\footnote{http://stp.lingfil.uu.se/~nivre/research/Penn2Malt.html} tool using head rules of \newcite{zhang08}. We use the Xinhua portion of Chinese Gigaword Version 5.0  \footnote{https://catalog.ldc.upenn.edu/LDC2011T13} as our source for unlabelled data. We noticed that the unlabelled data we used actually contains the Xinhua portion of the CTB5; to avoid potential conflict we removed them from the unlabelled data. After the pre-processing, our Chinese unlabelled data consists of 20 million sentences which are roughly 450 million words. We use ZPar\footnote{https://github.com/frcchang/zpar} v0.7.5 as our pre-processing tool. The word segmentor of ZPar is trained on the CTB5 training set. Table \ref{table:ctb_datasets_stats} gives some statistics about the corpus.

We use a modified version of the Mate transition-based parser in our experiments. We enhance the parser with our DLM-based features; other than this we used the parser's default setting. The part-of-speech tags are supplied by Mate parser's internal tagger. The baselines are trained only on the initial training set. In most of our experiments, DLMs are extracted from data annotated by the base model of Mate parser. For the evaluation on higher quality DLMs, the unlabelled data is additionally tagged and parsed by Berkeley parser \cite{petrov07} and is converted to dependency trees with the same tools as for gold data. 

We report both labelled (LAS) and unlabelled (UAS) attachment scores for our evaluation. The punctuation marks are excluded for our English and Chinese in-domain evaluations. For English evaluation on our main evaluation corpus we include the punctuations. The significance levels are marked due to their p-values, we use * and ** to represent the p-value of 0.05 and 0.01 levels respectively.

\section{Empirical Results} \label{section:dlm-results}

\begin{figure}[t]
	\begin{center}
		\begin{tikzpicture}
		\pgfplotsset{
			width=6cm,
			title=\tiny a) English,
			xmin=0.5,xmax=3.5,
			xtick={1,2,3},
			ymin=91,ymax=91.5,
			xlabel=Value of N,
			ylabel=Labeled Attachment Score (\%)
		}
		\begin{axis}

		\addplot[smooth,mark=triangle*,mark options={fill=white}]
		coordinates {(1,91.43)(2,91.14)(3,91.22)};
		\addlegendentry{\tiny Single DLM}
		
		\addplot[smooth,mark=*, mark options={fill=white}]
		coordinates{(1,91.43)(2,91.27)(3,91.26)};
		\addlegendentry{\tiny Multiple DLMs}
		
		%baseline
		\addplot[smooth,dashed] 
		coordinates{(0,91.05) (100,91.05)};
		\addlegendentry{\tiny Baseline}
		
		\end{axis}
		
		\end{tikzpicture}
		\begin{tikzpicture}
		\pgfplotsset{
			width=6cm,
			title=\tiny b) Chinese,
			xmin=0.5,xmax=4.5,
			xtick={1,2,3,4},
			ymin=78.7,ymax=80.7,
			xlabel=Value of N,
			ylabel=Labeled Attachment Score (\%)
		}
		\begin{axis}

		\addplot[smooth,mark=triangle*,mark options={fill=white}]
		coordinates {(1,79.85)(2,79.42)(3,79.06)};
		\addlegendentry{\tiny Single DLM}
		
		\addplot[smooth,mark=*, mark options={fill=white}]
		coordinates{(1,79.85)(2,79.97)(3,80.11)(4,79.73)};
		\addlegendentry{\tiny Multiple DLMs}
		
		%baseline
		\addplot[smooth,dashed] 
		coordinates{(0,78.95) (100,78.95)};
		\addlegendentry{\tiny Baseline}
		
		\end{axis}
		
		\end{tikzpicture}
	\end{center}
	\caption{\label{figure:dlm-ngram} Effects (LAS) of different number of DLMs on English and Chinese development sets.}
\end{figure}
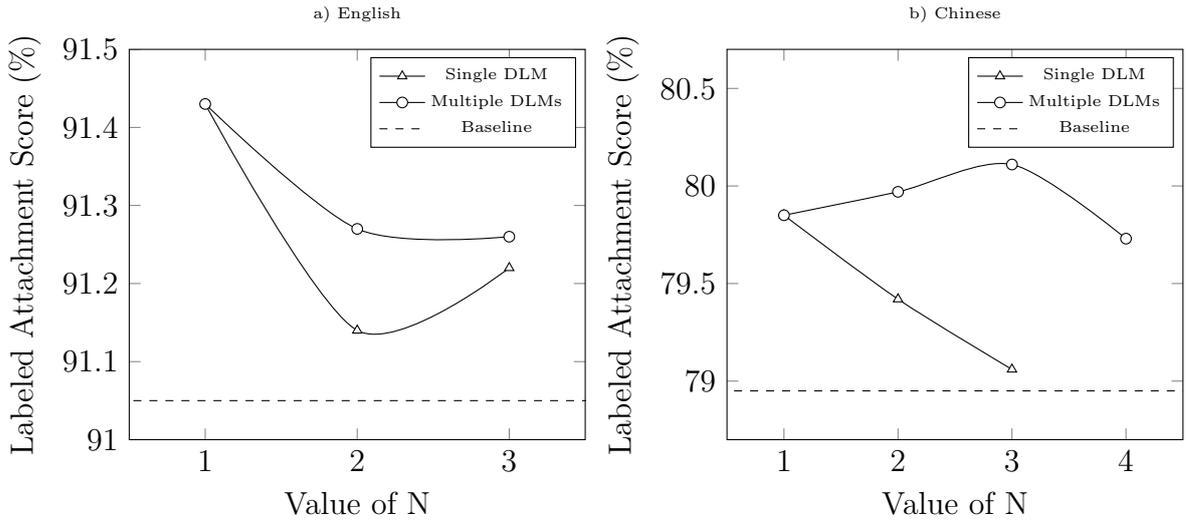

\textbf{Parsing with Single DLM.} We first evaluate the effect of the single DLM for both English and Chinese. We generate the unigram, bigram and trigram DLMs from 5 million auto-annotated sentences of the individual language. We then retrain the parser by providing different DLMs to generate new models. The lines marked with triangles in Figure \ref{figure:dlm-ngram} shows the results of our new models. Unigram DLM achieved the largest improvements for both English and Chinese. The unigram model achieved 0.38\% labelled improvement for English and the improvement for Chinese is 0.9\%.

\textbf{Parsing with Multiple DLMs.} We then evaluate the parser with multiple DLMs. We use DLMs up to N-gram to retrain the parser. Take N=2 as an example, we use both unigram and bigram DLMs for retraining. This setting allows the parser to explore multiple DLMs at the same time. We plot our multi-DLM results by lines marked with the circle in Figure \ref{figure:dlm-ngram} a) and b) for English and Chinese respectively. As we can see from the figures, the best setting for English remains the same, the parser does not gain additional improvement from the bigram and trigram. For Chinese, the improvement increased when more DLMs are used. We achieved the largest improvement by using unigram, bigram and trigram DLMs at the same time (N=3). By using multiple DLMs we achieved a 1.16\% gain on Chinese.

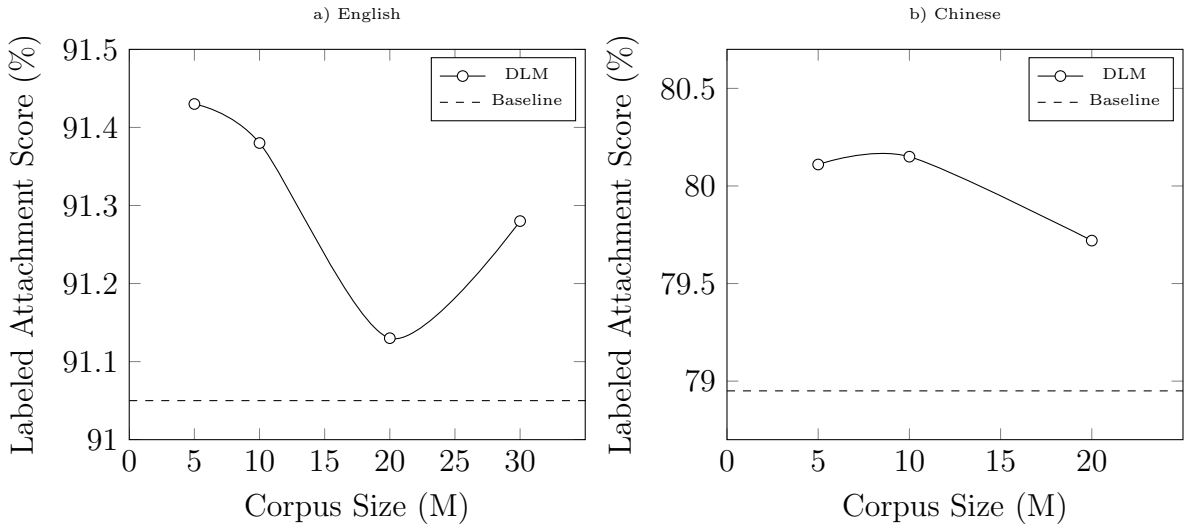
\begin{figure}[t]
	\begin{center}
		\begin{tikzpicture}
		\pgfplotsset{
			width=6cm,
			title=\tiny a) English,
			xmin=0,xmax=35,
			xtick={0,5,...,30},
			ymin=91,ymax=91.5,
			xlabel=Corpus Size (M),
			ylabel=Labeled Attachment Score (\%)
		}
		\begin{axis}

		\addplot[smooth,mark=*, mark options={fill=white}]
		coordinates{(5,91.43)(10,91.38)(20,91.13)(30,91.28)};
		\addlegendentry{\tiny DLM}
		
		%baseline
		\addplot[smooth,dashed] 
		coordinates{(0,91.05) (100,91.05)};
		\addlegendentry{\tiny Baseline}
		
		\end{axis}
		
		\end{tikzpicture}
		\begin{tikzpicture}
		\pgfplotsset{
			width=6cm,
			title=\tiny b) Chinese,
			xmin=0,xmax=25,
			xtick={0,5,...,20},
			ymin=78.7,ymax=80.7,
			xlabel=Corpus Size (M),
			ylabel=Labeled Attachment Score (\%)
		}
		\begin{axis}

		\addplot[smooth,mark=*, mark options={fill=white}]
		coordinates{(5,80.11)(10,80.15)(20,79.72)};
		\addlegendentry{\tiny DLM}
		
		%baseline
		\addplot[smooth,dashed] 
		coordinates{(0,78.95) (100,78.95)};
		\addlegendentry{\tiny Baseline}
		
		\end{axis}
		
		\end{tikzpicture}
	\end{center}
	\caption{\label{figure:dlm-size} Effects (LAS) of DLMs extracted from different size (in million sentences)  of corpus on English and Chinese development sets.}
\end{figure}

\textbf{Extracting DLMs from Larger datasets.} To determine the optimal corpus size to build DLMs we extract DLMs from different size corpora. We start with 10 million sentences and increase the size in steps until all the unlabelled data (30 million for English and 20 million for Chinese) are used. We compare our results with the best result achieved by the DLMs extracted from 5 million annotations in Figure \ref{figure:dlm-size}. The results on English data suggest that the DLMs generated from larger corpora do not gain additional improvement when compared to the one that used 5 million sentences. The Chinese results show a moderate additional gain of 0.04\% when compared to the previous best result. The effects indicate that 5 million sentences might already be enough for generating reasonably good DLMs.

\textbf{Extracting DLMs from High Quality Data.} 
To evaluate the influence of the quality of the input corpus for building the DLMs, we experiment in addition with DLMs extracted from high-quality corpora. 
The higher quality corpora are prepared by parsing unlabelled sentences with the Mate parser and the Berkeley parser. We add only the sentences that are parsed identically by both parsers to the high-quality corpus.
For Chinese, only 1 million sentences that consist of 5 tokens in average have the same syntactic structures assigned by the two parsers. Unfortunately, this amount is not sufficient for the experiments as their average sentence length is in stark contrast with the training data (27.1 tokens).
For English, we obtained 7 million sentences with an average sentence length of 16.9 tokens.
To get an impression of the quality, we parse the development set with those parsers. When the parsers agree, the parse trees have an accuracy of 97\% (LAS), while the labelled scores of both parsers are around 91\%. 
This indicates that parse trees where both parsers return the same tree have a higher accuracy.  The DLMs extracted from 7 million higher quality sentences achieved a labelled accuracy of 91.56\% which is 0.13\% higher than the best result achieved by DLMs extracted from single parsed sentences. In total, the new model outperforms the baseline by 0.51\%, with an error reduction rate of 5.7\%.

\begin{table}[t]
\begin{center}
\begin{tabular}{l|l|l|l|l}
\hline \bf System &\bf Beam&\bf POS &\bf LAS&\bf UAS \\ \hline
\newcite {zhang11}&32& 97.44&90.95&93.00\\ 
\newcite {bohnet2012eacl}&80& 97.44&91.19&93.27\\ 
\newcite{martins2013turning}&N/A& 97.44&90.55&92.89\\
\newcite{zhang2014enforcing}&N/A& 97.44&91.02&93.22\\
\newcite{chen2014neural}&1&N/A&89.60&91.80\\ 
\newcite{dyer2015slstm}&1&97.30&90.90&93.10\\ 
\newcite{weiss2015neural}&8& 97.44&92.05&93.99\\ 
\newcite{andor2016globally}&32&97.44&92.79&94.61\\
\newcite{dozat2017deep}&N/A&N/A&\bf 94.6&\bf 95.8\\
\hline
\newcite{chen2012utilizing} Baseline @&8&N/A&N/A&92.10\\
\newcite{chen2012utilizing} DLM @&8&N/A&N/A&92.76\\
Our Baseline @&40&97.33&92.44&93.38\\
\hline
Our Baseline&40&97.36&90.95&93.08\\
&80&97.34&91.05&93.28\\
&150&97.34&91.05&93.29\\ \hline
 Our DLM&40&97.38&91.41**&93.59**\\ 
 &80&97.39&91.47**&93.65**\\
 &150&97.42&91.56**&93.74**\\
 \hline

\end{tabular}
\end{center}
\caption[Comparing our DLM enhanced results with top performing parsers on English.]{\label{table:dlm-encompare} Comparing our DLM enhanced results with top performing parsers on English.  (@ results on \newcite{yamada03} conversion.)}
\end{table} 

\begin{table}[t]
	\begin{center}
		\begin{tabular}{l|l|l|l|l}
			\hline \bf System &\bf Beam&\bf POS&\bf LAS&\bf UAS \\ \hline
			\newcite{hatori2011joint}&64&93.94&N/A&81.33\\ 
			\newcite{li2012joint}&N/A&94.60&79.01&81.67\\ 
			\newcite{chen2013feature}&N/A&N/A&N/A&83.08\\ 
			\newcite{chen2015feature}&N/A&93.61&N/A&82.94\\ \hline
			Our Baseline&40&93.99&78.49&81.52\\
			&80&94.02&78.48&81.58\\
			&150&93.98&78.96&82.11\\ \hline
			Our DLM&40&94.27&79.42**&82.51**\\
			&80&94.39&79.79**&82.79**\\
			&150&94.40&\bf80.21**&\bf83.28**\\
			\hline
		\end{tabular}
	\end{center}
	\caption{\label{table:dlm-cncompare} Comparing our DLM enhanced results with top performing parsers on Chinese.}
\end{table} 

\textbf{Evaluating on Test Sets.}
We apply the best settings tuned on the development sets to the test sets. The best setting for English is the unigram DLM derived from the double parsed sentences. Table \ref{table:dlm-encompare} presents our results and top performing dependency parsers which were evaluated on the same English dataset. Our approach surpasses our baseline by 0.46/0.51\% (LAS/UAS) and is only lower than the three best neural network systems. When using a larger beam of 150, our system achieved a more competitive result. To have an idea of the performance difference between our baseline and that of \newcite{chen2012utilizing}, we include the accuracy of Mate parser on the same \newcite{yamada03} conversion used by \newcite{chen2012utilizing}. Our baseline is 0.64\% higher than their enhanced result and is 1.28\% higher than their baseline. This confirms that our approach is evaluated on a much stronger parser.  
For Chinese, we extracted the DLMs from 10 million sentences parsed by the Mate parser and using the unigram, bigram and the trigram DLMs together. Table \ref{table:dlm-cncompare} shows the results of our approach and a number of the best Chinese parsers. Our system gained a large improvement of 0.93/0.98\% for labelled and unlabelled attachment scores. Our scores with the default beam size (40) are competitive and are 0.2\% higher than the best reported result \cite{chen2013feature} when increasing the beam size to 150. 
Moreover, we gained improvements up to 0.42\% for part-of-speech tagging on Chinese tests, and our tagging accuracies for English are constantly higher than the baselines.

\begin{table}[t]
	\begin{center}
		\begin{tabular}{|l|ll|ll|}
			\cline{2-5} 
			\multicolumn{1}{c|}{}&\multicolumn{2}{|c|}{\bf DLM}&\multicolumn{2}{|c|}{\bf Baseline} \\ \cline{2-5}
			\multicolumn{1}{c|}{}&\bf LAS&\bf UAS&\bf LAS&\bf UAS\\\hline
			\sc Weblogs&79.77**&85.88**&78.99&85.1\\
			\sc Newsgroups&76.21**&83.7**&75.3&82.88\\
			\sc Reviews&75.47*&83.01&75.07&82.68\\
			\sc Answers&73.49&81.62*&73.08&81.15\\  \hline
			\sc Conll&90.43**&92.8**&90.07&92.4\\\hline
			
		\end{tabular}
	\end{center}
	\caption{\label{table:dlm-en-out-of-domain} The results of our DLM approach on English main evaluation corpus.}
\end{table}

\textbf{Results on English Main Evaluation Corpus.}
Finally, we apply our best English setting to our main evaluation corpus. We first extract new DLMs from the double parsed annotations of the LTH conversion, as LTH conversion is used in our main evaluation corpus. We then retain the parser with newly generated DLMs and apply the model to all five test domains (\textsc{Conll, Weblogs, Newsgroups, Reviews} and \textsc{Answers}). Table \ref{table:dlm-en-out-of-domain} shows the results of our best model and the baselines. Our newly trained model outperforms the baseline in all of the domains for both labelled and unlabelled accuracies. The largest improvements of 0.91\% and 0.82\% is achieved on \textsc{Newsgroups} domain for labelled and unlabelled accuracy respectively. On average our approach achieved 0.6\% labelled and unlabelled improvements for four target domains. The enhanced model also improved the source domain accuracy by 0.36\% and 0.4\% for labelled and unlabelled scores respectively.

\section{Analysis}\label{section:dlm-analysis}
In this section, we analyse the improvements achieved by our DLM-enhanced models. We analyse both English and Chinese results. For English, we analyse the results of our main evaluation corpus, as the corpus contains both in-domain and out-of-domain data. This allows us to compare the source domain and target domain results in a unified framework. We analyse the \textsc{Conll} in-domain test set and a combined out-of-domain dataset which consists of the \textsc{Weblogs, Newsgroups, Reviews} and \textsc{Answers} domain test sets. For Chinese, we analyse the in-domain test set to find out the sources of the improvements. We apply the token and sentence level analysis for both languages. The token level analysis includes the accuracy assessment of individual labels and the improvements comparison of known and unknown words. The sentence level analysis consists of assessments on four factors: sentence lengths, the number of unknown words, the number of prepositions and the number of conjunctions. For each of the factors, we group the sentences based on their properties assessed by each factor, we then calculate for each group the percentage of sentences that are improved, worsened and unchanged in accuracy. The improvements of each group can then be visualised by the gaps between improved and worsened sentences.

\subsection{English Analysis}

\subsubsection{Token Level Analysis}

%label evaluation
\begin{figure}[t]
	\begin{center}
		\begin{tikzpicture}
		\begin{axis}[
		title=\tiny a) In-domain Test Set,
		ymin=-0.5, ymax=1.8,
		width=12cm,height=5cm,
		ybar,ybar interval =0.7,
		ylabel=Accuracy Change (\%),
		enlargelimits=0.01,
		legend style={at={(0.5,-0.15)},anchor=north,legend columns=-1},
		symbolic x coords={NMOD,P,PMOD,SBJ,OBJ,ROOT,ADV,NAME,VC,COORD,TMP,DEP,CONJ,LOC,AMOD},
		xtick=data,
		x tick label style={font=\tiny,rotate=45,anchor=east}
		]
		\addplot coordinates{(NMOD,0.3) (P,0.1) (PMOD,0.2) (SBJ,0.1) (OBJ,0.5) (ROOT,0.3) (ADV,1.0) (NAME,0.5) (VC,-0.1) (COORD,0.5) (TMP,0.9) (DEP,-0.3) (CONJ,1.1) (LOC,0.7) (AMOD,-0.3) }; 
		%\addlegendentry{\tiny Recall}
		
		\addplot coordinates{(NMOD,0.3) (P,0.0) (PMOD,0.0) (SBJ,0.4) (OBJ,0.3) (ROOT,0.3) (ADV,0.4) (NAME,-0.4) (VC,0.4) (COORD,0.6) (TMP,-0.2) (DEP,-0.1) (CONJ,1.2) (LOC,1.2) (AMOD,1.4) }; 
		%\addlegendentry{\tiny Precision}
		
		\addplot coordinates{(NMOD,0.3) (P,0.1) (PMOD,0.1) (SBJ,0.2) (OBJ,0.4) (ROOT,0.3) (ADV,0.7) (NAME,0.0) (VC,0.2) (COORD,0.5) (TMP,0.3) (DEP,-0.2) (CONJ,1.2) (LOC,1.0) (AMOD,0.5) }; 
		%\addlegendentry{\tiny F-score}
		
		\end{axis}
		%NMOD cnt:15286 R/P/F95.7/95.6/95.7 Base R/P/F95.4/95.3/95.4 diff: F:0.3 R:0.3 P:0.3
		%P cnt:6831 R/P/F99.9/99.7/99.8 Base R/P/F99.8/99.7/99.7 diff: F:0.1 R:0.1 P:0.0
		%PMOD cnt:5469 R/P/F95.5/95.5/95.5 Base R/P/F95.3/95.5/95.4 diff: F:0.1 R:0.2 P:0.0
		%SBJ cnt:4332 R/P/F96.8/97.5/97.1 Base R/P/F96.7/97.1/96.9 diff: F:0.2 R:0.1 P:0.4
		%OBJ cnt:3073 R/P/F95.1/93.7/94.4 Base R/P/F94.6/93.4/94.0 diff: F:0.4 R:0.5 P:0.3
		%ROOT cnt:2399 R/P/F96.3/96.3/96.3 Base R/P/F96.0/96.0/96.0 diff: F:0.3 R:0.3 P:0.3
		%ADV cnt:2091 R/P/F84.1/75.5/79.8 Base R/P/F83.1/75.1/79.1 diff: F:0.7 R:1.0 P:0.4
		%NAME cnt:2002 R/P/F94.7/94.0/94.3 Base R/P/F94.2/94.4/94.3 diff: F:0.0 R:0.5 P:-0.4
		%VC cnt:1804 R/P/F97.4/97.4/97.4 Base R/P/F97.5/97.0/97.2 diff: F:0.2 R:-0.1 P:0.4
		%COORD cnt:1374 R/P/F92.8/94.2/93.5 Base R/P/F92.3/93.6/93.0 diff: F:0.5 R:0.5 P:0.6
		%TMP cnt:1341 R/P/F84.6/86.8/85.7 Base R/P/F83.7/87.0/85.4 diff: F:0.3 R:0.9 P:-0.2
		%DEP cnt:1302 R/P/F86.0/85.4/85.7 Base R/P/F86.3/85.5/85.9 diff: F:-0.2 R:-0.3 P:-0.1
		%CONJ cnt:1103 R/P/F91.6/91.2/91.4 Base R/P/F90.5/90.0/90.2 diff: F:1.2 R:1.1 P:1.2
		%LOC cnt:955 R/P/F84.1/87.5/85.8 Base R/P/F83.4/86.3/84.8 diff: F:1.0 R:0.7 P:1.2
		%AMOD cnt:913 R/P/F77.0/84.5/80.7 Base R/P/F77.3/83.1/80.2 diff: F:0.5 R:-0.3 P:1.4
		
		\end{tikzpicture}
		\begin{tikzpicture}
		\begin{axis}[
		title=\tiny b) Out-of-domain Test Set,
		ymin=-0.5, ymax=1.8,
		width=12cm,height=5cm,
		ybar,ybar interval =0.7,
		ylabel=Accuracy Change (\%),
		enlargelimits=0.01,
		legend style={at={(0.5,-0.2)},anchor=north,legend columns=-1},
		symbolic x coords={NMOD,P,SBJ,PMOD,ROOT,OBJ,ADV,COORD,VC,CONJ,DEP,PRD,AMOD,TMP,IM},
		xtick=data,
		x tick label style={font=\tiny,rotate=45,anchor=east}
		]
		\addplot coordinates{(NMOD,0.5) (P,0.2) (SBJ,0.2) (PMOD,0.7) (ROOT,0.2) (OBJ,1.0) (ADV,0.0) (COORD,0.3) (VC,0.8) (CONJ,0.7) (DEP,0.7) (PRD,1.7) (AMOD,0.3) (TMP,0.4) (IM,-0.2) }; 
		\addlegendentry{\tiny Recall}
		
		\addplot coordinates{(NMOD,0.6) (P,0.1) (SBJ,0.2) (PMOD,0.4) (ROOT,0.2) (OBJ,1.2) (ADV,0.3) (COORD,0.5) (VC,0.4) (CONJ,0.3) (DEP,0.3) (PRD,0.5) (AMOD,1.0) (TMP,-0.2) (IM,0.3) }; 
		\addlegendentry{\tiny Precision}
		
		\addplot coordinates{(NMOD,0.6) (P,0.1) (SBJ,0.3) (PMOD,0.6) (ROOT,0.2) (OBJ,1.1) (ADV,0.2) (COORD,0.4) (VC,0.7) (CONJ,0.5) (DEP,0.5) (PRD,1.1) (AMOD,0.6) (TMP,0.1) (IM,0.1) }; 
		\addlegendentry{\tiny F-score}
		
		\end{axis}
		%NMOD cnt:24656 R/P/F89.8/88.0/88.9 Base R/P/F89.3/87.4/88.3 diff: F:0.6 R:0.5 P:0.6
		%P cnt:11977 R/P/F92.9/98.8/95.8 Base R/P/F92.7/98.7/95.7 diff: F:0.1 R:0.2 P:0.1
		%SBJ cnt:9199 R/P/F92.2/88.1/90.2 Base R/P/F92.0/87.9/89.9 diff: F:0.3 R:0.2 P:0.2
		%PMOD cnt:9195 R/P/F90.3/90.3/90.3 Base R/P/F89.6/89.9/89.7 diff: F:0.6 R:0.7 P:0.4
		%ROOT cnt:6345 R/P/F86.0/86.0/86.0 Base R/P/F85.8/85.8/85.8 diff: F:0.2 R:0.2 P:0.2
		%OBJ cnt:6291 R/P/F88.1/70.7/79.4 Base R/P/F87.1/69.5/78.3 diff: F:1.1 R:1.0 P:1.2
		%ADV cnt:6088 R/P/F67.1/69.6/68.4 Base R/P/F67.1/69.3/68.2 diff: F:0.2 R:0.0 P:0.3
		%COORD cnt:4751 R/P/F81.7/89.4/85.6 Base R/P/F81.4/88.9/85.2 diff: F:0.4 R:0.3 P:0.5
		%VC cnt:4028 R/P/F89.9/94.0/92.0 Base R/P/F89.1/93.6/91.3 diff: F:0.7 R:0.8 P:0.4
		%CONJ cnt:3455 R/P/F86.3/86.2/86.2 Base R/P/F85.6/85.9/85.7 diff: F:0.5 R:0.7 P:0.3
		%DEP cnt:2866 R/P/F25.2/59.6/42.4 Base R/P/F24.5/59.3/41.9 diff: F:0.5 R:0.7 P:0.3
		%PRD cnt:2827 R/P/F45.3/89.9/67.6 Base R/P/F43.6/89.4/66.5 diff: F:1.1 R:1.7 P:0.5
		%AMOD cnt:2511 R/P/F57.6/72.9/65.2 Base R/P/F57.3/71.9/64.6 diff: F:0.6 R:0.3 P:1.0
		%TMP cnt:2387 R/P/F65.4/66.8/66.1 Base R/P/F65.0/67.0/66.0 diff: F:0.1 R:0.4 P:-0.2
		%IM cnt:1692 R/P/F97.9/97.5/97.7 Base R/P/F98.1/97.2/97.6 diff: F:0.1 R:-0.2 P:0.3
		
		\end{tikzpicture}
	\end{center}
	\caption{\label{figure:analysis-label-dlm-en} The English performance comparison between the DLM approach and the baseline on major labels.}
\end{figure}
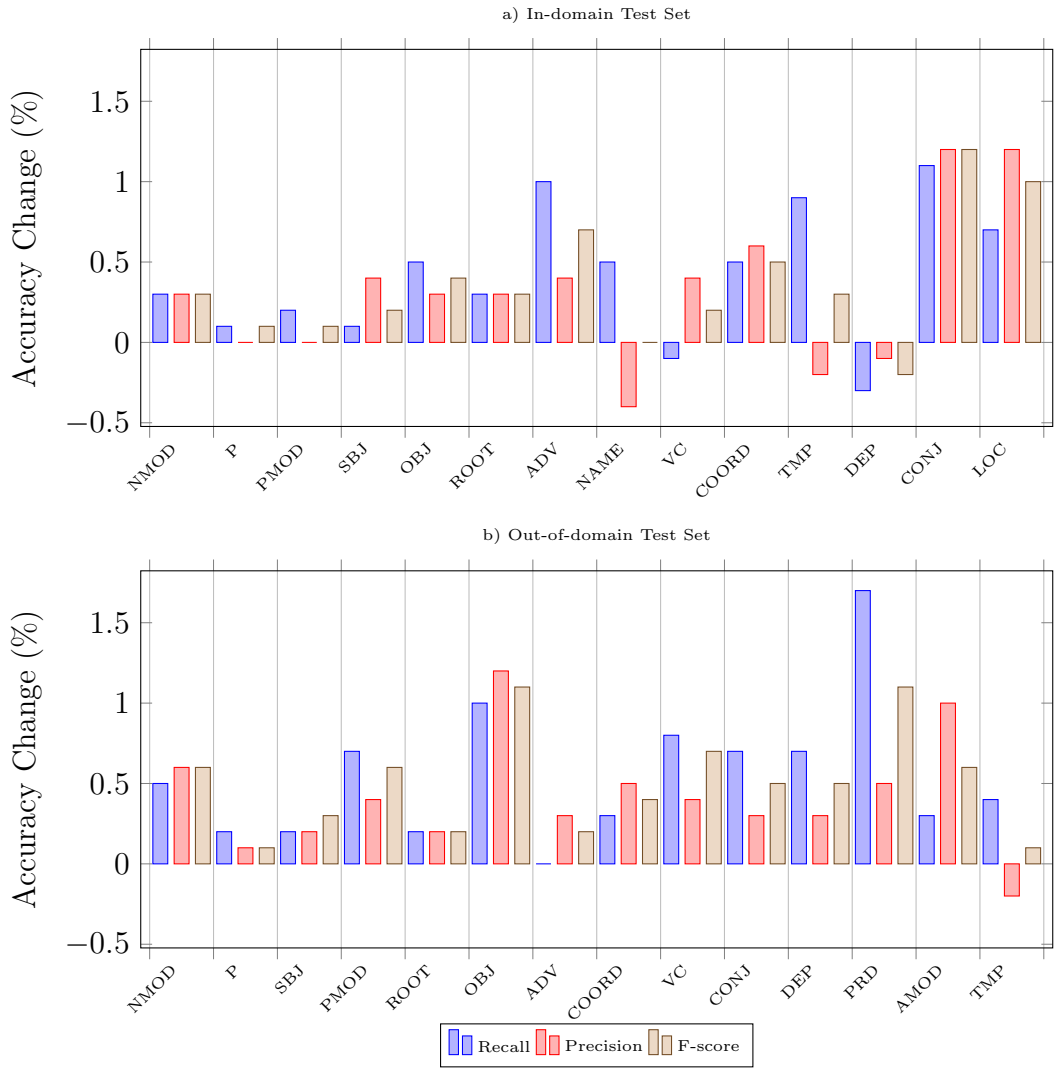

%Confution matric
\begin{table}[t]
\begin{center}
\begin{tabular}{l|r|r}
\hline
\bf Confusion & \bf Baseline & \bf DLM\\\hline
\footnotesize NMOD $\rightarrow$ \footnotesize ADV & 105 & 104\\
\footnotesize NMOD $\rightarrow$ \footnotesize OBJ & 37 & 28\\
\footnotesize NMOD $\rightarrow$ \footnotesize AMOD & 41 & 33\\
\footnotesize NMOD $\rightarrow$ \footnotesize DEP & 112 & 109\\
\footnotesize NMOD $\rightarrow$ \footnotesize NAME & 61 & 66\\
\footnotesize NMOD $\rightarrow$ \footnotesize APPO & 34 & 34\\
\footnotesize NMOD $\rightarrow$ \footnotesize PMOD & 83 & 76\\
\footnotesize NMOD $\rightarrow$ \footnotesize SBJ,TMP,CONJ & 75 & 72\\\hline
\footnotesize SBJ $\rightarrow$ \footnotesize NMOD & 33 & 29\\
\footnotesize SBJ $\rightarrow$ \footnotesize PMOD & 28 & 29\\\hline
\footnotesize OBJ $\rightarrow$ \footnotesize NMOD & 48 & 41\\\hline
\footnotesize PMOD $\rightarrow$ \footnotesize NMOD & 67 & 56\\
\footnotesize PMOD $\rightarrow$ \footnotesize OBJ & 34 & 37\\
\footnotesize PMOD $\rightarrow$ \footnotesize APPO,TMP & 46 & 45\\\hline
\footnotesize ROOT $\rightarrow$ \footnotesize NMOD & 22 & 19\\\hline
\footnotesize ADV $\rightarrow$ \footnotesize LOC & 33 & 28\\
\footnotesize ADV $\rightarrow$ \footnotesize NMOD & 103 & 93\\
\footnotesize ADV $\rightarrow$ \footnotesize TMP & 43 & 45\\
\footnotesize ADV $\rightarrow$ \footnotesize MNR,APPO,AMOD & 74 & 69\\\hline
\footnotesize CONJ $\rightarrow$ \footnotesize NMOD & 41 & 35\\\hline
\footnotesize DEP $\rightarrow$ \footnotesize NMOD & 74 & 82\\\hline
\footnotesize NAME $\rightarrow$ \footnotesize NMOD & 55 & 52\\\hline
\footnotesize TMP $\rightarrow$ \footnotesize LOC & 33 & 28\\
\footnotesize TMP $\rightarrow$ \footnotesize NMOD & 36 & 37\\
\footnotesize TMP $\rightarrow$ \footnotesize ADV & 84 & 80\\\hline
\footnotesize LOC $\rightarrow$ \footnotesize NMOD & 58 & 47\\
\footnotesize LOC $\rightarrow$ \footnotesize ADV & 59 & 62\\\hline
\end{tabular}
\end{center}
\caption{\label{table:analysis-confusion-dlm-in} The confusion matrix of dependency labels, compared between the DLM approach and the baseline on the in-domain test set.}
\end{table}

%Confution matric
\begin{table}
\begin{center}
\begin{tabular}{l|r|r}
\hline
\bf Confusion & \bf Baseline & \bf DLM\\\hline
\footnotesize NMOD $\rightarrow$ \footnotesize ADV & 235 & 219\\
\footnotesize NMOD $\rightarrow$ \footnotesize LOC & 162 & 156\\
\footnotesize NMOD $\rightarrow$ \footnotesize HYPH & 198 & 200\\
\footnotesize NMOD $\rightarrow$ \footnotesize NAME & 569 & 559\\
\footnotesize NMOD $\rightarrow$ \footnotesize PMOD & 187 & 174\\
\footnotesize NMOD $\rightarrow$ \footnotesize HMOD & 217 & 218\\
\footnotesize NMOD $\rightarrow$ \footnotesize ROOT,OBJ,SBJ,DEP & 491 & 470\\\hline
\footnotesize P $\rightarrow$ \footnotesize HYPH & 162 & 170\\
\footnotesize P $\rightarrow$ \footnotesize NAME,NMOD & 233 & 221\\\hline
\footnotesize SBJ $\rightarrow$ \footnotesize NMOD & 169 & 156\\
\footnotesize SBJ $\rightarrow$ \footnotesize OBJ & 132 & 107\\\hline
\footnotesize OBJ $\rightarrow$ \footnotesize NMOD & 218 & 191\\
\footnotesize OBJ $\rightarrow$ \footnotesize SBJ & 117 & 106\\\hline
\footnotesize PMOD $\rightarrow$ \footnotesize NMOD & 290 & 279\\
\footnotesize PMOD $\rightarrow$ \footnotesize OBJ & 122 & 108\\\hline
\footnotesize ROOT $\rightarrow$ \footnotesize NMOD & 235 & 240\\
\footnotesize ROOT $\rightarrow$ \footnotesize OBJ,SBJ & 256 & 244\\\hline
\footnotesize ADV $\rightarrow$ \footnotesize MNR & 150 & 156\\
\footnotesize ADV $\rightarrow$ \footnotesize AMOD & 152 & 134\\
\footnotesize ADV $\rightarrow$ \footnotesize LOC & 227 & 210\\
\footnotesize ADV $\rightarrow$ \footnotesize NMOD & 382 & 362\\
\footnotesize ADV $\rightarrow$ \footnotesize TMP & 182 & 204\\
\footnotesize ADV $\rightarrow$ \footnotesize DIR & 118 & 110\\\hline
\footnotesize COORD $\rightarrow$ \footnotesize NMOD & 164 & 143\\
\footnotesize COORD $\rightarrow$ \footnotesize ROOT & 102 & 96\\\hline
\footnotesize VC $\rightarrow$ \footnotesize OPRD & 114 & 83\\\hline
\footnotesize CONJ $\rightarrow$ \footnotesize NMOD & 132 & 113\\\hline
\footnotesize DEP $\rightarrow$ \footnotesize ROOT & 190 & 186\\
\footnotesize DEP $\rightarrow$ \footnotesize OBJ & 267 & 244\\
\footnotesize DEP $\rightarrow$ \footnotesize SBJ & 403 & 394\\
\footnotesize DEP $\rightarrow$ \footnotesize NMOD & 382 & 392\\
\footnotesize DEP $\rightarrow$ \footnotesize TMP & 176 & 183\\
\footnotesize DEP $\rightarrow$ \footnotesize ADV & 142 & 133\\\hline
\footnotesize AMOD $\rightarrow$ \footnotesize ADV & 169 & 179\\
\footnotesize AMOD $\rightarrow$ \footnotesize NMOD & 265 & 270\\
\footnotesize AMOD $\rightarrow$ \footnotesize HYPH & 104 & 106\\\hline
\footnotesize TMP $\rightarrow$ \footnotesize ADV & 280 & 283\\
\footnotesize TMP $\rightarrow$ \footnotesize NMOD & 133 & 122\\\hline
\footnotesize PRD $\rightarrow$ \footnotesize OBJ & 854 & 834\\
\footnotesize PRD $\rightarrow$ \footnotesize ADV,VC & 255 & 238\\\hline
\end{tabular}
\end{center}
\caption{\label{table:analysis-confusion-dlm-out} The confusion matrix of dependency labels, compared between the DLM approach and the baseline on the out-of-domain test sets.}
\end{table}

\textbf{Individual Label Accuracy.} We first analyse accuracy changes of most frequent labels of our in-domain and out-of-domain test sets. As we can see from Figure \ref{figure:analysis-label-dlm-en} the most frequent labels of in-domain data are slightly different from that of out-of-domain data. Label NAME (name-internal link) and LOC (locative adverbial) that frequently showed in the in-domain set is less frequent in out-of-domain data. Instead, the out-of-domain data have more PRD (predicative complement) and AMOD (modifier of adjective or adverbial) than in-domain data. In term of the improvements of individual labels, they both show improvements on most of the labels. They achieved improvements of at least 0.4\% on label OBJ (object), COORD (coordination), CONJ (conjunct). More precisely, the DLM model achieved large improvements of more than 1\% for in-domain data on CONJ (conjunct) and LOC (locative adverbial) and gained moderate improvements of more than 0.4\% on OBJ (object), COORD (coordination) and ADV (adverbial). While for out-of-domain data, our approach gained more than 1\% f-scores on OBJ (object) and PRD (predicative complement), and improved three major modifiers (NMOD, PMOD and AMOD), VC (verb chain), COORD (coordination), CONJ (conjunct) and DEP (unclassified) for more than 0.4\%. Table \ref{table:analysis-confusion-dlm-in} and table \ref{table:analysis-confusion-dlm-out} show the confusion matrices of the dependency labels on in-domain and out-of-domain test sets respectively. 

%corpus UNK
\begin{table}[t]
	\begin{center}
		\begin{adjustbox}{max width=15cm}
		\begin{tabular}{|l|r|rr|rr|r|rr|rr|}
			\cline{2-11}
			\multicolumn{1}{c|}{}&\multicolumn{5}{|c|}{\bf In-domain Test Set}&\multicolumn{5}{|c|}{\bf Out-of-domain Test Set}\\
			\cline{2-11}
			\multicolumn{1}{c|}{}&& \multicolumn{2}{|c|}{\bf DLM} &\multicolumn{2}{|c|}{\bf Baseline}&& \multicolumn{2}{|c|}{\bf DLM} &\multicolumn{2}{|c|}{\bf Baseline}\\
			\hline
			&\bf Tokens &\bf LAS&\bf UAS&\bf LAS&\bf UAS&\bf Tokens &\bf LAS&\bf UAS&\bf LAS&\bf UAS\\
			\hline
			\bf Known &56640&90.4&92.8&90.1&92.4&101616&77.8&84.7&77.1&84.1\\
			%diff: LAS:0.3 UAS:0.4   %diff: LAS:0.7 UAS:0.6
			\bf Unknown &1036&91.1&93.5&90.1&93.1&6055&62.0&72.3&61.4&71.9\\
			%diff: LAS:1.0 UAS:0.4   %diff: LAS:0.6 UAS:0.4
			\hline
			\bf All &57676&90.4&92.8&90.1&92.4&107671&76.9&84.0&76.3&83.4\\
			%diff: LAS:0.3 UAS:0.4   %diff: LAS:0.6 UAS:0.6
			\hline
		\end{tabular}
	\end{adjustbox}
	\end{center}
	\caption{\label{table:analysis-corpusunk-dlm-en} The English accuracy comparison between the DLM approach and the baseline on unknown words.}
\end{table}

%number of tokens
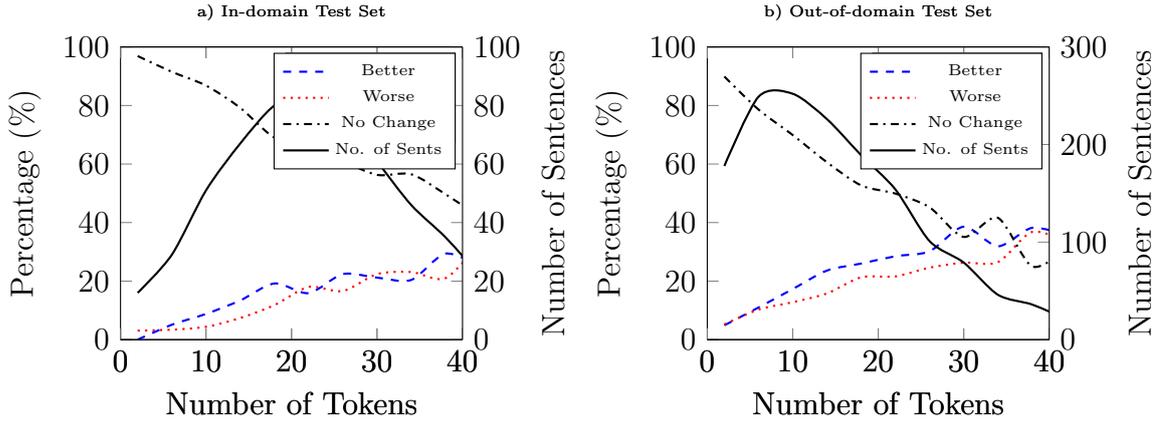
\begin{figure}[t]
	\begin{center}
		\begin{tikzpicture}
		\pgfplotsset{
			width=4.5cm,
			title=\tiny a) In-domain Test Set,
			xmin=0,xmax=40,
			xlabel=Number of Tokens}
		\begin{axis}[
		axis y line*=left,
		ymin=0,ymax=100,
		ylabel=Percentage (\%)]
		%Better
		\addplot[smooth,thick,dashed,color=blue] coordinates {(2,0.0) (6,5.1) (10,8.8) (14,13.4) (18,19.2) (22,15.9) (26,22.4) (30,21.2) (34,20.4) (38,29.4) (42,25.0) };\label{better}
		\addlegendentry{\tiny Better}
		
		%Worse
		\addplot[smooth,thick,dotted,color=red] coordinates{(2,3.1) (6,3.4) (10,4.4) (14,7.4) (18,11.8) (22,18.0) (26,16.6) (30,22.4) (34,23.1) (38,21.0) (42,33.0) };\label{worse}
		\addlegendentry{\tiny Worse}
		
		%No Change
		\addplot[smooth,thick,dashdotted] coordinates{(2,96.9) (6,91.5) (10,86.8) (14,79.2) (18,69.0) (22,66.1) (26,61.0) (30,56.3) (34,56.5) (38,49.7) (42,42.0) };\label{nochange}
		\addlegendentry{\tiny No Change}
		
		\end{axis}
		\begin{axis}[
		axis y line*=right,
		ymin=0,ymax=100,
		ylabel=Number of Sentences]
		\addlegendimage{/pgfplots/refstyle=better}\addlegendentry{\tiny Better}
		\addlegendimage{/pgfplots/refstyle=worse}\addlegendentry{\tiny Worse}
		\addlegendimage{/pgfplots/refstyle=nochange}\addlegendentry{\tiny No Change}
		%All Sent
		\addplot[smooth,thick,solid] coordinates{(2,16.0) (6,29.0) (10,51.0) (14,67.0) (18,80.0) (22,84.0) (26,77.0) (30,61.0) (34,46.0) (38,35.0) (42,22.0) };\addlegendentry{\tiny No. of Sents}
		
		\end{axis}
		%2 Total:64 LAS better/worse/nochange:0/2/62 0.0/3.1/96.9 Diff:-3.1
		%6 Total:117 LAS better/worse/nochange:6/4/107 5.1/3.4/91.5 Diff:1.7
		%10 Total:205 LAS better/worse/nochange:18/9/178 8.8/4.4/86.8 Diff:4.4
		%14 Total:269 LAS better/worse/nochange:36/20/213 13.4/7.4/79.2 Diff:6.0
		%18 Total:323 LAS better/worse/nochange:62/38/223 19.2/11.8/69.0 Diff:7.4
		%22 Total:339 LAS better/worse/nochange:54/61/224 15.9/18.0/66.1 Diff:-2.1
		%26 Total:308 LAS better/worse/nochange:69/51/188 22.4/16.6/61.0 Diff:5.8
		%30 Total:245 LAS better/worse/nochange:52/55/138 21.2/22.4/56.3 Diff:-1.2
		%34 Total:186 LAS better/worse/nochange:38/43/105 20.4/23.1/56.5 Diff:-2.7
		%38 Total:143 LAS better/worse/nochange:42/30/71 29.4/21.0/49.7 Diff:8.4
		%42 Total:88 LAS better/worse/nochange:22/29/37 25.0/33.0/42.0 Diff:-8.0
		
		\end{tikzpicture}
		\begin{tikzpicture}
		\pgfplotsset{
			width=4.5cm,
			title=\tiny b) Out-of-domain Test Set,
			xmin=0,xmax=40,
			xlabel=Number of Tokens}
		\begin{axis}[
		axis y line*=left,
		ymin=0,ymax=100,
		ylabel=Percentage (\%)]
		%Better
		\addplot[smooth,thick,dashed,color=blue] coordinates {(2,4.9) (6,11.0) (10,17.3) (14,23.6) (18,26.0) (22,28.5) (26,30.2) (30,38.6) (34,31.9) (38,38.2) (42,35.6) };\label{better}
		\addlegendentry{\tiny Better}
		
		%Worse
		\addplot[smooth,thick,dotted,color=red] coordinates{(2,5.2) (6,10.3) (10,12.8) (14,15.8) (18,21.3) (22,21.6) (26,24.6) (30,26.3) (34,26.5) (38,36.8) (42,33.3) };\label{worse}
		\addlegendentry{\tiny Worse}
		
		%No Change
		\addplot[smooth,thick,dashdotted] coordinates{(2,89.9) (6,78.6) (10,69.9) (14,60.6) (18,52.7) (22,49.9) (26,45.2) (30,35.1) (34,41.6) (38,25.0) (42,31.0) };\label{nochange}
		\addlegendentry{\tiny No Change}
		
		\end{axis}
		\begin{axis}[
		axis y line*=right,
		ymin=0,ymax=300,
		ylabel=Number of Sentences]
		\addlegendimage{/pgfplots/refstyle=better}\addlegendentry{\tiny Better}
		\addlegendimage{/pgfplots/refstyle=worse}\addlegendentry{\tiny Worse}
		\addlegendimage{/pgfplots/refstyle=nochange}\addlegendentry{\tiny No Change}
		%All Sent
		\addplot[smooth,thick,solid] coordinates{(2,178.0) (6,249.0) (10,252.0) (14,226.0) (18,189.0) (22,154.0) (26,101.0) (30,79.0) (34,46.0) (38,36.0) (42,21.0) };\addlegendentry{\tiny No. of Sents}
		
		\end{axis}
		%2 Total:713 LAS better/worse/nochange:35/37/641 4.9/5.2/89.9 Diff:-0.3
		%6 Total:996 LAS better/worse/nochange:110/103/783 11.0/10.3/78.6 Diff:0.7
		%10 Total:1010 LAS better/worse/nochange:175/129/706 17.3/12.8/69.9 Diff:4.5
		%14 Total:906 LAS better/worse/nochange:214/143/549 23.6/15.8/60.6 Diff:7.8
		%18 Total:759 LAS better/worse/nochange:197/162/400 26.0/21.3/52.7 Diff:4.7
		%22 Total:617 LAS better/worse/nochange:176/133/308 28.5/21.6/49.9 Diff:6.9
		%26 Total:407 LAS better/worse/nochange:123/100/184 30.2/24.6/45.2 Diff:5.6
		%30 Total:316 LAS better/worse/nochange:122/83/111 38.6/26.3/35.1 Diff:12.3
		%34 Total:185 LAS better/worse/nochange:59/49/77 31.9/26.5/41.6 Diff:5.4
		%38 Total:144 LAS better/worse/nochange:55/53/36 38.2/36.8/25.0 Diff:1.4
		%42 Total:87 LAS better/worse/nochange:31/29/27 35.6/33.3/31.0 Diff:2.3
		
		\end{tikzpicture}
	\end{center}
	\caption{\label{figure:analysis-sentlength-dlm-en} The English comparison between the DLM approach and the baseline on different number of tokens per sentence. }
\end{figure}

\textbf{Unknown Words Accuracy.} The unknown words rate for the in-domain test set is much lower than that of the out-of-domain one. For the in-domain test set, only 1,000 tokens are unknown and surprisingly both the DLM model and the base model have a better accuracy on the unknown words. Our DLM model achieved labelled improvement of 1\% on the unknown words which is 3 times than the gain for that of known words (0.3\%). While the unlabelled improvement for both known and unknown words are exactly the same 0.4\%. The larger improvement on out-of-domain data is achieved on the known words, with a 0.1\%-0.2\% small difference when compared to that of unknown words. A detailed comparison can be found in Table \ref{table:analysis-corpusunk-dlm-en}.

\subsubsection{Sentence Level Analysis}

%number of unknown words
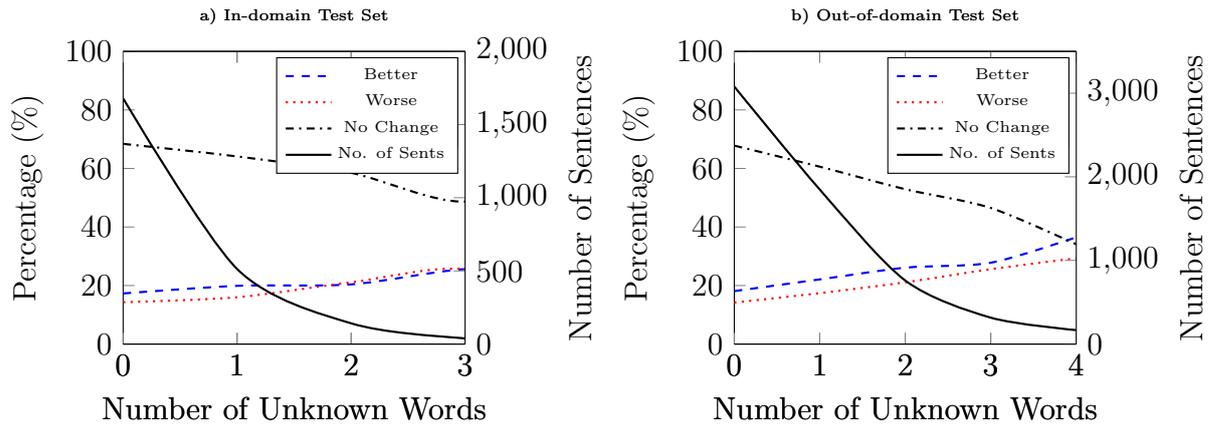
\begin{figure}[t]
	\begin{center}
		\begin{tikzpicture}
		\pgfplotsset{
			width=4.5cm,
			title=\tiny a) In-domain Test Set,
			xmin=0,xmax=3,
			xlabel=Number of Unknown Words}
		\begin{axis}[
		axis y line*=left,
		ymin=0,ymax=100,
		ylabel=Percentage (\%)]
		%Better
		\addplot[smooth,thick,dashed,color=blue] coordinates {(0,17.3) (1,19.9) (2,20.4) (3,25.6) (4,25.0) };\label{better}
		\addlegendentry{\tiny Better}
		
		%Worse
		\addplot[smooth,thick,dotted,color=red] coordinates{(0,14.3) (1,16.0) (2,21.1) (3,25.6) (4,12.5) };\label{worse}
		\addlegendentry{\tiny Worse}
		
		%No Change
		\addplot[smooth,thick,dashdotted] coordinates{(0,68.4) (1,64.1) (2,58.5) (3,48.7) (4,62.5) };\label{nochange}
		\addlegendentry{\tiny No Change}
		
		\end{axis}
		\begin{axis}[
		axis y line*=right,
		ymin=0,ymax=2000,
		ylabel=Number of Sentences]
		\addlegendimage{/pgfplots/refstyle=better}\addlegendentry{\tiny Better}
		\addlegendimage{/pgfplots/refstyle=worse}\addlegendentry{\tiny Worse}
		\addlegendimage{/pgfplots/refstyle=nochange}\addlegendentry{\tiny No Change}
		%All Sent
		\addplot[smooth,thick,solid] coordinates{(0,1679.0) (1,513.0) (2,142.0) (3,39.0) (4,16.0) };\addlegendentry{\tiny No. of Sents}
		
		\end{axis}
		%0 Total:1679 LAS better/worse/nochange:290/240/1149 17.3/14.3/68.4 Diff:3.0
		%1 Total:513 LAS better/worse/nochange:102/82/329 19.9/16.0/64.1 Diff:3.9
		%2 Total:142 LAS better/worse/nochange:29/30/83 20.4/21.1/58.5 Diff:-0.7
		%3 Total:39 LAS better/worse/nochange:10/10/19 25.6/25.6/48.7 Diff:0.0
		%4 Total:16 LAS better/worse/nochange:4/2/10 25.0/12.5/62.5 Diff:12.5
		
		\end{tikzpicture}
		\begin{tikzpicture}
		\pgfplotsset{
			width=4.5cm,
			title=\tiny b) Out-of-domain Test Set,
			xmin=0,xmax=4,
			xlabel=Number of Unknown Words}
		\begin{axis}[
		axis y line*=left,
		ymin=0,ymax=100,
		ylabel=Percentage (\%)]
		%Better
		\addplot[smooth,thick,dashed,color=blue] coordinates {(0,18.1) (1,22.1) (2,26.1) (3,27.8) (4,36.5) };\label{better}
		\addlegendentry{\tiny Better}
		
		%Worse
		\addplot[smooth,thick,dotted,color=red] coordinates{(0,14.2) (1,17.4) (2,21.1) (3,25.6) (4,29.3) };\label{worse}
		\addlegendentry{\tiny Worse}
		
		%No Change
		\addplot[smooth,thick,dashdotted] coordinates{(0,67.8) (1,60.6) (2,52.9) (3,46.5) (4,34.1) };\label{nochange}
		\addlegendentry{\tiny No Change}
		
		\end{axis}
		\begin{axis}[
		axis y line*=right,
		ymin=0,ymax=3500,
		ylabel=Number of Sentences]
		\addlegendimage{/pgfplots/refstyle=better}\addlegendentry{\tiny Better}
		\addlegendimage{/pgfplots/refstyle=worse}\addlegendentry{\tiny Worse}
		\addlegendimage{/pgfplots/refstyle=nochange}\addlegendentry{\tiny No Change}
		%All Sent
		\addplot[smooth,thick,solid] coordinates{(0,3079.0) (1,1848.0) (2,760.0) (3,316.0) (4,167.0) };\addlegendentry{\tiny No. of Sents}
		
		\end{axis}
		%0 Total:3079 LAS better/worse/nochange:556/436/2087 18.1/14.2/67.8 Diff:3.9
		%1 Total:1848 LAS better/worse/nochange:408/321/1119 22.1/17.4/60.6 Diff:4.7
		%2 Total:760 LAS better/worse/nochange:198/160/402 26.1/21.1/52.9 Diff:5.0
		%3 Total:316 LAS better/worse/nochange:88/81/147 27.8/25.6/46.5 Diff:2.2
		%4 Total:167 LAS better/worse/nochange:61/49/57 36.5/29.3/34.1 Diff:7.2
		
		\end{tikzpicture}
	\end{center}
	\caption{\label{figure:analysis-sentunk-dlm-en} The English comparison between the DLM approach and the baseline on different number of unknown words per sentence. }
\end{figure}

\textbf{Sentence Length.} Figure \ref{figure:analysis-sentlength-dlm-en} shows our analysis on sentence length. The analysis of in-domain data shows the DLM model mostly helped the sentences consisting of 10-20 tokens. For sentences shorter than 10 tokens the DLM model even shows some negative effects. We suggest this might because for in-domain parsing the base model is already able to achieve a high accuracy on short sentences thus they are harder to improve. When sentences are longer than 20 tokens, the rates for both improved and worsened sentences varies, but the overall positive and negative effects are similar. In terms of the analysis on out-of-domain set, positive effects of more than 4.5\% can be found in sentences that have a length of 10-35 tokens, but not in sentences shorter than 10 tokens.

\textbf{Unknown Words.} As stated before, the in-domain test set contains fewer unknown words. In fact, most of the sentences do not contain unknown words or only have one unknown word. The DLM model achieved 3\% gain for the former and 3.9\% gain for the latter. For analysis of the out-of-domain data, our DLM model showed similar gains of around 5\% for all the classes. Figure \ref{figure:analysis-sentunk-dlm-en} shows our analysis on unknown words.

%number of prepositions
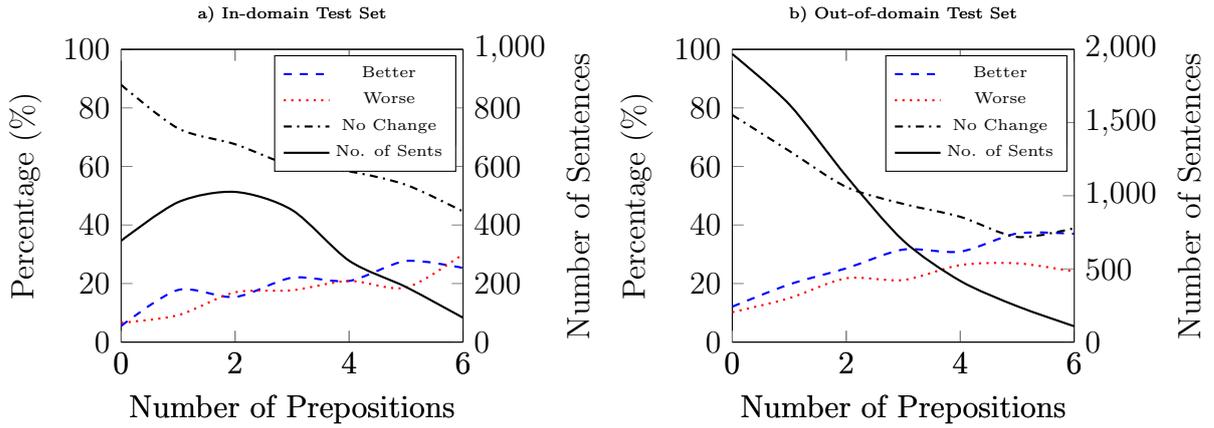
\begin{figure}[t]
	\begin{center}
		\begin{tikzpicture}
		\pgfplotsset{
			width=4.5cm,
			title=\tiny a) In-domain Test Set,
			xmin=0,xmax=6,
			xlabel=Number of Prepositions}
		\begin{axis}[
		axis y line*=left,
		ymin=0,ymax=100,
		ylabel=Percentage (\%)]
		%Better
		\addplot[smooth,thick,dashed,color=blue] coordinates {(0,5.5) (1,17.8) (2,15.4) (3,22.0) (4,20.9) (5,27.7) (6,25.3) };\label{better}
		\addlegendentry{\tiny Better}
		
		%Worse
		\addplot[smooth,thick,dotted,color=red] coordinates{(0,6.6) (1,9.2) (2,17.0) (3,17.7) (4,20.9) (5,18.6) (6,30.1) };\label{worse}
		\addlegendentry{\tiny Worse}
		
		%No Change
		\addplot[smooth,thick,dashdotted] coordinates{(0,87.9) (1,73.0) (2,67.6) (3,60.3) (4,58.3) (5,53.7) (6,44.6) };\label{nochange}
		\addlegendentry{\tiny No Change}
		
		\end{axis}
		\begin{axis}[
		axis y line*=right,
		ymin=0,ymax=1000,
		ylabel=Number of Sentences]
		\addlegendimage{/pgfplots/refstyle=better}\addlegendentry{\tiny Better}
		\addlegendimage{/pgfplots/refstyle=worse}\addlegendentry{\tiny Worse}
		\addlegendimage{/pgfplots/refstyle=nochange}\addlegendentry{\tiny No Change}
		%All Sent
		\addplot[smooth,thick,solid] coordinates{(0,346.0) (1,478.0) (2,513.0) (3,451.0) (4,278.0) (5,188.0) (6,83.0) };\addlegendentry{\tiny No. of Sents}
		
		\end{axis}
		%0 Total:346 LAS better/worse/nochange:19/23/304 5.5/6.6/87.9 Diff:-1.1
		%1 Total:478 LAS better/worse/nochange:85/44/349 17.8/9.2/73.0 Diff:8.6
		%2 Total:513 LAS better/worse/nochange:79/87/347 15.4/17.0/67.6 Diff:-1.6
		%3 Total:451 LAS better/worse/nochange:99/80/272 22.0/17.7/60.3 Diff:4.3
		%4 Total:278 LAS better/worse/nochange:58/58/162 20.9/20.9/58.3 Diff:0.0
		%5 Total:188 LAS better/worse/nochange:52/35/101 27.7/18.6/53.7 Diff:9.1
		%6 Total:83 LAS better/worse/nochange:21/25/37 25.3/30.1/44.6 Diff:-4.8
		
		\end{tikzpicture}
		\begin{tikzpicture}
		\pgfplotsset{
			width=4.5cm,
			title=\tiny b) Out-of-domain Test Set,
			xmin=0,xmax=6,
			xlabel=Number of Prepositions}
		\begin{axis}[
		axis y line*=left,
		ymin=0,ymax=100,
		ylabel=Percentage (\%)]
		%Better
		\addplot[smooth,thick,dashed,color=blue] coordinates {(0,12.1) (1,19.7) (2,25.2) (3,31.6) (4,30.9) (5,37.1) (6,37.0) };\label{better}
		\addlegendentry{\tiny Better}
		
		%Worse
		\addplot[smooth,thick,dotted,color=red] coordinates{(0,10.2) (1,15.0) (2,21.8) (3,21.2) (4,26.3) (5,26.9) (6,24.1) };\label{worse}
		\addlegendentry{\tiny Worse}
		
		%No Change
		\addplot[smooth,thick,dashdotted] coordinates{(0,77.6) (1,65.3) (2,53.0) (3,47.2) (4,42.8) (5,35.9) (6,38.9) };\label{nochange}
		\addlegendentry{\tiny No Change}
		
		\end{axis}
		\begin{axis}[
		axis y line*=right,
		ymin=0,ymax=2000,
		ylabel=Number of Sentences]
		\addlegendimage{/pgfplots/refstyle=better}\addlegendentry{\tiny Better}
		\addlegendimage{/pgfplots/refstyle=worse}\addlegendentry{\tiny Worse}
		\addlegendimage{/pgfplots/refstyle=nochange}\addlegendentry{\tiny No Change}
		%All Sent
		\addplot[smooth,thick,solid] coordinates{(0,1968.0) (1,1624.0) (2,1134.0) (3,693.0) (4,418.0) (5,245.0) (6,108.0) };\addlegendentry{\tiny No. of Sents}
		
		\end{axis}
		%0 Total:1968 LAS better/worse/nochange:239/201/1528 12.1/10.2/77.6 Diff:1.9
		%1 Total:1624 LAS better/worse/nochange:320/243/1061 19.7/15.0/65.3 Diff:4.7
		%2 Total:1134 LAS better/worse/nochange:286/247/601 25.2/21.8/53.0 Diff:3.4
		%3 Total:693 LAS better/worse/nochange:219/147/327 31.6/21.2/47.2 Diff:10.4
		%4 Total:418 LAS better/worse/nochange:129/110/179 30.9/26.3/42.8 Diff:4.6
		%5 Total:245 LAS better/worse/nochange:91/66/88 37.1/26.9/35.9 Diff:10.2
		%6 Total:108 LAS better/worse/nochange:40/26/42 37.0/24.1/38.9 Diff:12.9
		
		\end{tikzpicture}
	\end{center}
	\caption{\label{figure:analysis-sentin-dlm-en} The English comparison between the DLM approach and the baseline on different number of prepositions per sentence. }
\end{figure}

%number of conjunctions
\begin{figure}[t]
	\begin{center}
		\begin{tikzpicture}
		\pgfplotsset{
			width=4.5cm,
			title=\tiny a) In-domain Test Set,
			xtick={0,1,2},
			xmin=0,xmax=2,
			xlabel=Number of Conjunctions}
		\begin{axis}[
		axis y line*=left,
		ymin=0,ymax=100,
		ylabel=Percentage (\%)]
		%Better
		\addplot[smooth,thick,dashed,color=blue] coordinates {(0,17.0) (1,18.8) (2,22.7) (3,23.3) };\label{better}
		\addlegendentry{\tiny Better}
		
		%Worse
		\addplot[smooth,thick,dotted,color=red] coordinates{(0,12.9) (1,17.4) (2,22.7) (3,30.0) };\label{worse}
		\addlegendentry{\tiny Worse}
		
		%No Change
		\addplot[smooth,thick,dashdotted] coordinates{(0,70.1) (1,63.8) (2,54.5) (3,46.7) };\label{nochange}
		\addlegendentry{\tiny No Change}
		
		\end{axis}
		\begin{axis}[
		axis y line*=right,
		ymin=0,ymax=1500,
		ylabel=Number of Sentences]
		\addlegendimage{/pgfplots/refstyle=better}\addlegendentry{\tiny Better}
		\addlegendimage{/pgfplots/refstyle=worse}\addlegendentry{\tiny Worse}
		\addlegendimage{/pgfplots/refstyle=nochange}\addlegendentry{\tiny No Change}
		%All Sent
		\addplot[smooth,thick,solid] coordinates{(0,1423.0) (1,719.0) (2,220.0) (3,30.0) };\addlegendentry{\tiny No. of Sents}
		
		\end{axis}
		%0 Total:1423 LAS better/worse/nochange:242/184/997 17.0/12.9/70.1 Diff:4.1
		%1 Total:719 LAS better/worse/nochange:135/125/459 18.8/17.4/63.8 Diff:1.4
		%2 Total:220 LAS better/worse/nochange:50/50/120 22.7/22.7/54.5 Diff:0.0
		%3 Total:30 LAS better/worse/nochange:7/9/14 23.3/30.0/46.7 Diff:-6.7
		
		\end{tikzpicture}
		\begin{tikzpicture}
		\pgfplotsset{
			width=4.5cm,
			title=\tiny b) Out-of-domain Test Set,
			xmin=0,xmax=3,
			xlabel=Number of Conjunctions}
		\begin{axis}[
		axis y line*=left,
		ymin=0,ymax=100,
		ylabel=Percentage (\%)]
		%Better
		\addplot[smooth,thick,dashed,color=blue] coordinates {(0,17.4) (1,25.2) (2,31.2) (3,41.3) };\label{better}
		\addlegendentry{\tiny Better}
		
		%Worse
		\addplot[smooth,thick,dotted,color=red] coordinates{(0,13.5) (1,19.5) (2,28.1) (3,28.0) };\label{worse}
		\addlegendentry{\tiny Worse}
		
		%No Change
		\addplot[smooth,thick,dashdotted] coordinates{(0,69.1) (1,55.3) (2,40.7) (3,30.7) };\label{nochange}
		\addlegendentry{\tiny No Change}
		
		\end{axis}
		\begin{axis}[
		axis y line*=right,
		ymin=0,ymax=4000,
		ylabel=Number of Sentences]
		\addlegendimage{/pgfplots/refstyle=better}\addlegendentry{\tiny Better}
		\addlegendimage{/pgfplots/refstyle=worse}\addlegendentry{\tiny Worse}
		\addlegendimage{/pgfplots/refstyle=nochange}\addlegendentry{\tiny No Change}
		%All Sent
		\addplot[smooth,thick,solid] coordinates{(0,3640.0) (1,1878.0) (2,615.0) (3,150.0) };\addlegendentry{\tiny No. of Sents}
		
		\end{axis}
		%0 Total:3640 LAS better/worse/nochange:635/490/2515 17.4/13.5/69.1 Diff:3.9
		%1 Total:1878 LAS better/worse/nochange:473/366/1039 25.2/19.5/55.3 Diff:5.7
		%2 Total:615 LAS better/worse/nochange:192/173/250 31.2/28.1/40.7 Diff:3.1
		%3 Total:150 LAS better/worse/nochange:62/42/46 41.3/28.0/30.7 Diff:13.3
		
		\end{tikzpicture}
	\end{center}
	\caption{\label{figure:analysis-sentcc-dlm-en} The English comparison between the DLM approach and the baseline on different number of conjunctions per sentence. }
\end{figure}
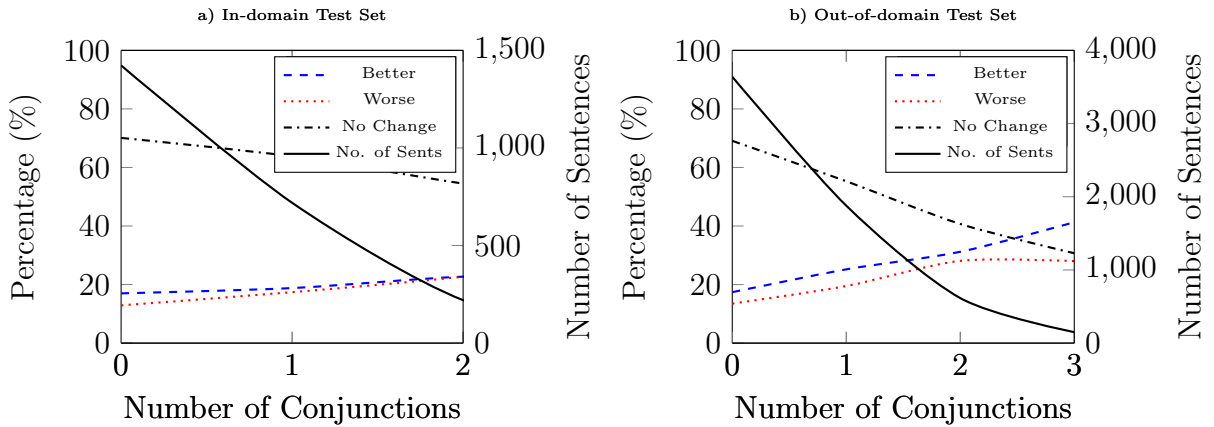

\textbf{Prepositions.} The number of prepositions analysis for in-domain data does not show a clear picture of where the improvement comes from. The rates of sentences parsed better and sentences parsed worse varies, cf. Figure \ref{figure:analysis-sentin-dlm-en}. While the analysis for out-of-domain showed a clear increased gap between sentences have better accuracies and the sentences have lowered accuracies when the number of prepositions increases. The largest gap of 10\% is achieved on sentences that have at least 5 prepositions.

\textbf{Conjunctions.} Figure \ref{figure:analysis-sentcc-dlm-en} shows our analysis of the different number of conjunctions. For in-domain test set, the DLM model gained 4\% for sentences do not have conjunctions and the number decreased when the number of conjunctions increases. For the out-of-domain test set the enhanced model gained around 4\% for sentences have up to 2 conjunctions, after that, the gap increased to 13\% for sentences have 3 conjunctions.

%examples
\begin{table}
\begin{center}
\begin{tabular}{p{\linewidth}}
\hline
\hline
\colorbox{blue!30}{\footnotesize ``}$_{20_{\textsc{p}}}^{1}$ \footnotesize It$_{3_{\textsc{sbj}}}^{2}$ \colorbox{blue!30}{\footnotesize seems}$_{20_{\textsc{obj}}}^{3}$ \footnotesize to$_{3_{\textsc{adv}}}^{4}$ \footnotesize me$_{4_{\textsc{pmod}}}^{5}$ \footnotesize that$_{3_{\textsc{prd}}}^{6 \colorbox{red!30}{\tiny p}}$ \footnotesize a$_{8_{\textsc{nmod}}}^{7}$ \colorbox{blue!30}{\footnotesize story}$_{11_{\textsc{sbj}}}^{8}$ \footnotesize like$_{8_{\textsc{nmod}}}^{9 \colorbox{red!30}{\tiny p}}$ \colorbox{blue!30}{\footnotesize this}$_{9_{\textsc{pmod}}}^{10}$ \colorbox{blue!30}{\footnotesize breaks}$_{6_{\textsc{sub}}}^{11}$ \footnotesize just$_{13_{\textsc{pmod}}}^{12}$ \colorbox{green!30}{\footnotesize before}$_{11_{\textsc{tmp}}}^{13 \colorbox{red!30}{\tiny p}}$ \footnotesize every$_{17_{\textsc{nmod}}}^{14}$ \footnotesize important$_{17_{\textsc{nmod}}}^{15}$ \footnotesize Cocom$_{17_{\textsc{nmod}}}^{16}$ \footnotesize meeting$_{13_{\textsc{pmod}}}^{17}$ \footnotesize ,$_{20_{\textsc{p}}}^{18}$ \footnotesize ''$_{20_{\textsc{p}}}^{19}$ \colorbox{blue!30}{\footnotesize said}$_{0_{\textsc{root}}}^{20}$ \footnotesize a$_{23_{\textsc{nmod}}}^{21}$ \footnotesize Washington$_{23_{\textsc{nmod}}}^{22}$ \footnotesize lobbyist$_{20_{\textsc{sbj}}}^{23}$ \footnotesize for$_{23_{\textsc{nmod}}}^{24 \colorbox{red!30}{\tiny p}}$ \footnotesize a$_{26_{\textsc{nmod}}}^{25}$ \footnotesize number$_{24_{\textsc{pmod}}}^{26}$ \footnotesize of$_{26_{\textsc{nmod}}}^{27 \colorbox{red!30}{\tiny p}}$ \footnotesize U.S.$_{30_{\textsc{nmod}}}^{28}$ \footnotesize computer$_{30_{\textsc{nmod}}}^{29}$ \footnotesize companies$_{27_{\textsc{pmod}}}^{30}$ \colorbox{blue!30}{\footnotesize .}$_{20_{\textsc{p}}}^{31}$\\\hline

\footnotesize The$_{2_{\textsc{nmod}}}^{1}$ \colorbox{blue!30}{\footnotesize games}$_{17_{\textsc{sbj}}}^{2}$ \colorbox{blue!30}{\footnotesize Bronx}$_{4_{\textsc{nmod}}}^{3}$ \footnotesize children$_{5_{\textsc{sbj}}}^{4}$ \colorbox{blue!30}{\footnotesize played}$_{2_{\textsc{nmod}}}^{5}$ \colorbox{blue!30}{\footnotesize (}$_{7_{\textsc{p}}}^{6}$ \footnotesize holding$_{2_{\textsc{prn}}}^{7}$ \footnotesize kids$_{7_{\textsc{obj}}}^{8}$ \footnotesize down$_{7_{\textsc{prt}}}^{9}$ \footnotesize and$_{7_{\textsc{coord}}}^{10 \colorbox{red!30}{\tiny c}}$ \footnotesize stripping$_{10_{\textsc{conj}}}^{11}$ \footnotesize them$_{11_{\textsc{obj}}}^{12}$ \colorbox{blue!30}{\footnotesize ,}$_{7_{\textsc{p}}}^{13}$ \colorbox{yellow!30}{\footnotesize for}$_{7_{\textsc{dep}}}^{14 \colorbox{red!30}{\tiny p}}$ \footnotesize example$_{14_{\textsc{pmod}}}^{15}$ \colorbox{blue!30}{\footnotesize )}$_{7_{\textsc{p}}}^{16}$ \colorbox{blue!30}{\footnotesize seem}$_{0_{\textsc{root}}}^{17}$ \footnotesize tame$_{17_{\textsc{prd}}}^{18}$ \footnotesize by$_{17_{\textsc{adv}}}^{19 \colorbox{red!30}{\tiny p}}$ \footnotesize today$_{23_{\textsc{nmod}}}^{20}$ \footnotesize 's$_{20_{\textsc{suffix}}}^{21}$ \footnotesize crack$_{23_{\textsc{nmod}}}^{22}$ \footnotesize standards$_{19_{\textsc{pmod}}}^{23}$ \footnotesize ,$_{17_{\textsc{p}}}^{24}$ \footnotesize but$_{17_{\textsc{coord}}}^{25 \colorbox{red!30}{\tiny c}}$ \footnotesize Ms.$_{27_{\textsc{title}}}^{26}$ \footnotesize Cunningham$_{28_{\textsc{sbj}}}^{27}$ \footnotesize makes$_{25_{\textsc{conj}}}^{28}$ \footnotesize it$_{28_{\textsc{obj}}}^{29}$ \footnotesize all$_{31_{\textsc{dep}}}^{30}$ \footnotesize sound$_{28_{\textsc{oprd}}}^{31}$ \footnotesize like$_{31_{\textsc{prd}}}^{32 \colorbox{red!30}{\tiny p}}$ \footnotesize a$_{35_{\textsc{nmod}}}^{33}$ \footnotesize great$_{35_{\textsc{nmod}}}^{34}$ \footnotesize adventure$_{32_{\textsc{pmod}}}^{35}$ \colorbox{blue!30}{\footnotesize .}$_{17_{\textsc{p}}}^{36}$\\\hline

\footnotesize This$_{3_{\textsc{nmod}}}^{1}$ \footnotesize role$_{3_{\textsc{nmod}}}^{2}$ \footnotesize reversal$_{4_{\textsc{sbj}}}^{3}$ \colorbox{blue!30}{\footnotesize holds}$_{0_{\textsc{root}}}^{4}$ \footnotesize true$_{4_{\textsc{oprd}}}^{5}$ \footnotesize ,$_{4_{\textsc{p}}}^{6}$ \footnotesize as$_{8_{\textsc{amod}}}^{7 \colorbox{red!30}{\tiny p}}$ \footnotesize well$_{4_{\textsc{adv}}}^{8}$ \footnotesize ,$_{4_{\textsc{p}}}^{9}$ \footnotesize for$_{4_{\textsc{adv}}}^{10 \colorbox{red!30}{\tiny p}}$ \colorbox{blue!30}{\footnotesize his}$_{16_{\textsc{nmod}}}^{11}$ \colorbox{blue!30}{\footnotesize three}$_{16_{\textsc{nmod}}}^{12}$ \colorbox{blue!30}{\footnotesize liberal}$_{16_{\textsc{nmod}}}^{13}$ \footnotesize and$_{13_{\textsc{coord}}}^{14 \colorbox{red!30}{\tiny c}}$ \colorbox{blue!30}{\footnotesize moderate}$_{14_{\textsc{conj}}}^{15}$ \colorbox{blue!30}{\footnotesize allies}$_{10_{\textsc{pmod}}}^{16}$ \colorbox{blue!30}{\footnotesize ,}$_{16_{\textsc{p}}}^{17}$ \colorbox{yellow!30}{\footnotesize Justices}$_{20_{\textsc{title}}}^{18}$ \footnotesize Thurgood$_{20_{\textsc{name}}}^{19 \colorbox{red!30}{\tiny u}}$ \colorbox{blue!30}{\footnotesize Marshall}$_{16_{\textsc{appo}}}^{20}$ \footnotesize ,$_{20_{\textsc{p}}}^{21}$ \footnotesize Harry$_{23_{\textsc{name}}}^{22}$ \footnotesize Blackmun$_{20_{\textsc{coord}}}^{23 \colorbox{red!30}{\tiny u}}$ \footnotesize and$_{23_{\textsc{coord}}}^{24 \colorbox{red!30}{\tiny c}}$ \footnotesize John$_{26_{\textsc{name}}}^{25}$ \footnotesize Stevens$_{24_{\textsc{conj}}}^{26}$ \colorbox{blue!30}{\footnotesize .}$_{4_{\textsc{p}}}^{27}$\\\hline

\colorbox{blue!30}{\footnotesize Harvard}$_{6_{\textsc{name}}}^{1}$ \colorbox{blue!30}{\footnotesize Law}$_{6_{\textsc{name}}}^{2}$ \colorbox{blue!30}{\footnotesize School}$_{6_{\textsc{name}}}^{3}$ \colorbox{blue!30}{\footnotesize Professor}$_{6_{\textsc{name}}}^{4}$ \footnotesize Laurence$_{6_{\textsc{name}}}^{5}$ \colorbox{blue!30}{\footnotesize Tribe}$_{7_{\textsc{sbj}}}^{6}$ \footnotesize says$_{0_{\textsc{root}}}^{7}$ \footnotesize there$_{9_{\textsc{sbj}}}^{8}$ \footnotesize is$_{7_{\textsc{obj}}}^{9}$ \footnotesize a$_{16_{\textsc{nmod}}}^{10}$ \colorbox{blue!30}{\footnotesize ``}$_{14_{\textsc{p}}}^{11}$ \footnotesize generation$_{14_{\textsc{hmod}}}^{12}$ \footnotesize -$_{12_{\textsc{hyph}}}^{13}$ \colorbox{blue!30}{\footnotesize skipping}$_{16_{\textsc{nmod}}}^{14}$ \colorbox{blue!30}{\footnotesize ''}$_{14_{\textsc{p}}}^{15}$ \footnotesize flavor$_{9_{\textsc{prd}}}^{16}$ \footnotesize to$_{9_{\textsc{adv}}}^{17}$ \footnotesize current$_{19_{\textsc{nmod}}}^{18}$ \footnotesize dissents$_{17_{\textsc{pmod}}}^{19 \colorbox{red!30}{\tiny u}}$ \footnotesize .$_{7_{\textsc{p}}}^{20}$\\\hline

\colorbox{blue!30}{\footnotesize While}$_{12_{\textsc{adv}}}^{1 \colorbox{red!30}{\tiny p}}$ \footnotesize there$_{3_{\textsc{sbj}}}^{2}$ \footnotesize are$_{1_{\textsc{sub}}}^{3}$ \colorbox{blue!30}{\footnotesize some}$_{9_{\textsc{nmod}}}^{4}$ \colorbox{blue!30}{\footnotesize popular}$_{9_{\textsc{nmod}}}^{5}$ \colorbox{blue!30}{\footnotesize action}$_{9_{\textsc{nmod}}}^{6}$ \footnotesize and$_{6_{\textsc{coord}}}^{7 \colorbox{red!30}{\tiny c}}$ \colorbox{blue!30}{\footnotesize drama}$_{7_{\textsc{conj}}}^{8}$ \colorbox{blue!30}{\footnotesize series}$_{3_{\textsc{prd}}}^{9}$ \colorbox{blue!30}{\footnotesize ,}$_{12_{\textsc{p}}}^{10}$ \colorbox{green!30}{\footnotesize few}$_{12_{\textsc{sbj}}}^{11}$ \colorbox{blue!30}{\footnotesize boast}$_{0_{\textsc{root}}}^{12}$ \footnotesize the$_{15_{\textsc{nmod}}}^{13}$ \footnotesize high$_{15_{\textsc{nmod}}}^{14}$ \colorbox{blue!30}{\footnotesize culture}$_{12_{\textsc{obj}}}^{15}$ \footnotesize and$_{15_{\textsc{coord}}}^{16 \colorbox{red!30}{\tiny c}}$ \colorbox{blue!30}{\footnotesize classy}$_{19_{\textsc{nmod}}}^{17}$ \colorbox{blue!30}{\footnotesize production}$_{19_{\textsc{nmod}}}^{18}$ \colorbox{blue!30}{\footnotesize values}$_{16_{\textsc{conj}}}^{19}$ \footnotesize one$_{21_{\textsc{sbj}}}^{20}$ \colorbox{blue!30}{\footnotesize might}$_{15_{\textsc{nmod}}}^{21}$ \footnotesize expect$_{21_{\textsc{vc}}}^{22}$ \colorbox{blue!30}{\footnotesize .}$_{12_{\textsc{p}}}^{23}$\\\hline

\footnotesize The$_{2_{\textsc{nmod}}}^{1}$ \footnotesize question$_{3_{\textsc{sbj}}}^{2}$ \footnotesize is$_{0_{\textsc{root}}}^{3}$ \footnotesize ,$_{3_{\textsc{p}}}^{4}$ \colorbox{blue!30}{\footnotesize if}$_{33_{\textsc{adv}}}^{5 \colorbox{red!30}{\tiny p}}$ \footnotesize group$_{7_{\textsc{nmod}}}^{6}$ \footnotesize conflicts$_{9_{\textsc{sbj}}}^{7}$ \footnotesize still$_{9_{\textsc{tmp}}}^{8}$ \footnotesize exist$_{5_{\textsc{sub}}}^{9}$ \colorbox{blue!30}{\footnotesize (}$_{11_{\textsc{p}}}^{10}$ \colorbox{green!30}{\footnotesize as}$_{9_{\textsc{prn}}}^{11 \colorbox{red!30}{\tiny p}}$ \colorbox{blue!30}{\footnotesize undeniably}$_{14_{\textsc{adv}}}^{12}$ \footnotesize they$_{14_{\textsc{sbj}}}^{13}$ \footnotesize do$_{11_{\textsc{sub}}}^{14}$ \colorbox{blue!30}{\footnotesize )}$_{11_{\textsc{p}}}^{15}$ \colorbox{blue!30}{\footnotesize ,}$_{5_{\textsc{p}}}^{16}$ \colorbox{blue!30}{\footnotesize and}$_{5_{\textsc{coord}}}^{17 \colorbox{red!30}{\tiny c}}$ \colorbox{blue!30}{\footnotesize if}$_{17_{\textsc{conj}}}^{18 \colorbox{red!30}{\tiny p}}$ \footnotesize Mr.$_{20_{\textsc{title}}}^{19}$ \footnotesize Mason$_{22_{\textsc{nmod}}}^{20}$ \footnotesize 's$_{20_{\textsc{suffix}}}^{21}$ \footnotesize type$_{26_{\textsc{sbj}}}^{22}$ \footnotesize of$_{22_{\textsc{nmod}}}^{23 \colorbox{red!30}{\tiny p}}$ \footnotesize ethnic$_{25_{\textsc{nmod}}}^{24}$ \footnotesize humor$_{23_{\textsc{pmod}}}^{25}$ \footnotesize is$_{18_{\textsc{sub}}}^{26}$ \footnotesize passe$_{26_{\textsc{prd}}}^{27 \colorbox{red!30}{\tiny u}}$ \footnotesize ,$_{33_{\textsc{p}}}^{28}$ \footnotesize then$_{33_{\textsc{adv}}}^{29}$ \footnotesize what$_{32_{\textsc{nmod}}}^{30}$ \footnotesize other$_{32_{\textsc{nmod}}}^{31}$ \footnotesize means$_{35_{\textsc{obj}}}^{32}$ \colorbox{blue!30}{\footnotesize do}$_{3_{\textsc{prd}}}^{33}$ \footnotesize we$_{33_{\textsc{sbj}}}^{34}$ \footnotesize have$_{33_{\textsc{vc}}}^{35}$ \footnotesize for$_{32_{\textsc{nmod}}}^{36 \colorbox{red!30}{\tiny p}}$ \footnotesize letting$_{36_{\textsc{pmod}}}^{37}$ \footnotesize off$_{37_{\textsc{prt}}}^{38}$ \footnotesize steam$_{37_{\textsc{obj}}}^{39}$ \colorbox{blue!30}{\footnotesize ?}$_{33_{\textsc{p}}}^{40}$\\\hline

\footnotesize The$_{3_{\textsc{nmod}}}^{1}$ \footnotesize main$_{3_{\textsc{nmod}}}^{2}$ \footnotesize thing$_{4_{\textsc{sbj}}}^{3}$ \colorbox{blue!30}{\footnotesize was}$_{0_{\textsc{root}}}^{4}$ \footnotesize portfolio$_{6_{\textsc{nmod}}}^{5}$ \footnotesize insurance$_{4_{\textsc{prd}}}^{6}$ \colorbox{blue!30}{\footnotesize ,}$_{6_{\textsc{p}}}^{7}$ \colorbox{blue!30}{\footnotesize ''}$_{6_{\textsc{p}}}^{8}$ \footnotesize a$_{12_{\textsc{nmod}}}^{9}$ \footnotesize mechanical$_{12_{\textsc{nmod}}}^{10}$ \footnotesize trading$_{12_{\textsc{nmod}}}^{11}$ \colorbox{blue!30}{\footnotesize system}$_{6_{\textsc{appo}}}^{12}$ \colorbox{blue!30}{\footnotesize intended}$_{12_{\textsc{appo}}}^{13}$ \footnotesize to$_{13_{\textsc{oprd}}}^{14}$ \footnotesize protect$_{14_{\textsc{im}}}^{15}$ \footnotesize an$_{17_{\textsc{nmod}}}^{16}$ \footnotesize investor$_{15_{\textsc{obj}}}^{17}$ \footnotesize against$_{15_{\textsc{adv}}}^{18 \colorbox{red!30}{\tiny p}}$ \footnotesize losses$_{18_{\textsc{pmod}}}^{19}$ \colorbox{blue!30}{\footnotesize .}$_{4_{\textsc{p}}}^{20}$ \colorbox{blue!30}{\footnotesize ``}$_{4_{\textsc{p}}}^{21}$\\\hline

\footnotesize At$_{3_{\textsc{prd}}}^{1 \colorbox{red!30}{\tiny p}}$ \footnotesize stake$_{1_{\textsc{pmod}}}^{2}$ \footnotesize is$_{0_{\textsc{root}}}^{3}$ \footnotesize what$_{16_{\textsc{obj}}}^{4}$ \footnotesize Mike$_{6_{\textsc{name}}}^{5}$ \footnotesize Swavely$_{16_{\textsc{sbj}}}^{6}$ \footnotesize ,$_{6_{\textsc{p}}}^{7}$ \footnotesize Compaq$_{10_{\textsc{nmod}}}^{8}$ \footnotesize 's$_{8_{\textsc{suffix}}}^{9}$ \footnotesize president$_{6_{\textsc{appo}}}^{10}$ \footnotesize of$_{10_{\textsc{nmod}}}^{11 \colorbox{red!30}{\tiny p}}$ \footnotesize North$_{13_{\textsc{name}}}^{12}$ \footnotesize America$_{14_{\textsc{nmod}}}^{13}$ \footnotesize operations$_{11_{\textsc{pmod}}}^{14}$ \footnotesize ,$_{6_{\textsc{p}}}^{15}$ \footnotesize calls$_{28_{\textsc{nmod}}}^{16}$ \colorbox{blue!30}{\footnotesize ``}$_{20_{\textsc{p}}}^{17}$ \colorbox{blue!30}{\footnotesize the}$_{20_{\textsc{nmod}}}^{18}$ \colorbox{blue!30}{\footnotesize Holy}$_{20_{\textsc{name}}}^{19}$ \colorbox{blue!30}{\footnotesize Grail}$_{16_{\textsc{oprd}}}^{20 \colorbox{red!30}{\tiny u}}$ \colorbox{blue!30}{\footnotesize of}$_{20_{\textsc{nmod}}}^{21 \colorbox{red!30}{\tiny p}}$ \footnotesize the$_{24_{\textsc{nmod}}}^{22}$ \footnotesize computer$_{24_{\textsc{nmod}}}^{23}$ \colorbox{blue!30}{\footnotesize industry}$_{21_{\textsc{pmod}}}^{24}$ \colorbox{blue!30}{\footnotesize ''}$_{20_{\textsc{p}}}^{25}$ \colorbox{blue!30}{\footnotesize --}$_{28_{\textsc{p}}}^{26}$ \footnotesize the$_{28_{\textsc{nmod}}}^{27}$ \footnotesize search$_{3_{\textsc{sbj}}}^{28}$ \footnotesize for$_{28_{\textsc{nmod}}}^{29 \colorbox{red!30}{\tiny p}}$ \footnotesize ``$_{29_{\textsc{p}}}^{30}$ \footnotesize a$_{33_{\textsc{nmod}}}^{31}$ \footnotesize real$_{33_{\textsc{nmod}}}^{32}$ \footnotesize computer$_{29_{\textsc{pmod}}}^{33}$ \footnotesize in$_{33_{\textsc{loc}}}^{34 \colorbox{red!30}{\tiny p}}$ \footnotesize a$_{36_{\textsc{nmod}}}^{35}$ \footnotesize package$_{34_{\textsc{pmod}}}^{36}$ \footnotesize so$_{38_{\textsc{amod}}}^{37}$ \colorbox{blue!30}{\footnotesize small}$_{36_{\textsc{appo}}}^{38}$ \footnotesize you$_{40_{\textsc{sbj}}}^{39}$ \footnotesize can$_{38_{\textsc{amod}}}^{40}$ \footnotesize take$_{40_{\textsc{vc}}}^{41}$ \footnotesize it$_{41_{\textsc{obj}}}^{42}$ \footnotesize everywhere$_{41_{\textsc{loc}}}^{43}$ \footnotesize .$_{3_{\textsc{p}}}^{44}$ \footnotesize ''$_{3_{\textsc{p}}}^{45}$\\\hline

\hline
\end{tabular}
\end{center}
\caption[The example sentences that have been improved by the DLM approach when compared to the baseline on in-domain test set.]{\label{table:analysis-examples-dlm-in} The example sentences that have been improved by the DLM approach when compared to the baseline  on in-domain test set. In which the dependency head/relation of a token are marked as the subscript, while the superscript is the index of token. The unknown words, prepositions and conjunctions are highlighted with \colorbox{red!30}{u}, \colorbox{red!30}{p} and \colorbox{red!30}{c} respectively. We highlight the different levels of the improvements achieved by our dlm model on the dependency edges by different colours. In which the \colorbox{blue!30}{blue} colour means both head and label are corrected, the \colorbox{yellow!30}{yellow} colour means only the head is corrected and the \colorbox{green!30}{green} colour means only the label is corrected.}
\end{table}

%examples
\begin{table}
\begin{center}
\begin{tabular}{p{\linewidth}}
\hline
\hline
\colorbox{green!30}{\footnotesize 2.}$_{12_{\textsc{dep}}}^{1 \colorbox{red!30}{\tiny u}}$ \colorbox{blue!30}{\footnotesize To}$_{12_{\textsc{prp}}}^{2}$ \footnotesize make$_{2_{\textsc{im}}}^{3}$ \footnotesize the$_{5_{\textsc{nmod}}}^{4}$ \footnotesize body$_{3_{\textsc{obj}}}^{5}$ \footnotesize fully$_{7_{\textsc{mnr}}}^{6}$ \colorbox{blue!30}{\footnotesize absorb}$_{3_{\textsc{oprd}}}^{7}$ \colorbox{blue!30}{\footnotesize proteins}$_{7_{\textsc{obj}}}^{8}$ \colorbox{yellow!30}{\footnotesize better}$_{7_{\textsc{mnr}}}^{9}$ \footnotesize ,$_{12_{\textsc{p}}}^{10}$ \footnotesize we$_{12_{\textsc{sbj}}}^{11}$ \footnotesize should$_{0_{\textsc{root}}}^{12}$ \footnotesize eat$_{12_{\textsc{vc}}}^{13}$ \footnotesize plenty$_{13_{\textsc{obj}}}^{14}$ \footnotesize of$_{14_{\textsc{nmod}}}^{15 \colorbox{red!30}{\tiny p}}$ \colorbox{blue!30}{\footnotesize food}$_{15_{\textsc{pmod}}}^{16}$ \colorbox{blue!30}{\footnotesize containing}$_{16_{\textsc{appo}}}^{17}$ \colorbox{yellow!30}{\footnotesize vitamin}$_{19_{\textsc{nmod}}}^{18 \colorbox{red!30}{\tiny u}}$ \colorbox{blue!30}{\footnotesize B1}$_{17_{\textsc{obj}}}^{19 \colorbox{red!30}{\tiny u}}$ \colorbox{blue!30}{\footnotesize and}$_{19_{\textsc{coord}}}^{20 \colorbox{red!30}{\tiny c}}$ \footnotesize vitamin$_{22_{\textsc{nmod}}}^{21 \colorbox{red!30}{\tiny u}}$ \footnotesize C$_{20_{\textsc{conj}}}^{22}$ \footnotesize .$_{12_{\textsc{p}}}^{23}$\\\hline

\footnotesize But$_{3_{\textsc{dep}}}^{1 \colorbox{red!30}{\tiny c}}$ \footnotesize everyone$_{3_{\textsc{sbj}}}^{2}$ \footnotesize needs$_{0_{\textsc{root}}}^{3}$ \footnotesize to$_{3_{\textsc{oprd}}}^{4}$ \footnotesize recognize$_{4_{\textsc{im}}}^{5}$ \footnotesize that$_{5_{\textsc{obj}}}^{6 \colorbox{red!30}{\tiny p}}$ \footnotesize Arlington$_{9_{\textsc{nmod}}}^{7}$ \footnotesize 's$_{7_{\textsc{suffix}}}^{8}$ \colorbox{blue!30}{\footnotesize decision}$_{16_{\textsc{sbj}}}^{9}$ \footnotesize not$_{11_{\textsc{adv}}}^{10}$ \footnotesize to$_{9_{\textsc{nmod}}}^{11}$ \footnotesize pursue$_{11_{\textsc{im}}}^{12}$ \colorbox{blue!30}{\footnotesize a}$_{15_{\textsc{nmod}}}^{13}$ \colorbox{blue!30}{\footnotesize balanced}$_{15_{\textsc{nmod}}}^{14}$ \colorbox{blue!30}{\footnotesize community}$_{12_{\textsc{obj}}}^{15}$ \colorbox{blue!30}{\footnotesize means}$_{6_{\textsc{sub}}}^{16}$ \colorbox{blue!30}{\footnotesize that}$_{16_{\textsc{obj}}}^{17 \colorbox{red!30}{\tiny p}}$ \footnotesize housing$_{19_{\textsc{sbj}}}^{18}$ \colorbox{blue!30}{\footnotesize will}$_{17_{\textsc{sub}}}^{19}$ \footnotesize end$_{19_{\textsc{vc}}}^{20}$ \footnotesize up$_{20_{\textsc{prt}}}^{21}$ \footnotesize somewhere$_{20_{\textsc{loc}}}^{22}$ \footnotesize else$_{22_{\textsc{amod}}}^{23}$ \footnotesize ,$_{22_{\textsc{p}}}^{24}$ \colorbox{green!30}{\footnotesize presumably}$_{26_{\textsc{pmod}}}^{25}$ \footnotesize in$_{22_{\textsc{amod}}}^{26 \colorbox{red!30}{\tiny p}}$ \colorbox{blue!30}{\footnotesize outlying}$_{28_{\textsc{nmod}}}^{27}$ \colorbox{blue!30}{\footnotesize counties}$_{26_{\textsc{pmod}}}^{28}$ \footnotesize .$_{3_{\textsc{p}}}^{29}$\\\hline

\footnotesize In$_{5_{\textsc{adv}}}^{1 \colorbox{red!30}{\tiny p}}$ \footnotesize some$_{3_{\textsc{nmod}}}^{2}$ \footnotesize respects$_{1_{\textsc{pmod}}}^{3}$ \footnotesize ,$_{5_{\textsc{p}}}^{4}$ \footnotesize is$_{0_{\textsc{root}}}^{5}$ \footnotesize n't$_{5_{\textsc{adv}}}^{6}$ \footnotesize that$_{5_{\textsc{sbj}}}^{7}$ \footnotesize essentially$_{5_{\textsc{adv}}}^{8}$ \footnotesize what$_{14_{\textsc{obj}}}^{9}$ \colorbox{blue!30}{\footnotesize No}$_{11_{\textsc{name}}}^{10}$ \footnotesize Va$_{12_{\textsc{nmod}}}^{11}$ \footnotesize jursidictions$_{13_{\textsc{sbj}}}^{12 \colorbox{red!30}{\tiny u}}$ \footnotesize are$_{5_{\textsc{prd}}}^{13}$ \colorbox{blue!30}{\footnotesize doing}$_{13_{\textsc{vc}}}^{14}$ \footnotesize -$_{14_{\textsc{p}}}^{15}$ \colorbox{blue!30}{\footnotesize favoring}$_{14_{\textsc{adv}}}^{16}$ \footnotesize non-residential$_{18_{\textsc{nmod}}}^{17}$ \colorbox{blue!30}{\footnotesize development}$_{16_{\textsc{obj}}}^{18}$ \colorbox{blue!30}{\footnotesize and}$_{16_{\textsc{coord}}}^{19 \colorbox{red!30}{\tiny c}}$ \footnotesize letting$_{19_{\textsc{conj}}}^{20}$ \footnotesize other$_{22_{\textsc{nmod}}}^{21}$ \colorbox{blue!30}{\footnotesize jursidictions}$_{20_{\textsc{obj}}}^{22 \colorbox{red!30}{\tiny u}}$ \colorbox{green!30}{\footnotesize handle}$_{20_{\textsc{oprd}}}^{23}$ \footnotesize the$_{25_{\textsc{nmod}}}^{24}$ \footnotesize residential$_{23_{\textsc{obj}}}^{25}$ \footnotesize ?$_{5_{\textsc{p}}}^{26}$\\\hline

\colorbox{blue!30}{\footnotesize Her}$_{4_{\textsc{nmod}}}^{1}$ \footnotesize ``$_{4_{\textsc{p}}}^{2}$ \footnotesize Rubble$_{4_{\textsc{name}}}^{3}$ \colorbox{green!30}{\footnotesize Division}$_{6_{\textsc{sbj}}}^{4}$ \footnotesize ''$_{4_{\textsc{p}}}^{5}$ \colorbox{blue!30}{\footnotesize mixes}$_{0_{\textsc{root}}}^{6}$ \footnotesize such$_{9_{\textsc{nmod}}}^{7}$ \footnotesize disparate$_{9_{\textsc{nmod}}}^{8}$ \colorbox{blue!30}{\footnotesize materials}$_{6_{\textsc{obj}}}^{9}$ \footnotesize as$_{9_{\textsc{nmod}}}^{10 \colorbox{red!30}{\tiny p}}$ \footnotesize ink$_{13_{\textsc{nmod}}}^{11}$ \footnotesize -$_{13_{\textsc{nmod}}}^{12}$ \footnotesize jet$_{14_{\textsc{nmod}}}^{13}$ \colorbox{blue!30}{\footnotesize prints}$_{10_{\textsc{pmod}}}^{14}$ \colorbox{blue!30}{\footnotesize pasted}$_{14_{\textsc{appo}}}^{15 \colorbox{red!30}{\tiny u}}$ \colorbox{yellow!30}{\footnotesize on}$_{15_{\textsc{loc}}}^{16 \colorbox{red!30}{\tiny p}}$ \footnotesize board$_{16_{\textsc{pmod}}}^{17}$ \footnotesize ,$_{14_{\textsc{p}}}^{18}$ \footnotesize foam$_{20_{\textsc{nmod}}}^{19}$ \footnotesize rubber$_{14_{\textsc{coord}}}^{20}$ \footnotesize ,$_{20_{\textsc{p}}}^{21}$ \footnotesize galvanized$_{23_{\textsc{nmod}}}^{22}$ \footnotesize steel$_{20_{\textsc{coord}}}^{23}$ \footnotesize ,$_{23_{\textsc{p}}}^{24}$ \footnotesize concrete$_{23_{\textsc{coord}}}^{25}$ \footnotesize ,$_{25_{\textsc{p}}}^{26}$ \footnotesize steel$_{28_{\textsc{nmod}}}^{27}$ \footnotesize rebar$_{25_{\textsc{coord}}}^{28 \colorbox{red!30}{\tiny u}}$ \footnotesize and$_{28_{\textsc{coord}}}^{29 \colorbox{red!30}{\tiny c}}$ \footnotesize bungee$_{31_{\textsc{nmod}}}^{30 \colorbox{red!30}{\tiny u}}$ \footnotesize cords$_{29_{\textsc{conj}}}^{31 \colorbox{red!30}{\tiny u}}$ \colorbox{blue!30}{\footnotesize .}$_{6_{\textsc{p}}}^{32}$\\\hline

\footnotesize they$_{2_{\textsc{sbj}}}^{1}$ \footnotesize were$_{0_{\textsc{root}}}^{2}$ \colorbox{green!30}{\footnotesize convinced}$_{2_{\textsc{prd}}}^{3}$ \colorbox{green!30}{\footnotesize that}$_{3_{\textsc{amod}}}^{4 \colorbox{red!30}{\tiny p}}$ \colorbox{blue!30}{\footnotesize if}$_{20_{\textsc{adv}}}^{5 \colorbox{red!30}{\tiny p}}$ \footnotesize only$_{8_{\textsc{adv}}}^{6}$ \footnotesize they$_{8_{\textsc{sbj}}}^{7}$ \colorbox{blue!30}{\footnotesize could}$_{5_{\textsc{sub}}}^{8}$ \footnotesize speak$_{8_{\textsc{vc}}}^{9}$ \footnotesize to$_{9_{\textsc{adv}}}^{10 \colorbox{red!30}{\tiny p}}$ \footnotesize an$_{12_{\textsc{nmod}}}^{11}$ \footnotesize American$_{10_{\textsc{pmod}}}^{12}$ \colorbox{blue!30}{\footnotesize ,}$_{20_{\textsc{p}}}^{13}$ \footnotesize Abather$_{19_{\textsc{nmod}}}^{14 \colorbox{red!30}{\tiny u}}$ \footnotesize 's$_{14_{\textsc{suffix}}}^{15}$ \footnotesize charred$_{19_{\textsc{nmod}}}^{16}$ \footnotesize and$_{16_{\textsc{coord}}}^{17 \colorbox{red!30}{\tiny c}}$ \footnotesize mangled$_{17_{\textsc{conj}}}^{18 \colorbox{red!30}{\tiny u}}$ \footnotesize flesh$_{20_{\textsc{sbj}}}^{19}$ \colorbox{blue!30}{\footnotesize would}$_{4_{\textsc{sub}}}^{20}$ \footnotesize make$_{20_{\textsc{vc}}}^{21}$ \footnotesize their$_{23_{\textsc{nmod}}}^{22}$ \footnotesize case$_{21_{\textsc{obj}}}^{23}$ \colorbox{blue!30}{\footnotesize ,}$_{2_{\textsc{p}}}^{24}$ \colorbox{blue!30}{\footnotesize but}$_{2_{\textsc{coord}}}^{25 \colorbox{red!30}{\tiny c}}$ \footnotesize they$_{27_{\textsc{sbj}}}^{26}$ \footnotesize had$_{25_{\textsc{conj}}}^{27}$ \footnotesize never$_{27_{\textsc{tmp}}}^{28}$ \footnotesize gotten$_{27_{\textsc{vc}}}^{29}$ \footnotesize past$_{29_{\textsc{adv}}}^{30 \colorbox{red!30}{\tiny p}}$ \footnotesize the$_{34_{\textsc{nmod}}}^{31}$ \footnotesize Jordanian$_{34_{\textsc{nmod}}}^{32 \colorbox{red!30}{\tiny u}}$ \footnotesize security$_{34_{\textsc{nmod}}}^{33}$ \footnotesize guards$_{30_{\textsc{pmod}}}^{34}$ \footnotesize .$_{2_{\textsc{p}}}^{35}$\\\hline

\colorbox{yellow!30}{\footnotesize -}$_{3_{\textsc{p}}}^{1}$ \footnotesize Dr.$_{3_{\textsc{title}}}^{2}$ \colorbox{blue!30}{\footnotesize Seuss}$_{0_{\textsc{root}}}^{3 \colorbox{red!30}{\tiny u}}$ \colorbox{blue!30}{\footnotesize ,}$_{3_{\textsc{p}}}^{4}$ \footnotesize ``$_{3_{\textsc{p}}}^{5}$ \footnotesize One$_{7_{\textsc{nmod}}}^{6}$ \colorbox{yellow!30}{\footnotesize Fish}$_{3_{\textsc{coord}}}^{7}$ \footnotesize ,$_{7_{\textsc{p}}}^{8}$ \colorbox{green!30}{\footnotesize Two}$_{10_{\textsc{nmod}}}^{9}$ \colorbox{green!30}{\footnotesize Fish}$_{7_{\textsc{coord}}}^{10}$ \footnotesize ,$_{10_{\textsc{p}}}^{11}$ \footnotesize Red$_{13_{\textsc{nmod}}}^{12}$ \colorbox{blue!30}{\footnotesize Fish}$_{10_{\textsc{coord}}}^{13}$ \colorbox{blue!30}{\footnotesize ,}$_{13_{\textsc{p}}}^{14}$ \footnotesize Blue$_{16_{\textsc{nmod}}}^{15}$ \colorbox{blue!30}{\footnotesize Fish}$_{13_{\textsc{coord}}}^{16}$ \footnotesize ''$_{3_{\textsc{p}}}^{17}$\\\hline

\footnotesize But$_{2_{\textsc{dep}}}^{1 \colorbox{red!30}{\tiny c}}$ \footnotesize let$_{0_{\textsc{root}}}^{2}$ \footnotesize s$_{2_{\textsc{obj}}}^{3}$ \footnotesize hope$_{2_{\textsc{oprd}}}^{4}$ \footnotesize for$_{4_{\textsc{prp}}}^{5 \colorbox{red!30}{\tiny p}}$ \footnotesize their$_{7_{\textsc{nmod}}}^{6}$ \footnotesize sake$_{5_{\textsc{pmod}}}^{7}$ \colorbox{blue!30}{\footnotesize (}$_{7_{\textsc{p}}}^{8}$ \colorbox{blue!30}{\footnotesize and}$_{7_{\textsc{coord}}}^{9 \colorbox{red!30}{\tiny c}}$ \footnotesize the$_{11_{\textsc{nmod}}}^{10}$ \colorbox{blue!30}{\footnotesize sake}$_{9_{\textsc{conj}}}^{11}$ \footnotesize of$_{11_{\textsc{nmod}}}^{12 \colorbox{red!30}{\tiny p}}$ \colorbox{blue!30}{\footnotesize all}$_{15_{\textsc{nmod}}}^{13}$ \colorbox{blue!30}{\footnotesize space}$_{15_{\textsc{nmod}}}^{14}$ \colorbox{blue!30}{\footnotesize lovers}$_{12_{\textsc{pmod}}}^{15}$ \colorbox{green!30}{\footnotesize out}$_{15_{\textsc{loc}}}^{16}$ \colorbox{green!30}{\footnotesize there}$_{16_{\textsc{amod}}}^{17}$ \colorbox{blue!30}{\footnotesize )}$_{7_{\textsc{p}}}^{18}$ \colorbox{blue!30}{\footnotesize that}$_{4_{\textsc{obj}}}^{19 \colorbox{red!30}{\tiny p}}$ \footnotesize they$_{21_{\textsc{sbj}}}^{20}$ \footnotesize can$_{19_{\textsc{sub}}}^{21}$ \footnotesize redefine$_{21_{\textsc{vc}}}^{22}$ \footnotesize their$_{24_{\textsc{nmod}}}^{23}$ \footnotesize image$_{22_{\textsc{obj}}}^{24}$ \footnotesize and$_{22_{\textsc{coord}}}^{25 \colorbox{red!30}{\tiny c}}$ \footnotesize rekindle$_{25_{\textsc{conj}}}^{26}$ \footnotesize the$_{28_{\textsc{nmod}}}^{27}$ \footnotesize hope$_{26_{\textsc{obj}}}^{28}$ \footnotesize of$_{28_{\textsc{nmod}}}^{29 \colorbox{red!30}{\tiny p}}$ \footnotesize space$_{31_{\textsc{nmod}}}^{30}$ \footnotesize colonization$_{29_{\textsc{pmod}}}^{31 \colorbox{red!30}{\tiny u}}$ \footnotesize again$_{26_{\textsc{adv}}}^{32}$ \footnotesize .$_{2_{\textsc{p}}}^{33}$\\\hline

\colorbox{blue!30}{\footnotesize Flex}$_{0_{\textsc{root}}}^{1 \colorbox{red!30}{\tiny u}}$ \footnotesize your$_{3_{\textsc{nmod}}}^{2}$ \colorbox{blue!30}{\footnotesize muscles}$_{1_{\textsc{obj}}}^{3}$ \colorbox{blue!30}{\footnotesize ,}$_{1_{\textsc{p}}}^{4}$ \colorbox{blue!30}{\footnotesize friend}$_{6_{\textsc{nmod}}}^{5}$ \colorbox{yellow!30}{\footnotesize Libra}$_{1_{\textsc{voc}}}^{6 \colorbox{red!30}{\tiny u}}$ \colorbox{blue!30}{\footnotesize ,}$_{1_{\textsc{p}}}^{7}$ \colorbox{blue!30}{\footnotesize and}$_{1_{\textsc{coord}}}^{8 \colorbox{red!30}{\tiny c}}$ \footnotesize prepare$_{8_{\textsc{conj}}}^{9}$ \footnotesize for$_{9_{\textsc{adv}}}^{10 \colorbox{red!30}{\tiny p}}$ \footnotesize a$_{14_{\textsc{nmod}}}^{11}$ \footnotesize relatively$_{13_{\textsc{amod}}}^{12}$ \footnotesize easy$_{14_{\textsc{nmod}}}^{13}$ \footnotesize ride$_{10_{\textsc{pmod}}}^{14}$ \colorbox{blue!30}{\footnotesize .}$_{1_{\textsc{p}}}^{15}$\\\hline

\colorbox{blue!30}{\footnotesize Bending}$_{5_{\textsc{sbj}}}^{1}$ \footnotesize to$_{1_{\textsc{dir}}}^{2 \colorbox{red!30}{\tiny p}}$ \colorbox{blue!30}{\footnotesize the}$_{4_{\textsc{nmod}}}^{3}$ \colorbox{blue!30}{\footnotesize right}$_{2_{\textsc{pmod}}}^{4}$ \colorbox{blue!30}{\footnotesize indicates}$_{0_{\textsc{root}}}^{5}$ \footnotesize :$_{5_{\textsc{p}}}^{6}$ \colorbox{blue!30}{\footnotesize "}$_{5_{\textsc{p}}}^{7 \colorbox{red!30}{\tiny u}}$ \footnotesize I$_{9_{\textsc{sbj}}}^{8}$ \colorbox{blue!30}{\footnotesize know}$_{5_{\textsc{obj}}}^{9}$ \footnotesize the$_{12_{\textsc{nmod}}}^{10}$ \footnotesize right$_{12_{\textsc{nmod}}}^{11}$ \footnotesize way$_{9_{\textsc{obj}}}^{12}$ \footnotesize to$_{12_{\textsc{nmod}}}^{13}$ \footnotesize request$_{13_{\textsc{im}}}^{14}$ \footnotesize You$_{14_{\textsc{obj}}}^{15}$ \colorbox{blue!30}{\footnotesize .}$_{5_{\textsc{p}}}^{16}$ \colorbox{blue!30}{\footnotesize "}$_{5_{\textsc{p}}}^{17 \colorbox{red!30}{\tiny u}}$\\\hline

\colorbox{blue!30}{\footnotesize SO}$_{14_{\textsc{adv}}}^{1}$ \colorbox{blue!30}{\footnotesize ,}$_{14_{\textsc{p}}}^{2}$ \colorbox{blue!30}{\footnotesize IF}$_{14_{\textsc{adv}}}^{3 \colorbox{red!30}{\tiny p}}$ \footnotesize YOU$_{5_{\textsc{sbj}}}^{4}$ \colorbox{blue!30}{\footnotesize WANT}$_{3_{\textsc{sub}}}^{5}$ \footnotesize A$_{7_{\textsc{nmod}}}^{6}$ \footnotesize BURGER$_{5_{\textsc{obj}}}^{7}$ \footnotesize AND$_{7_{\textsc{coord}}}^{8 \colorbox{red!30}{\tiny c}}$ \footnotesize FRIES$_{8_{\textsc{conj}}}^{9 \colorbox{red!30}{\tiny u}}$ \colorbox{blue!30}{\footnotesize ,}$_{14_{\textsc{p}}}^{10}$ \footnotesize WELL$_{14_{\textsc{dep}}}^{11}$ \footnotesize ,$_{14_{\textsc{p}}}^{12}$ \colorbox{blue!30}{\footnotesize IT}$_{14_{\textsc{sbj}}}^{13}$ \colorbox{blue!30}{\footnotesize IS}$_{0_{\textsc{root}}}^{14}$ \colorbox{yellow!30}{\footnotesize OK}$_{14_{\textsc{prd}}}^{15}$ \colorbox{blue!30}{\footnotesize .}$_{14_{\textsc{p}}}^{16}$\\\hline

\hline
\end{tabular}
\end{center}
\caption[The example sentences that have been improved by the DLM approach when compared to the baseline on out-of-domain test sets.]{\label{table:analysis-examples-dlm-out} The example sentences that have been improved by the DLM approach when compared to the baseline on out-of-domain test sets. In which the dependency head/relation of a token are marked as the subscript, while the superscript is the index of token. The unknown words, prepositions and conjunctions are highlighted with \colorbox{red!30}{u}, \colorbox{red!30}{p} and \colorbox{red!30}{c} respectively. We highlight the different levels of the improvements achieved by our dlm model on the dependency edges by different colours. In which the \colorbox{blue!30}{blue} colour means both head and label are corrected, the \colorbox{yellow!30}{yellow} colour means only the head is corrected and the \colorbox{green!30}{green} colour means only the label is corrected.}
\end{table}

\textbf{Example Sentences.} Table \ref{table:analysis-examples-dlm-in} and table \ref{table:analysis-examples-dlm-out} show some example sentences that have been improved largely by our DLM-based approaches on the English in-domain and out-of-domain test sets respectively.

\subsection{Analysis for Chinese}

\subsubsection{Token Level Analysis}
%label evaluation
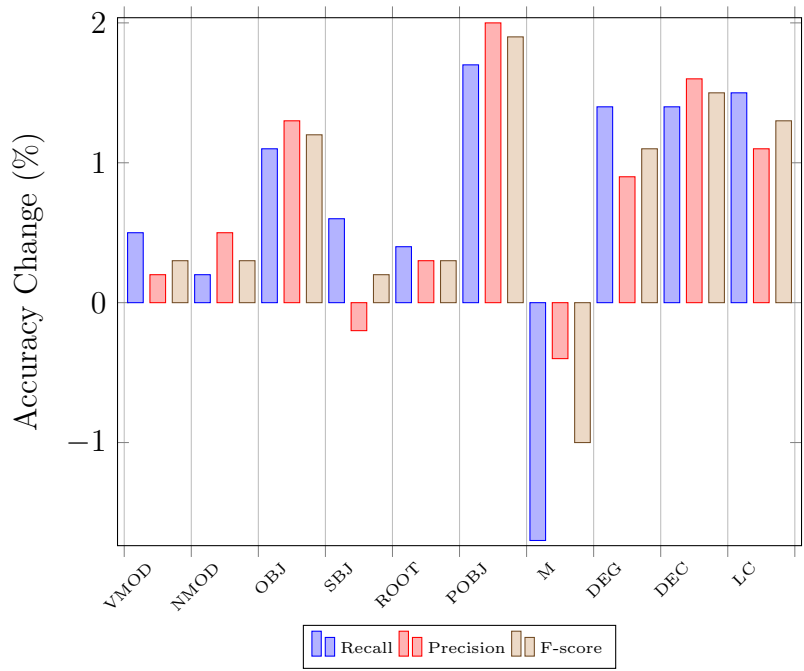
\begin{figure}[t]
	\begin{center}
		\begin{tikzpicture}
		\begin{axis}[
		width=9cm,height=7cm,
		ybar,ybar interval =0.7,
		ylabel=Accuracy Change (\%),
		enlargelimits=0.01,
		legend style={at={(0.5,-0.15)},anchor=north,legend columns=-1},
		symbolic x coords={VMOD,NMOD,OBJ,SBJ,ROOT,POBJ,M,DEG,DEC,LC,NA},
		xtick=data,
		x tick label style={font=\tiny,rotate=45,anchor=east}
		]
		\addplot coordinates{(VMOD,0.5) (NMOD,0.2) (OBJ,1.1) (SBJ,0.6) (ROOT,0.4) (POBJ,1.7) (M,-1.7) (DEG,1.4) (DEC,1.4) (LC,1.5) (NA,0.0)}; 
		\addlegendentry{\tiny Recall}
		
		\addplot coordinates{(VMOD,0.2) (NMOD,0.5) (OBJ,1.3) (SBJ,-0.2) (ROOT,0.3) (POBJ,2.0) (M,-0.4) (DEG,0.9) (DEC,1.6) (LC,1.1) (NA,0.0)}; 
		\addlegendentry{\tiny Precision}
		
		\addplot coordinates{(VMOD,0.3) (NMOD,0.3) (OBJ,1.2) (SBJ,0.2) (ROOT,0.3) (POBJ,1.9) (M,-1.0) (DEG,1.1) (DEC,1.5) (LC,1.3)(NA,0.0) }; 
		\addlegendentry{\tiny F-score}
		
		\end{axis}
		%VMOD cnt:18480 R/P/F89.9/91.7/90.8 Base R/P/F89.4/91.5/90.5 diff: F:0.3 R:0.5 P:0.2
		%NMOD cnt:15095 R/P/F94.7/92.2/93.4 Base R/P/F94.5/91.7/93.1 diff: F:0.3 R:0.2 P:0.5
		%OBJ cnt:3997 R/P/F81.0/79.8/80.4 Base R/P/F79.9/78.5/79.2 diff: F:1.2 R:1.1 P:1.3
		%SBJ cnt:3310 R/P/F82.8/80.0/81.4 Base R/P/F82.2/80.2/81.2 diff: F:0.2 R:0.6 P:-0.2
		%ROOT cnt:1915 R/P/F76.7/76.7/76.7 Base R/P/F76.3/76.4/76.4 diff: F:0.3 R:0.4 P:0.3
		%POBJ cnt:1772 R/P/F83.0/82.8/82.9 Base R/P/F81.3/80.8/81.0 diff: F:1.9 R:1.7 P:2.0
		%M cnt:1736 R/P/F92.3/94.1/93.2 Base R/P/F94.0/94.5/94.2 diff: F:-1.0 R:-1.7 P:-0.4
		%DEG cnt:1281 R/P/F87.3/85.3/86.3 Base R/P/F85.9/84.4/85.2 diff: F:1.1 R:1.4 P:0.9
		%DEC cnt:1179 R/P/F76.5/77.3/76.9 Base R/P/F75.1/75.7/75.4 diff: F:1.5 R:1.4 P:1.6
		%LC cnt:828 R/P/F93.5/93.1/93.3 Base R/P/F92.0/92.0/92.0 diff: F:1.3 R:1.5 P:1.1
		
		\end{tikzpicture}
	\end{center}
	\caption{\label{figure:analysis-label-dlm-cn} The Chinese performance comparison between the DLM approach and the baseline on major labels.}
\end{figure}

%Confution matric
\begin{table}[t]
\begin{center}
\begin{tabular}{l|r|r}
\hline
\bf Confusion & \bf Baseline & \bf DLM\\\hline
\footnotesize VMOD $\rightarrow$ \footnotesize POBJ & 125 & 127\\
\footnotesize VMOD $\rightarrow$ \footnotesize ROOT & 305 & 305\\
\footnotesize VMOD $\rightarrow$ \footnotesize NMOD & 579 & 508\\
\footnotesize VMOD $\rightarrow$ \footnotesize OBJ & 411 & 384\\
\footnotesize VMOD $\rightarrow$ \footnotesize SBJ & 307 & 309\\
\footnotesize VMOD $\rightarrow$ \footnotesize DEC,DEG & 126 & 119\\\hline
\footnotesize NMOD $\rightarrow$ \footnotesize VMOD & 369 & 374\\
\footnotesize NMOD $\rightarrow$ \footnotesize SBJ & 153 & 161\\
\footnotesize NMOD $\rightarrow$ \footnotesize POBJ,DEC,M,OBJ & 242 & 208\\\hline
\footnotesize SBJ $\rightarrow$ \footnotesize NMOD & 218 & 214\\
\footnotesize SBJ $\rightarrow$ \footnotesize VMOD & 186 & 182\\
\footnotesize SBJ $\rightarrow$ \footnotesize OBJ & 104 & 93\\
\footnotesize SBJ $\rightarrow$ \footnotesize POBJ & 47 & 50\\\hline
\footnotesize OBJ $\rightarrow$ \footnotesize VMOD & 282 & 279\\
\footnotesize OBJ $\rightarrow$ \footnotesize NMOD & 160 & 140\\
\footnotesize OBJ $\rightarrow$ \footnotesize SBJ & 107 & 108\\
\footnotesize OBJ $\rightarrow$ \footnotesize POBJ,ROOT,DEG & 192 & 175\\\hline
\footnotesize ROOT $\rightarrow$ \footnotesize VMOD & 299 & 285\\
\footnotesize ROOT $\rightarrow$ \footnotesize OBJ & 74 & 77\\\hline
\footnotesize POBJ $\rightarrow$ \footnotesize NMOD,VMOD,OBJ,SBJ & 270 & 241\\\hline
\footnotesize M $\rightarrow$ \footnotesize NMOD & 58 & 83\\\hline
\footnotesize DEG $\rightarrow$ \footnotesize DEC,VMOD,OBJ & 151 & 136\\\hline
\footnotesize DEC $\rightarrow$ \footnotesize NMOD,VMOD,OBJ,DEG & 224 & 215\\\hline
\footnotesize LC $\rightarrow$ \footnotesize VMOD & 31 & 26\\\hline
\end{tabular}
\end{center}
\caption{\label{table:analysis-confusion-dlm-cn} The confusion matrix of dependency labels, compared between the DLM approach and the baseline on Chinese test set.}
\end{table}

\textbf{Individual Label Accuracy.} The Chinese dataset has a smaller label set than that of English, the 10 most frequent labels already cover 97\% of the test set. We illustrate accuracy changes of individual labels in Figure \ref{figure:analysis-label-dlm-cn}. Our DLM model improved all major labels, the only exception is the label M (dependent of measure word, such as in words `` \includegraphics[height=0.8em]{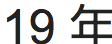} " (19 years),`` \includegraphics[height=0.8em]{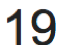} " is the dependent of the measure word `` \includegraphics[height=0.8em]{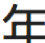} ") which showed a 1\% decreasement in f-score. Our model achieved the largest improvement of 1.9\% on POBJ (object of preposition), large improvements of more than 1\% can be also found for label OBJ (object), DEG (dependent of associative DE), DEC (dependent of DE in a relative-clause) and LC (Child of localizer). For all other labels, moderate improvements of 0.2\%-0.3\% are achieved by our method. Table \ref{table:analysis-confusion-dlm-cn} shows the confusion matrix of the dependency labels on the Chinese test set.

%corpus UNK
\begin{table}[t]
	\begin{center}
		\begin{tabular}{|l|r|rr|rr|}
			\cline{3-6}
			\multicolumn{2}{c|}{}& \multicolumn{2}{|c|}{\bf DLM} &\multicolumn{2}{|c|}{\bf Baseline}\\
			\cline{2-6}
			\multicolumn{1}{c|}{}&\bf Tokens &\bf LAS&\bf UAS&\bf LAS&\bf UAS\\
			\hline
			\bf Known &39636&80.1&82.9&79.1&81.9\\%diff: LAS:1.0 UAS:1.0
			\bf Unknown &3137&71.3&77.7&71.3&77.5\\%diff: LAS:0.0 UAS:0.2
			\hline
			\bf All &42773&79.4&82.5&78.5&81.6\\%diff: LAS:0.9 UAS:0.9
			\hline
		\end{tabular}
	\end{center}
	\caption{\label{table:analysis-corpusunk-dlm-cn} The Chinese accuracy comparison between the DLM approach and the baseline on unknown words.}
\end{table}

\textbf{Unknown Words Accuracy.} Table \ref{table:analysis-corpusunk-dlm-cn} shows our analysis of the unknown words accuracies. Our DLM model improved mainly the known words, with 1\% large gains for both labelled and unlabelled accuracies. While our model did not improve the labelled accuracy of the unknown words, the model only achieved a small 0.2\% improvement on the unlabelled score. This is an indication that the Chinese unknown words are very hard to improve without the manually annotated examples.

\subsubsection{Sentence Level Analysis}

%number of tokens
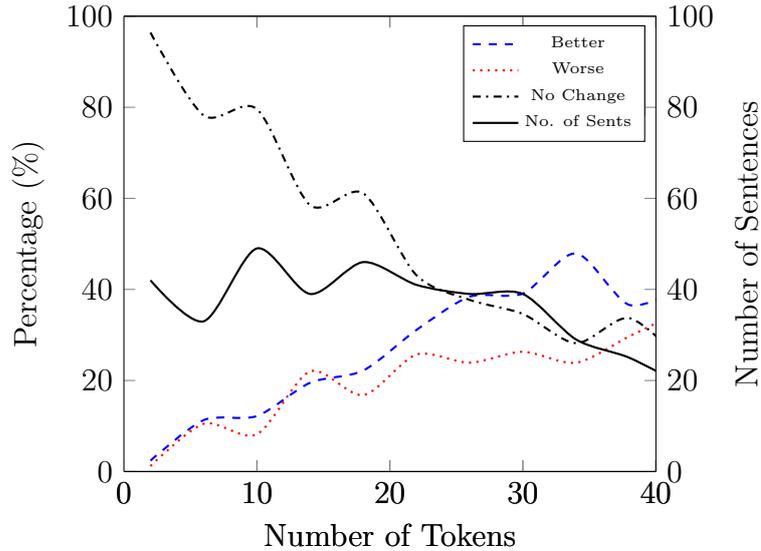
\begin{figure}[t]
	\begin{center}
		\begin{tikzpicture}
		\pgfplotsset{
			xmin=0,xmax=40,
			xlabel=Number of Tokens}
		\begin{axis}[
		axis y line*=left,
		ymin=0,ymax=100,
		ylabel=Percentage (\%)]
		%Better
		\addplot[smooth,thick,dashed,color=blue] coordinates {(2,2.4) (6,11.3) (10,12.2) (14,19.5) (18,22.2) (22,31.1) (26,38.4) (30,39.1) (34,47.9) (38,36.6) (42,40.5) };\label{better}
		\addlegendentry{\tiny Better}
		
		%Worse
		\addplot[smooth,thick,dotted,color=red] coordinates{(2,1.2) (6,10.5) (10,8.2) (14,22.0) (18,16.8) (22,25.7) (26,23.9) (30,26.3) (34,23.9) (38,29.7) (42,35.4) };\label{worse}
		\addlegendentry{\tiny Worse}
		
		%No Change
		\addplot[smooth,thick,dashdotted] coordinates{(2,96.4) (6,78.2) (10,79.6) (14,58.5) (18,61.1) (22,43.1) (26,37.7) (30,34.6) (34,28.2) (38,33.7) (42,24.1) };\label{nochange}
		\addlegendentry{\tiny No Change}
		
		\end{axis}
		\begin{axis}[
		axis y line*=right,
		ymin=0,ymax=100,
		ylabel=Number of Sentences]
		\addlegendimage{/pgfplots/refstyle=better}\addlegendentry{\tiny Better}
		\addlegendimage{/pgfplots/refstyle=worse}\addlegendentry{\tiny Worse}
		\addlegendimage{/pgfplots/refstyle=nochange}\addlegendentry{\tiny No Change}
		%All Sent
		\addplot[smooth,thick,solid] coordinates{(2,42.0) (6,33.0) (10,49.0) (14,39.0) (18,46.0) (22,41.0) (26,39.0) (30,39.0) (34,29.0) (38,25.0) (42,19.0) };\addlegendentry{\tiny No. of Sents}
		
		\end{axis}
		%2 Total:169 LAS better/worse/nochange:4/2/163 2.4/1.2/96.4 Diff:1.2
		%6 Total:133 LAS better/worse/nochange:15/14/104 11.3/10.5/78.2 Diff:0.8
		%10 Total:196 LAS better/worse/nochange:24/16/156 12.2/8.2/79.6 Diff:4.0
		%14 Total:159 LAS better/worse/nochange:31/35/93 19.5/22.0/58.5 Diff:-2.5
		%18 Total:185 LAS better/worse/nochange:41/31/113 22.2/16.8/61.1 Diff:5.4
		%22 Total:167 LAS better/worse/nochange:52/43/72 31.1/25.7/43.1 Diff:5.4
		%26 Total:159 LAS better/worse/nochange:61/38/60 38.4/23.9/37.7 Diff:14.5
		%30 Total:156 LAS better/worse/nochange:61/41/54 39.1/26.3/34.6 Diff:12.8
		%34 Total:117 LAS better/worse/nochange:56/28/33 47.9/23.9/28.2 Diff:24.0
		%38 Total:101 LAS better/worse/nochange:37/30/34 36.6/29.7/33.7 Diff:6.9
		%42 Total:79 LAS better/worse/nochange:32/28/19 40.5/35.4/24.1 Diff:5.1
		
		\end{tikzpicture}
	\end{center}
	\caption{\label{figure:analysis-sentlength-dlm-cn} The Chinese comparison between the DLM approach and the baseline on different number of tokens per sentence.}
\end{figure}

\textbf{Sentence Length.} As shown in Figure \ref{figure:analysis-sentlength-dlm-cn}, the Chinese sentences are evenly distributed in the classes of different sentence length. Our model had limited effects on sentences less than 20 tokens but showed large gains on sentences longer than that. The enhanced model achieved a gain of 5\% on sentences of 20 tokens and the improvement increases until reaching the largest gain (24\%) at the class of 35 tokens/sentence. Overall the major improvements of Chinese data were achieved on sentences that have at least 20 tokens.

\textbf{Unknown Words.} We skip the unknown words factor for our Chinese sentence level analysis. This is due to the finding from our token level analysis, which suggests our model did not improve the accuracy of the unknown words. Thus it is not necessary for us to conduct further evaluation of this factor.

%number of prepositions
\begin{figure}[t]
	\begin{center}
		\begin{tikzpicture}
		\pgfplotsset{
			xmin=0,xmax=3,
			xtick={0,1,2,3},
			xlabel=Number of Prepositions}
		\begin{axis}[
		axis y line*=left,
		ymin=0,ymax=100,
		ylabel=Percentage (\%)]
		%Better
		\addplot[smooth,thick,dashed,color=blue] coordinates {(0,19.3) (1,33.9) (2,35.1) (3,37.2) (4,53.8) (5,62.5) (6,60.0) };\label{better}
		\addlegendentry{\tiny Better}
		
		%Worse
		\addplot[smooth,thick,dotted,color=red] coordinates{(0,15.7) (1,23.5) (2,32.5) (3,35.4) (4,30.8) (5,25.0) (6,20.0) };\label{worse}
		\addlegendentry{\tiny Worse}
		
		%No Change
		\addplot[smooth,thick,dashdotted] coordinates{(0,65.1) (1,42.6) (2,32.5) (3,27.4) (4,15.4) (5,12.5) (6,20.0) };\label{nochange}
		\addlegendentry{\tiny No Change}
		
		\end{axis}
		\begin{axis}[
		axis y line*=right,
		ymin=0,ymax=1000,
		ylabel=Number of Sentences]
		\addlegendimage{/pgfplots/refstyle=better}\addlegendentry{\tiny Better}
		\addlegendimage{/pgfplots/refstyle=worse}\addlegendentry{\tiny Worse}
		\addlegendimage{/pgfplots/refstyle=nochange}\addlegendentry{\tiny No Change}
		%All Sent
		\addplot[smooth,thick,solid] coordinates{(0,887.0) (1,584.0) (2,268.0) (3,113.0) (4,39.0) (5,16.0) (6,5.0) };\addlegendentry{\tiny No. of Sents}
		
		\end{axis}
		%0 Total:887 LAS better/worse/nochange:171/139/577 19.3/15.7/65.1 Diff:3.6
		%1 Total:584 LAS better/worse/nochange:198/137/249 33.9/23.5/42.6 Diff:10.4
		%2 Total:268 LAS better/worse/nochange:94/87/87 35.1/32.5/32.5 Diff:2.6
		%3 Total:113 LAS better/worse/nochange:42/40/31 37.2/35.4/27.4 Diff:1.8
		%4 Total:39 LAS better/worse/nochange:21/12/6 53.8/30.8/15.4 Diff:23.0
		%5 Total:16 LAS better/worse/nochange:10/4/2 62.5/25.0/12.5 Diff:37.5
		%6 Total:5 LAS better/worse/nochange:3/1/1 60.0/20.0/20.0 Diff:40.0
		
		\end{tikzpicture}
	\end{center}
	\caption{\label{figure:analysis-sentin-dlm-cn} The Chinese comparison between the DLM approach and the baseline on different number of prepositions per sentence.}
\end{figure}
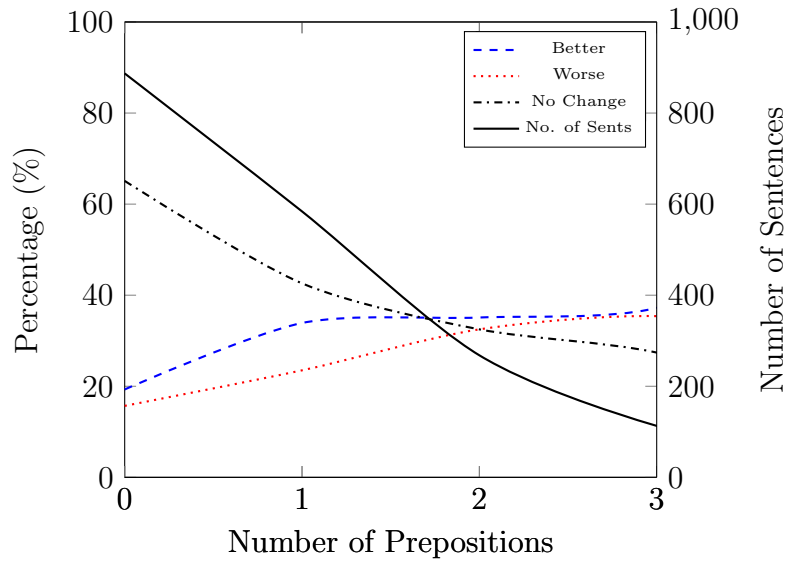

\textbf{Prepositions.} As shown in Figure \ref{figure:analysis-sentin-dlm-cn} most Chinese sentences have no or only single prepositions. The DLM model achieved an improvement of 3.6\% for sentences do not contain a preposition. For sentences that contain single preposition, our model achieved 10.4\% gain. The gain decreased largely when more prepositions are found in the sentences.

%number of conjunctions
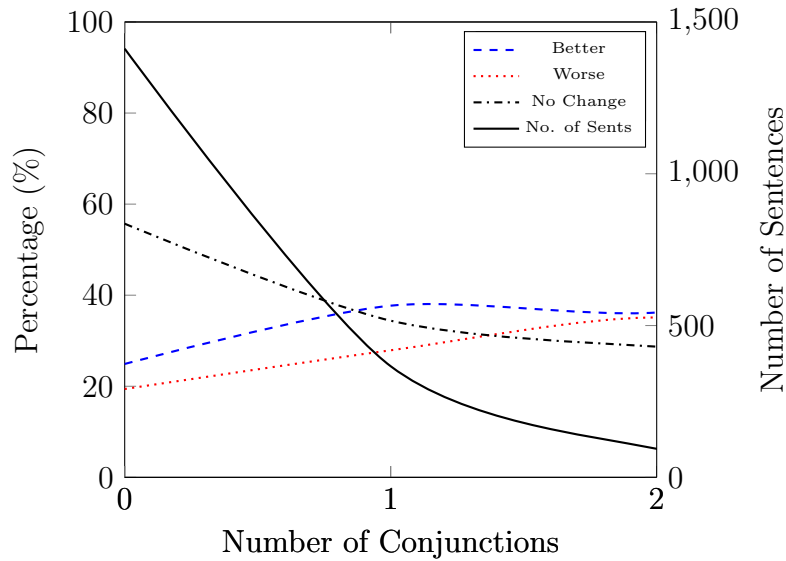
\begin{figure}[t]
	\begin{center}
		\begin{tikzpicture}
		\pgfplotsset{
			xmin=0,xmax=2,
			xtick={0,1,2},
			xlabel=Number of Conjunctions}
		\begin{axis}[
		axis y line*=left,
		ymin=0,ymax=100,
		ylabel=Percentage (\%)]
		%Better
		\addplot[smooth,thick,dashed,color=blue] coordinates {(0,24.9) (1,37.7) (2,36.2) (3,45.5) };\label{better}
		\addlegendentry{\tiny Better}
		
		%Worse
		\addplot[smooth,thick,dotted,color=red] coordinates{(0,19.4) (1,27.9) (2,35.1) (3,27.3) };\label{worse}
		\addlegendentry{\tiny Worse}
		
		%No Change
		\addplot[smooth,thick,dashdotted] coordinates{(0,55.7) (1,34.4) (2,28.7) (3,27.3) };\label{nochange}
		\addlegendentry{\tiny No Change}
		
		\end{axis}
		\begin{axis}[
		axis y line*=right,
		ymin=0,ymax=1500,
		ylabel=Number of Sentences]
		\addlegendimage{/pgfplots/refstyle=better}\addlegendentry{\tiny Better}
		\addlegendimage{/pgfplots/refstyle=worse}\addlegendentry{\tiny Worse}
		\addlegendimage{/pgfplots/refstyle=nochange}\addlegendentry{\tiny No Change}
		%All Sent
		\addplot[smooth,thick,solid] coordinates{(0,1412.0) (1,366.0) (2,94.0) (3,22.0) };\addlegendentry{\tiny No. of Sents}
		
		\end{axis}
		%0 Total:1412 LAS better/worse/nochange:352/274/786 24.9/19.4/55.7 Diff:5.5
		%1 Total:366 LAS better/worse/nochange:138/102/126 37.7/27.9/34.4 Diff:9.8
		%2 Total:94 LAS better/worse/nochange:34/33/27 36.2/35.1/28.7 Diff:1.1
		%3 Total:22 LAS better/worse/nochange:10/6/6 45.5/27.3/27.3 Diff:18.2
		
		\end{tikzpicture}
	\end{center}
	\caption{\label{figure:analysis-sentcc-dlm-cn} The Chinese comparison between the DLM approach and the baseline on different number of conjunctions per sentence.}
\end{figure}

\textbf{Conjunctions.} The curves of our analysis on the different number of conjunctions (Figure \ref{figure:analysis-sentcc-dlm-cn}) are nearly identical to that of prepositions. For sentences that do not have conjunction a gain of 5.5\% is achieved and the improvement for sentences containing a single conjunction is much larger (9.8\%). The improvement dropped for sentences containing 2 conjunctions.

\section{Chapter Summary}\label{section:dlm-conclusion}
In this chapter, we adapted the dependency language models (DLM) approach of \newcite{chen2012utilizing} to a strong transition-based parser. We integrated a small number of DLM-based features into the parser to allow the parser to explore DLMs extracted from a large auto-parsed corpus. We evaluated the parser with single and multiple DLMs extracted from corpora of different size and quality to improve the in-domain accuracy of the English and Chinese texts. The English model enhanced by a unigram DLM extracted from double parsed high-quality sentences achieved statistically significant improvements of 0.46\% and 0.51\% for labelled and unlabelled accuracies respectively. Our results outperform most of the latest systems and are close to the state-of-the-art. By using all unigram, bigram and trigram DLMs in our Chinese experiments, we achieved large improvements of 0.93\% and 0.98\% for both labelled and unlabelled scores. When increasing the beam size to 150, our system outperforms the best reported results by 0.2\%. In addition to that, our approach gained an improvement of 0.4\% on Chinese part-of-speech tagging.

We further evaluate our approach on our main evaluation corpus. The method is tested on both in-domain and out-of-domain parsing. Our DLM-based approach achieved large improvement on all five domains evaluated (\textsc{Conll, Weblogs, Newsgroups, Reviews, Answers}). We achieved the labelled and unlabelled improvements of up to 0.91\% and 0.82\% on \textsc{Newsgroups} domain. On average we achieved 0.6\% gains for both labelled and unlabelled scores on four out-of-domain test sets. We also improved the in-domain accuracy by 0.36\% (LAS) and 0.4\% (UAS).

The analysis on our English main evaluation corpus suggests that the DLM model behaves differently on in-domain and out-of-domain parsing for a number of factors. Firstly, the DLM model achieved the largest improvement on label CONJ (conjunct) and LOC (locative adverbial) for in-domain parsing, while the largest improvement for out-of-domain dataset is contributed by OBJ (object) and PRD (predicative complement). Secondly, the DLM model improved more on unknown words for in-domain data but for out-of-domain text, DLM model delivered larger gains on known words. Thirdly, the analysis on sentence level shows that our model achieved most improvement on sentences of a length between 10 and 20, the range is wider (10-35) for out-of-domain data. 

We also analysed the Chinese results. The analysis shows the improvement on Chinese data is mainly contributed by the objects (OBJ, POBJ), dependent of DE (DEC, DEG) and children of localizer (LC). The DLM model only shows a large improvement on the known words, it nearly does not affect the unknown words accuracy. The DLM model mostly helped the sentences that have at least 20 tokens.

\chapter{Conclusions}\label{chapter:conclusion}
In this last chapter, we summarise the work of this thesis. In this thesis, we evaluated three semi-supervised techniques (co-training, self-training and dependency language models) on out-of-domain dependency parsing. The evaluations on various domains and languages demonstrated the effectiveness and robustness of all three techniques. We believe we have achieved the initial goals of this thesis.

As introduced in Chapter \ref{chapter:intro}, our goals for this thesis are to answer the following research questions:

\begin{enumerate}
	\item \rQone
	\item \rQtwo
	\item \rQthree
	\item \rQfour
	\item \rQfive
	\item \rQsix
	\item \rQseven
\end{enumerate}

In the following sections, we answer all the questions in turns. Section \ref{section:conclusion:cotrain} summarises our work on agreement based co-training and tri-training, we answer questions 1 and 2 in this section. In Section \ref{section:conclusion:selftrain} we conclude our evaluations on English and multi-lingual confidence-based self-training; questions 3 and 4 are answered in this section. We discuss our work on dependency language models in Section \ref{section:conclusion:dlm} and answer the last three questions.

\section{Conclusions on Co-training}\label{section:conclusion:cotrain}
In this section, we discuss our work on agreement based co-training (Chapter \ref{chapter:cotrain}) and answer two research questions related to our co-training evaluation.

\subsection{\rQone}
To answer this question we evaluated the agreement based co-training approach with four popular off-the-shelf parsers (Malt parser \cite{nivre2009non}, MST parser \cite{mcdonald2006online}, Mate parser \cite{bohnet2013joint} and Turbo parser \cite{martins2010turbo}). We pair the Mate parser with the rest of three parsers to create three co-training settings. The unlabelled data is double parsed by the parser pairs and the sentences that are annotated the same by both parsers are used as additional training data. New models are created by retraining the Mate parser on training data boosted by different parser pairs. All the enhanced models achieved large gains when compared to the baselines. The largest improvement of 1.1\% is achieved by the Mate and Malt parsers. An additional 0.27\% is achieved when we omit the short sentences from the additional training data. Our results demonstrated the effectiveness of the agreement-based co-training on out-of-domain parsing. The off-the-shelf parsers have proved their suitability on this task.

\subsection{\rQtwo}
The tri-training different from the normal co-training by retraining the evaluation learner on additional training data agreed by other two learners. In total, three learners are required, to form the tri-training we used the Malt, MST parsers as the source learners and the Mate parser is used as the evaluation learner. The tri-trained model outperforms the best normal co-training setting on all the experiments, thus is more effective. A large 1.6\% improvement is achieved on the development set when compared to the baseline. We further evaluate the tri-training approach on four test domains. It achieved largest labelled and unlabelled improvements of 1.8\% and 0.58\% respectively. On average it achieved 1.5\% (LAS) and 0.4\% (UAS) for all four test domains. Our results not only confirmed the tri-training is more effective than normal co-training but also demonstrated the merit of tri-training on multiple tested domains.

\section{Conclusions on Self-training}\label{section:conclusion:selftrain}
In this section, we discuss our work on confidence-based self-training (Chapter \ref{chapter:selftrain} and \ref{chapter:multiselftrain}) and answer two relevant questions.

\subsection{\rQthree}\label{subsection:researchquestion3-self}
We start with the hypothesis that the selection of high-quality auto-annotated data is the pre-condition of the successful use of self-training on dependency parsing. To obtain the high-quality additional training data we introduced two confidence-based methods that are able to detect high accuracy annotations. We compared our confidence-based self-training with the random selection-based self-training and the baseline. The random selection-based self-training is \textit{not} able to gain statistically significant improvement which is in line with previous work. Both confidence-based methods achieved large improvements on all three web domain test sets and the additional \textsc{Chemical} domain evaluation. For web domain, our method achieved up to 0.8\% gains for both labelled and unlabelled scores. On average both methods improved the baseline by 0.6\% (LAS and UAS). The evaluation on the \textsc{Chemical} domain resulted in larger improvements of up to 1.4\% (LAS) and 1.2\% (UAS). The evaluation on different domains confirmed our hypothesis.

\subsection{\rQfour}
We demonstrated the effectiveness of our confidence-based self-training for English dependency parsing in the last question, cf. Section \ref{subsection:researchquestion3-self}. To assess the multi-lingual capacity of our confidence-based self-training, we evaluated it on nine languages (\textsc{Arabic, Basque, French, German, Hebrew, Hungarian, Korean, Polish, Swedish}) corpora. We evaluated on a unified setting for all the languages, the results show our method is able to achieve statistically significant improvements on five languages (\textsc{Basque, German, Hungarian, Korean} and \textsc{Swedish}). Our self-training approach achieved the largest labelled and unlabelled accuracy gain of 2.14\% and 1.79\% on \textsc{Korean}. The average improvements achieved by our method on five languages are 0.87\% (LAS) and 0.78\% (UAS). We further analyse the result of a negative effect (\textsc{French}) introduced by our method to assess the reason why self-training did not work. The analysis suggests the large difference between unlabelled data and the training data is likely to be the main reason disqualifies the self-training. Overall, our evaluations show that confidence-based self-training can be successfully applied to multi-lingual dependency parsing.

\section{Conclusions on Dependency Language Models}\label{section:conclusion:dlm}
In this section, we discuss our findings on dependency language models (Chapter \ref{chapter:dlm}) and answer the last three research questions.

\subsection{\rQfive}
To answer this question, we applied the dependency language models (DLM) to the Mate transition-based parser. We successfully integrated the DLM-based features to the transition-based parser by using a modified version of \newcite{chen2012utilizing}'s original templates for the graph-based parser. The evaluations on English and Chinese in-domain parsing confirmed the effectiveness of dependency language models on the Mate parser. We improved a strong English baseline by 0.46\% and 0.51\% for labelled and unlabelled accuracies respectively. For Chinese, we achieved the state-of-the-art accuracy and gained large improvements of 0.93\% (LAS) and 0.98\% (UAS). The results show a strong evidence that dependency language models can be adapted successfully to a strong transition-based parser.

\subsection{\rQsix}
To address this question, we applied our approach to four web domain texts (\textsc{Weblogs, Newsgroups, Reviews, Answers}). We achieved the largest labelled and unlabelled improvements of 0.91\% and 0.82\% on \textsc{Newsgroups} domain. And on average we achieved 0.6\% gains for both labelled and unlabelled scores. The evaluations on multiple domains advised that DLM-based approach is an effective technique for domain adaptation tasks.

\subsection{\rQseven}
The evaluations on both English and Chinese suggest \textit{no} large additional gains can be achieved by using DLMs extracted from corpus larger than 5 million sentences. In fact, in most of the cases, the best model is achieved by using DLMs extracted from 5 million sentences. The evaluation of using DLMs extracted from high-quality data, on the other hand, surpasses the best results achieved by normal quality DLMs. Overall, the quality of the auto-labelled data used to generate DLMs is more important than the quantity.

\section{Chapter Summary}
In this chapter, we summarised our work of this thesis by answering seven research questions that we introduced in Chapter \ref{chapter:intro}. We successfully answered all the questions using our findings in the previous chapters.

\backmatter 
\bibliographystyle{acl2016} 
\bibliography{thesis}

\begin{thebibliography}{}

\bibitem[\protect\citename{Andor \bgroup et al.\egroup
  }2016]{andor2016globally}
Daniel Andor, Chris Alberti, David Weiss, Aliaksei Severyn, Alessandro Presta,
  Kuzman Ganchev, Slav Petrov, and Michael Collins.
\newblock 2016.
\newblock Globally normalized transition-based neural networks.
\newblock In {\em Proceedings of the 54th Annual Meeting of the Association for
  Computational Linguistics}, pages 2442--2452, Berlin, Germany. Association
  for Computational Linguistics.

\bibitem[\protect\citename{Bengio \bgroup et al.\egroup }2003]{bengio03a}
Yoshua Bengio, R{\'e}jean Ducharme, Pascal Vincent, and Christian Janvin.
\newblock 2003.
\newblock A neural probabilistic language model.
\newblock {\em Journal of Machine Learning Research}, 3:1137--1155.

\bibitem[\protect\citename{Bj{\"o}rkelund \bgroup et al.\egroup
  }2014]{bjorkelund2014spmrl}
Anders Bj{\"o}rkelund, {\"O}zlem {\c{C}}etino{\u{g}}lu, Agnieszka
  Fale{\'{n}}ska, Rich{\'a}rd Farkas, Thomas Mueller, Wolfgang Seeker, and
  Zsolt Sz{\'a}nt{\'o}.
\newblock 2014.
\newblock The {IMS-Wroc{\l}aw-Szeged-CIS} entry at the {SPMRL} 2014 shared
  task: Reranking and morphosyntax meet unlabeled data.
\newblock In {\em Proceedings of the Shared Task on Statistical Parsing of
  Morphologically Rich Languages}, Dublin, Ireland. Dublin City University.

\bibitem[\protect\citename{Blum and Mitchell}1998]{blum98}
Avrim Blum and Tom Mitchell.
\newblock 1998.
\newblock Combining labeled and unlabeled data with co-training.
\newblock In {\em Proceedings of the Eleventh Annual Conference on
  Computational Learning Theory}, pages 92--100, Madison, Wisconsin, USA.
  Association for Computing Machinery.

\bibitem[\protect\citename{Bohnet and Kuhn}2012]{bohnet2012eacl}
Bernd Bohnet and Jonas Kuhn.
\newblock 2012.
\newblock The best of both worlds -- a graph-based completion model for
  transition-based parsers.
\newblock In {\em Proceedings of the 13th Conference of the European Chapter of
  the Association for Computational Linguistics}, pages 77--87, Avignon,
  France. Association for Computational Linguistics.

\bibitem[\protect\citename{Bohnet and Nivre}2012]{bohnet2012emnlp}
Bernd Bohnet and Joakim Nivre.
\newblock 2012.
\newblock A transition-based system for joint part-of-speech tagging and
  labeled non-projective dependency parsing.
\newblock In {\em Proceedings of the 2012 Joint Conference on Empirical Methods
  in Natural Language Processing and Computational Natural Language Learning},
  pages 1455--1465, Jeju Island, Korea. Association for Computational
  Linguistics.

\bibitem[\protect\citename{Bohnet \bgroup et al.\egroup }2013]{bohnet2013joint}
Bernd Bohnet, Joakim Nivre, Igor Boguslavsky, Richard Farkas, Filip Ginter, and
  Jan Hajic.
\newblock 2013.
\newblock Joint morphological and syntactic analysis for richly inflected
  languages.
\newblock {\em Transactions of the Association of Computational Linguistics},
  1:415--428.

\bibitem[\protect\citename{Bohnet}2010]{bohnet10}
Bernd Bohnet.
\newblock 2010.
\newblock Top accuracy and fast dependency parsing is not a contradiction.
\newblock In {\em Proceedings of the 23rd International Conference on
  Computational Linguistics}, pages 89--97, Beijing, China. Association for
  Computational Linguistics.

\bibitem[\protect\citename{Brown \bgroup et al.\egroup }1992]{brown1992class}
Peter~F Brown, Peter~V Desouza, Robert~L Mercer, Vincent J~Della Pietra, and
  Jenifer~C Lai.
\newblock 1992.
\newblock Class-based n-gram models of natural language.
\newblock {\em Computational linguistics}, 18(4):467--479.

\bibitem[\protect\citename{Carreras}2007]{carreras07}
Xavier Carreras.
\newblock 2007.
\newblock Experiments with a higher-order projective dependency parser.
\newblock In {\em Proceedings of the 2007 Joint Conference on Empirical Methods
  in Natural Language Processing and Computational Natural Language Learning},
  pages 957--961, Prague, Czech Republic. Association for Computational
  Linguistics.

\bibitem[\protect\citename{Cerisara}2014]{cerisara2014spmrl}
Christophe Cerisara.
\newblock 2014.
\newblock Semi-supervised experiments at {LORIA} for the {SPMRL} 2014 shared
  task.
\newblock In {\em Proceedings of the Shared Task on Statistical Parsing of
  Morphologically Rich Languages}, Dublin, Ireland. Dublin City University.

\bibitem[\protect\citename{Charniak and Johnson}2005]{charniak2005}
Eugene Charniak and Mark Johnson.
\newblock 2005.
\newblock Coarse-to-fine n-best parsing and maxent discriminative reranking.
\newblock In {\em Proceedings of the 43rd Annual Meeting of the Association for
  Computational Linguistics}, pages 173--180, Ann Arbor, Michigan, USA.
  Association for Computational Linguistics.

\bibitem[\protect\citename{Charniak}1997]{charniak1997statistical}
Eugene Charniak.
\newblock 1997.
\newblock Statistical parsing with a context-free grammar and word statistics.
\newblock In {\em Proceedings of the Fourteenth National Conference on
  Artificial Intelligence and Ninth Conference on Innovative Applications of
  Artificial Intelligence}, pages 598--603, Providence, Rhode Island.
  Association for the Advancement of Artificial Intelligence.

\bibitem[\protect\citename{Charniak}2000]{charniak00}
Eugene Charniak.
\newblock 2000.
\newblock A maximum-entropy-inspired parser.
\newblock In {\em Proceedings of the 1st Meeting of the North American Chapter
  of the Association for Computational Linguistics}, pages 132--139, Seattle,
  Washington, USA. Association for Computational Linguistics.

\bibitem[\protect\citename{Chelba \bgroup et al.\egroup
  }2013]{chelba13onebillion}
Ciprian Chelba, Tomas Mikolov, Mike Schuster, Qi~Ge, Thorsten Brants, and
  Philipp Koehn.
\newblock 2013.
\newblock One billion word benchmark for measuring progress in statistical
  language modeling.
\newblock Technical Report 41880, Google.

\bibitem[\protect\citename{Chen and Manning}2014]{chen2014neural}
Danqi Chen and Christopher Manning.
\newblock 2014.
\newblock A fast and accurate dependency parser using neural networks.
\newblock In {\em Proceedings of the 2014 Conference on Empirical Methods in
  Natural Language Processing}, pages 740--750, Doha, Qatar. Association for
  Computational Linguistics.

\bibitem[\protect\citename{Chen \bgroup et al.\egroup }2008]{chen2008learning}
Wenliang Chen, Youzheng Wu, and Hitoshi Isahara.
\newblock 2008.
\newblock Learning reliable information for dependency parsing adaptation.
\newblock In {\em Proceedings of the 22nd International Conference on
  Computational Linguistics}, pages 113--120, Manchester, UK. Association for
  Computational Linguistics.

\bibitem[\protect\citename{Chen \bgroup et al.\egroup }2012]{chen2012utilizing}
Wenliang Chen, Min Zhang, and Haizhou Li.
\newblock 2012.
\newblock Utilizing dependency language models for graph-based dependency
  parsing models.
\newblock In {\em Proceedings of the 50th Annual Meeting of the Association for
  Computational Linguistics}, pages 213--222, Jeju Island, Korea. Association
  for Computational Linguistics.

\bibitem[\protect\citename{Chen \bgroup et al.\egroup }2013]{chen2013feature}
Wenliang Chen, Min Zhang, and Yue Zhang.
\newblock 2013.
\newblock Semi-supervised feature transformation for dependency parsing.
\newblock In {\em Proceedings of the 2013 Conference on Empirical Methods in
  Natural Language Processing}, pages 1303--1313, Seattle, Washington, USA.
  Association for Computational Linguistics.

\bibitem[\protect\citename{Chen \bgroup et al.\egroup }2015]{chen2015feature}
Wenliang Chen, Min Zhang, and Yue Zhang.
\newblock 2015.
\newblock Distributed feature representations for dependency parsing.
\newblock {\em IEEE/ACM Transactions on Audio, Speech, and Language
  Processing}, 23(3):451--460.

\bibitem[\protect\citename{Chrupala}2011]{chrupala2011lda}
Grzegorz Chrupala.
\newblock 2011.
\newblock Efficient induction of probabilistic word classes with {LDA}.
\newblock In {\em Proceedings of 5th International Joint Conference on Natural
  Language Processing}, pages 363--372, Chiang Mai, Thailand. Asian Federation
  of Natural Language Processing.

\bibitem[\protect\citename{Collins and Singer}1999]{collins99}
Michael Collins and Yoram Singer.
\newblock 1999.
\newblock Unsupervised models for named entity classification.
\newblock In {\em Proceedings of the 1999 Joint SIGDAT Conference on Empirical
  Methods in Natural Language Processing and Very Large Corpora}, pages
  100--110, College Park, Maryland, USA. Association for Computational
  Linguistics.

\bibitem[\protect\citename{Collins}1999]{collins1999phd}
Micheal Collins.
\newblock 1999.
\newblock {\em Head-Driven Statistical Models for Natural Language Parsing}.
\newblock {Ph.D.} thesis, University of Pennsylvania.

\bibitem[\protect\citename{Crammer \bgroup et al.\egroup
  }2006]{crammer2006online}
Koby Crammer, Ofer Dekel, Joseph Keshet, Shai Shalev-Shwartz, and Yoram Singer.
\newblock 2006.
\newblock Online passive-aggressive algorithms.
\newblock {\em J. Mach. Learn. Res.}, 7:551--585, December.

\bibitem[\protect\citename{Crammer \bgroup et al.\egroup
  }2009]{crammer2009adaptive}
Koby Crammer, Alex Kulesza, and Mark Dredze.
\newblock 2009.
\newblock Adaptive regularization of weight vectors.
\newblock In {\em Advances in Neural Information Processing Systems},
  volume~22, pages 414--422, Vancouver, Canada. Curran Associates, Inc.

\bibitem[\protect\citename{de Marneffe \bgroup et al.\egroup
  }2006]{demarneffe06}
Marie-Catherine de~Marneffe, Bill Maccartney, and Christopher Manning.
\newblock 2006.
\newblock Generating typed dependency parses from phrase structure parses.
\newblock In {\em Proceedings of the Fifth International Conference on Language
  Resources and Evaluation}, Genoa, Italy. European Language Resources
  Association.

\bibitem[\protect\citename{Dozat and Manning}2017]{dozat2017deep}
Timothy Dozat and Christopher Manning.
\newblock 2017.
\newblock Deep biaffine attention for neural dependency parsing.
\newblock In {\em Proceedings of the 5th International Conference on Learning
  Representations}, Toulon, France.

\bibitem[\protect\citename{Dredze \bgroup et al.\egroup
  }2008]{dredze2008confidence}
Mark Dredze, Koby Crammer, and Fernando Pereira.
\newblock 2008.
\newblock Confidence-weighted linear classification.
\newblock In {\em Proceedings of the 25th international conference on Machine
  learning}, pages 264--271, Helsinki, Finland. Association for Computing
  Machinery.

\bibitem[\protect\citename{Dyer \bgroup et al.\egroup }2015]{dyer2015slstm}
Chris Dyer, Miguel Ballesteros, Wang Ling, Austin Matthews, and A.~Noah Smith.
\newblock 2015.
\newblock Transition-based dependency parsing with stack long short-term
  memory.
\newblock In {\em Proceedings of the 53rd Annual Meeting of the Association for
  Computational Linguistics and the 7th International Joint Conference on
  Natural Language Processing}, pages 334--343, Beijing, China. Association for
  Computational Linguistics.

\bibitem[\protect\citename{Francis and Kucera}1979]{browncorpus1979}
W.~N. Francis and H.~Kucera.
\newblock 1979.
\newblock Brown corpus manual: Manual of information to accompany a standard
  corpus of present-day edited american english, for use with digital
  computers.
\newblock Technical report, Department of Linguistics, Brown University.

\bibitem[\protect\citename{Goldman and Zhou}2000]{goldman00}
Sally~A. Goldman and Yan Zhou.
\newblock 2000.
\newblock Enhancing supervised learning with unlabeled data.
\newblock In {\em Proceedings of the Seventeenth International Conference on
  Machine Learning}, pages 327--334, San Francisco, California, USA. Morgan
  Kaufmann Publishers Inc.

\bibitem[\protect\citename{Goutam and Ambati}2011]{goutam2011exploring}
Rahul Goutam and Ram~Bharat Ambati.
\newblock 2011.
\newblock Exploring self training for {Hindi} dependency parsing.
\newblock In {\em Proceedings of 5th International Joint Conference on Natural
  Language Processing}, pages 1452--1456, Chiang Mai, Thailand. Asian
  Federation of Natural Language Processing.

\bibitem[\protect\citename{Haji{\v{c}} \bgroup et al.\egroup
  }2009]{hajic09conll}
Jan Haji{\v{c}}, Massimiliano Ciaramita, Richard Johansson, Daisuke Kawahara,
  Ant{\`o}nia~Maria Mart{\'i}, Llu{\'i}s M{\`a}rquez, Adam Meyers, Joakim
  Nivre, Sebastian Pad{\'o}, Jan {\v{S}}tep{\'a}nek, Pavel Stra{\v{n}}{\'a}k,
  Mihai Surdeanu, Nianwen Xue, and Yi~Zhang.
\newblock 2009.
\newblock The {CoNLL}-2009 shared task: Syntactic and semantic dependencies in
  multiple languages.
\newblock In {\em Proceedings of the Thirteenth Conference on Computational
  Natural Language Learning: Shared Task}, pages 1--18, Boulder, Colorado.
  Association for Computational Linguistics.

\bibitem[\protect\citename{Hatori \bgroup et al.\egroup }2011]{hatori2011joint}
Jun Hatori, Takuya Matsuzaki, Yusuke Miyao, and Jun'ichi Tsujii.
\newblock 2011.
\newblock Incremental joint {POS} tagging and dependency parsing in {Chinese}.
\newblock In {\em Proceedings of 5th International Joint Conference on Natural
  Language Processing}, pages 1216--1224, Chiang Mai, Thailand. Asian
  Federation of Natural Language Processing.

\bibitem[\protect\citename{Huang \bgroup et al.\egroup }2009]{huang09}
Liang Huang, Wenbin Jiang, and Qun Liu.
\newblock 2009.
\newblock Bilingually-constrained (monolingual) shift-reduce parsing.
\newblock In {\em Proceedings of the 2009 Conference on Empirical Methods in
  Natural Language Processing}, pages 1222--1231, Singapore. Association for
  Computational Linguistics.

\bibitem[\protect\citename{Johansson and Nugues}2007]{johansson2007extended}
Richard Johansson and Pierre Nugues.
\newblock 2007.
\newblock Extended constituent-to-dependency conversion for {English}.
\newblock In {\em Proceedings of the 16th Nordic Conference of Computational
  Linguistics}, pages 105--112, Tartu, Estonia. University of Tartu.

\bibitem[\protect\citename{Kahane \bgroup et al.\egroup }1998]{kahane98acl}
Sylvain Kahane, Alexis Nasr, and Owen Rambow.
\newblock 1998.
\newblock Pseudo-projectivity: A polynomially parsable non-projective
  dependency grammar.
\newblock In {\em Proceedings of the 17th International Conference on
  Computational Linguistics}, pages 646--652, Montreal, Quebec, Canada.
  Association for Computational Linguistics.

\bibitem[\protect\citename{Kawahara and Uchimoto}2008]{kawahara2008learning}
Daisuke Kawahara and Kiyotaka Uchimoto.
\newblock 2008.
\newblock Learning reliability of parses for domain adaptation of dependency
  parsing.
\newblock In {\em Proceedings of the Third International Joint Conference on
  Natural Language Processing}, pages 709--714, Hyderabad, India. Association
  for Computational Linguistics.

\bibitem[\protect\citename{Khan \bgroup et al.\egroup }2013]{khan13towards}
Mohammad Khan, Markus Dickinson, and Sandra K{\"u}bler.
\newblock 2013.
\newblock Towards domain adaptation for parsing web data.
\newblock In {\em Proceedings of the International Conference Recent Advances
  in Natural Language Processing}, pages 357--364, Hissar, Bulgaria. INCOMA
  Ltd.

\bibitem[\protect\citename{Klein and D.~Manning}2003]{klein03}
Dan Klein and Christopher D.~Manning.
\newblock 2003.
\newblock Accurate unlexicalized parsing.
\newblock In {\em Proceedings of the 41st Annual Meeting on Association for
  Computational Linguistics}, pages 423--430, Sapporo, Japan. Association for
  Computational Linguistics.

\bibitem[\protect\citename{Koo and Collins}2010]{koo10acl}
Terry Koo and Michael Collins.
\newblock 2010.
\newblock Efficient third-order dependency parsers.
\newblock In {\em Proceedings of the 48th Annual Meeting of the Association for
  Computational Linguistics}, pages 1--11, Uppsala, Sweden. Association for
  Computational Linguistics.

\bibitem[\protect\citename{Koo \bgroup et al.\egroup }2008]{koo08}
Terry Koo, Xavier Carreras, and Michael Collins.
\newblock 2008.
\newblock Simple semi-supervised dependency parsing.
\newblock In {\em Proceedings of the 46th Annual Meeting of the Association for
  Computational Linguistics: Human Language Technologies}, pages 595--603,
  Columbus, Ohio, USA. Association for Computational Linguistics.

\bibitem[\protect\citename{Le~Roux \bgroup et al.\egroup }2012]{le2012dcu}
Joseph Le~Roux, Jennifer Foster, Joachim Wagner, Rasul Samad Zadeh~Kaljahi, and
  Anton Bryl.
\newblock 2012.
\newblock {DCU-Paris13} systems for the {SANCL} 2012 shared task.
\newblock In {\em Proceedings of the First Workshop on Syntactic Analysis of
  Non-Canonical Language}, Montreal, Quebec, Canada.

\bibitem[\protect\citename{Li \bgroup et al.\egroup }2012]{li2012joint}
Zhenghua Li, Min Zhang, Wanxiang Che, and Ting Liu.
\newblock 2012.
\newblock A separately passive-aggressive training algorithm for joint {POS}
  tagging and dependency parsing.
\newblock In {\em Proceedings of the 24th International Conference on
  Computational Linguistics}, pages 1681--1698, Mumbai, India. Association for
  Computational Linguistics.

\bibitem[\protect\citename{Liang}2005]{liang05master}
Percy Liang.
\newblock 2005.
\newblock Semi-supervised learning for natural language.
\newblock Master's thesis, Massachusetts Institute of Technology.

\bibitem[\protect\citename{Malouf and Noord}2004]{malouf04wide}
Robert Malouf and Gertjan Noord.
\newblock 2004.
\newblock Wide coverage parsing with stochastic attribute value grammars.
\newblock In {\em Proceedings of the IJCNLP-04 Workshop on Beyond Shallow
  Analyses - Formalisms and statistical modeling for deep analyses}, Hainan,
  China. Asian Federation of Natural Language Processing.

\bibitem[\protect\citename{Martins \bgroup et al.\egroup
  }2010]{martins2010turbo}
Andre Martins, Noah Smith, Eric Xing, Pedro Aguiar, and Mario Figueiredo.
\newblock 2010.
\newblock {Turbo} parsers: Dependency parsing by approximate variational
  inference.
\newblock In {\em Proceedings of the 2010 Conference on Empirical Methods in
  Natural Language Processing}, pages 34--44, Cambridge, Massachusetts, USA.
  Association for Computational Linguistics.

\bibitem[\protect\citename{Martins \bgroup et al.\egroup
  }2013]{martins2013turning}
Andre Martins, Miguel Almeida, and A.~Noah Smith.
\newblock 2013.
\newblock Turning on the {Turbo}: Fast third-order non-projective {Turbo}
  parsers.
\newblock In {\em Proceedings of the 51st Annual Meeting of the Association for
  Computational Linguistics}, pages 617--622, Sofia, Bulgaria. Association for
  Computational Linguistics.

\bibitem[\protect\citename{McClosky \bgroup et al.\egroup
  }2006a]{mcclosky06naacl}
David McClosky, Eugene Charniak, and Mark Johnson.
\newblock 2006a.
\newblock Effective self-training for parsing.
\newblock In {\em Proceedings of the Human Language Technology Conference of
  the North American Chapter of the Association for Computational Linguistics},
  pages 152--159, New York, USA. Association for Computational Linguistics.

\bibitem[\protect\citename{McClosky \bgroup et al.\egroup
  }2006b]{mcclosky2006reranking}
David McClosky, Eugene Charniak, and Mark Johnson.
\newblock 2006b.
\newblock Reranking and self-training for parser adaptation.
\newblock In {\em Proceedings of the 21st International Conference on
  Computational Linguistics and 44th Annual Meeting of the Association for
  Computational Linguistics}, pages 337--344, Sydney, Australia. Association
  for Computational Linguistics.

\bibitem[\protect\citename{McDonald and Pereira}2006]{mcdonald2006online}
Ryan McDonald and Fernando Pereira.
\newblock 2006.
\newblock Online learning of approximate dependency parsing algorithms.
\newblock In {\em Proceedings of the 11th Conference of the European Chapter of
  the Association for Computational Linguistics}, pages 81--88, Trento, Italy.
  Association for Computational Linguistics.

\bibitem[\protect\citename{McDonald \bgroup et al.\egroup }2005]{mcdonald05acl}
Ryan McDonald, Koby Crammer, and Fernando Pereira.
\newblock 2005.
\newblock Online large-margin training of dependency parsers.
\newblock In {\em Proceedings of the 43rd Annual Meeting of the Association for
  Computational Linguistics}, pages 91--98, Ann Arbor, Michigan, USA.
  Association for Computational Linguistics.

\bibitem[\protect\citename{Mejer and Crammer}2012]{mejer2012}
Avihai Mejer and Koby Crammer.
\newblock 2012.
\newblock Are you sure? confidence in prediction of dependency tree edges.
\newblock In {\em Proceedings of the 2012 Conference of the North American
  Chapter of the Association for Computational Linguistics: Human Language
  Technologies}, pages 573--576, Montreal, Quebec, Canada. Association for
  Computational Linguistics.

\bibitem[\protect\citename{Mikolov \bgroup et al.\egroup
  }2013]{mikolov2013distributed}
Tomas Mikolov, Ilya Sutskever, Kai Chen, Greg~S Corrado, and Jeff Dean.
\newblock 2013.
\newblock Distributed representations of words and phrases and their
  compositionality.
\newblock In {\em Advances in neural information processing systems},
  volume~26, pages 3111--3119. Curran Associates, Inc.

\bibitem[\protect\citename{Mirroshandel \bgroup et al.\egroup
  }2012]{mirroshandel12}
Abolghasem~Seyed Mirroshandel, Alexis Nasr, and Joseph Le~Roux.
\newblock 2012.
\newblock Semi-supervised dependency parsing using lexical affinities.
\newblock In {\em Proceedings of the 50th Annual Meeting of the Association for
  Computational Linguistics}, pages 777--785, Jeju Island, Korea. Association
  for Computational Linguistics.

\bibitem[\protect\citename{Nivre \bgroup et al.\egroup }2007a]{nivre07conll}
Joakim Nivre, Johan Hall, Sandra K{\"u}bler, Ryan McDonald, Jens Nilsson,
  Sebastian Riedel, and Deniz Yuret.
\newblock 2007a.
\newblock The {CoNLL} 2007 shared task on dependency parsing.
\newblock In {\em Proceedings of the 2007 Joint Conference on Empirical Methods
  in Natural Language Processing and Computational Natural Language Learning},
  pages 915--932, Prague, Czech Republic. Association for Computational
  Linguistics.

\bibitem[\protect\citename{Nivre \bgroup et al.\egroup }2007b]{nivre07nle}
Joakim Nivre, Johan Hall, Jens Nilsson, Atanas Chanev, G{\"u}l{\c{s}}en
  Eryi{\v{g}}it, Sandra K{\"u}bler, Svetoslav Marinov, and Erwin Marsi.
\newblock 2007b.
\newblock Maltparser: A language-independent system for data-driven dependency
  parsing.
\newblock {\em Natural Language Engineering}, 13(2):95--135.

\bibitem[\protect\citename{Nivre}2004]{nivre2004incrementality}
Joakim Nivre.
\newblock 2004.
\newblock Incrementality in deterministic dependency parsing.
\newblock In {\em Proceedings of the Workshop on Incremental Parsing: Bringing
  Engineering and Cognition Together}, pages 50--57. Association for
  Computational Linguistics.

\bibitem[\protect\citename{Nivre}2007]{nivre07naacl}
Joakim Nivre.
\newblock 2007.
\newblock Incremental non-projective dependency parsing.
\newblock In {\em Proceedings of the Human Language Technologies 2007: The
  Conference of the North American Chapter of the Association for Computational
  Linguistics}, pages 396--403, Rochester, New York, USA. Association for
  Computational Linguistics.

\bibitem[\protect\citename{Nivre}2009]{nivre2009non}
Joakim Nivre.
\newblock 2009.
\newblock Non-projective dependency parsing in expected linear time.
\newblock In {\em Proceedings of the Joint Conference of the 47th Annual
  Meeting of the ACL and the 4th International Joint Conference on Natural
  Language Processing}, pages 351--359, Singapore. Association for
  Computational Linguistics.

\bibitem[\protect\citename{P.~Marcus \bgroup et al.\egroup }1993]{marcus93}
Mitchell P.~Marcus, Beatrice Santorini, and Mary Ann~Marcinkiewicz.
\newblock 1993.
\newblock Building a large annotated corpus of {English}: The {Penn Treebank}.
\newblock {\em Computational Linguistics}, 19(2):313--330.

\bibitem[\protect\citename{Pekar \bgroup et al.\egroup
  }2014]{pekar2014exploring}
Viktor Pekar, Juntao Yu, Mohab Elkaref, and Bernd Bohnet.
\newblock 2014.
\newblock Exploring options for fast domain adaptation of dependency parsers.
\newblock In {\em Proceedings of the First Joint Workshop on Statistical
  Parsing of Morphologically Rich Languages and Syntactic Analysis of
  Non-Canonical Languages}, pages 54--65, Dublin, Ireland. Dublin City
  University.

\bibitem[\protect\citename{Pennington \bgroup et al.\egroup
  }2014]{pennington2014glove}
Jeffrey Pennington, Richard Socher, and Christopher Manning.
\newblock 2014.
\newblock Glove: Global vectors for word representation.
\newblock In {\em Proceedings of the 2014 Conference on Empirical Methods in
  Natural Language Processing}, pages 1532--1543, Doha, Qatar. Association for
  Computational Linguistics.

\bibitem[\protect\citename{Petrov and Klein}2007]{petrov07}
Slav Petrov and Dan Klein.
\newblock 2007.
\newblock Improved inference for unlexicalized parsing.
\newblock In {\em Proceedings of the Human Language Technologies 2007: The
  Conference of the North American Chapter of the Association for Computational
  Linguistics}, pages 404--411, Rochester, New York, USA. Association for
  Computational Linguistics.

\bibitem[\protect\citename{Petrov and McDonald}2012]{petrov2012overview}
Slav Petrov and Ryan McDonald.
\newblock 2012.
\newblock Overview of the 2012 shared task on parsing the web.
\newblock In {\em Proceedings of the First Workshop on Syntactic Analysis of
  Non-Canonical Language}, Montreal, Quebec, Canada.

\bibitem[\protect\citename{Plank and S{\o}gaard}2013]{plank2013experiments}
Barbara Plank and Anders S{\o}gaard.
\newblock 2013.
\newblock Experiments in newswire-to-law adaptation of graph-based dependency
  parsers.
\newblock In {\em Evaluation of Natural Language and Speech Tools for Italian},
  pages 70--76, Berlin, Heidelberg. Springer.

\bibitem[\protect\citename{Plank and van Noord}2011]{plank2011effective}
Barbara Plank and Gertjan van Noord.
\newblock 2011.
\newblock Effective measures of domain similarity for parsing.
\newblock In {\em Proceedings of the 49th Annual Meeting of the Association for
  Computational Linguistics: Human Language Technologies}, pages 1566--1576,
  Portland, Oregon, USA. Association for Computational Linguistics.

\bibitem[\protect\citename{Plank}2011]{bplank2011phd}
Barbara Plank.
\newblock 2011.
\newblock {\em Domain Adaptation for Parsing}.
\newblock {Ph.D.} thesis, University of Groningen.

\bibitem[\protect\citename{Pradhan \bgroup et al.\egroup }2011]{pradhan11conll}
Sameer Pradhan, Lance Ramshaw, Mitchell Marcus, Martha Palmer, Ralph
  Weischedel, and Nianwen Xue.
\newblock 2011.
\newblock {CoNLL}-2011 shared task: Modeling unrestricted coreference in
  {OntoNotes}.
\newblock In {\em Proceedings of the Fifteenth Conference on Computational
  Natural Language Learning: Shared Task}, pages 1--27, Portland, Oregon, USA.
  Association for Computational Linguistics.

\bibitem[\protect\citename{Pradhan \bgroup et al.\egroup }2012]{pradhan12conll}
Sameer Pradhan, Alessandro Moschitti, Nianwen Xue, Olga Uryupina, and Yuchen
  Zhang.
\newblock 2012.
\newblock {CoNLL}-2012 shared task: Modeling multilingual unrestricted
  coreference in {OntoNotes}.
\newblock In {\em Proceedings of the 2012 Joint Conference on Empirical Methods
  in Natural Language Processing and Computational Natural Language Learning:
  Shared Task}, pages 1--40, Jeju Island, Korea. Association for Computational
  Linguistics.

\bibitem[\protect\citename{Pyysalo \bgroup et al.\egroup }2006]{pyysalo2006}
Sampo Pyysalo, Tapio Salakoski, Sophie Aubin, and Adeline Nazarenko.
\newblock 2006.
\newblock Lexical adaptation of link grammar to the biomedical sublanguage: A
  comparative evaluation of three approaches.
\newblock {\em BMC Bioinformatics}, 7(Suppl 3).

\bibitem[\protect\citename{Reichart and Rappoport}2007]{reichart2007self}
Roi Reichart and Ari Rappoport.
\newblock 2007.
\newblock Self-training for enhancement and domain adaptation of statistical
  parsers trained on small datasets.
\newblock In {\em Proceedings of the 45th Annual Meeting of the Association of
  Computational Linguistics}, pages 616--623, Prague, Czech Republic.
  Association for Computational Linguistics.

\bibitem[\protect\citename{Sagae and Tsujii}2007]{sagae07}
Kenji Sagae and Jun'ichi Tsujii.
\newblock 2007.
\newblock Dependency parsing and domain adaptation with {LR} models and parser
  ensembles.
\newblock In {\em Proceedings of the 2007 Joint Conference on Empirical Methods
  in Natural Language Processing and Computational Natural Language Learning},
  pages 1044--1050, Prague, Czech Republic. Association for Computational
  Linguistics.

\bibitem[\protect\citename{Sagae}2010]{sagae2010self}
Kenji Sagae.
\newblock 2010.
\newblock Self-training without reranking for parser domain adaptation and its
  impact on semantic role labeling.
\newblock In {\em Proceedings of the 2010 Workshop on Domain Adaptation for
  Natural Language Processing}, pages 37--44, Uppsala, Sweden. Association for
  Computational Linguistics.

\bibitem[\protect\citename{Sarkar}2001]{sarkar01}
Anoop Sarkar.
\newblock 2001.
\newblock Applying co-training methods to statistical parsing.
\newblock In {\em Proceedings of the Second Meeting of the North American
  Chapter of the Association for Computational Linguistics}, pages 1--8,
  Pittsburgh, Pennsylvania, USA. Association for Computational Linguistics.

\bibitem[\protect\citename{Seddah \bgroup et al.\egroup
  }2013]{seddah2013overview}
Djam{\'e} Seddah, Reut Tsarfaty, Sandra K{\"u}bler, Marie Candito, D.~Jinho
  Choi, Rich{\'a}rd Farkas, Jennifer Foster, Iakes Goenaga, Koldo
  Gojenola~Galletebeitia, Yoav Goldberg, Spence Green, Nizar Habash, Marco
  Kuhlmann, Wolfgang Maier, Joakim Nivre, Adam Przepi{\'o}rkowski, Ryan Roth,
  Wolfgang Seeker, Yannick Versley, Veronika Vincze, Marcin Woli{\'{n}}ski,
  Alina Wr{\'o}blewska, and Villemonte~Eric de~la Clergerie.
\newblock 2013.
\newblock Overview of the {SPMRL} 2013 shared task: A cross-framework
  evaluation of parsing morphologically rich languages.
\newblock In {\em Proceedings of the Fourth Workshop on Statistical Parsing of
  Morphologically-Rich Languages}, pages 146--182, Seattle, Washington, USA.
  Association for Computational Linguistics.

\bibitem[\protect\citename{Seddah \bgroup et al.\egroup
  }2014]{seddah2014introducing}
Djam{\'e} Seddah, Sandra K{\"u}bler, and Reut Tsarfaty.
\newblock 2014.
\newblock Introducing the {SPMRL} 2014 shared task on parsing
  morphologically-rich languages.
\newblock In {\em Proceedings of the First Joint Workshop on Statistical
  Parsing of Morphologically Rich Languages and Syntactic Analysis of
  Non-Canonical Languages}, pages 103--109, Dublin, Ireland. Dublin City
  University.

\bibitem[\protect\citename{Shen \bgroup et al.\egroup }2008]{shen2008new}
Libin Shen, Jinxi Xu, and Ralph Weischedel.
\newblock 2008.
\newblock A new string-to-dependency machine translation algorithm with a
  target dependency language model.
\newblock In {\em Proceedings of the 46th Annual Meeting of the Association for
  Computational Linguistics: Human Language Technologies}, pages 577--585,
  Columbus, Ohio, USA. Association for Computational Linguistics.

\bibitem[\protect\citename{S{\o}gaard and Plank}2012]{sogaard12sancl}
Anders S{\o}gaard and Barbara Plank.
\newblock 2012.
\newblock Parsing the web as covariate shift.
\newblock In {\em Proceedings of the First Workshop on Syntactic Analysis of
  Non-Canonical Language}, Montreal, Quebec, Canada.

\bibitem[\protect\citename{S{\o}gaard and Rish{\o}j}2010]{sogaard2010}
Anders S{\o}gaard and Christian Rish{\o}j.
\newblock 2010.
\newblock Semi-supervised dependency parsing using generalized tri-training.
\newblock In {\em Proceedings of the 23rd International Conference on
  Computational Linguistics}, pages 1065--1073, Beijing, China. Association for
  Computational Linguistics.

\bibitem[\protect\citename{Steedman \bgroup et al.\egroup
  }2003]{steedman2003semi}
Mark Steedman, Miles Osborne, Anoop Sarkar, Stephen Clark, Rebecca Hwa, Julia
  Hockenmaier, Paul Ruhlen, Steven Baker, and Jeremiah Crim.
\newblock 2003.
\newblock Bootstrapping statistical parsers from small datasets.
\newblock In {\em Proceedings of the 10th Conference of the European Chapter of
  the Association for Computational Linguistics}, pages 331--338, Budapest,
  Hungary. Association for Computational Linguistics.

\bibitem[\protect\citename{Surdeanu \bgroup et al.\egroup
  }2008]{surdeanu08conll}
Mihai Surdeanu, Richard Johansson, Adam Meyers, Llu{\'i}s M{\`a}rquez, and
  Joakim Nivre.
\newblock 2008.
\newblock The {CoNLL} 2008 shared task on joint parsing of syntactic and
  semantic dependencies.
\newblock In {\em Proceedings of the Twelfth Conference on Computational
  Natural Language Learning}, pages 159--177, Manchester, England. Association
  for Computational Linguistics.

\bibitem[\protect\citename{Szolovits}2003]{szolovits2003adding}
Peter Szolovits.
\newblock 2003.
\newblock Adding a medical lexicon to an {English} parser.
\newblock {\em AMIA Annual Symposium Proceedings}, pages 639--643.

\bibitem[\protect\citename{Tiedemann}2012]{tiedemann12}
J{\"o}rg Tiedemann.
\newblock 2012.
\newblock Parallel data, tools and interfaces in {OPUS}.
\newblock In {\em Proceedings of the Eighth International Conference on
  Language Resources and Evaluation}, pages 2214--2218, Istanbul, Turkey.
  European Language Resources Association.

\bibitem[\protect\citename{Weiss \bgroup et al.\egroup }2015]{weiss2015neural}
David Weiss, Chris Alberti, Michael Collins, and Slav Petrov.
\newblock 2015.
\newblock Structured training for neural network transition-based parsing.
\newblock In {\em Proceedings of the 53rd Annual Meeting of the Association for
  Computational Linguistics and the 7th International Joint Conference on
  Natural Language Processing}, pages 323--333, Beijing, China. Association for
  Computational Linguistics.

\bibitem[\protect\citename{Xue \bgroup et al.\egroup }2005]{xue05}
Naiwen Xue, Fei Xia, Fu-dong Chiou, and Marta Palmer.
\newblock 2005.
\newblock The {Penn Chinese TreeBank}: Phrase structure annotation of a large
  corpus.
\newblock {\em Natural Language Engineering}, 11(2):207--238.

\bibitem[\protect\citename{Yamada and Matsumoto}2003]{yamada03}
Hiroyasu Yamada and Yuji Matsumoto.
\newblock 2003.
\newblock Statistical dependency analysis with support vector machines.
\newblock In {\em Proceedings of the 8th International Workshop on Parsing
  Technologies}, pages 195--206, Nancy, France. Lorraine Laboratory for
  Research in Information Technology and its Applications.

\bibitem[\protect\citename{Yu and Bohnet}2015]{yu2015depling}
Juntao Yu and Bernd Bohnet.
\newblock 2015.
\newblock Exploring confidence-based self-training for multilingual dependency
  parsing in an under-resourced language scenario.
\newblock In {\em Proceedings of the Third International Conference on
  Dependency Linguistics}, pages 350--358, Uppsala, Sweden. Uppsala University.

\bibitem[\protect\citename{Yu and Bohnet}2017]{yu2017iwpt}
Juntao Yu and Bernd Bohnet.
\newblock 2017.
\newblock Dependency language models for transition-based dependency parsing.
\newblock In {\em Proceedings of the 15th International Conference on Parsing
  Technologies}, pages 11--17, Pisa, Italy. Association for Computational
  Linguistics.

\bibitem[\protect\citename{Yu \bgroup et al.\egroup }2015]{yu2015iwpt}
Juntao Yu, Mohab Elkaref, and Bernd Bohnet.
\newblock 2015.
\newblock Domain adaptation for dependency parsing via self-training.
\newblock In {\em Proceedings of the 14th International Conference on Parsing
  Technologies}, pages 1--10, Bilbao, Spain. Association for Computational
  Linguistics.

\bibitem[\protect\citename{Zhang and Clark}2008]{zhang08}
Yue Zhang and Stephen Clark.
\newblock 2008.
\newblock A tale of two parsers: Investigating and combining graph-based and
  transition-based dependency parsing.
\newblock In {\em Proceedings of the 2008 Conference on Empirical Methods in
  Natural Language Processing}, pages 562--571, Honolulu, Hawaii, USA.
  Association for Computational Linguistics.

\bibitem[\protect\citename{Zhang and McDonald}2014]{zhang2014enforcing}
Hao Zhang and Ryan McDonald.
\newblock 2014.
\newblock Enforcing structural diversity in cube-pruned dependency parsing.
\newblock In {\em Proceedings of the 52nd Annual Meeting of the Association for
  Computational Linguistics}, pages 656--661, Baltimore, Maryland, USA.
  Association for Computational Linguistics.

\bibitem[\protect\citename{Zhang and Nivre}2011]{zhang11}
Yue Zhang and Joakim Nivre.
\newblock 2011.
\newblock Transition-based dependency parsing with rich non-local features.
\newblock In {\em Proceedings of the 49th Annual Meeting of the Association for
  Computational Linguistics: Human Language Technologies}, pages 188--193,
  Portland, Oregon, USA. Association for Computational Linguistics.

\bibitem[\protect\citename{Zhang \bgroup et al.\egroup }2012]{zhang12hit}
Meishan Zhang, Wanxiang Che, Yijia Liu, Zhenghua Li, and Ting Liu.
\newblock 2012.
\newblock {HIT} dependency parsing: Bootstrap aggregating heterogeneous
  parsers.
\newblock In {\em Proceedings of the First Workshop on Syntactic Analysis of
  Non-Canonical Language}, Montreal, Quebec, Canada.

\bibitem[\protect\citename{Zhou and Li}2005]{zhou2005tri}
Zhi-Hua Zhou and Ming Li.
\newblock 2005.
\newblock Tri-training: exploiting unlabeled data using three classifiers.
\newblock {\em IEEE Transactions on Knowledge and Data Engineering},
  17(11):1529--1541.

\bibitem[\protect\citename{Zhou \bgroup et al.\egroup
  }2011]{zhou2011exploiting}
Guangyou Zhou, Jun Zhao, Kang Liu, and Li~Cai.
\newblock 2011.
\newblock Exploiting web-derived selectional preference to improve statistical
  dependency parsing.
\newblock In {\em Proceedings of the 49th Annual Meeting of the Association for
  Computational Linguistics: Human Language Technologies}, pages 1556--1565,
  Portland, Oregon, USA. Association for Computational Linguistics.

\bibitem[\protect\citename{Zhu}2005]{zhu2005semi}
Xiaojin Zhu.
\newblock 2005.
\newblock Semi-supervised learning literature survey.
\newblock Technical Report 1530, Computer Sciences, University of
  Wisconsin-Madison.

\end{thebibliography}

\end{document}